%% file: main.tex
\documentclass{article}

\usepackage{microtype}
\usepackage{graphicx}
\usepackage{subcaption}
\usepackage{booktabs} %

\usepackage{hyperref}
\usepackage{caption}

\usepackage{enumitem}

\setlist[itemize]{nosep}

\usepackage[accepted]{icml2026}

\usepackage{amsmath}
\usepackage{amssymb}
\usepackage{mathtools}
\usepackage{amsthm}

\usepackage[capitalize,noabbrev]{cleveref}

\input{preamble}

\theoremstyle{plain}

\theoremstyle{definition}

\theoremstyle{remark}

\usepackage[textsize=tiny]{todonotes}

\icmltitlerunning{\acronym}

\begin{document}

\twocolumn[
  \icmltitle{{\color{nvidiagreen}Ada}ptive {\color{nvidiagreen}Vo}lumetric {\color{nvidiagreen}M}echanical {\color{nvidiagreen}P}roperty Fields Invariant to Resolution}

  \icmlsetsymbol{equal}{*}

  \begin{icmlauthorlist}
    \icmlauthor{Rishit Dagli}{nv,uoft}
    \icmlauthor{Donglai Xiang}{nv}
    \icmlauthor{Vismay Modi}{nv}
    \icmlauthor{Xuning Yang}{nv}\\
    \icmlauthor{Gavriel State}{nv}
    \icmlauthor{David I.W. Levin}{nv,uoft}
    \icmlauthor{Maria Shugrina}{nv}
  \end{icmlauthorlist}
  \centering{\url{https://research.nvidia.com/labs/sil/projects/adavomp/}}

  \icmlaffiliation{nv}{NVIDIA}
  \icmlaffiliation{uoft}{University of Toronto}

  \icmlcorrespondingauthor{Rishit Dagli}{rdagli@nvidia.com}

  \icmlkeywords{Physics-based Modeling, 3D Dynamics, Simulation, Interactive Worlds}

  \vskip 0.3in
]

\printAffiliationsAndNotice{}  %

\begin{figure*}[t]
    \centering
    \includegraphics{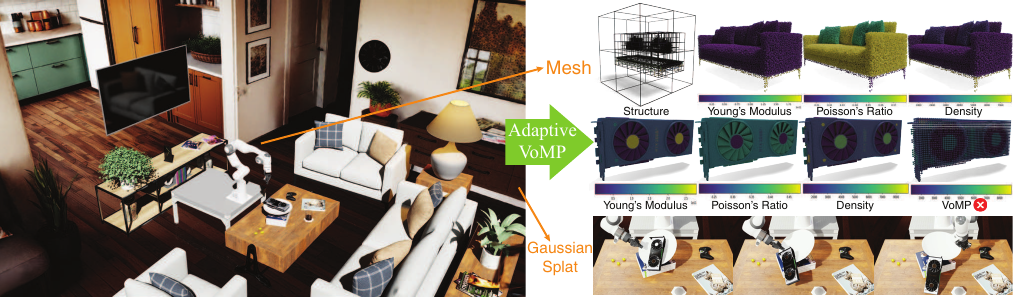}
    \caption{\textbf{\acronym}\ generates high-resolution physically accurate volumetric mechanical property fields with detailed parts across 3D representations, enabling their use in building realistic interactive worlds and deformable simulations. We simulate a robot interacting with a high-resolution GPU and the sofa or pillows being stable under gravity in this Gaussian splat + mesh environment. (\video{01 :13})}
    \label{fig:teaser}
\end{figure*}

\begin{abstract}
\input{text/abstract}
\end{abstract}

\input{text/intro}

\input{text/related}

\input{text/method}
\input{text/results}

\section*{Acknowledgements}

We thank Gilles Daviet for help in setting up some of the simulations. We thank Jean-Francois Lafleche for help with rendering. We thank Beau Perschall, Katherine Cheung for help in using the datasets. We thank Ruchik Thaker for help in releasing the code and data. We thank Andre Pradhana, Anka He Chen, Anita Hu, Charles Loop, Clement Fuji Tsang, Francis Williams, Hexu Zhao and Ken Museth for insightful discussions.

\section*{Impact Statement}

This paper studies conditional generation of volumetric mechanical property fields from geometric and visual cues. A potential positive impact is to reduce the cost of building simulation-ready digital assets by providing a learned prior over physically plausible, spatially varying materials, which may benefit downstream tasks such as simulation and interactive scene generation. A risk with our model like many other models is that it can be misused to create realistic digital content or deepfakes. The risk is misuse of predicted properties in safety-critical decisions without validation. Our outputs are learned estimates and can be wrong under distribution shift, partial observability, or atypical materials; using them as a substitute for measurement, testing, or certified engineering analysis could lead to unsafe designs or incorrect conclusions. We view the method as a tool for accelerating asset preparation and providing initialization for downstream pipelines, not as a replacement for verification.

\clearpage
\nocite{polyscope}
\bibliography{example_paper}
\bibliographystyle{icml2026}

\newpage
\appendix
\onecolumn

\makeatletter
\def\addcontentsline#1#2#3{%
  \addtocontents{#1}{\protect\contentsline{#2}{#3}{\thepage}{\@currentHref}{}}}
\makeatother

\begin{center}
    {\LARGE \bfseries Supplementary Material for {\color{nvidiagreen}Ada}ptive {\color{nvidiagreen}Vo}lumetric {\color{nvidiagreen}M}echanical {\color{nvidiagreen}P}roperty Fields Invariant to Resolution}
\end{center}
\vspace{2em}

\section*{Supplementary Contents}
\startcontents
\printcontents{l}{1}{\setcounter{tocdepth}{2}}

\clearpage
\twocolumn

\input{text/appendix/main}

\stopcontents

\end{document}

%% file: preamble.tex
\usepackage{layouts}
\usepackage[T1]{fontenc}
\usepackage{graphicx}
\usepackage{booktabs}
\usepackage{subcaption}
\usepackage{adjustbox}
\usepackage{multirow}
\usepackage{wrapfig}
\usepackage{layouts}
\usepackage{float}
\usepackage{array}
\usepackage{tabularx}
\usepackage[table]{xcolor}
\usepackage{tikz}
\usepackage{tcolorbox}
\usepackage{color}
\usepackage{amsthm}
\usepackage{tikz}
\usepackage{colortbl}
\usepackage{soul}
\usepackage{multicol}
\usepackage{multirow}
\usepackage{titletoc}
\usepackage{csquotes}

\titlecontents{lsection}[1.5em]{\smallskip\bfseries}{\contentslabel{1.5em}}{}{\titlerule*[0.5pc]{.}\contentspage}
\titlecontents{lsubsection}[4em]{}{\contentslabel{2.5em}}{}{\titlerule*[0.5pc]{.}\contentspage}
\usepackage{tikz}
\usepackage{makecell}
\usepackage{pifont}
\usepackage{svg}
\usepackage[flushleft]{threeparttable}
\usepackage{titletoc}

\definecolor{nvidiagreen}{HTML}{76B900}
\definecolor{nvidiagreen_light}{HTML}{E1FFAD}
\definecolor{my_green}{RGB}{51,102,0}
\definecolor{my_red}{RGB}{204, 0, 0}
\newcommand{\stdfmt}[1]{{\textbf{\scriptsize\textcolor{gray}{($\pm$#1)}}}}

\newcommand{\acronym}{\textsc{AdaVoMP}}
\newcommand{\vacronym}{\textsc{SAV}}
\newcommand{\gvt}{\textsc{GVT}}
\input{math_commands}

\definecolor{mainblue}{HTML}{1f77b4}  %
\definecolor{lightblue}{HTML}{CCE5FF}  %
\definecolor{mainorange}{HTML}{ff7f0e}  %
\definecolor{lightorange}{HTML}{FFE5CC}  %

\newcommand{\video}[1]{%
  \includegraphics[height=0.6\baselineskip]{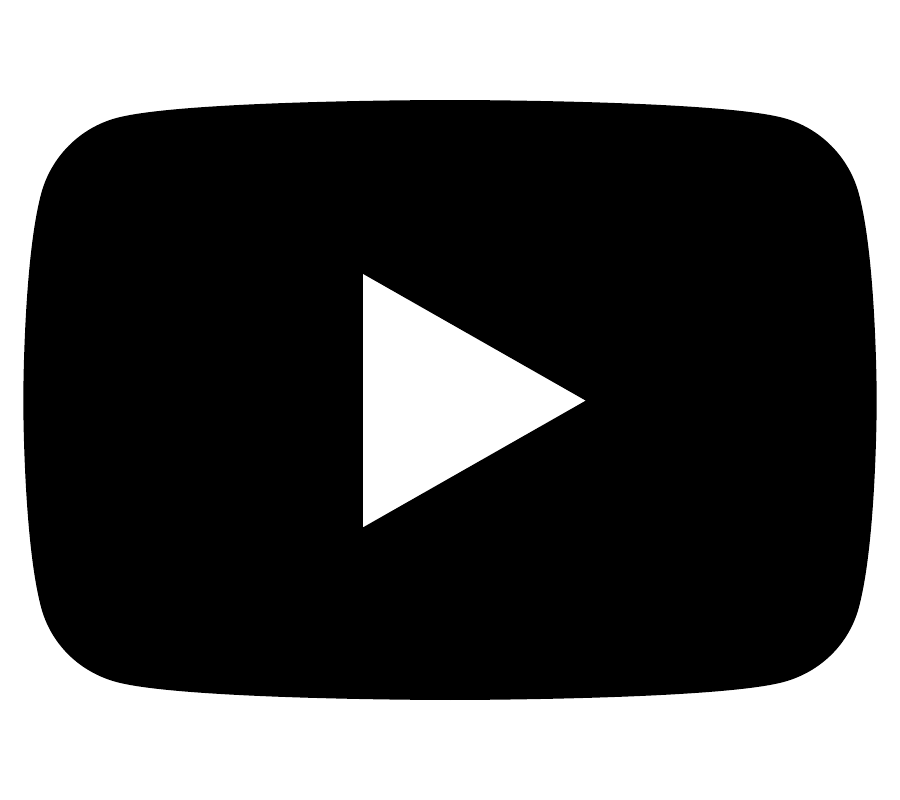}%
  {\color[HTML]{76B900}: #1}%
}

\newcommand{\cC}{\mathcal{C}}

\newcommand{\cL}{\mathcal{L}}
\newcommand{\cM}{\mathcal{M}}

\newcommand{\cP}{\mathcal{P}}

\newcommand{\cV}{\mathcal{V}}

\newcommand{\mattriplet}{($E$, $\nu$, $\rho$)}

\newcommand{\ourmodel}{\textsc{AdaVoMP}}

%% file: math_commands.tex
\usepackage{amsmath,amsfonts,bm}
\usepackage{mathtools}

\def\eqref#1{equation~\ref{#1}}

\def\1{\bm{1}}

\DeclareMathAlphabet{\mathsfit}{\encodingdefault}{\sfdefault}{m}{sl}
\SetMathAlphabet{\mathsfit}{bold}{\encodingdefault}{\sfdefault}{bx}{n}

\newcommand{\bzero}{\mathbf{0}}

\newcommand{\bi}{\mathbf{i}}
\newcommand{\be}{\mathbf{e}}
\newcommand{\bh}{\mathbf{h}}
\newcommand{\bu}{\mathbf{u}}

\newcommand{\bx}{\mathbf{x}}

\newcommand{\bz}{\mathbf{z}}

%% file: text/abstract.tex
Accurate mechanical properties (or materials) Young's modulus ($E$), Poisson's ratio ($\nu$) and density ($\rho$) are essential
for reliable physics simulation of digital worlds, but most 3D assets lack this information.
We propose {\ourmodel}, a method for predicting accurate dense spatially-varying {\mattriplet} for input 3D objects across representations, improving the resolution, accuracy, and memory efficiency over the state-of-the-art. The foundation of our technique is a sparse and \emph{adaptive} voxel structure {\vacronym} that efficiently represents both the input 3D shape and the material field output.
We replace the fixed-voxel model of the most accurate prior method, VoMP, with a novel sparse transformer encoder-decoder model that 
learns to generate a unique {\vacronym} autoregressively for every input shape to represent its materials, achieving a resolution $16^3\times$ higher than prior art. Experiments show that {\acronym} estimates more accurate volumetric properties, even with lesser test-time compute than all prior art. This allows us to convert high-resolution complex 3D objects into simulation-ready assets, resulting in realistic deformable simulations.

%% file: text/intro.tex
\section{Introduction}

Surging interest in robotics is amplifying the demand for 
realistic digital environments, suitable for training robotic agents with physics simulation in the loop.
However, constructing such environments remains labor-intensive. Typical 3D scenes, authored, generated or captured from photos,
lack the parameters necessary for physics simulation,
notably the mechanical properties, Young's modulus ($E$), Poisson's ratio ($\nu$) and density ($\rho$),
all of which are \emph{spatially-varying} and must be defined \emph{throughout each object's volume} to ensure accurate
simulation of real-world behaviors. Accurately assigning such properties manually is difficult to impossible, and measuring
real-world objects do not scale with the demand for digital simulation. Recent works \cite{dagli2025vomppredictingvolumetricmechanical,le2025pixie} propose
techniques that learn to predict spatially varying material properties for 3D objects automatically, but
are limited in either their accuracy or resolution. 

We propose {\ourmodel}, a method for predicting accurate spatially-varying mechanical properties {\mattriplet} for input 3D shapes
using an adaptive structure, improving the resolution, accuracy, and memory efficiency over state-of-the-art. Our method replaces the fixed voxel model of the most accurate prior method, VoMP
\cite{dagli2025vomppredictingvolumetricmechanical}, with a novel sparse transformer encoder-decoder model that 
learns to generate a unique adaptive structure for every input shape to compactly represent its material distribution. This allows us to operate at the max
resolution of $1024^3$, compared to $64^3$ of prior art \cite{dagli2025vomppredictingvolumetricmechanical,le2025pixie}. 
We introduce an adaptive structure that uses only a few voxels for constant material regions (e.g.\ the couch armrests in Fig.\ref{fig:teaser}),
concentrating the model's predictive capacity in more challenging regions and sharp material boundaries, achieving much finer predictions than VoMP (e.g.\ GPU density in Fig.\ref{fig:teaser}). 

To accomplish this, we introduce sparse adaptive voxel trees (\vacronym) to represent the input shape and to autoregressively generate its material field. 
Unlike prior art \cite{xiang2025structured3dlatentsscalable, dagli2025vomppredictingvolumetricmechanical}, we aggregate multi-view visual features of the 3D input
into a more efficient \emph{adaptive} structure. We introduce a learned sparse transformer encoder for processing this input while attending across multi-level voxels, and
a jointly trained sparse transformer generator, that learns to \emph{autoregressively} output materials as a compact {\vacronym} representation. 
We introduce a generative mechanism that predicts both \emph{structure} (per-voxel "\textsc{Keep}"/"\textsc{Subdivide}"/"\textsc{Empty}") and \emph{material} values at every level. All models are trained jointly on a dataset auto-labeled with a VLM-based pipeline, similar to prior techniques. 
In summary, our contributions are as follows:
    \begin{itemize}
        \item A \textbf{sparse adaptive voxel (\vacronym)} representation for 3D shapes and materials, designed for transformer-based processing and generation, and efficient querying (\S\ref{sec:save}).
        \item An \textbf{Adaptive Geometry Transformer} that embeds adaptive DINO feature trees with unified coordinate embeddings and sparse windowed attention (\S\ref{sec:method_encoder}). 
        \item A \textbf{novel Generator design with a autoregressive mechanism} for generating {\vacronym}s coarse-to-fine (\S\ref{sec:method_decoder}).
        \item A \textbf{training formulation} for autoregressive generation, combining multi-scales supervision, teacher forcing and explicit empty space negatives (\S\ref{sec:method_training}).
        \item Significantly higher resolution and accuracy over state-of-the-art mechanical property prediction methods, advancing simulatable environment authoring (\S\ref{sec:results}).
        \item Extensive ablations of model design and scale (\S\ref{sec:app_ablations}.).
    \end{itemize}

\paragraph{Conflict of Interest Disclosure.} The authors are employed by NVIDIA, which leads the development of VoMP, which was among the ones evaluated in this paper.

%% file: text/related.tex
\section {Related Works}
\label{sec:rw}

To accurately predict dynamic behavior, deformable simulations rely on constitutive (or material) models (e.g. Neo-Hookean, St. Venant Kirchoff), which require parameter fields for Young's modulus, Poisson Ratio, and density \mattriplet. Physically accurate parameters are portable across diverse material models, enabling consistent simulation results; in contrast, methods optimizing for computational speed often require modifying those parameters to mitigate numerical instability~\cite{xpbd,mpm}.

\paragraph{Inverse Physics vs. Static Inference.}\label{sec:rw2} 
Material parameters can be obtained via expensive real-world measurement or inverse optimization. Inverse physics methods~\cite{physdreamer, huang2024dreamphysicslearningphysicsbased3d, Liu_2025_CVPR, cleach2023differentiablephysicssimulationdynamicsaugmented, liu2024physics3d, lin2025omniphysgs} optimize parameters from video or priors but suffer from overfitting, simulator-dependence~\cite{mpm, le2025pixie}, and poor scalability. In contrast, feed-forward methods like~\citet{dagli2025vomppredictingvolumetricmechanical} and ours, learn from ground truth material datasets and infer volumetric parameters from static scenes, enabling rapid run-time inference.

\paragraph{Mechanical Property Datasets.}
Predicting volumetric mechanical properties from shape and appearance alone is difficult for learning-based methods, largely due to limited datasets ~\cite{gao2022objectfolder20multisensoryobject, downs2022googlescannedobjectshighquality, chen2025matpredictdatasetbenchmarklearning} and noisy data ~\cite{lin2018learning}, overfit to a simulator ~\cite{mishra2024latticemldatadrivenapplicationpredicting, Xie__2025, belikov2015material}, or coarsely annotated~\cite{ahmed20253dcompat200, slim20233dcompat, 10.1007/978-3-031-20074-8_7}, or limited to rigid objects~\cite{cao2025physx-3d}. High-quality physical data remains difficult to collect~\cite{ASTM_D638_2022, ASTM_E8_E8M_2024, pai2000robotics, loveday2004tensile}. Our model can be trained with part-segmented 3D assets datasets which have mechanical properties, and thus we reuse the VLM data annotation from prior art \cite{dagli2025vomppredictingvolumetricmechanical}.

\paragraph{Inferring Materials for Static Scenes.}
Approaches based on NeRF and Gaussian splats~\cite{mildenhall2020nerf, 10.1145/3592433}, including ~\cite{zhai2024physicalpropertyunderstandinglanguageembedded,shuai2025pugszeroshotphysicalunderstanding}, optimize feature fields that often focus on the surface regions, and cannot model the internal volume. VLM-based methods~\cite{liu2024physgen, Chen_2025_CVPR, lin2025phys4dgenphysicscompliant4dgeneration} allow single-image inference but are computationally heavy and reliant on external segmentation. Other works annotating 3D data~\cite{cao2025sophylearninggeneratesimulationready, cao2025physx-3d, zhao2024efficient_aka_automated, zhao2024automated, le2025pixie, liu2024physics3d} often target surface properties, or new shape generation, rather than the volumetric augmentation of existing assets. In contrast to these techniques, our method predicts \emph{volumetric} materials for existing shapes. 

\paragraph{Comparison to Feed-Forward Methods.} \label{sec:rw4} 
We build on top of VoMP~\cite{dagli2025vomppredictingvolumetricmechanical} by replacing its fixed-resolution grids with a sparse, adaptive voxel tree (\Cref{sec:save}). Similar to VoMP~\cite{dagli2025vomppredictingvolumetricmechanical}, Pixie~\cite{le2025pixie} also operates on a fixed resolution grid.
This allows us to generate coarse-to-fine predictions, scaling to significantly higher effective resolutions in complex regions.
While adaptive feature voxel structures are not new~\cite{takikawa2021neural}, we make the observation that they are especially well-suited
for representing volumetric material distributions, which often contain large homogeneous regions (e.g.\ metal bedframe).  
Our model is trained to output the least number of voxels, such that if queried with points within the geometry, it would yield correct mechanical properties.
Unlike general spatial data structures (e.g., Octrees, OpenVDB)~\cite{deng2025efficientautoregressiveshapegeneration, Museth2013OpenVDB} or adaptive discretization models with fixed multi-resolution input~\cite{choudhury2025acceleratingvisiontransformersadaptive}, our sparse representation (\S~\ref{sec:save}) is autoregressively refined specifically for material prediction rather than geometry.
Furthermore, we provide a parameterization of building the structure that is differentiable, allowing us to train with such a representation.
There exists some recent work~\cite{deng2025efficientautoregressiveshapegeneration} taht propose an autoregressive formulation that operates over an octree by serializing into a 1D sequence of discrete tokens, which the model generates autoregressively. In contrast, our method maintains the explicit 3D spatial structure throughout the generation process

%% file: text/method.tex
\section{\vacronym: Sparse Adaptive Voxels}
\label{sec:save}

The foundation of our technique is \vacronym, a sparse adaptive voxel representation that we use to encode both
the input 3D shape and the output spatially varying materials.
By efficiently representing geometry and materials in adaptive structures, we can allocate less compute to predict areas of piecewise constant
materials, common in everyday objects (the wooden surface, the metal frame), while recursively refining only the fine heterogeneous regions and boundaries,
enabling our model to predict material fields at $G^3 = 1024^3$ resolution, much higher than
$64^3$ for VoMP ~\cite{dagli2025vomppredictingvolumetricmechanical} and Pixie~\cite{le2025pixie}.

Unlike general-purpose spatial data structures such as octrees or OpenVDB~\cite{Museth2013OpenVDB}, which subdivide based on geometric criteria or explicit thresholds, {\vacronym} is autoregressively learned to optimize for \emph{material prediction}. 
While VoMP~\cite{dagli2025vomppredictingvolumetricmechanical} and TRELLIS~\cite{Xiang_2025_CVPR} use sparse voxel representations, they operate at a single fixed resolution, requiring all active voxels to be processed at the finest level, which is a prohibitive cost when scaling to high resolutions for volumetric material fields.
In contrast, {\vacronym} is both sparse \emph{and} adaptive: it stores voxels at multiple resolution levels simultaneously, allocating finer voxels only where material heterogeneity demands them, while representing homogeneous regions with single coarse voxels regardless of their spatial extent.

\subsection{Definitions}

Here we define {\vacronym} structure and basic properties that we utilize during training of our model.
For a bounded 3D domain $\Omega$ (e.g. $[-0.5, 0.5)^3$), 
 \vacronym\ represents a spatially-varying feature field $\mathcal{F}:\Omega\to\mathbb{R}^d$ using an adaptive voxel tree $\mathcal{T}$ whose leaf voxels form an axis-aligned partition of $\Omega$, but may reside at different resolution levels. 
 Each voxel stores its level $\ell\in\{0,\dots,L_{\max}\}$, where $0$ is the finest level, and an integer grid index $\bi\in\{0,\dots,G_\ell-1\}^3$ (exponentiation denotes the cartesian product), where $G_\ell:=G/2^\ell$ and $L_{\max}:=\log_2 G$. 
 To enable our Transformers to attend across resolutions, we also map each voxel (level $\ell$, index $\bi$) to its \emph{unified} coordinates:
\vspace{-0.5em}
\begin{equation}
\bu_{\ell,\bi} := 2^\ell \bi \in \{0,\dots,G-1\}^3,
\label{eq:unified_coords}
\end{equation}
where $2^\ell \bi$ denotes element-wise scalar multiplication $2^\ell \bi = (2^\ell i_x, 2^\ell i_y, 2^\ell i_z)$, mapping each voxel to the finest-resolution grid.
Given a voxel with index $\bi=(i_x,i_y,i_z)$, we further encode its relative position within the parent voxel using its discrete octant id,
$o(\bi)\in\ \{0,\dots,7\}$:
\vspace{-0.5em}
\begin{equation}
o(\bi) := (i_x \bmod 2) + 2(i_y \bmod 2) + 4(i_z \bmod 2).
\label{eq:octant_id}
\end{equation}

\vspace*{-1.0em}

 Each leaf voxel of level and index $(\ell, \bi)$ stores a constant feature vector $\be_{\ell,\bi}\in\mathbb{R}^{d}$, inducing a directly queryable piecewise-constant field for spatial queries $\mathbf{x} \in \Omega$, denoted: $\mathcal{T}(\mathbf{x}) := \be_{\ell',\bi'}$, where $(\ell',\bi')$ is the leaf voxel containing $\mathbf{x}$.
We implement querying and construction operations using coordinate-based sparse tensors for $\ell$ and $\bi$, and use a hierarchical hash lookup for fast batched queries. Refer to~\Cref{sec:savedetails} for details on our memory-efficient GPU implementation.

\subsection{Representing the Input Shape}\label{sec:sav:input}

Our goal is to predict material distributions for diverse 3D representations, and we adopt the
methodology of VoMP \cite{dagli2025vomppredictingvolumetricmechanical}, requiring that the 3D input shape be voxelized and
renderable from multiple viewpoints, with no other assumptions. First, we discretize the object's occupied volume, normalized to $\Omega \subset [-0.5, 0.5)^3$, into a base grid of resolution $G = 2^{10}$. Then, we aggregate multi-view DINOv3~\cite{simeoni2025dinov3} patch-token features over this volumetric voxelization,
with a few critical differences from prior art. First, to avoid excessive feature averaging that dilutes details, we adopt a depth-attenuated averaging of projected features
throughout the voxel structure, in contrast to uniform averaging from prior work ~\cite{Wang_2023_CVPR, Dutt_2024_CVPR, Xiang_2025_CVPR, dagli2025vomppredictingvolumetricmechanical}. 
\begin{wrapfigure}{r}{0.4\linewidth}
  \vspace{-0.1em}
  \centering
  \includegraphics{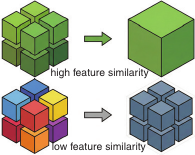}
  \label{fig:inset_dino}
  \vspace{-1.25em}
\end{wrapfigure}
Second, after aggregating features in a fixed resolution voxel structure, we progressively voxels with similar features (see inline figure) into an adaptive and more efficient {\vacronym} structure $\mathcal{T}^{\mathrm{in}}$,
the input to our model. 
See \Cref{sec:app_baking_dino},~\Cref{sec:app_dataset} for details.

\subsection{Representing the Material Distribution}\label{sec:sav:materials}

We denote the target (ground truth) volumetric material field by $\cM:\Omega\to\mathbb{R}^3$, where $\cM(\mathbf{x})=(E(\mathbf{x}),\nu(\mathbf{x}),\rho(\mathbf{x}))$. We represent $\cM$ as a {\vacronym} material tree $\mathcal{T}^{\cM}$ storing material vectors $\mathbf{m}_V\in\mathbb{R}^3$ as features of each voxel $V=(\ell, \bi)$. 
We construct $\mathcal{T}^{\cM}$ (detailed in~\Cref{alg:topdown-tree}) by: first, aggregating ground truth materials from finest to coarsest, computing the mean and range at each level; second, we traverse coarse-to-fine and subdivide a voxel $V$ only when the material variation within it exceeds a tolerance $\boldsymbol{\tau}$ (computed over its finest-level descendants); otherwise, we keep $V$ and store the descendant 
mean $\mathbf{m}_V := \frac{1}{|\mathrm{desc}(V)|}\sum_{U\in\mathrm{desc}(V)} \mathbf{m}_U$, where $\mathrm{desc}(V)$ 
\begin{wrapfigure}{r}{0.4\linewidth}
  \vspace{0.1em}
  \centering
  \includegraphics{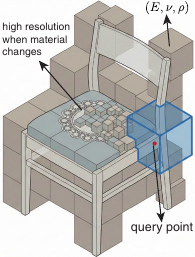}
  \label{fig:inset_material_boundary}
  \vspace{-3em}
\end{wrapfigure}
denotes the set of finest-level voxels contained in $V$. 
Consequently, partially specified trees remain well-defined \textit{i.e.} missing fine voxels in a region simply return the coarser averaged material for that region, which is a
direct mechanism for level-by-level supervision of structure and materials.

\section{Learning Adaptive Material Fields}
\label{sec:method}

\begin{figure*}
\centering
  \includegraphics[width=\textwidth]{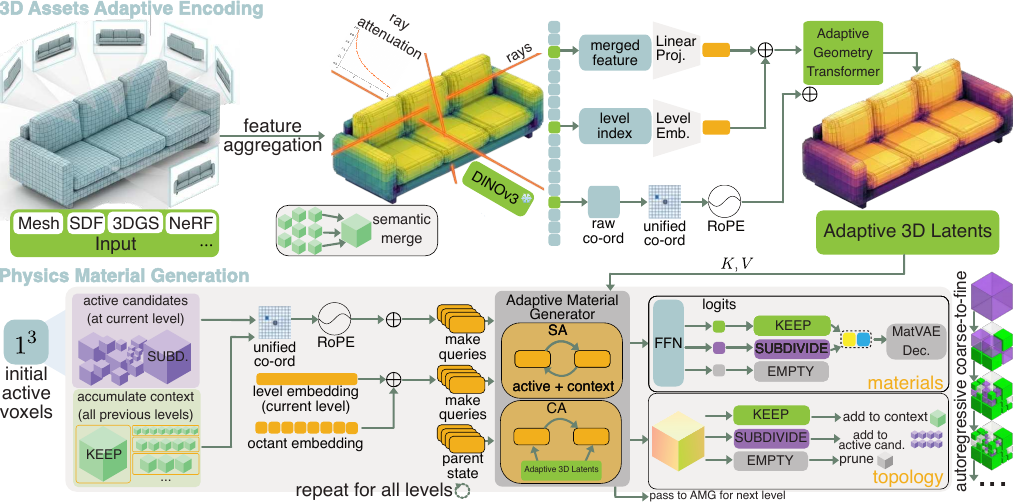}
  \label{fig:method}
  \caption{\textbf{Method Overview:} input shape is encoded as {\vacronym} (top left, \S\ref{sec:sav:input}), encoded (top right, \S\ref{sec:method_encoder}), and processed with our autoregressive Adaptive Material Generator (bottom, \S\ref{sec:method_decoder}), which is trained (\S\ref{sec:method_training}) to output material field as {\vacronym}. }
\end{figure*}

Our goal is to generate a physically accurate volumetric mechanical property field, given an input shape. We first encode 
the input shape, represented as a {\vacronym} $\mathcal{T}^{\mathrm{in}}$ (\S\ref{sec:sav:input}), with a trainable
Adaptive Geometry Transformer $\mathbf{E}$ (\Cref{sec:method_encoder}). The resulting latents condition
an Adaptive Material Generator $\mathbf{G}$ (\Cref{sec:method_decoder}), which outputs the final
material field, represented as {\vacronym} $\mathcal{T}^{\cM'}$, at an effective resolution of $G^3 = 1024^3$ without instantiating a dense grid.
The $\mathbf{G}$ model operates autoregressively, coarse-to-fine, by predicting both (i) adaptive structure (\emph{i.e.} what spatial regions need high resolutions) and (ii) per-voxel material latents.
Both models are trained jointly (\Cref{sec:method_training}), 
and supervised by ground-truth material trees, also represented as SAV $\mathcal{T}^{\cM}$ (\Cref{sec:sav:materials}).

\subsection{Adaptive Geometry Transformer}
\label{sec:method_encoder}
The input to our encoder $\textbf{E}$ is a {\vacronym} $\mathcal{T}^{\mathrm{in}}$, with aggregated DINOv3 features of the input shape (\Cref{sec:sav:input}) as its voxel features  $\be_{\ell,\bi}\in\mathbb{R}^{d_{\mathrm{in}}}$, $d_{\mathrm{in}}=1280$.
The mixed-level leaf voxels in $\mathcal{T}^{\mathrm{in}}$ form the input sparse token set of $\mathbf{E}$, where each voxel token at level $\ell$, index $\bi$ is embedded as:
\vspace{-0.5cm}
{%
\setlength{\abovedisplayskip}{0pt}%
\setlength{\belowdisplayskip}{0pt}%
\setlength{\abovedisplayshortskip}{0pt}%
\setlength{\belowdisplayshortskip}{0pt}%

\begin{equation}
\bh^0_{\ell,\bi} = W_{\mathrm{in}}\,\be_{\ell,\bi} + \be^{\mathrm{lvl}}_\ell,
\label{eq:agt_token_embed}
\end{equation}
}
where $W_{\mathrm{in}}$ is a linear projection and $\be^{\mathrm{lvl}}_\ell$ is a learned level embedding, $^0$ denotes the initial token embedding (layer 0) before transformer blocks. Additionally, for each token we inject positional information by applying RoPE~\cite{rope} on its unified coordinates $\bu_{\ell,\bi}$ (Eq.\ref{eq:unified_coords}) inside self-attention.  We then apply sparse 3D shifted-window self-attention~\cite{swin, swinv2, Xiang_2025_CVPR} in the unified coordinate system, following with a feed-forward network (FFN). Refer to ~\Cref{sec:app_training} for further details. This yields contextual latents $
\mathbf{E}(\mathcal{T}^{\mathrm{in}})$ that serve as conditioning for the Adaptive Material Generator at all generation levels. 

\subsection{Adaptive Material Generator}
\label{sec:method_decoder}

Our \emph{autoregressive} transformer model $\mathbf{G}$ generates $\mathcal{T}^{\cM'}$ coarse-to-fine, over resolution levels $\ell=L_{\max},\dots,0$. This allows natural test-time compute scaling,
yielding well-defined lower-resolution outputs for fewer iterations of $\mathbf{G}$. At each level $\ell$ we restrict all computation to an explicit sparse candidate set $\cC_\ell$ (the refinement frontier), rather than enumerating the full $G_\ell^3$ grid. For each candidate voxel $(\ell,\bi)\in\cC_\ell$, $\mathbf{G}$ outputs (i) structure logits over three actions, $\textsc{Empty}$, $\textsc{Keep}$, and $\textsc{Subdivide}$, and (ii) a latent material vector $\bz_{\ell,\bi}\in\mathbb{R}^2$ for non-empty voxels. The \textsc{Empty} action allows our model to explicitly predict empty space, unlike prior work~\cite{dagli2025vomppredictingvolumetricmechanical, lin2025phys4dgenphysicscompliant4dgeneration, le2025pixie, shuai2025pugszeroshotphysicalunderstanding, Feng_2024_CVPR}.

The transformer model $\mathbf{G}$ is shared across levels, where $\mathbf{G}(\cC_\ell)$ yields the
candidate set at the next level $\cC_{\ell-1}$ along with material latents $\bz_{\ell,\bi}$ for all $\textsc{Keep}$ voxels at  level $\ell$. In addition to its level $\ell$ and index $\bi$, each candidate $(\ell,\bi) \in \cC_\ell$ also contains its parent's hidden state $\bh_{\ell+1,\lfloor \bi/2 \rfloor}$ obtained from intermediate layers of prior application of $\mathbf{G}$ (see below). This context from previously subdivided voxels is needed because $\cC_\ell$ contains only the refinement frontier, so finer-level candidates would otherwise observe disconnected ``holes'' wherever coarser \textsc{Keep} voxels remain unsplit.
While the parent necessarily chose \textsc{Subdivide} for these candidates to exist, this context is essential for spatial coherence.
We initialize the coarsest candidate set as $\cC_{L_{\max}}=\{(\ell = L_{\max},\bi = (0,0,0))\}$,
with $\bzero$ parent state.

At each level, we construct an initial query embedding for each candidate $(\ell,\bi)\in\cC_\ell$ by combining its level, octant id (Eq.\ref{eq:octant_id}), and parent state $\bh_{\ell+1,\lfloor \bi/2 \rfloor}$:
\vspace{-0.5em}
\begin{equation}
\mathbf{q}_{\ell,\bi} = \be^{\mathrm{lvl}}_\ell + W_{\mathrm{oct}}\,\be^{\mathrm{oct}}_{o(\bi)} + W_{\mathrm{par}}\,\bh_{\ell+1,\lfloor \bi/2 \rfloor},
\label{eq:decoder_query}
\end{equation}

where $W_{\mathrm{oct}}$ and $W_{\mathrm{par}}$ are learned linear projections and $\be^{\mathrm{lvl}}_\ell$ is a learned level embedding (different from Eq.\ref{eq:agt_token_embed}). 
\begin{wrapfigure}{r}{0.6\linewidth}
  \vspace{0.1em}
  \centering
  \includegraphics[width=\linewidth]{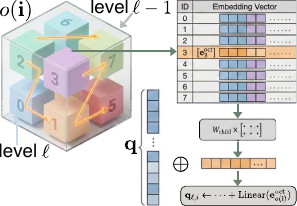}
  \label{fig:oct_id}
  \vspace{-1em}
\end{wrapfigure}
Each candidate also carries its unified coordinate (Eq.\ref{eq:unified_coords}) as its discrete sparse coordinate. We first apply cross-attention from candidates to the input latents $\mathbf{E}(\mathcal{T}^{\mathrm{in}})$ (\S\ref{sec:method_encoder}), then sparse windowed self-attention among candidates with RoPE~\cite{rope} on unified coordinates. See~\Cref{sec:results}, \Cref{sec:app_ablations} for ablation of these choices. From the resulting candidate hidden states $\bh_{\ell,\bi}$, FFN heads predict the structure action and the 2D material latent $\bz_{\ell, \bi}$ for every candidate.

\begin{wrapfigure}{r}{0.6\linewidth}
  \centering
  \includegraphics[width=\linewidth]{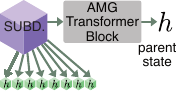}
  \label{fig:inset_parent}
  \vspace{-0.6em}
\end{wrapfigure}
If a candidate is predicted as \textsc{Subdivide} at level $\ell>0$, we include its eight children in the next candidate set $\cC_{\ell-1}$ together with the parent hidden state $\bh_{\ell,\bi}$.
Candidates predicted as \textsc{Empty} are discarded, and \textsc{Keep} voxels remain as leaves of the final material tree.
Although these voxels generate no children and thus do not directly appear in finer candidate sets $\cC_{\ell'}$ for $\ell'<\ell$, 
their information is propagated through the hidden state $\bh_{\ell,\bi}$ (Eq.\ref{eq:decoder_query}),
which encodes the broader spatial context,
including regions that were kept coarse.

At the finest level $\ell=0$, refinement terminates and all non-empty voxels are leaves.
See Alg.\ref{alg:amg_inference} for this coarse-to-fine decoding.

\paragraph{MatVAE.} To ensure generated properties are physically plausible, we incorporate the frozen MatVAE decoder from VoMP~\cite{dagli2025vomppredictingvolumetricmechanical}, predicting per-voxel latents $\bz_{\ell,\bi}\in\mathbb{R}^2$ in its latent space. The $\bz_{\ell,\bi}$ are mapped by MatVAE to \mattriplet, showing improved results (\Cref{sec:app_ablations}).

\subsection{Training}
\label{sec:method_training}

We train $\mathbf{E}$ and $\mathbf{G}$ end-to-end using teacher forcing, jointly supervising structure decisions and node materials. The ground truth $\mathcal{T}^{\cM}$ (\Cref{sec:sav:materials}) stores a material value at every node, where the \textsc{Subdivide} nodes store descendant means (\Cref{sec:save}). Teacher forcing deterministically fixes the breadth-first refinement schedule by replacing predicted subdivision decisions with ground-truth ones during training. Starting from $\cC_{L_{\max}}=\{(L_{\max},(0,0,0))\}$, we define:
\vspace{-0.5em}
\begin{equation}
\cC_{\ell-1} := \bigcup_{(\ell,\bi)\in\cC_\ell:\ s^{\star}_{\ell,\bi}=\textsc{Subdivide}} \mathrm{Children}(\ell,\bi),
\label{eq:teacher_forcing_candidates}
\end{equation}
for $\ell=L_{\max},\dots,1$, where $s^{\star}_{\ell,\bi}$ denotes the ground-truth structure label in $\mathcal{T}^{\cM}$. This construction expands all eight children of every subdividing voxel, ensuring that empty-space children are included as explicit negative candidates. We compute loss across all levels, weighted by $\omega_\ell := \gamma^\ell$, with $\gamma>1$, causing larger voxels to contribute more, and optimize the following overall objective: 
\vspace{-0.5em}
\begin{equation}
\cL = \lambda_{\mathrm{struct}}\,\cL_{\mathrm{struct}} + \lambda_{\mathrm{mat}}\,\cL_{\mathrm{mat}},
\label{eq:train_loss}
\end{equation}
where $\cL_{\mathrm{struct}}$ supervises structure actions and $\cL_{\mathrm{mat}}$ supervises materials, as detailed in \S~\ref{sec:app_training}.

\paragraph{Supervising Structure.}
Let $\cV^{\star}_\ell$ denote the grid indices of voxels present at level $\ell$ in $\mathcal{T}^{\cM}$. For a candidate $(\ell,\bi)\in\cC_\ell$, we define its ground truth structure decision as:
\begin{equation}
s^{\star}_{\ell,\bi} :=
\begin{cases}
\textsc{Empty}, & \bi\notin \cV^{\star}_\ell,\\
\textsc{Subdivide}, & \bi\in \cV^{\star}_\ell\ \text{and}\ \ell>0\ \text{and}\ \\
&\mathrm{Children}(\ell,\bi) \cap \cV^{\star}_{\ell-1} \neq \emptyset\\
\textsc{Keep}, & \text{otherwise.}
\end{cases}
\label{eq:gt_labels}
\end{equation}
Then, the structure loss is the candidate-count normalized, level-weighted negative log-likelihood over \emph{all} candidates across levels:
\vspace{-0.5em}
{\footnotesize
\begin{equation}
\cL_{\mathrm{struct}} =
\frac{1}{\sum_{\ell=0}^{L_{\max}} |\cC_\ell|}
\sum_{\ell=0}^{L_{\max}} \omega_\ell
\sum_{(\ell,\bi)\in\cC_\ell}
\Big(-\log p_{\ell,\bi}\big(s^{\star}_{\ell,\bi}\big)\Big),
\label{eq:struct_loss}
\end{equation}}
\noindent where the $\textsc{Empty}$, $\textsc{Subdivide}$, $\textsc{Keep}$ probabilities are computed as $p_{\ell,\bi}=\mathrm{softmax}(\mathbf{a}_{\ell,\bi})$ over the structure latents $\mathbf{a}_{\ell,\bi}$ output by $\mathbf{G}$, and $p_{\ell,\bi}(s^{\star}_{\ell,\bi})$ denotes selecting the probability of the ground truth label. $\omega^l$ is the weight for supervising predictions at level $l$.

\paragraph{Supervising Materials.}
For material supervision, we only penalize candidates whose ground-truth label is non-empty,
\begin{equation}
\cP_\ell := \{(\ell,\bi)\in\cC_\ell : s^{\star}_{\ell,\bi}\neq\textsc{Empty}\},
\end{equation}
since empty candidates have no well-defined material target. For each $(\ell,\bi)\in\cP_\ell$, we decode the predicted 2D latent through MatVAE to obtain a normalized triplet $\hat{\mathbf{m}}_{\ell,\bi}\in\mathbb{R}^3$ and compare it to the normalized target $\mathbf{m}^{\star}_{\ell,\bi}$ from $\mathcal{T}^{\cM}$:
{\setlength{\abovedisplayskip}{1pt}
\begin{equation}
\cL_{\mathrm{mat}} =
\frac{1}{\sum_{\ell=0}^{L_{\max}} |\cP_\ell|}
\sum_{\ell=0}^{L_{\max}} \omega_\ell
\sum_{(\ell,\bi)\in\cP_\ell}
\big\|\hat{\mathbf{m}}_{\ell,\bi}-\mathbf{m}^{\star}_{\ell,\bi}\big\|^2_{\boldsymbol{\Lambda}},
\label{eq:mat_loss}
\end{equation}}
where $\|\mathbf{v}\|^2_{\boldsymbol{\Lambda}}:=\mathbf{v}^\top\boldsymbol{\Lambda}\mathbf{v}$ with $\boldsymbol{\Lambda}=\mathrm{diag}(\lambda_E,\lambda_\nu,\lambda_\rho)$. See~\Cref{sec:app_training}
for more details. 

%% file: text/results.tex
\input{figures/results_mega_fig}
\section{Experiments and Results}\label{sec:results}
Quantitative and qualitative evaluation against prior art (\S\ref{sec:experiments_quantitative}) show significant improvements in accuracy and resolution. We further ablate 
our model size (\Cref{sec:experiments_model_test_time_compute_resolution_scaling}), showing that the gains are not solely due to increased model parameters. See \includegraphics[height=0.6\baselineskip]{assets/youtube-brands_svg-tex.pdf} video and~\Cref{sec:app_results} for additional results, ~\Cref{sec:app_ablations} for ablations, \Cref{sec:app:sim_results} for end-to-end evaluation running simulation of meshes and Gaussian splats
using our predicted materials. 

\subsection{Implementation Details}\label{sec:main_impl_details}
\input{figures/fig_qual}
\paragraph{Training and Parallelism.}
We train our models end-to-end in BF16 mixed precision, and develop and implementation that effectively performs Hybrid Sharded Data Parallelism (HSDP) \emph{i.e.} ZeRO-3~\cite{rajbhandari2020zeromemoryoptimizationstraining}/FSDP-2~\cite{zhao2023pytorchfsdpexperiencesscaling} + Distributed Data Parallelism (DDP) with sparse tensors and sparse operations based on top of Megatron-FSDP~\cite{shoeybi2020megatronlmtrainingmultibillionparameter}. All of our models are trained on a machine with 32$\times$A100-80 GB GPUs for 5 days. We present additional details in~\Cref{sec:app_training}.
\vspace{-0.4cm}
\paragraph{Datasets.}
Our dataset Geometry with Volumetric Trees (\gvt) builds on top of the GVM dataset~\cite{dagli2025vomppredictingvolumetricmechanical}, using the same assets slightly expanded by 61 objects. We follow 
the same VLM annotation using Qwen2.5-VL 72B~\cite{bai2025qwen25vltechnicalreport}, rendering in Omniverse~\cite{omniverse} and Blender~\cite{blender}. For image features we use DINOv3 ViT-H+/16~\cite{simeoni2025dinov3}.
Our material tree and feature tree creation were run on a machine with 128$\times$A100-80GB GPUs for two days each. See \Cref{sec:savedetails} for data details.
We train on 149.50M input tokens and 1.62B output tokens. We report results on two held-out evaluation sets. \textsc{GVT-Test} contains the same objects as the GVM test split in \cite{dagli2025vomppredictingvolumetricmechanical}. We additionally evaluate on \textsc{GVT-Hard}, a curated set of 50 objects by including an object if it contains at least one annotated mesh segment that is present under fine voxelization at $1024^3$ but is entirely skipped by a coarse $32^3$ voxelization. We share additional details in~\Cref{sec:app_dataset}.

\subsection{Quantitative and Qualitative Evaluation}
\label{sec:experiments_quantitative}

We evalute our performance against best recent methods VoMP~\cite{dagli2025vomppredictingvolumetricmechanical} and Pixie~\cite{le2025pixie}, as well as other baselines NeRF2Physics~\cite{zhai2024physicalpropertyunderstandinglanguageembedded}, PUGS~\cite{shuai2025pugszeroshotphysicalunderstanding} and Phys4DGen~\cite{lin2025phys4dgenphysicscompliant4dgeneration} in Tb.\ref{tab:voxel_object_metrics}. All metrics (\S\ref{app:metrics}), also used in ~\cite{dagli2025vomppredictingvolumetricmechanical}, show significant improvement over state of the art across
all three mechanical properties \mattriplet. Note that our method performs better even if evaluated at a lower effective resolution of $64^3$, matching many of the baselines. Because our method has significant resolution improvement, we further evaluate on a harder more detailed \textsc{GVT-Hard} dataset (\S\ref{sec:main_impl_details}) in Tb.\ref{tab:gvt_hard_object_avg}, showing an even larger gap in performance, with significant advantages offered by our {\ourmodel}.

From qualitative materials fields (Fig.\ref{fig:qual}, \S~\ref{sec:app_results}), we observe that NeRF2Physics~\cite{zhai2024physicalpropertyunderstandinglanguageembedded} and PUGS~\cite{shuai2025pugszeroshotphysicalunderstanding} have highly noisy estimates, Phys4DGen~\cite{lin2025phys4dgenphysicscompliant4dgeneration} mislabels segments and is unable to segment out complex objects, and Pixie~\cite{le2025pixie} consistently predicts softer materials and fails on complex objects. VoMP~\cite{dagli2025vomppredictingvolumetricmechanical} can accurately predict volumetric materials, but for high-resolution objects, VoMP completely misses a part of the object due to its low resolution.
We demonstrate high-fidelity end-to-end simulations on complex objects in~\Cref{fig:teaser,fig:teaser2,sec:app:sim_results} (\video{00:00}).

Further, we show on-par or better material validity (whether material is within physically measured material values) against VoMP (Tb.\ref{tab:validity}) on the GVM dataset from VoMP~\cite{dagli2025vomppredictingvolumetricmechanical}, and improved mass estimation on the ABO benchmark in Tb.\ref{tab:mass}.

\subsection{Model, Test-Time Compute, and Resolution Scaling}
\label{sec:experiments_model_test_time_compute_resolution_scaling}
\begin{figure}[tb]
    \centering
    \includegraphics[width=\linewidth]{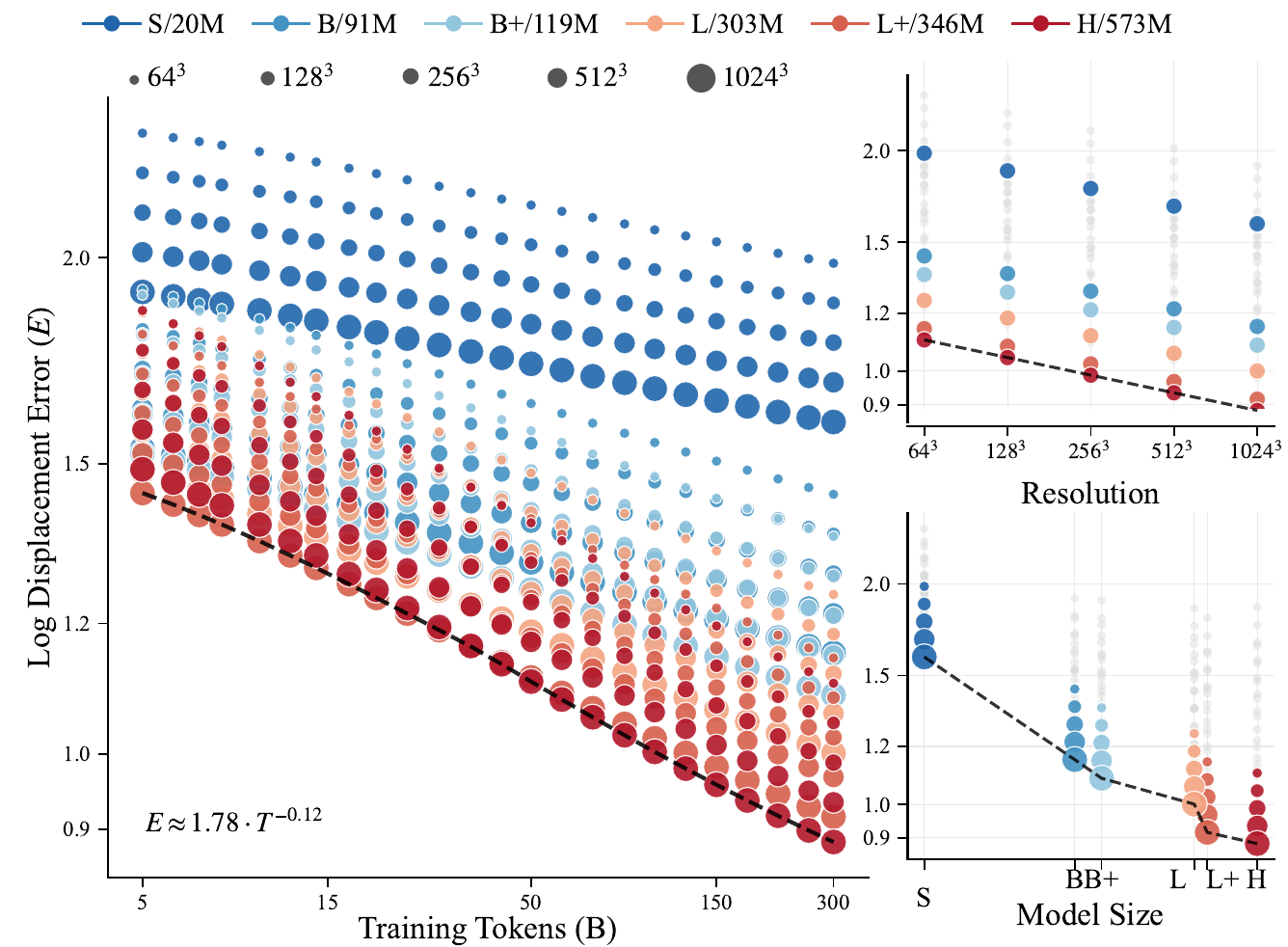}
    \caption{\textbf{Scaling Model, Training, and Test-time Compute.}
    \emph{Left:} Our method shown across three independent axes: training tokens, test-time compute (output resolution), and model size. We show displacement errors for Young's modulus ($E$) as a function of training tokens. Larger models achieve lower error at a fixed training budget and allocate additional test-time compute (higher resolution) consistently improves accuracy.
    \emph{Right:} Final training budget and show the error trend as a function of resolution (top) and model size (bottom). A detailed version of this plot is shown in~\Cref{fig:full_scaling}.}
    \label{fig:scaling}
\end{figure}

We scale the model to 0.6B parameters and train multiple sizes denoted \textsc{Small} (\textsc{S}), \textsc{Base} (\textsc{B}), \textsc{Base+} (\textsc{B+}), \textsc{Large} (\textsc{L}), \textsc{Large+} (\textsc{L+}), and \textsc{Huge} (\textsc{H}) in~\Cref{fig:scaling}. Apart from these model sizes, we further scale the model in~\Cref{sec:app_results}.
We find that our \textsc{B+} model, has similar parameters as VoMP but still outperforms VoMP (\Cref{tab:scale_object_avg}).

\subsection{Structure Efficiency}
\label{sec:experiments_structure_efficiency}
A key advantage of the adaptive \vacronym\ representation is its ability to reduce the number of stored voxels compared to a fixed-grid baseline while preserving material fidelity. We quantify this compactness by comparing the number of leaf nodes in ground-truth material trees to the voxel count that would result from a dense $64^3$ voxelization, and separately measure how faithfully our generated trees recover the ground-truth structure. Throughout, we report leaf-node counts restricted to levels $\ell\le 6$, so the finest cells match $64^3$ resolution.

\Cref{tab:structure_compactness} reports the compactness of the ground-truth \vacronym\ representation. On \textsc{GVT-Test}, ground-truth material trees require only $7.24\%$ of the occupied voxels of a dense $64^3$ voxelization (VoMP) when counting leaves at level $\ell\le 6$ (i.e., $64^3$ or coarser)
; on the full dataset the ratio is $10.54\%$. Per-object statistics reveal substantial variation (median $4.42\%$ on test, $1.97\%$ on full), indicating that many objects are highly compressible while a long tail of complex objects approaches the dense baseline.

\Cref{tab:structure_recovery} compares the generated material trees produced by our model to the ground-truth trees and VoMP~\cite{dagli2025vomppredictingvolumetricmechanical}. On \textsc{GVT-Test}, ground-truth trees contain $79.28\%$ as many leaves as the generated trees, i.e., generated trees use $26.14\%$ more leaves than the oracle structure at levels $\ell\le 6$. Combining this with the ground-truth compactness ratio of $7.24\%$ (\Cref{tab:structure_compactness}) implies that generated trees require $9.14\%$ of the occupied voxels of a dense $64^3$ voxelization, preserving the compactness advantage after learning.

\section{Discussion}

As a data-driven method, the accuracy and generalization of our model will improve with more available training data. The high-resolution prediction of mechanical properties by our method enables the approximation of anisotropic materials via multiscale modeling, but still cannot handle truly directional materials whose Young's moduli are spatial-varying tensor fields. Future work can also extend our predictions beyond linear elasticity to include yield strength, shear modulus, and thermal expansion. Our method predicts `true' material properties that work well for accurate simulators, but it would be useful to adapt to specific, often non-physical parameter scales for approximate, real-time simulators. Lastly, as our approach is designed for static 3D assets, we currently cannot incorporate dynamic physical cues available in video observations. These limitations point to interesting future directions.

\section{Conclusion}
{\ourmodel} predicts mechanical property fields for 3D assets at $16^3\times$ higher resolution than prior works while maintaining memory efficiency. 
Using surface-level visual appearance to transform 3D assets into volumetric, physically interactable entities, we obviate the need for manual parameter tuning, which is currently the bottleneck to realistic simulation at scale. We hope this work will become a foundational block of physical AI, opening the door to scalable pipelines for generating simulation-ready assets, training robotic agents with physics in the loop, to produce realistic dynamic 3D worlds and to produce realistic interactive worlds.

%% file: figures/results_mega_fig.tex
\begingroup

\setlength{\abovedisplayskip}{0pt}
\setlength{\belowdisplayskip}{0.5pt}
\setlength{\abovedisplayshortskip}{0pt}
  \setlength{\belowdisplayshortskip}{0.5pt}

\begin{figure*}[htbp!] %
    \centering
    
    \captionof{table}{\textbf{Mechanical Property Estimates} of our method significantly outperform the baselines on all metrics and marginally outperforms the baseline even with low test-time compute ($64^3$). Per-voxel error rate is first computed per object, then averaged across all objects in the test set to avoid weighing
   some objects more. Global voxel-level normalization yields similar results (\Cref{tab:voxel_avg_metrics}).}
   \label{tab:voxel_object_metrics}
    \resizebox{\textwidth}{!}{
    \input{figures/content/cont_tb_mech_properties_object}
   }
   \vspace{0.5em}

   \captionof{table}{\textbf{\textsc{GVT-Hard} at $1024^3$ (object-averaged).} Object-averaged errors on the challenging \textsc{GVT-Hard} subset.}
  \label{tab:gvt_hard_object_avg}
  \resizebox{\textwidth}{!}{
  \input{figures/content/cont_tb_hard_object}
  }

  \vspace{0.5em}

    \begin{minipage}[t]{0.47\linewidth}
        \centering
        \captionof{table}{\textbf{Mass Estimation.} Errors for estimating mass of objects on the ABO-500~\cite{Collins_2022_CVPR} dataset, the only existing benchmark, approximating the accuracy of our $\rho$ estimates.}
        \label{tab:mass}
            \resizebox{0.98\linewidth}{!}{
            \input{figures/content/cont_tb_mass}
        }
    \end{minipage}
    \hfill %
    \begin{minipage}[t]{0.47\linewidth}
        \centering
        \captionof{table}{\textbf{Material Validity.} We report mean values and relative errors (in \%) with the closest physically measured material range in Material Triplet Dataset~\cite{dagli2025vomppredictingvolumetricmechanical}.}
        \label{tab:validity}
        \resizebox{0.98\linewidth}{!}{
        \input{figures/content/cont_tb_validity}
        }
    \end{minipage}

    \vspace{0.5em}

     \begin{minipage}[t]{0.47\linewidth}
        \centering
        \captionof{table}{\textbf{Ground-truth \vacronym\ compactness.} Ratio of leaf nodes at $64^3$ resolution or coarser in the ground-truth material tree to the number of occupied voxels under dense $64^3$ voxelization.}
        \label{tab:structure_compactness}
         \resizebox{0.98\linewidth}{!}{
        \input{figures/content/tb_cont_compactness}
        }
    \end{minipage}
    \hfill %
    \begin{minipage}[t]{0.47\linewidth}
        \centering
        \captionof{table}{\textbf{Generated vs.\ ground-truth structure.} Ratios of aggregated counts on \textsc{GVT-Test}: leaf nodes at levels $\ell\le 6$ for trees, and occupied voxels for VoMP's dense $64^3$ voxelization.}
  \label{tab:structure_recovery}
        \resizebox{0.98\linewidth}{!}{
        \input{figures/content/cont_tb_structure}
        }
    \end{minipage}
    
\end{figure*}

\endgroup

%% file: figures/content/cont_tb_mech_properties_object.tex
\begin{tabular}{lrrrrrr}
    \toprule
    \rowcolor{nvidiagreen!15}Method & \multicolumn{2}{c}{Young's Modulus Pa ($E$)} & \multicolumn{2}{c}{Poisson's Ratio ($\nu$)} & \multicolumn{2}{c}{Density $\frac{kg}{m^3}$ ($\rho$)} \\
    \cmidrule(r){2-3} \cmidrule(r){4-5} \cmidrule(r){6-7}
    \rowcolor{nvidiagreen!15}& ALDE ($\downarrow$) & ALRE ($\downarrow$) & ADE ($\downarrow$) & ARE ($\downarrow$) & ADE ($\downarrow$) & ARE ($\downarrow$) \\
    \midrule
    \rowcolor{gray!15} \multicolumn{7}{l}{Evaluation at $64^3$ resolution.} \\
    NeRF2Physics~\cite{zhai2024physicalpropertyunderstandinglanguageembedded} & 2.8000 {\scriptsize{($\pm$1.05)}} & 0.1346 {\scriptsize{($\pm$0.05)}} & - & - & 1432.0343 {\scriptsize{($\pm$964.88)}} & 1.0365 {\scriptsize{($\pm$0.63)}} \\
    PUGS~\cite{shuai2025pugszeroshotphysicalunderstanding} & 3.3942 {\scriptsize{($\pm$1.72)}} & 0.1688 {\scriptsize{($\pm$0.10)}} & - & - & 3568.2150 {\scriptsize{($\pm$2839.13)}} & 3.2429 {\scriptsize{($\pm$3.56)}} \\
    Phys4DGen$^\star$~\cite{lin2025phys4dgenphysicscompliant4dgeneration} & 4.8967 {\scriptsize{($\pm$3.17)}} & 0.2227 {\scriptsize{($\pm$0.14)}} & 0.0407 {\scriptsize{($\pm$0.04)}} & 0.1467 {\scriptsize{($\pm$0.18)}} & 1865.5673 {\scriptsize{($\pm$2176.90)}} & 1.4394 {\scriptsize{($\pm$2.35)}} \\
    Pixie~\cite{le2025pixie} & 0.3986 {\scriptsize{($\pm$0.30)}} & 0.0446 {\scriptsize{($\pm$0.04)}} & 0.0259 {\scriptsize{($\pm$0.01)}} & 0.0869 {\scriptsize{($\pm$0.03)}} & \underline{141.7812} {\scriptsize{($\pm$163.40)}} & \underline{0.0917} {\scriptsize{($\pm$0.07)}} \\
    VoMP~\cite{dagli2025vomppredictingvolumetricmechanical} & \underline{0.3793} {\scriptsize{($\pm$0.29)}} & \underline{0.0409} {\scriptsize{($\pm$0.04)}} & \underline{0.0241} {\scriptsize{($\pm$0.01)}} & \underline{0.0818} {\scriptsize{($\pm$0.03)}} & 142.6949 {\scriptsize{($\pm$166.90)}} & 0.0921 {\scriptsize{($\pm$0.07)}} \\
    \midrule
    Ours-H (0.6B) & \textbf{0.3278} {\textbf{\scriptsize\textcolor{gray}{($\pm$0.26)}}} & \textbf{0.0340} {\textbf{\scriptsize\textcolor{gray}{($\pm$0.03)}}} & \textbf{0.0205} {\textbf{\scriptsize\textcolor{gray}{($\pm$0.01)}}} & \textbf{0.0680} {\textbf{\scriptsize\textcolor{gray}{($\pm$0.03)}}} & \textbf{127.3125} {\textbf{\scriptsize\textcolor{gray}{($\pm$150.83)}}} & \textbf{0.0842} {\textbf{\scriptsize\textcolor{gray}{($\pm$0.07)}}} \\
    \midrule
    \rowcolor{gray!15} \multicolumn{7}{l}{Evaluation at $1024^3$ resolution.} \\
    NeRF2Physics~\cite{zhai2024physicalpropertyunderstandinglanguageembedded} & 4.1273 {\scriptsize{($\pm$1.71)}} & 0.2064 {\scriptsize{($\pm$0.09)}} & - & - & 2578.3261 {\scriptsize{($\pm$1621.75)}} & 1.8734 {\scriptsize{($\pm$1.06)}} \\
    PUGS~\cite{shuai2025pugszeroshotphysicalunderstanding} & 5.6871 {\scriptsize{($\pm$2.53)}} & 0.2982 {\scriptsize{($\pm$0.13)}} & - & - & 6345.9184 {\scriptsize{($\pm$4012.23)}} & 5.3621 {\scriptsize{($\pm$4.94)}} \\
    Phys4DGen$^\star$~\cite{lin2025phys4dgenphysicscompliant4dgeneration} & 6.9145 {\scriptsize{($\pm$4.02)}} & 0.3576 {\scriptsize{($\pm$0.21)}} & 0.0732 {\scriptsize{($\pm$0.06)}} & 0.2624 {\scriptsize{($\pm$0.32)}} & 3187.4207 {\scriptsize{($\pm$3098.55)}} & 2.9127 {\scriptsize{($\pm$3.72)}} \\
    Pixie~\cite{le2025pixie} & 1.2264 {\scriptsize{($\pm$0.52)}} & 0.1372 {\scriptsize{($\pm$0.10)}} & 0.0413 {\scriptsize{($\pm$0.02)}} & 0.1396 {\scriptsize{($\pm$0.06)}} & 248.6735 {\scriptsize{($\pm$252.11)}} & 0.1568 {\scriptsize{($\pm$0.14)}} \\
    VoMP~\cite{dagli2025vomppredictingvolumetricmechanical} & 
    \underline{1.1371} {\scriptsize{($\pm$0.36)}} & 
    \underline{0.1226} {\scriptsize{($\pm$0.08)}} & 
    \underline{0.0289} {\scriptsize{($\pm$0.01)}} & 
    \underline{0.0965} {\scriptsize{($\pm$0.04)}} & 
    \underline{191.6284} {\scriptsize{($\pm$212.77)}} & 
    \underline{0.1216} {\scriptsize{($\pm$0.09)}} \\
    \midrule
    Ours-H (0.6B) & \textbf{0.8841} {\textbf{\scriptsize\textcolor{gray}{($\pm$0.27)}}} & \textbf{0.0917} {\textbf{\scriptsize\textcolor{gray}{($\pm$0.07)}}} & \textbf{0.0215} {\textbf{\scriptsize\textcolor{gray}{($\pm$0.01)}}} & \textbf{0.0714} {\textbf{\scriptsize\textcolor{gray}{($\pm$0.03)}}} & \textbf{158.4602} {\textbf{\scriptsize\textcolor{gray}{($\pm$176.28)}}} & \textbf{0.1048} {\textbf{\scriptsize\textcolor{gray}{($\pm$0.08)}}} \\
    \bottomrule
  \end{tabular}

%% file: figures/content/cont_tb_hard_object.tex
\begin{tabular}{lrrrrrr}
    \toprule
    \rowcolor{nvidiagreen!15}Method & \multicolumn{2}{c}{Young's Modulus Pa ($E$)} & \multicolumn{2}{c}{Poisson's Ratio ($\nu$)} & \multicolumn{2}{c}{Density $\frac{kg}{m^3}$ ($\rho$)} \\
    \cmidrule(r){2-3} \cmidrule(r){4-5} \cmidrule(r){6-7}
    \rowcolor{nvidiagreen!15}& ALDE ($\downarrow$) & ALRE ($\downarrow$) & ADE ($\downarrow$) & ARE ($\downarrow$) & ADE ($\downarrow$) & ARE ($\downarrow$) \\
    \midrule
    NeRF2Physics~\cite{zhai2024physicalpropertyunderstandinglanguageembedded} & 6.1600 {\scriptsize{($\pm$2.30)}} & 0.2960 {\scriptsize{($\pm$0.12)}} & - & - & 3718.4285 {\scriptsize{($\pm$2376.12)}} & 2.7319 {\scriptsize{($\pm$1.53)}} \\
    PUGS~\cite{shuai2025pugszeroshotphysicalunderstanding} & 9.0500 {\scriptsize{($\pm$3.80)}} & 0.4500 {\scriptsize{($\pm$0.18)}} & - & - & 8157.9374 {\scriptsize{($\pm$5482.55)}} & 7.4836 {\scriptsize{($\pm$5.38)}} \\
    Phys4DGen$^\star$~\cite{lin2025phys4dgenphysicscompliant4dgeneration} & 12.3100 {\scriptsize{($\pm$5.60)}} & 0.5600 {\scriptsize{($\pm$0.26)}} & 0.1082 {\scriptsize{($\pm$0.09)}} & 0.3900 {\scriptsize{($\pm$0.48)}} & 5179.6421 {\scriptsize{($\pm$4291.83)}} & 3.9738 {\scriptsize{($\pm$4.43)}} \\
    Pixie~\cite{le2025pixie} & 1.8950 {\scriptsize{($\pm$1.10)}} & 0.2120 {\scriptsize{($\pm$0.11)}} & 0.0492 {\scriptsize{($\pm$0.03)}} & 0.1650 {\scriptsize{($\pm$0.09)}} & 393.6274 {\scriptsize{($\pm$359.28)}} & 0.2586 {\scriptsize{($\pm$0.24)}} \\
    VoMP~\cite{dagli2025vomppredictingvolumetricmechanical} & \underline{1.6680} {\scriptsize{($\pm$0.98)}} & \underline{0.1800} {\scriptsize{($\pm$0.10)}} & \underline{0.0368} {\scriptsize{($\pm$0.02)}} & \underline{0.1250} {\scriptsize{($\pm$0.07)}} & \underline{348.1956} {\scriptsize{($\pm$317.42)}} & \underline{0.2239} {\scriptsize{($\pm$0.20)}} \\
    \midrule
    Ours-H (0.6B) & \textbf{1.2440} {\textbf{\scriptsize\textcolor{gray}{($\pm$0.44)}}} & \textbf{0.1290} {\textbf{\scriptsize\textcolor{gray}{($\pm$0.09)}}} & \textbf{0.0286} {\textbf{\scriptsize\textcolor{gray}{($\pm$0.02)}}} & \textbf{0.0950} {\textbf{\scriptsize\textcolor{gray}{($\pm$0.06)}}} & \textbf{241.8735} {\textbf{\scriptsize\textcolor{gray}{($\pm$224.18)}}} & \textbf{0.1573} {\textbf{\scriptsize\textcolor{gray}{($\pm$0.14)}}} \\
    \bottomrule
  \end{tabular}

%% file: figures/content/cont_tb_mass.tex
\begin{tabular}{lrrrr}
    \toprule
    \rowcolor{nvidiagreen!15}Method & ALDE ($\downarrow$) & ADE ($\downarrow$) & ARE ($\downarrow$) & MnRE \textbf{($\uparrow$)}\\
    \midrule
    NeRF2Physics~\cite{zhai2024physicalpropertyunderstandinglanguageembedded} & 0.736 & 12.725 & 1.040 & 0.564\\
    PUGS~\cite{shuai2025pugszeroshotphysicalunderstanding} & 0.661 & 9.461 & \underline{0.767} & 0.576\\
    Phys4DGen$^\star$~\cite{lin2025phys4dgenphysicscompliant4dgeneration} & 0.664 & 9.961 & 0.825 & 0.566\\
    Pixie~\cite{le2025pixie} & 0.654 & \underline{8.231} & 0.875 & \underline{0.584}\\
    VoMP~\cite{dagli2025vomppredictingvolumetricmechanical} & \underline{0.631} & 8.433 & 0.887 & 0.576\\
    \midrule
    Ours-\textsc{H} (0.6B) & \textbf{0.457} & \textbf{6.924} & \textbf{0.512} & \textbf{0.667}\\
    \bottomrule
\end{tabular}

%% file: figures/content/cont_tb_validity.tex
\begin{tabular}{lrrrr}
    \toprule
    \rowcolor{nvidiagreen!15}Method & $\log(E) (\downarrow)$ & $\nu (\downarrow)$ & $\rho (\downarrow)$ \\
    \midrule
    NeRF2Physics~\cite{zhai2024physicalpropertyunderstandinglanguageembedded}
    & 1.62 {\textbf{\scriptsize\textcolor{gray}{($\pm$4.96)}}} 
    & -- 
    & 19.75 {\textbf{\scriptsize\textcolor{gray}{($\pm$46.60)}}} \\
    PUGS~\cite{shuai2025pugszeroshotphysicalunderstanding}
    & 1.87 {\textbf{\scriptsize\textcolor{gray}{($\pm$4.50)}}} 
    & -- 
    & 13.24 {\textbf{\scriptsize\textcolor{gray}{($\pm$12.63)}}} \\
    Phys4DGen$^\star$~\cite{lin2025phys4dgenphysicscompliant4dgeneration}  
    & 1.77 {\textbf{\scriptsize\textcolor{gray}{($\pm$8.53)}}} 
    & \underline{0.85} {\textbf{\scriptsize\textcolor{gray}{($\pm$3.01)}}} 
    & 39.49 {\textbf{\scriptsize\textcolor{gray}{($\pm$35.47)}}} \\
    Pixie~\cite{le2025pixie}       
    & 11.90 {\textbf{\scriptsize\textcolor{gray}{($\pm$17.41)}}} 
    & 3.46 {\textbf{\scriptsize\textcolor{gray}{($\pm$4.42)}}} 
    & 46.58 {\textbf{\scriptsize\textcolor{gray}{($\pm$36.35)}}} \\
    VoMP~\cite{dagli2025vomppredictingvolumetricmechanical}
    & \underline{0.29} {\textbf{\scriptsize\textcolor{gray}{($\pm$1.23)}}} 
    & \textbf{0.00} {\textbf{\scriptsize\textcolor{gray}{($\pm$0.00)}}} 
    & \textbf{11.75} {\textbf{\scriptsize\textcolor{gray}{($\pm$4.02)}}} \\
    \midrule
    Ours-\textsc{H} (0.6B) & \textbf{0.28} {\textbf{\scriptsize\textcolor{gray}{($\pm$1.27)}}} & \textbf{0.00} {\textbf{\scriptsize\textcolor{gray}{($\pm$0.00)}}} & \underline{11.78} {\textbf{\scriptsize\textcolor{gray}{($\pm$3.92)}}}\\
    \bottomrule
  \end{tabular}

%% file: figures/content/tb_cont_compactness.tex
\begin{tabular}{lrr}
    \toprule
    \rowcolor{nvidiagreen!15}Metric & \textsc{GVT-Test} (166) & Full (1{,}719) \\
    \midrule
    GT leaves ($\ell\le 6$) & 622{,}022 & 6{,}282{,}369 \\
    $64^3$ voxels & 8{,}586{,}819 & 59{,}621{,}729 \\
    \midrule
    Overall ratio & 7.24\% & 10.54\% \\
    Per-object mean & 16.16\% \stdfmt{24.57} & 14.67\% \stdfmt{30.01} \\
    Median & 4.42\% & 1.97\% \\
    \bottomrule
  \end{tabular}

%% file: figures/content/cont_tb_structure.tex
\begin{tabular}{lr}
    \toprule
    \rowcolor{nvidiagreen!15}Comparison & Ratio \\
    \midrule
    GT leaves / Generated leaves & 79.28\% \\
    GT leaves / VoMP voxels ($64^3$) & 7.24\% \\
    Generated leaves / VoMP voxels ($64^3$) & 9.14\% \\
    \bottomrule
  \end{tabular}

%% file: figures/fig_qual.tex
\begin{figure*}[ht!]
	\centering
    \includegraphics[width=\textwidth]{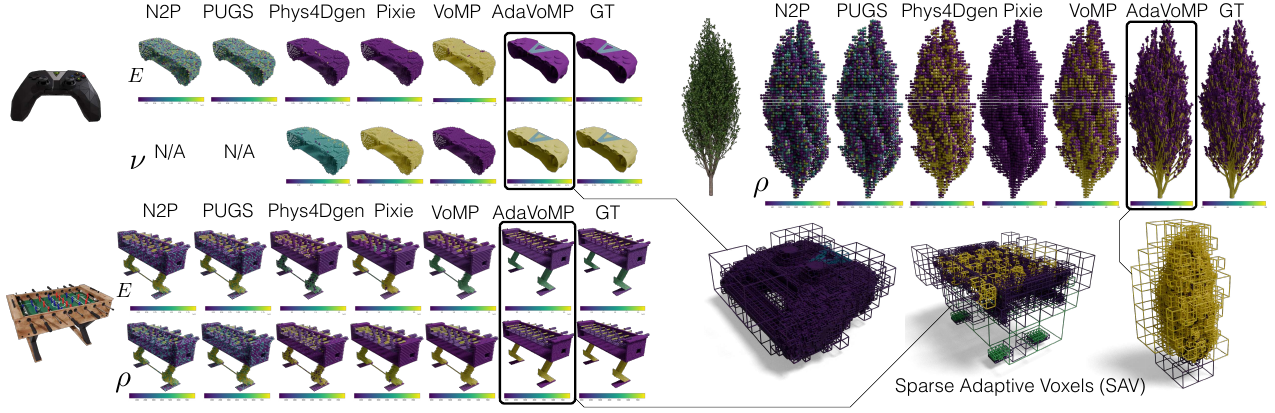}
	 \caption{\textbf{Qualitative Results:} comparing {\acronym} material predictions with prior works. These results are generated with our \textsc{H} model with the largest test-time compute. Note: Colorbar scales are different for each algorithm. \video{03:18}}\label{fig:qual}
\end{figure*}

%% file: text/appendix/main.tex
\input{text/appendix/results}
\input{text/appendix/sav}
\input{text/appendix/ablations}
\input{text/appendix/metrics}
\input{text/appendix/dataset}
\input{text/appendix/implementation}

%% file: text/appendix/results.tex
\begin{figure*}[tb]
    \centering
    \includegraphics{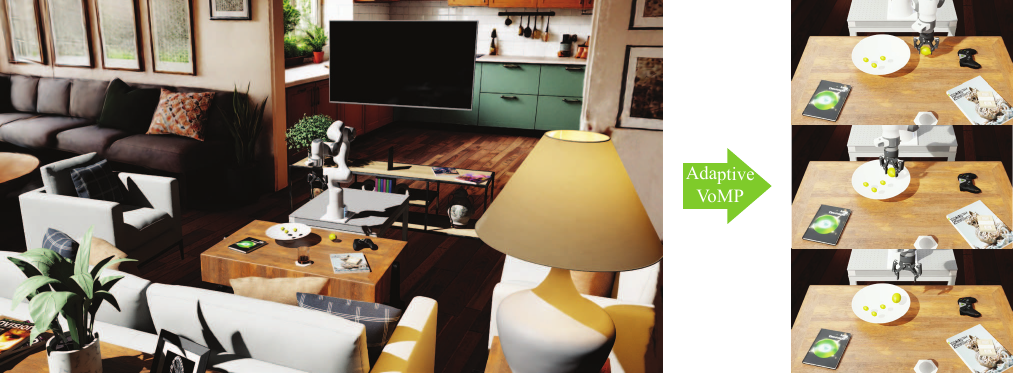}
    \caption{\textbf{Simulating Gaussian Splats and Meshes at Scale.} We show an elastodynamic simulation of a Gaussian Splat and a mesh scene with objects given mechanical properties generated by \acronym. We find that objects like the sofa and the pillows on the sofa are stable under gravity. Near the center of the scene, we simulate a robot~\cite{franka_description} which interacts with the fruits on the table producing realistic interactions. We integrate \acronym\ into RoboLab~\cite{yang2026robolabhighfidelitysimulationbenchmark} to generate this demo. The Gaussian Splat is generated with Marble~\cite{marble} and the robot is controlled by the $\pi_{0.5}$~\cite{pmlr-v305-black25a} Vision-Language-Action model (\video{04:27}).}
    \label{fig:teaser2}
\end{figure*}

\section{Additional Results}
\label{sec:app_results}

\subsection{Scaling Experiments}
\label{app:scaling}

We show full object-averaged scaling results in Tb.\ref{tab:scale_object_avg}. We complement~\Cref{fig:scaling} with a figure scaling the model, training, and test-time compute in~\Cref{fig:full_scaling}. We demonstrate experiments on how memory scales in~\Cref{fig:memory_scaling} with our framework as we scale model sizes and resolution. We show the dimensionality of generated {\vacronym} as we scale the resolution in~\Cref{fig:fractal_dimension}. We show the computational cost for scaling model parameters and resolution in~\Cref{fig:parameter_scaling}.

\input{figures/tb_scaling_object}

\input{figures/appendix_scaling}

\subsection{End-to-end Examples with Simulation} \label{sec:app:sim_results}

We qualitatively evaluate \ourmodel\ by using it to annotate volumetric mechanical fields for several meshes and 3D Gaussian Splats, and running physics simulation with these spatially varying \mattriplet\ values, resulting in realistic simulations without any hand-tweaks.

\input{figures/fig_sim_results}

\subsection{Additional Mechanical Property Prediction Results}

For completeness, we show voxel (not object) averaged results over regular and hard datasets in Tb.\ref{tab:voxel_avg_metrics} and Tb.\ref{tab:gvt_hard_voxel_avg}, 
complementing main paper tabulations in Tb.\ref{tab:voxel_object_metrics}, Tb.\ref{tab:gvt_hard_object_avg}. These are computed by averaging metrics over all voxels in the dataset.

\input{figures/tb_mech_properties_voxel}
\input{figures/tb_hard_voxels}

\subsection{Additional Mechanical Property Fields}

\begin{figure*}[tb]
    \centering
    \setlength{\tabcolsep}{0pt}
    \begin{tabular}{p{0.125\textwidth}p{0.125\textwidth}p{0.125\textwidth}p{0.125\textwidth}|p{0.125\textwidth}p{0.125\textwidth}p{0.125\textwidth}p{0.125\textwidth}}
    \rowcolor{nvidiagreen!15}\multicolumn{4}{c|}{\ourmodel} &
    \multicolumn{4}{c}{VoMP} \\
    \cline{1-4}\cline{5-8}
    \centering Object & \centering Young's Modulus ($E$, Pa) & \centering Poisson's Ratio ($\nu$) & \centering Density ($\rho, \frac{kg}{m^3}$) & \centering Object & \centering Young's Modulus ($E$, Pa) & \centering Poisson's Ratio ($\nu$) & \centering Density ($\rho, \frac{kg}{m^3}$) \\
    \end{tabular}
    \begin{tabular}{c|c}
\includegraphics[width=0.5\textwidth]{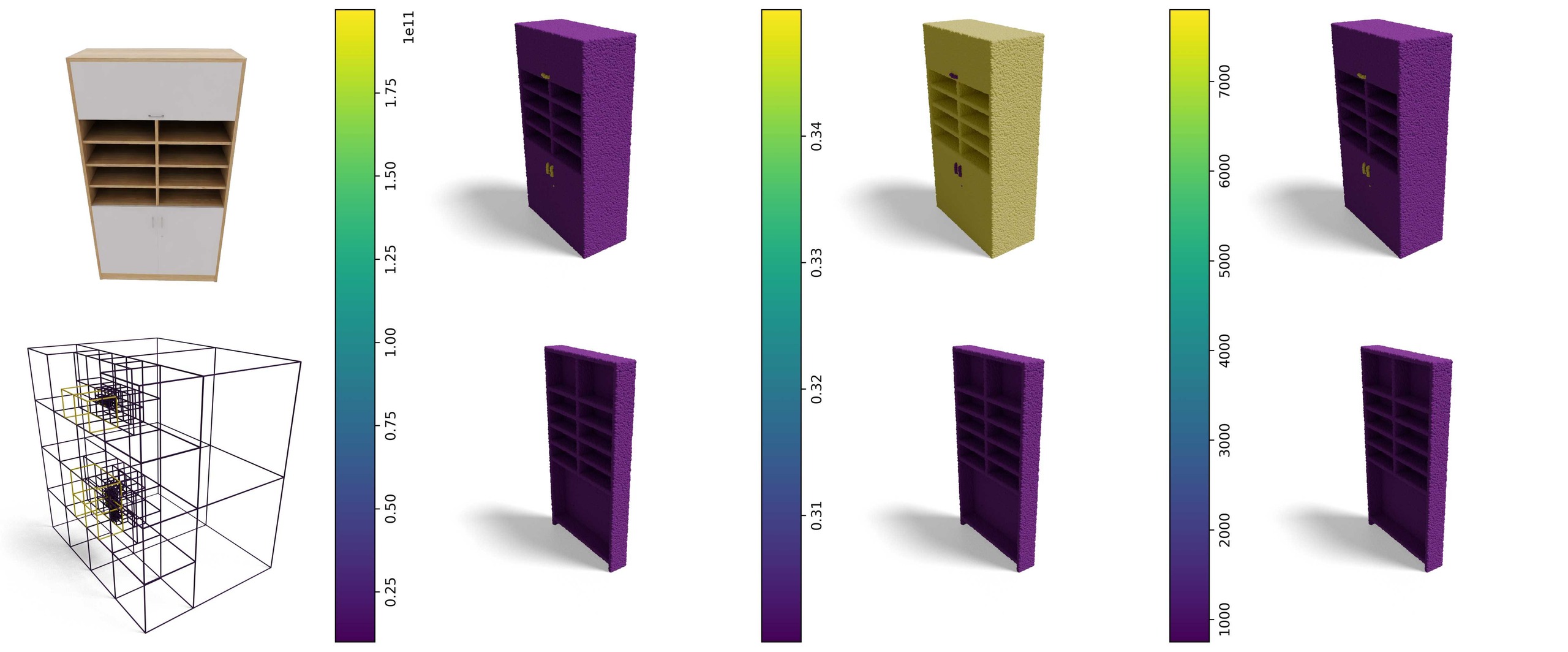} & \includegraphics[width=0.5\textwidth]{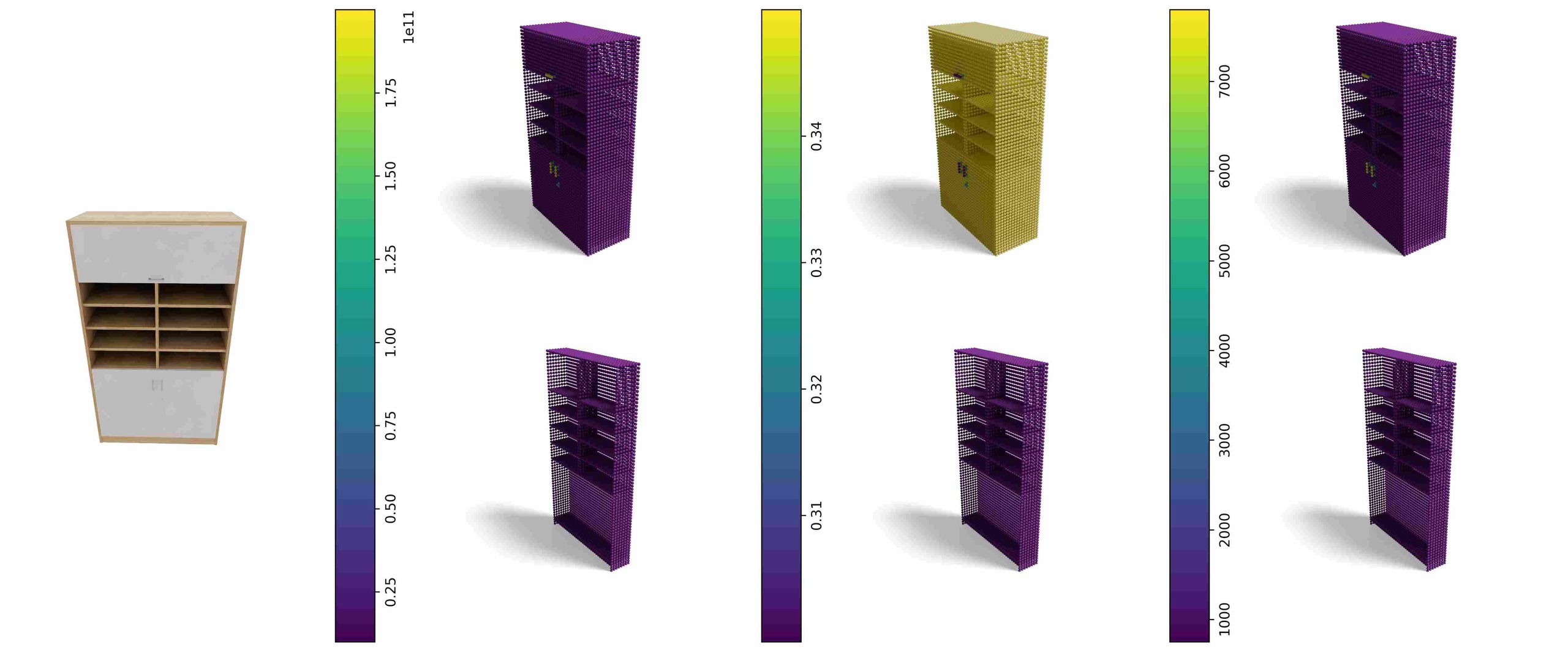} \\
\includegraphics[width=0.5\textwidth]{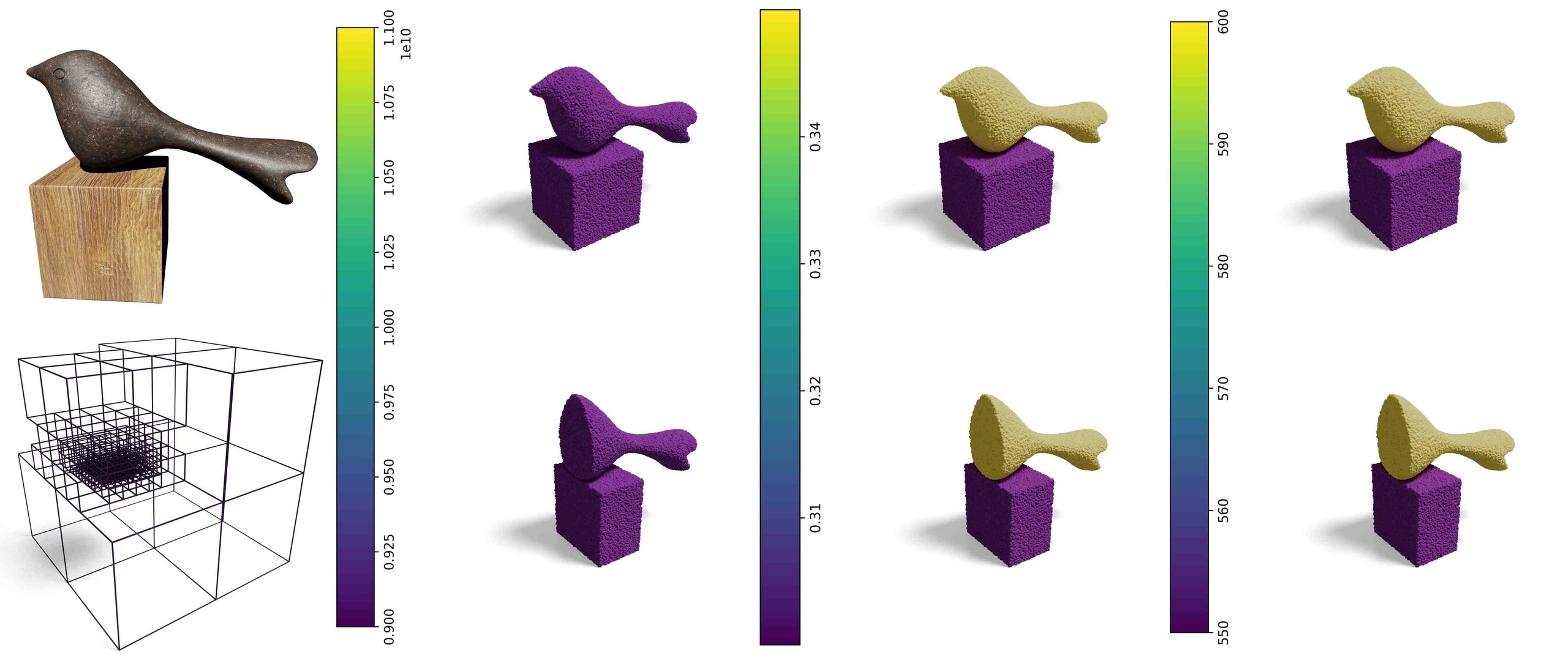} & \includegraphics[width=0.5\textwidth]{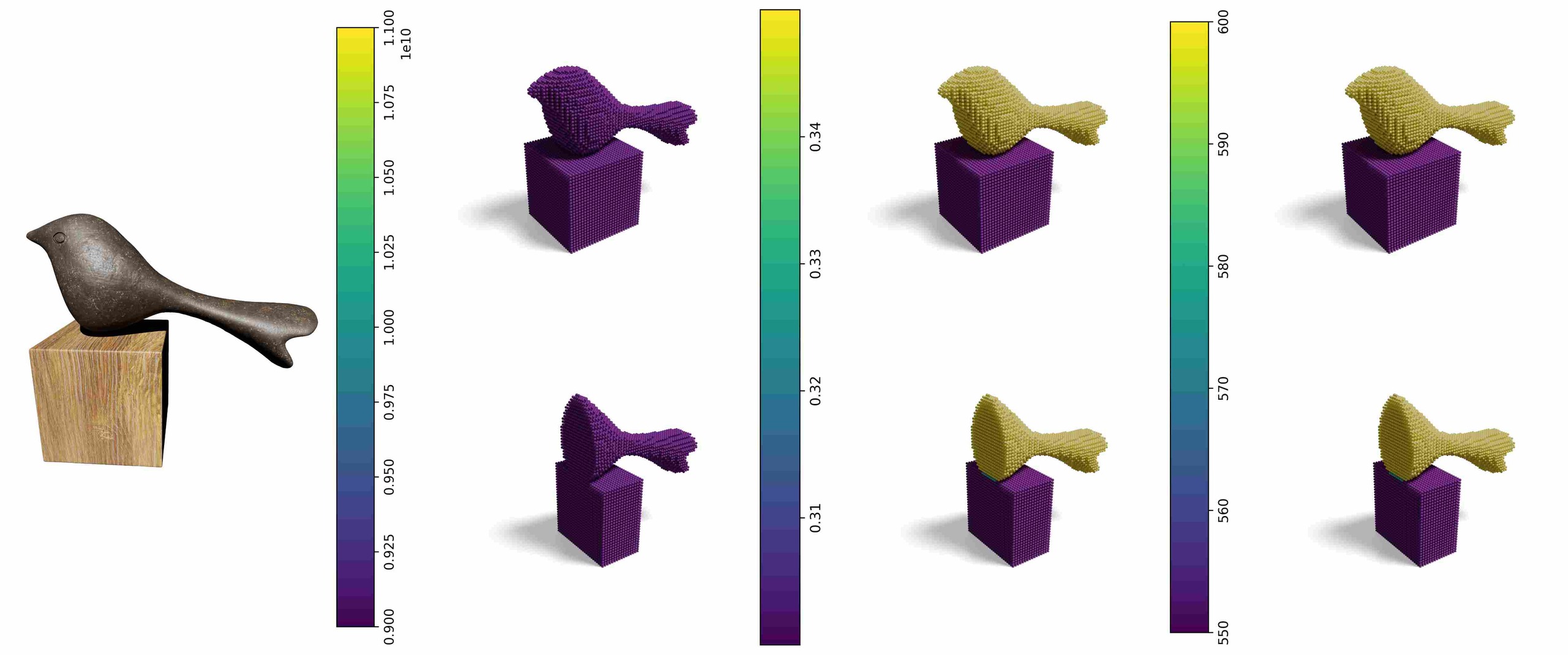} \\
    \end{tabular}
    \caption{\textbf{Inferred Mechanical Property Fields.} We show additional mechanical property fields and slice planes through mechanical property fields estimated by \ourmodel.}
    \label{fig:extrafields3}
\end{figure*}

\begin{figure*}[ht]
    \centering
    \setlength{\tabcolsep}{0pt}
    \begin{tabular}{p{0.125\textwidth}p{0.125\textwidth}p{0.125\textwidth}p{0.125\textwidth}|p{0.125\textwidth}p{0.125\textwidth}p{0.125\textwidth}p{0.125\textwidth}}
    \rowcolor{nvidiagreen!15}\multicolumn{4}{c|}{\ourmodel} &
    \multicolumn{4}{c}{VoMP} \\
    \cline{1-4}\cline{5-8}
    \centering Object & \centering Young's Modulus ($E$, Pa) & \centering Poisson's Ratio ($\nu$) & \centering Density ($\rho, \frac{kg}{m^3}$) & \centering Object & \centering Young's Modulus ($E$, Pa) & \centering Poisson's Ratio ($\nu$) & \centering Density ($\rho, \frac{kg}{m^3}$) \\
    \end{tabular}
    \begin{tabular}{c|c}
\includegraphics[width=0.5\textwidth]{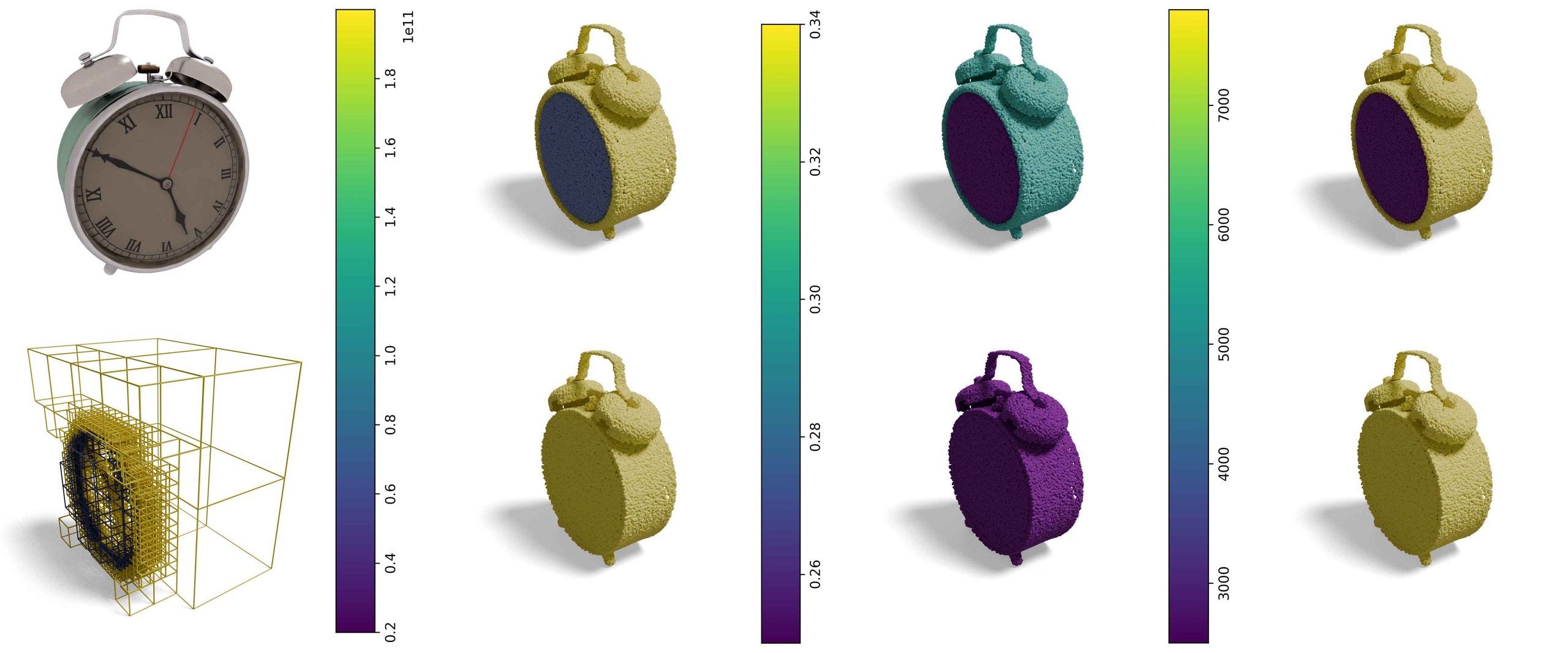} & \includegraphics[width=0.5\textwidth]{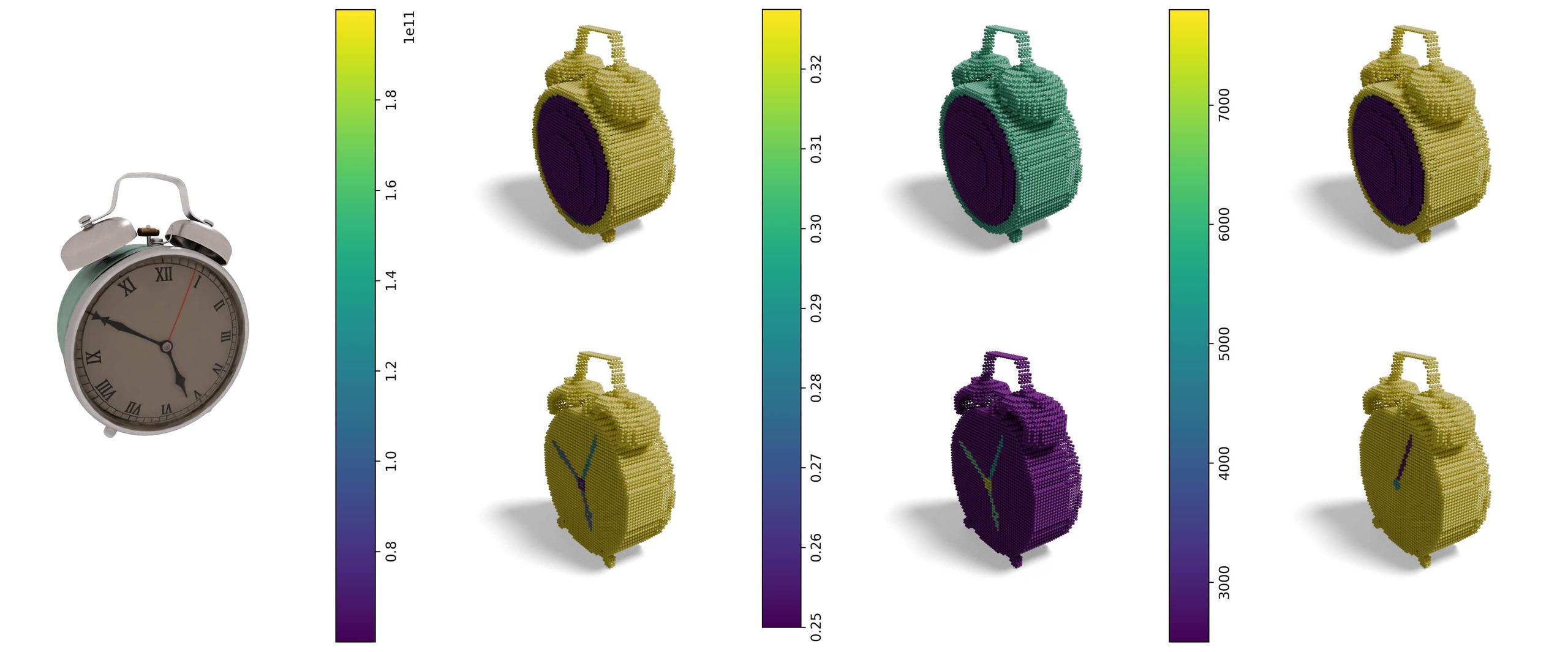} \\
\includegraphics[width=0.5\textwidth]{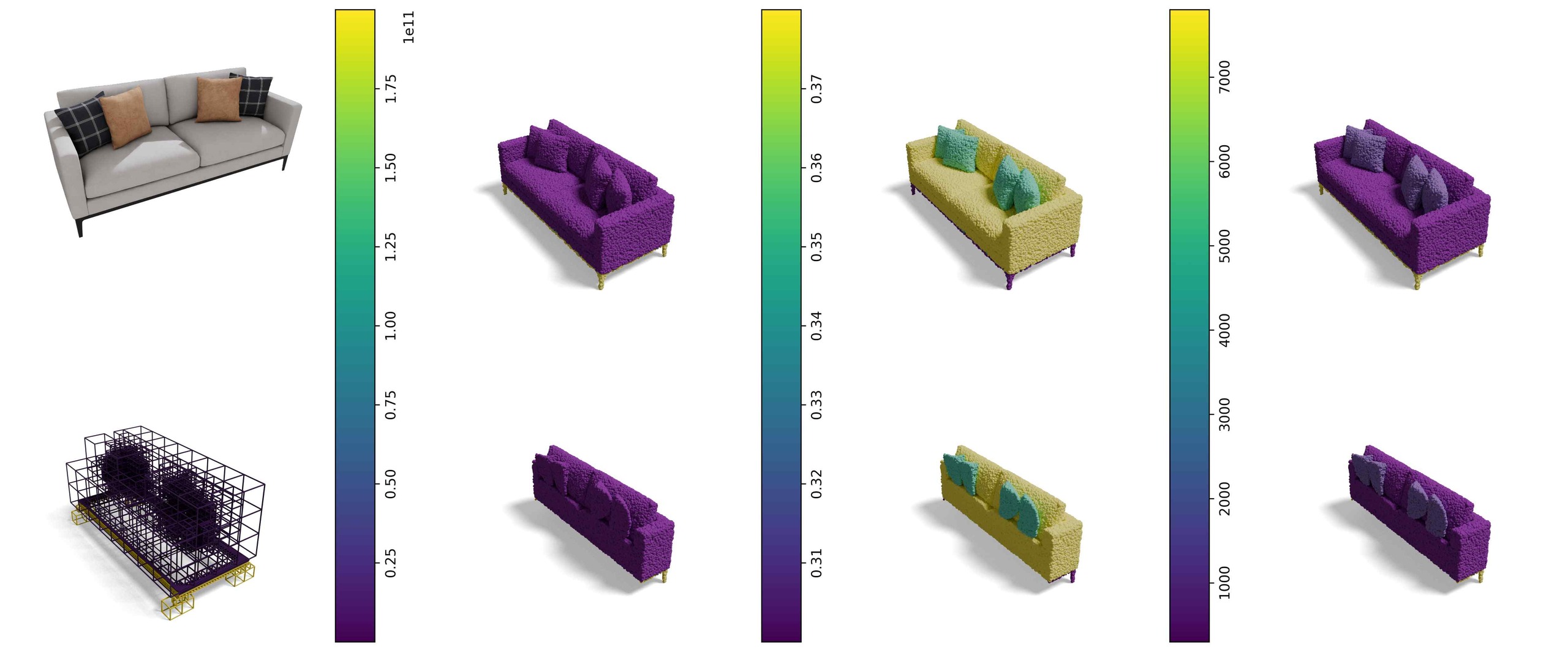} & \includegraphics[width=0.5\textwidth]{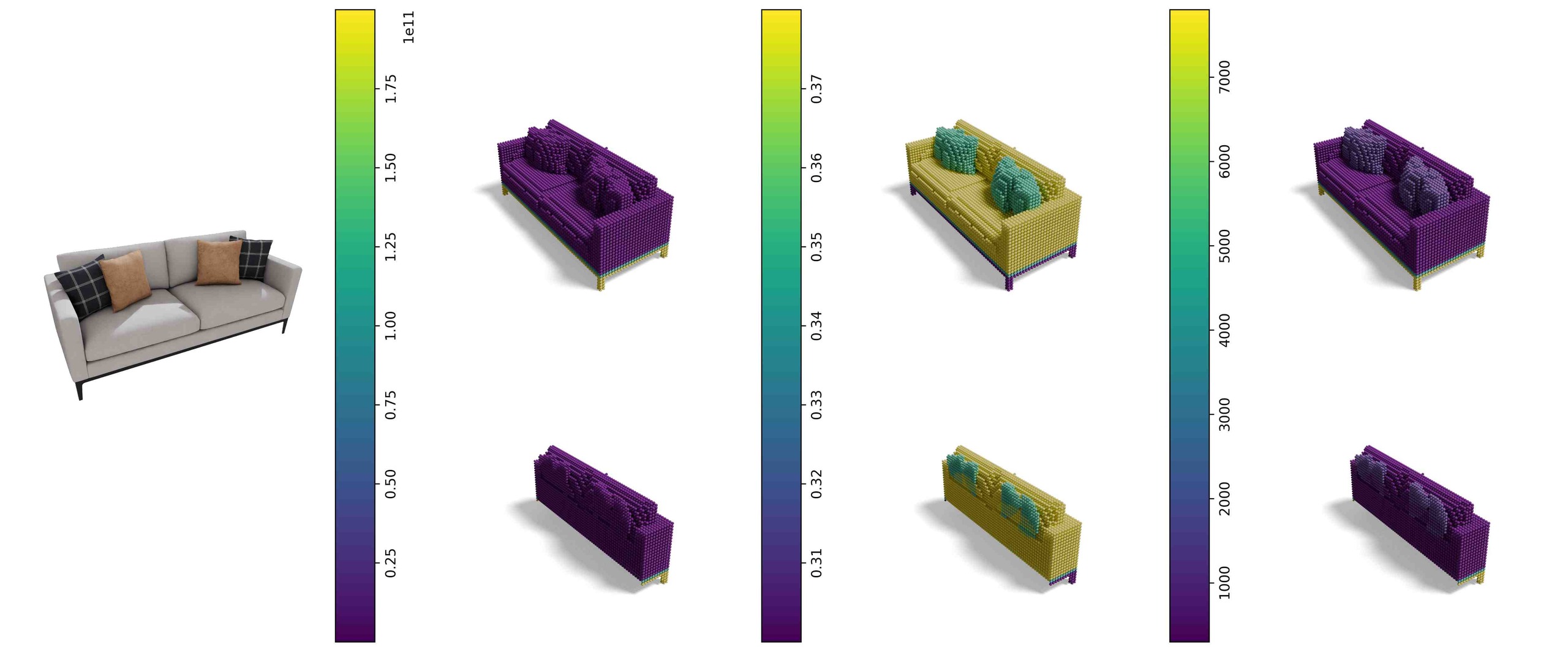} \\
\includegraphics[width=0.5\textwidth]{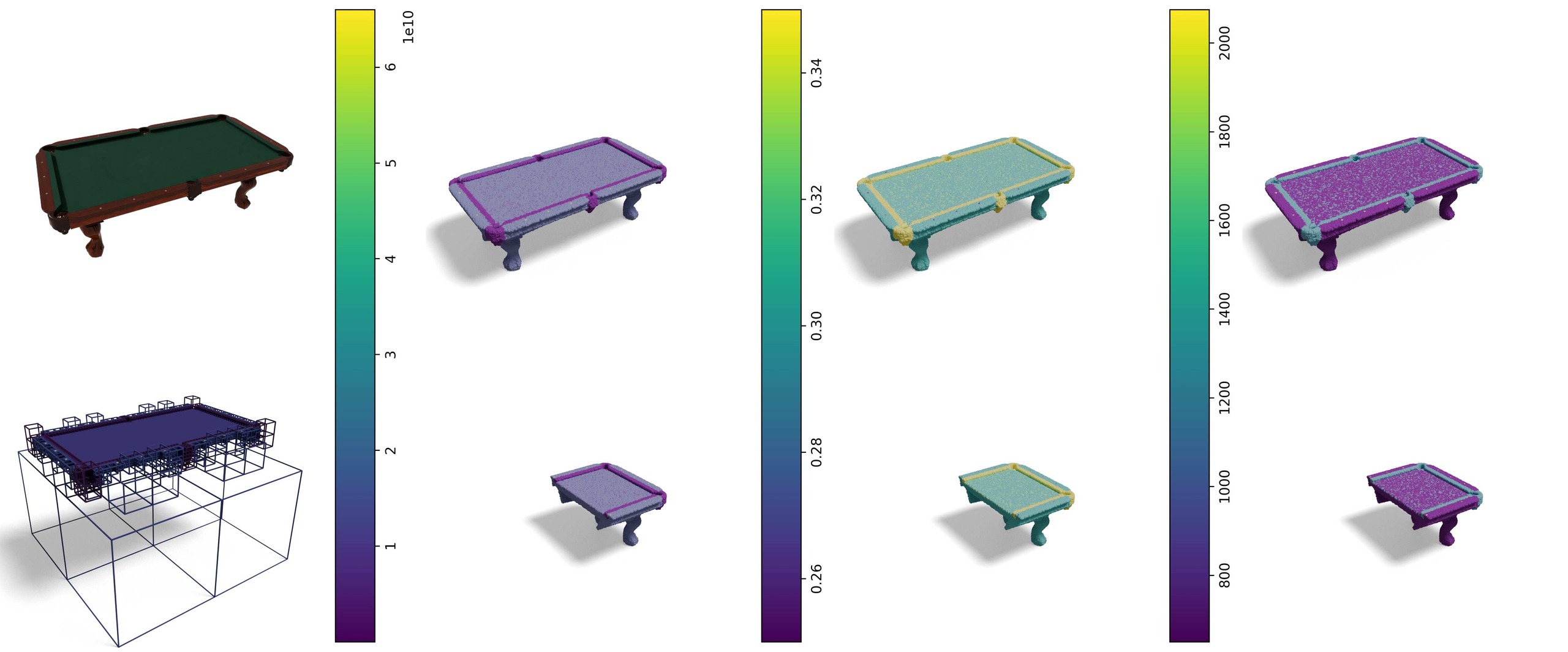} & \includegraphics[width=0.5\textwidth]{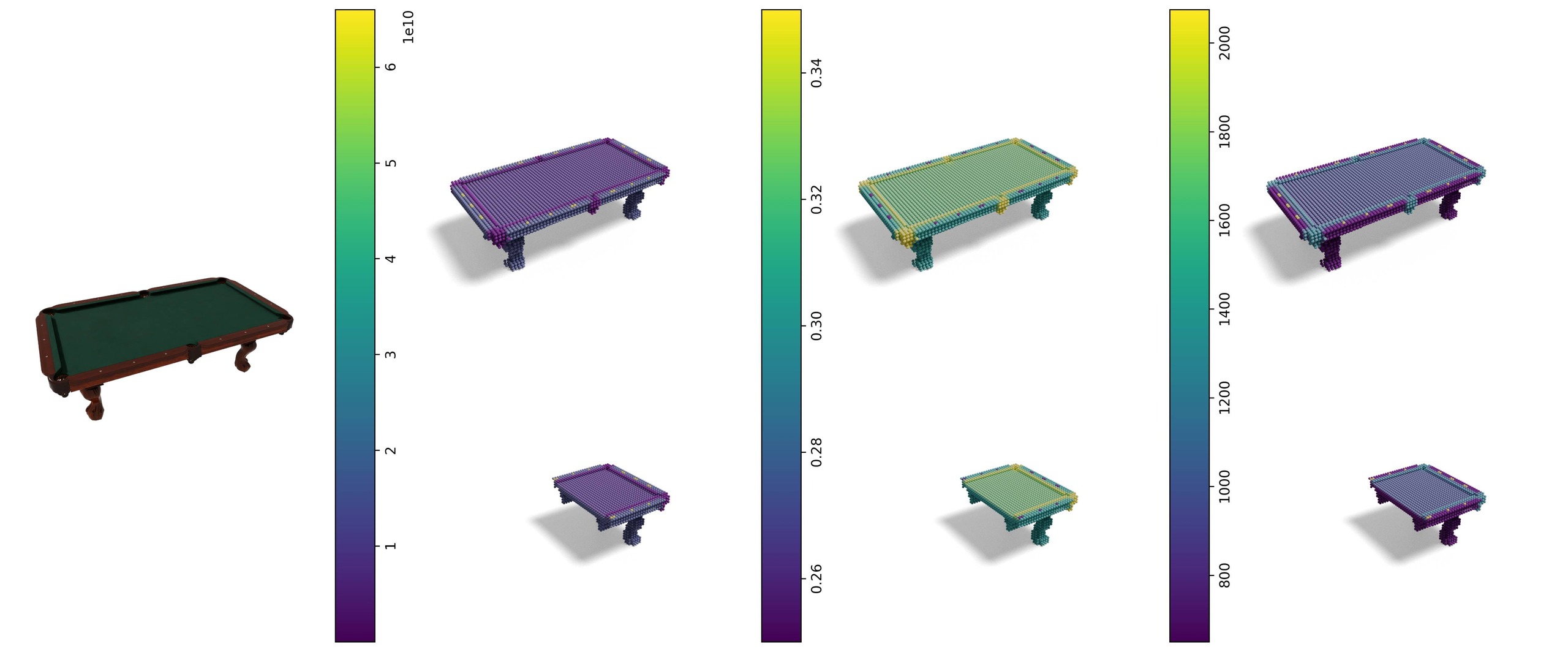} \\
\includegraphics[width=0.5\textwidth]{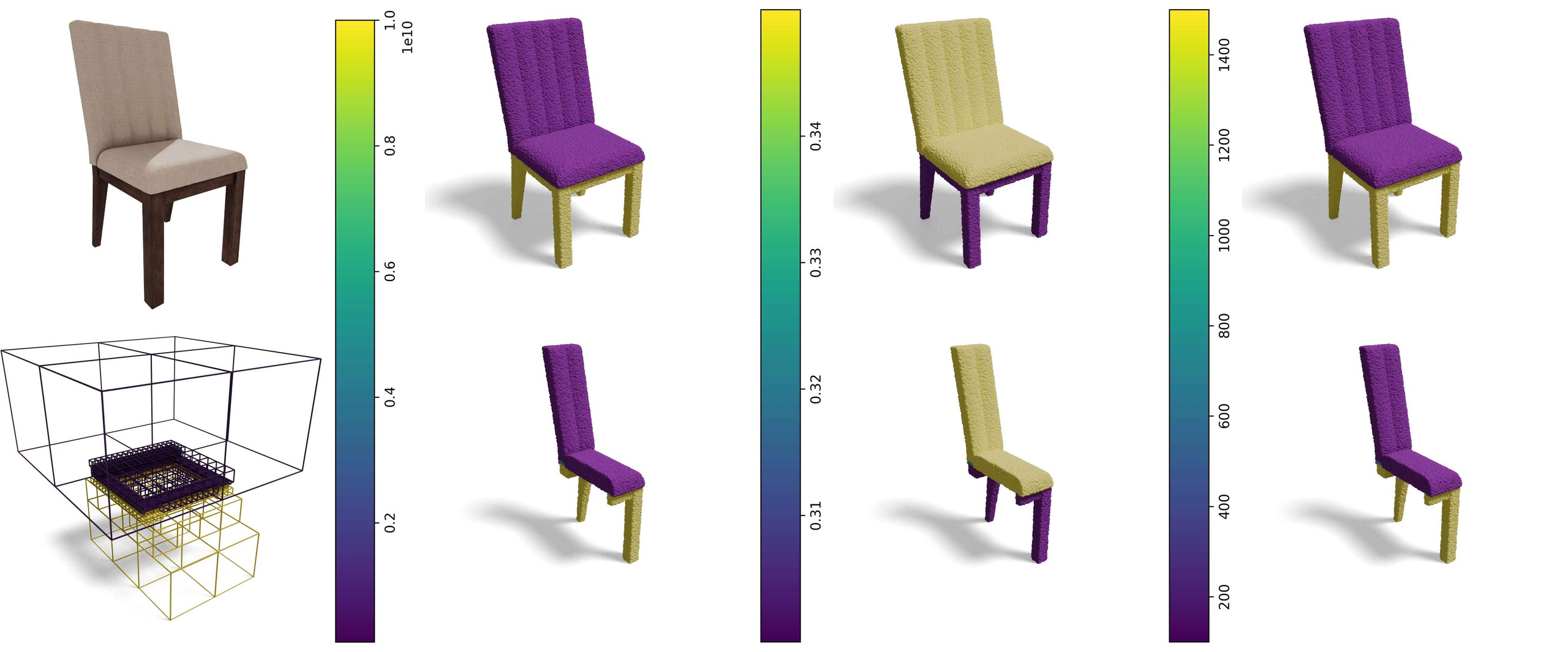} & \includegraphics[width=0.5\textwidth]{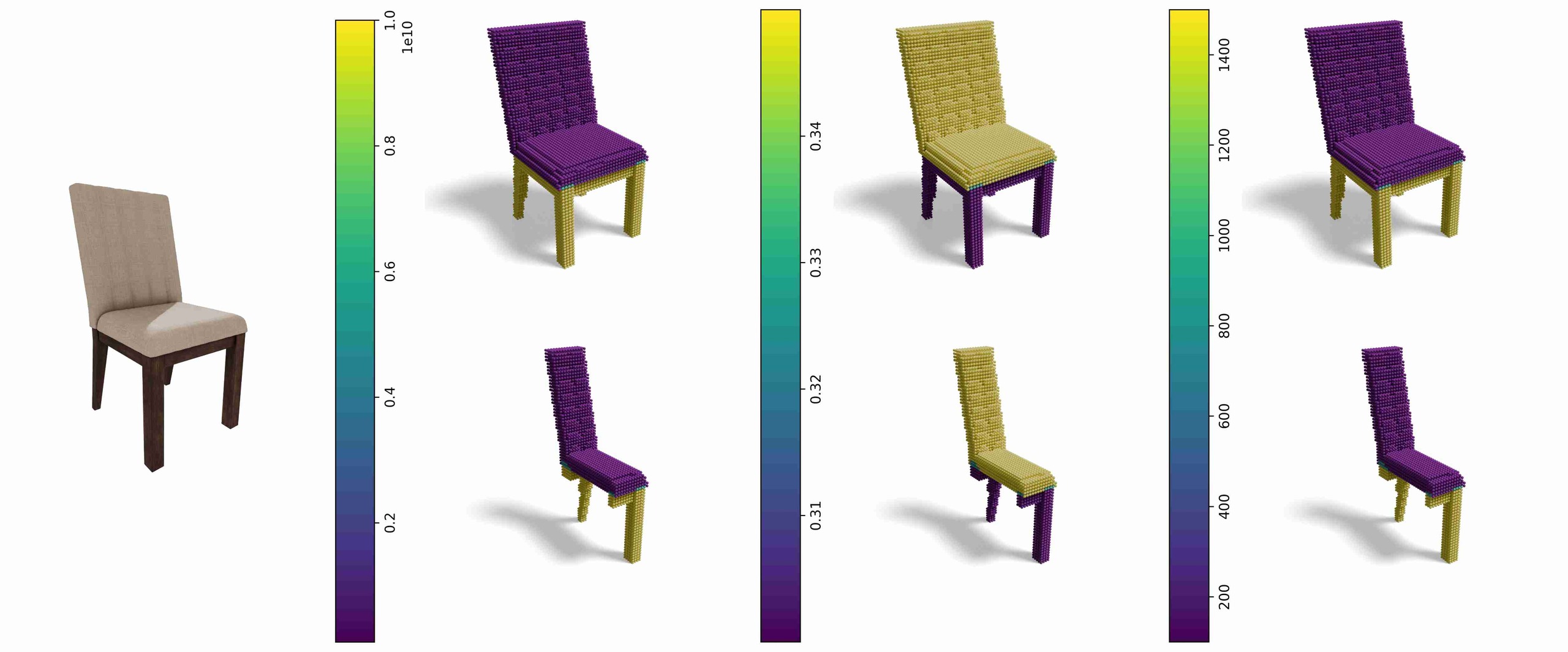} \\
\includegraphics[width=0.5\textwidth]{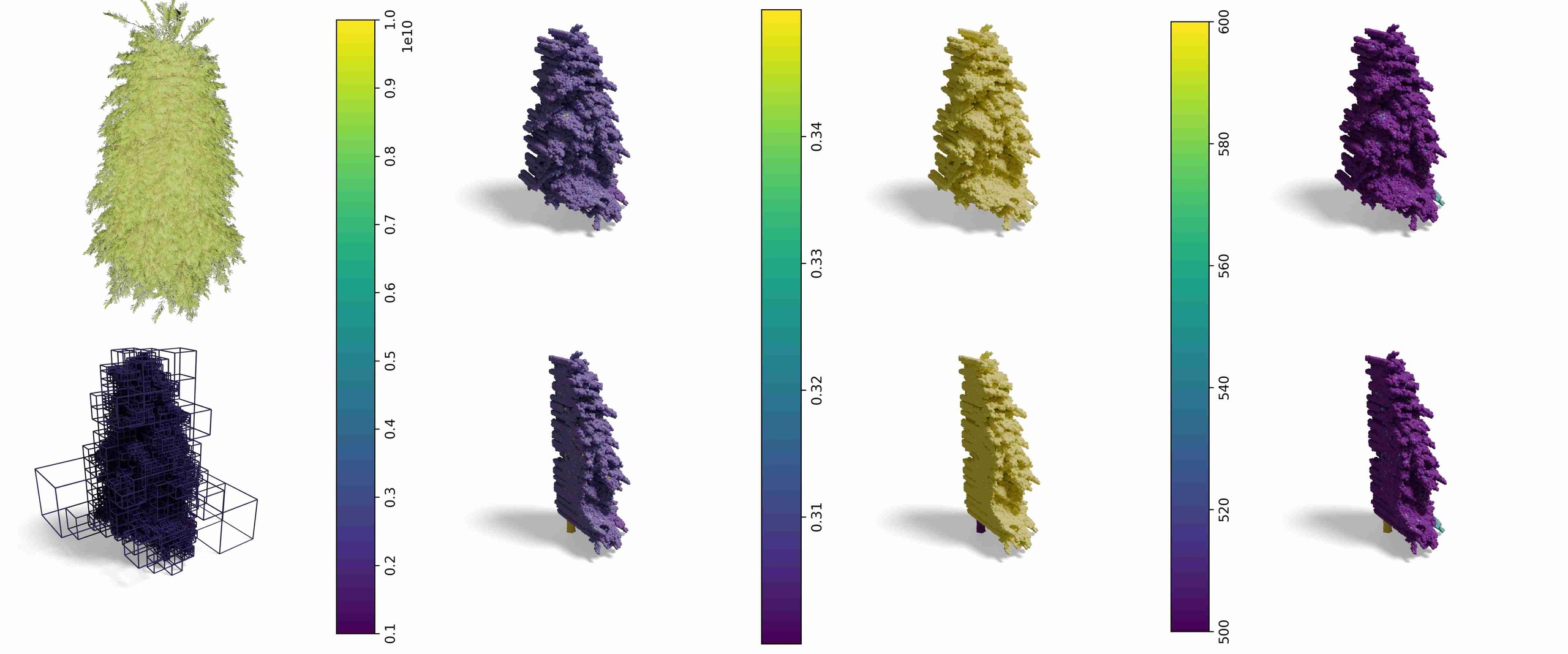} & \includegraphics[width=0.5\textwidth]{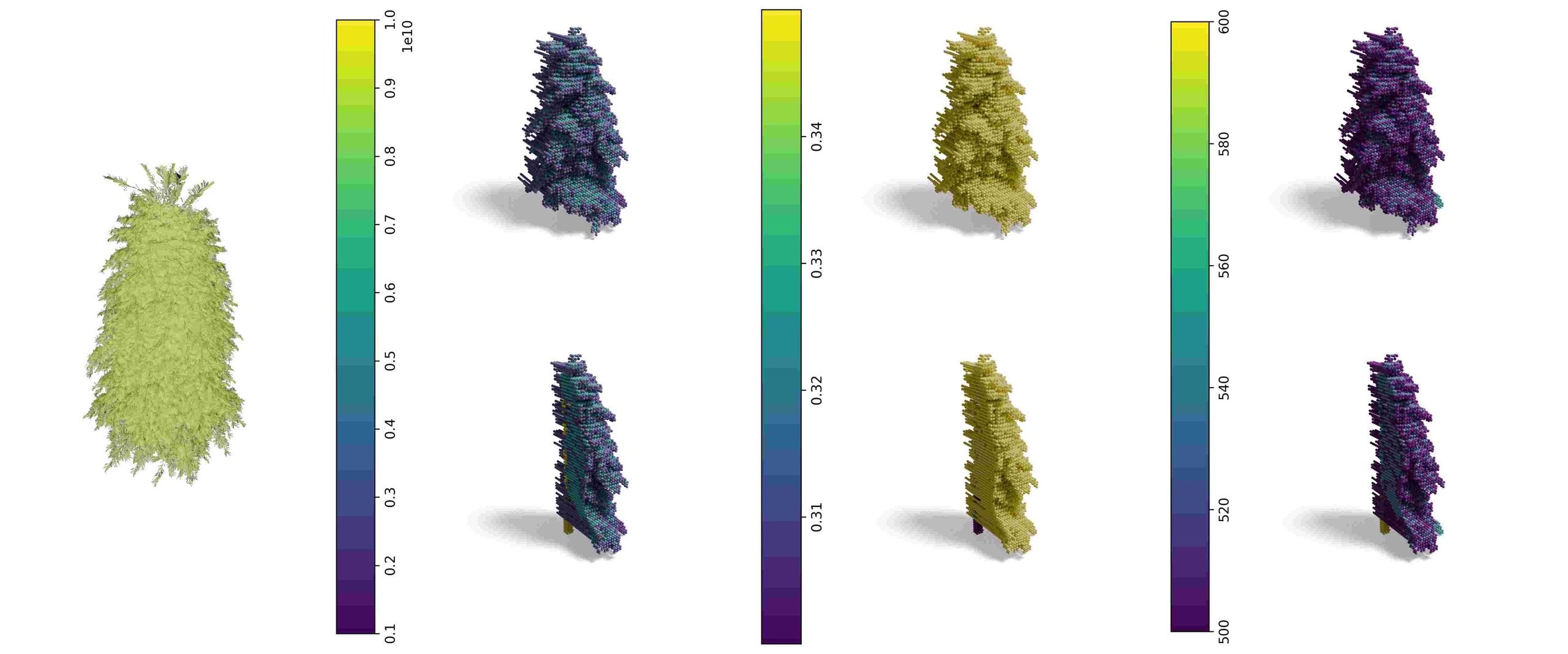}
    \end{tabular}
    \caption{\textbf{Inferred Mechanical Property Fields.} We show additional mechanical property fields and slice planes through mechanical property fields estimated by \ourmodel.}
    \label{fig:extrafields1}
\end{figure*}

\begin{figure*}[ht]
    \centering
    \setlength{\tabcolsep}{0pt}
    \begin{tabular}{p{0.125\textwidth}p{0.125\textwidth}p{0.125\textwidth}p{0.125\textwidth}|p{0.125\textwidth}p{0.125\textwidth}p{0.125\textwidth}p{0.125\textwidth}}
    \rowcolor{nvidiagreen!15}\multicolumn{4}{c|}{\ourmodel} &
    \multicolumn{4}{c}{VoMP} \\
    \cline{1-4}\cline{5-8}
    \centering Object & \centering Young's Modulus ($E$, Pa) & \centering Poisson's Ratio ($\nu$) & \centering Density ($\rho, \frac{kg}{m^3}$) & \centering Object & \centering Young's Modulus ($E$, Pa) & \centering Poisson's Ratio ($\nu$) & \centering Density ($\rho, \frac{kg}{m^3}$) \\
    \end{tabular}
    \begin{tabular}{c|c}
\includegraphics[width=0.5\textwidth]{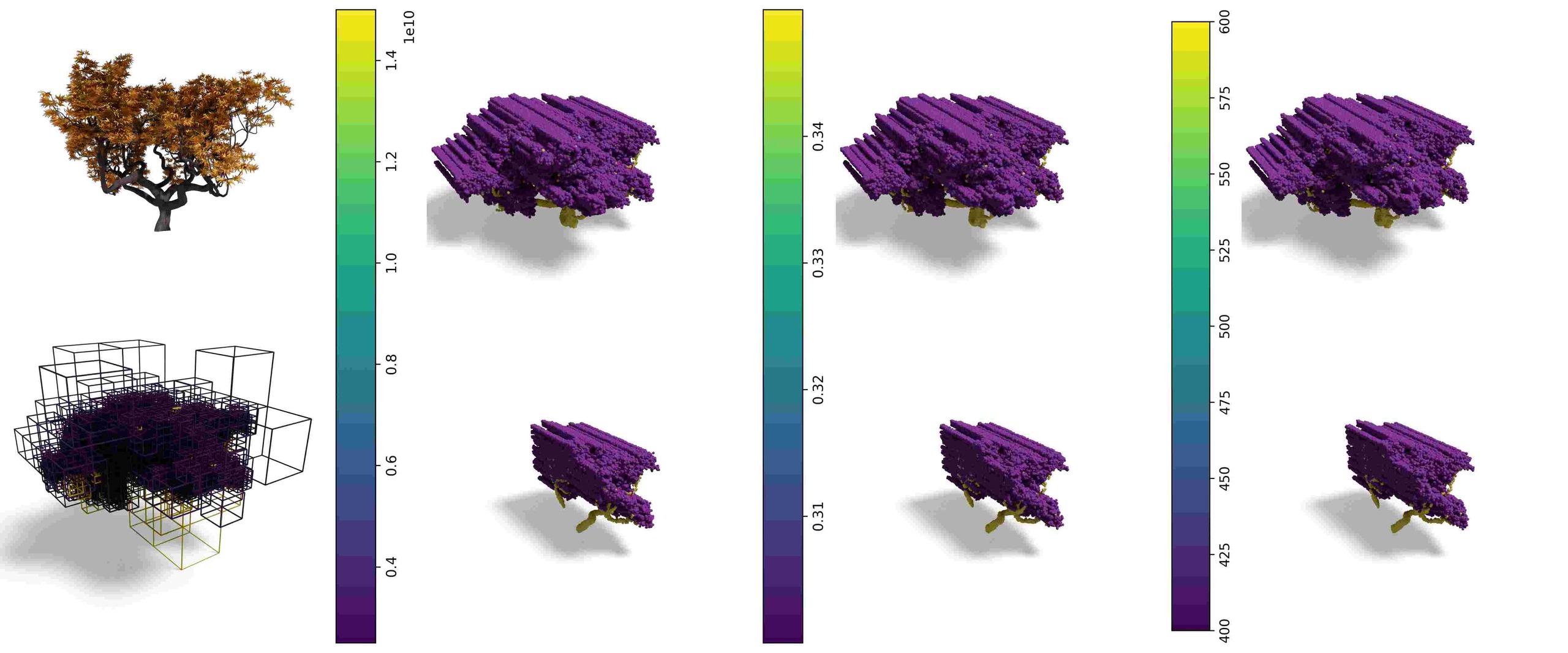} & \includegraphics[width=0.5\textwidth]{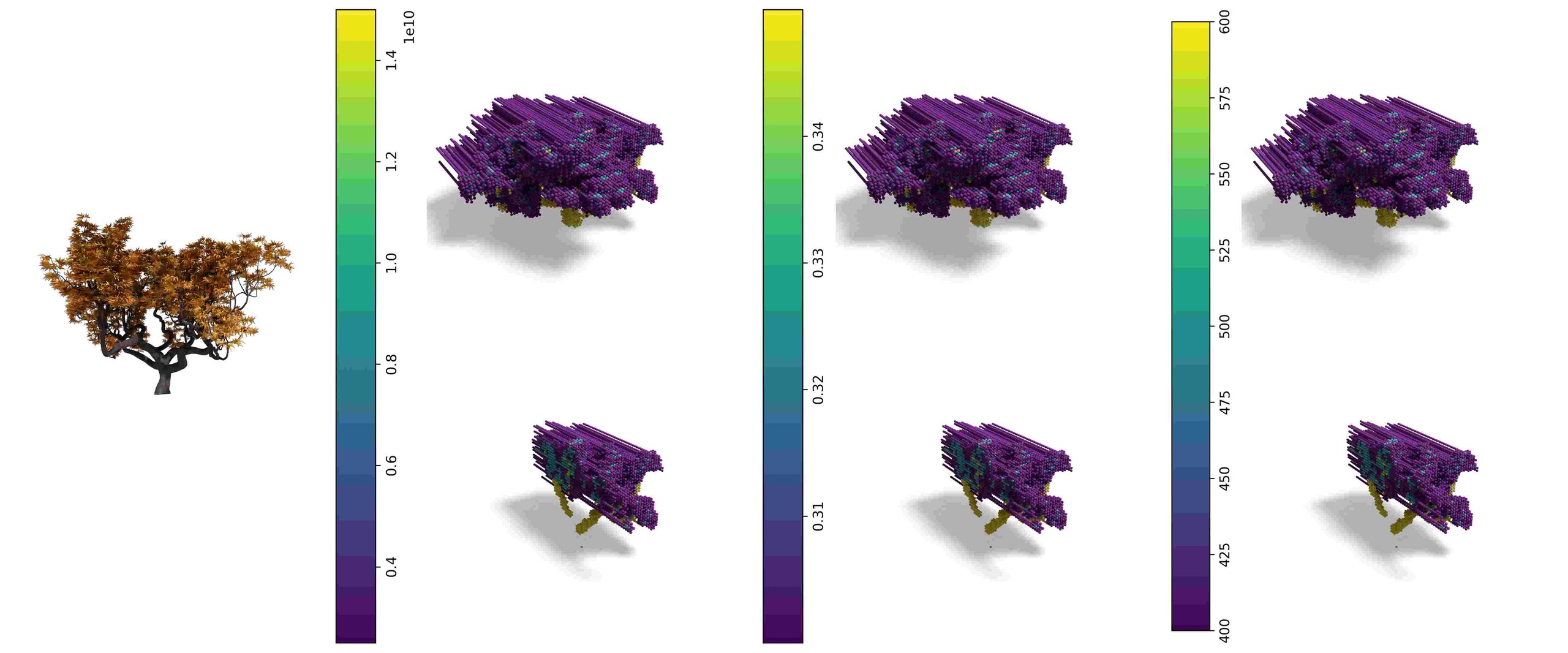} \\
\includegraphics[width=0.5\textwidth]{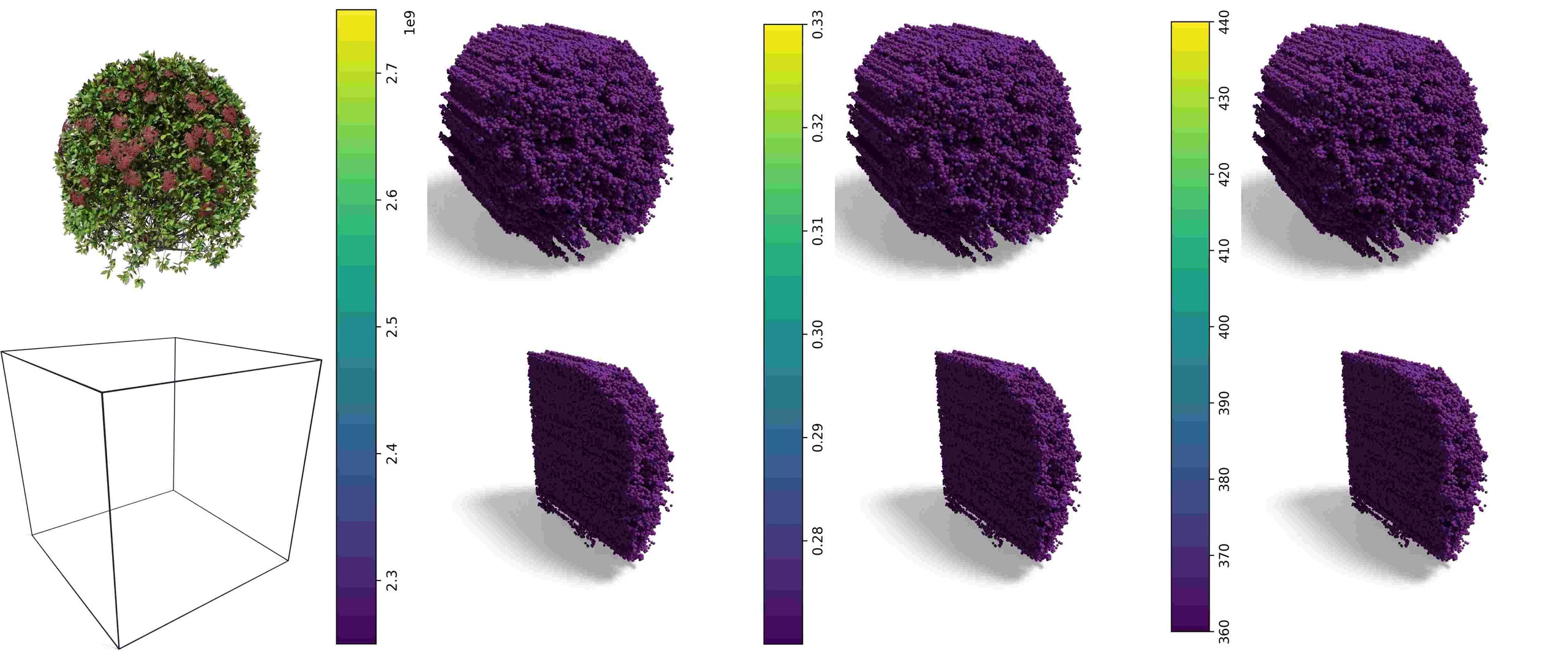} & \includegraphics[width=0.5\textwidth]{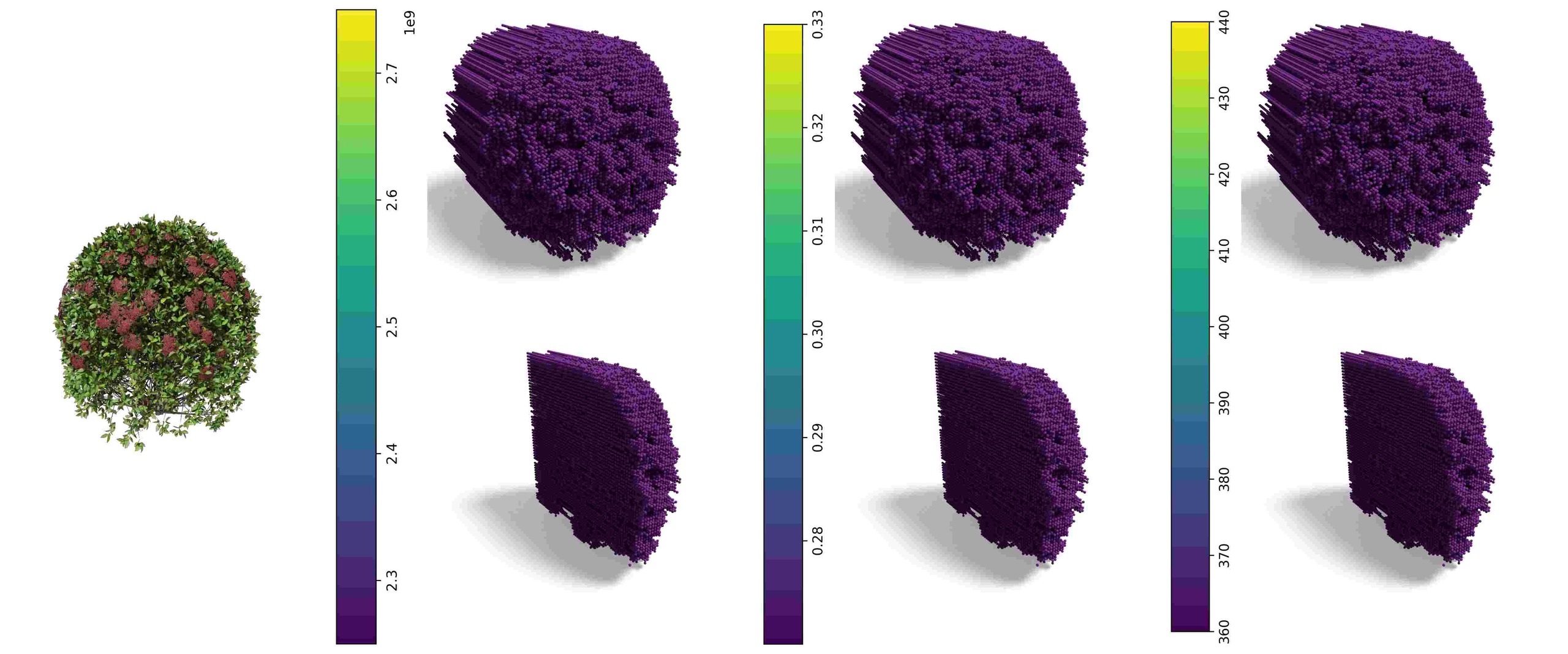} \\
\includegraphics[width=0.5\textwidth]{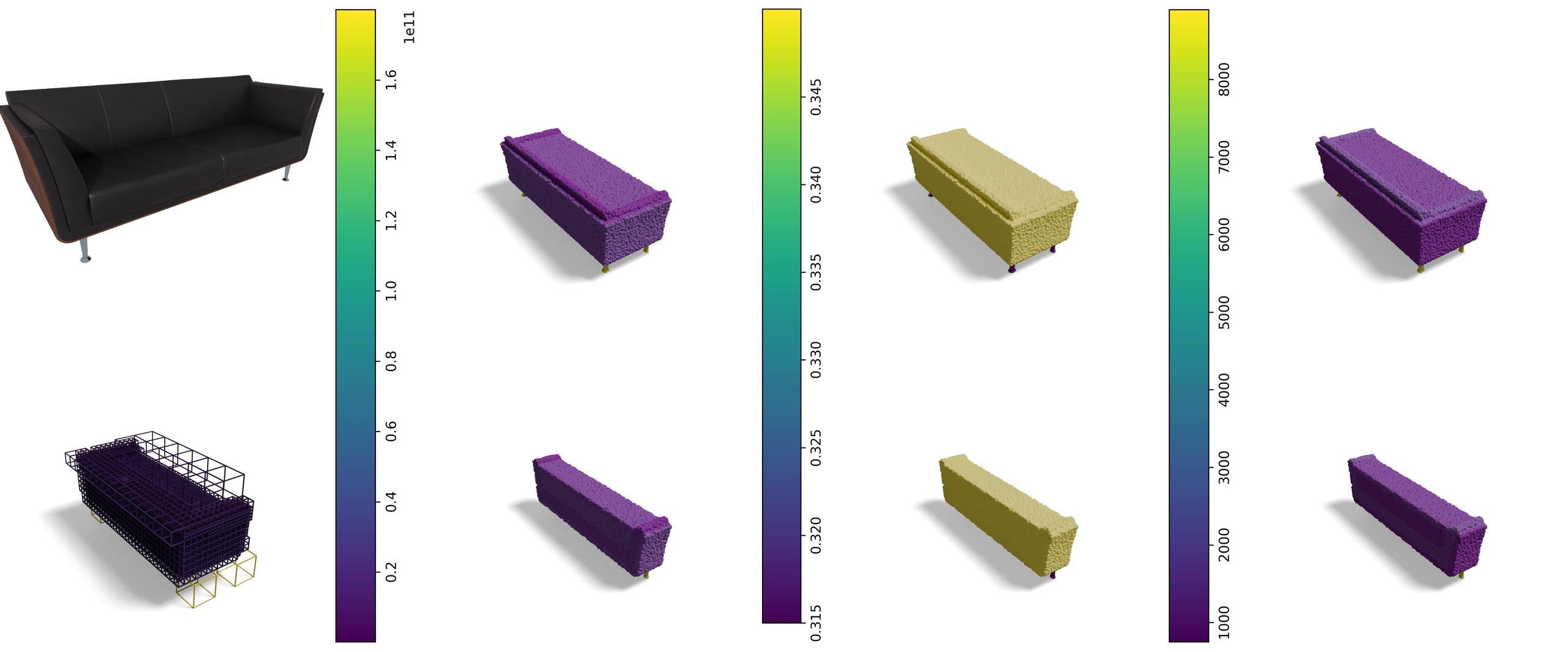} & \includegraphics[width=0.5\textwidth]{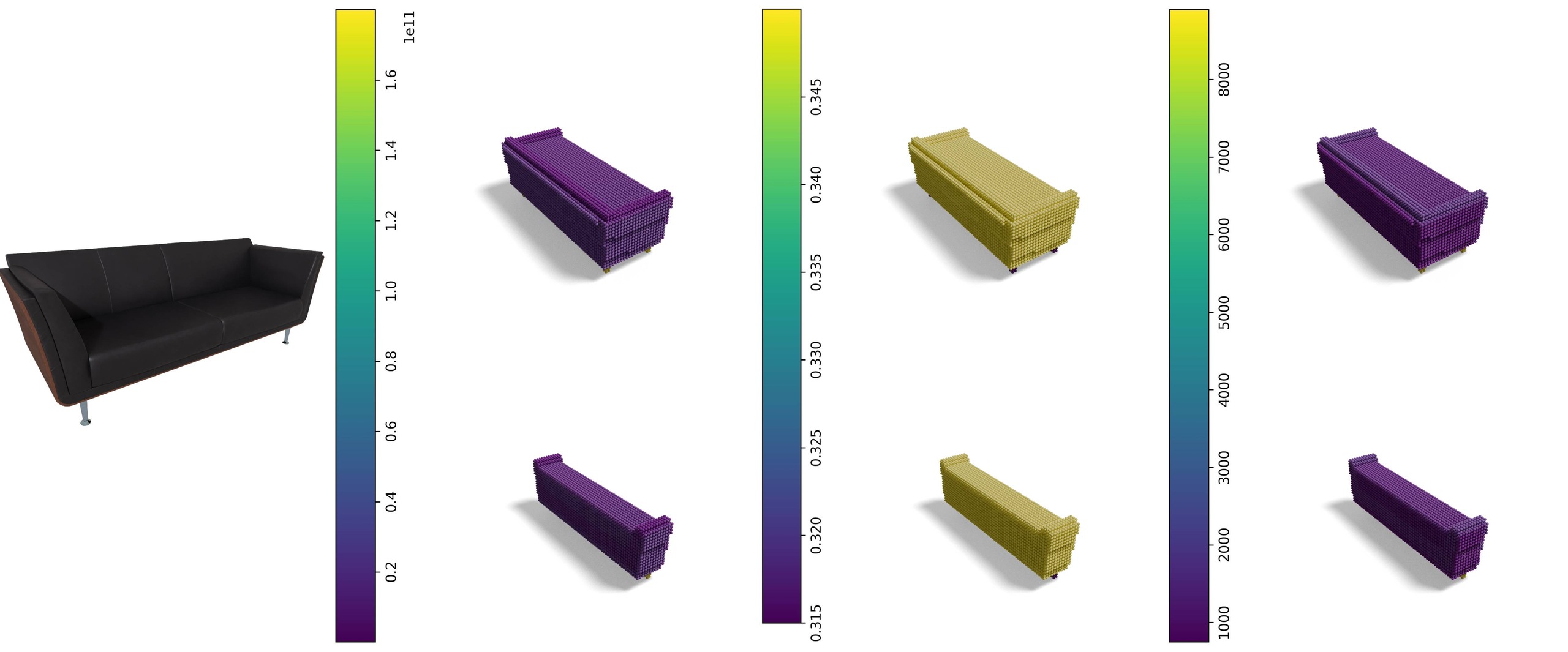} \\
\includegraphics[width=0.5\textwidth]{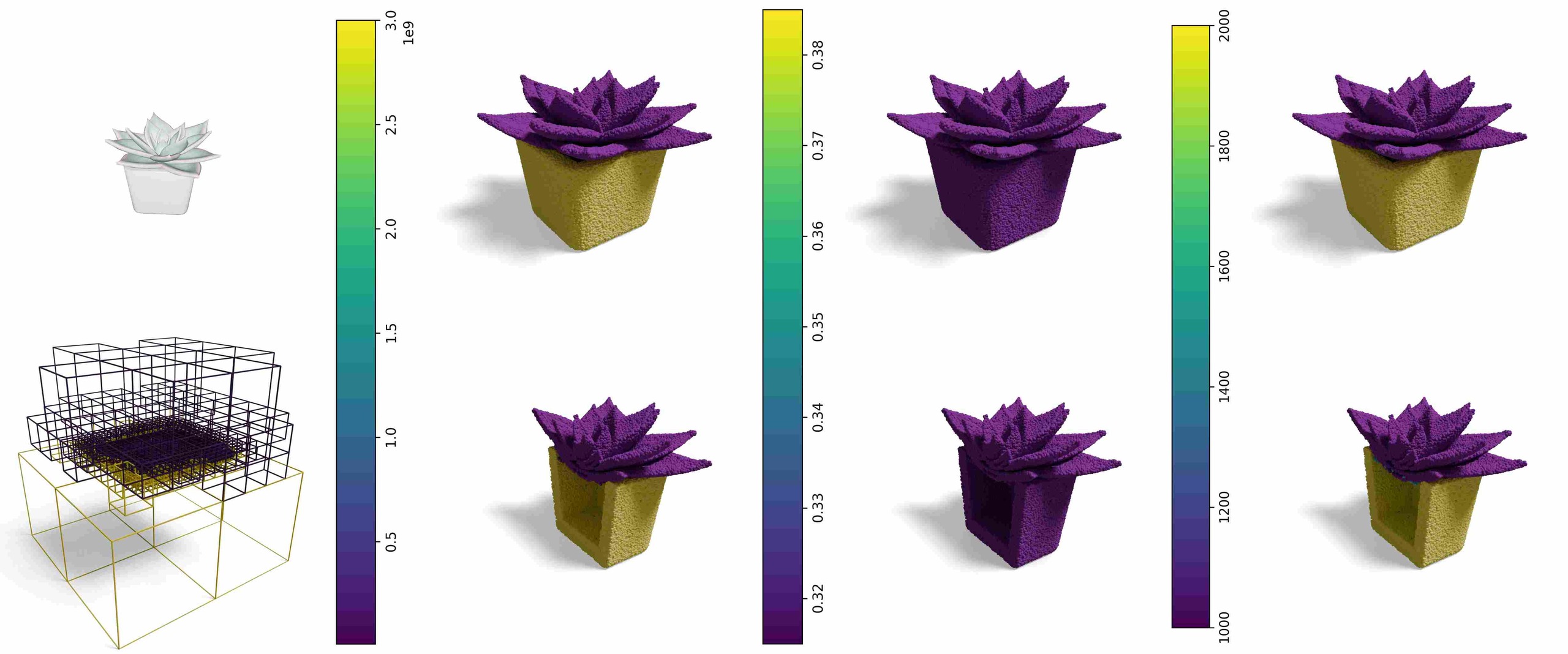} & \includegraphics[width=0.5\textwidth]{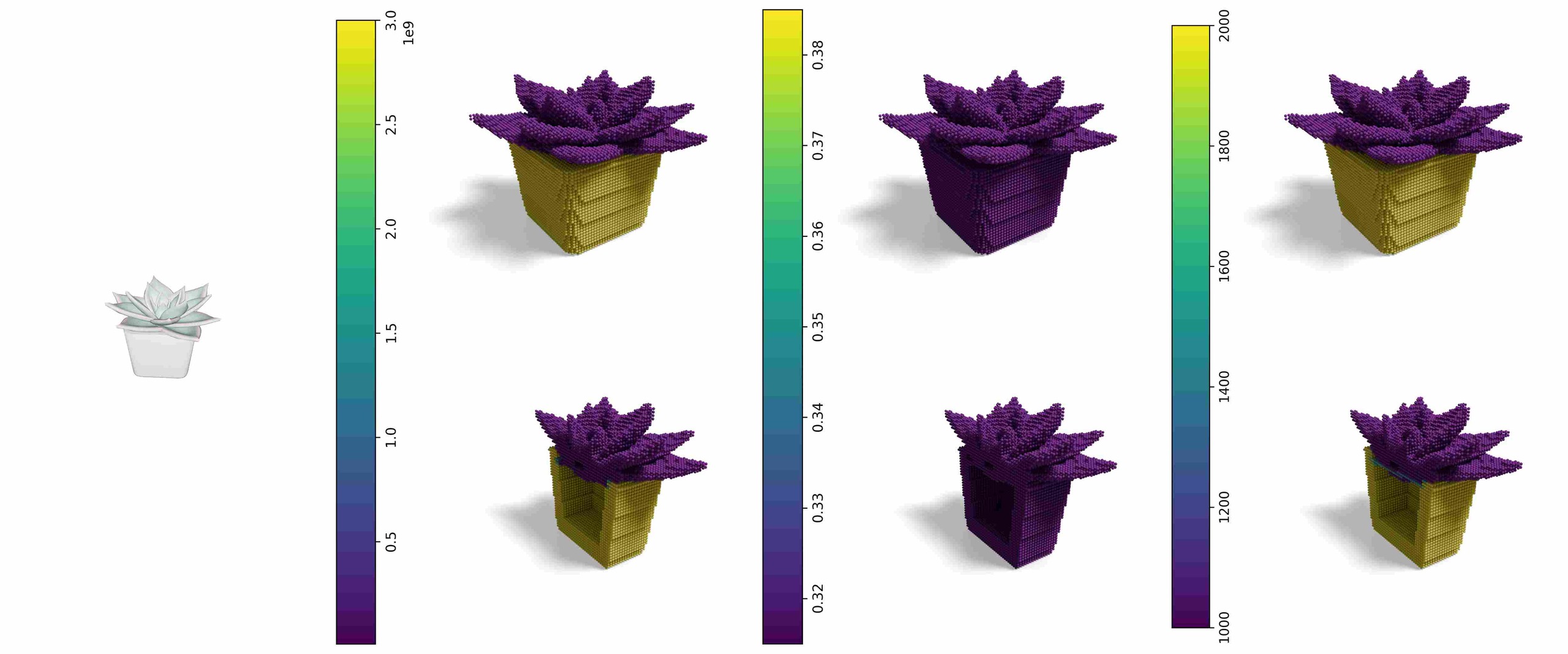} \\
\includegraphics[width=0.5\textwidth]{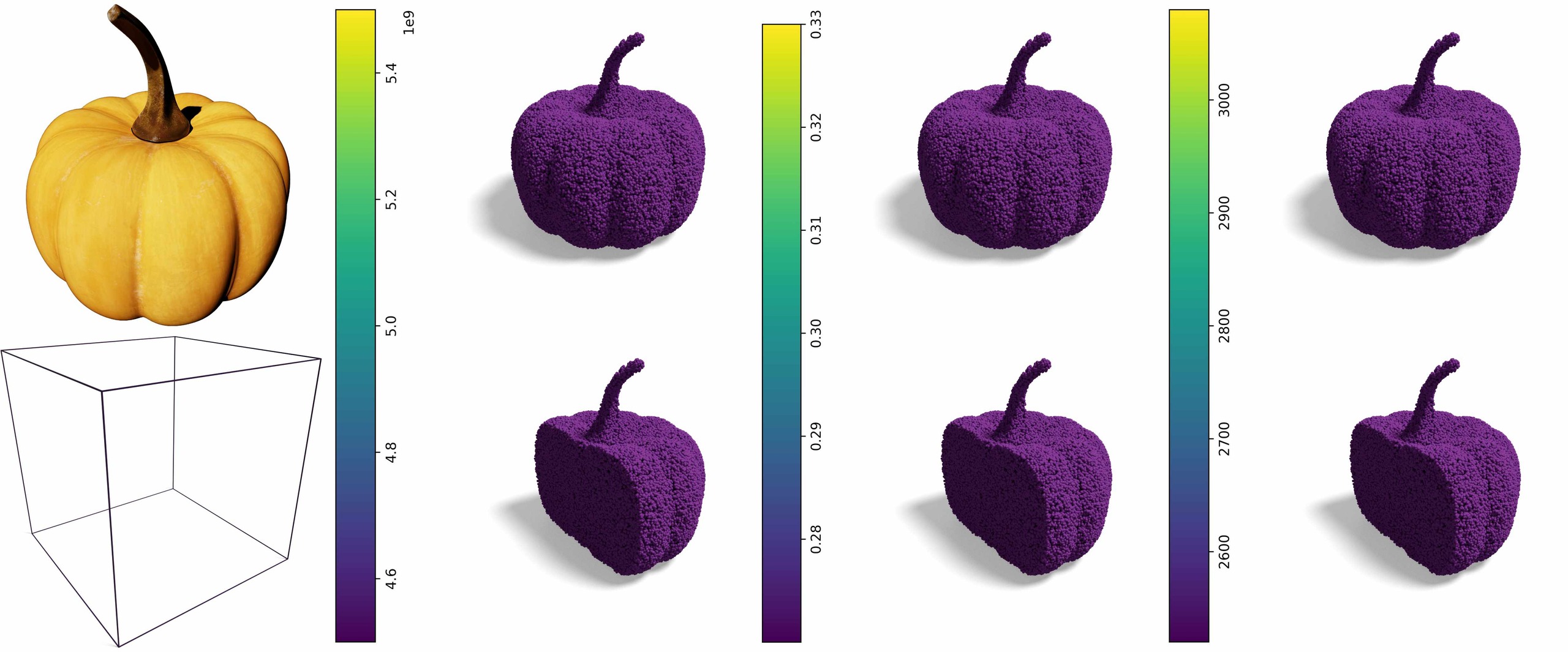} & \includegraphics[width=0.5\textwidth]{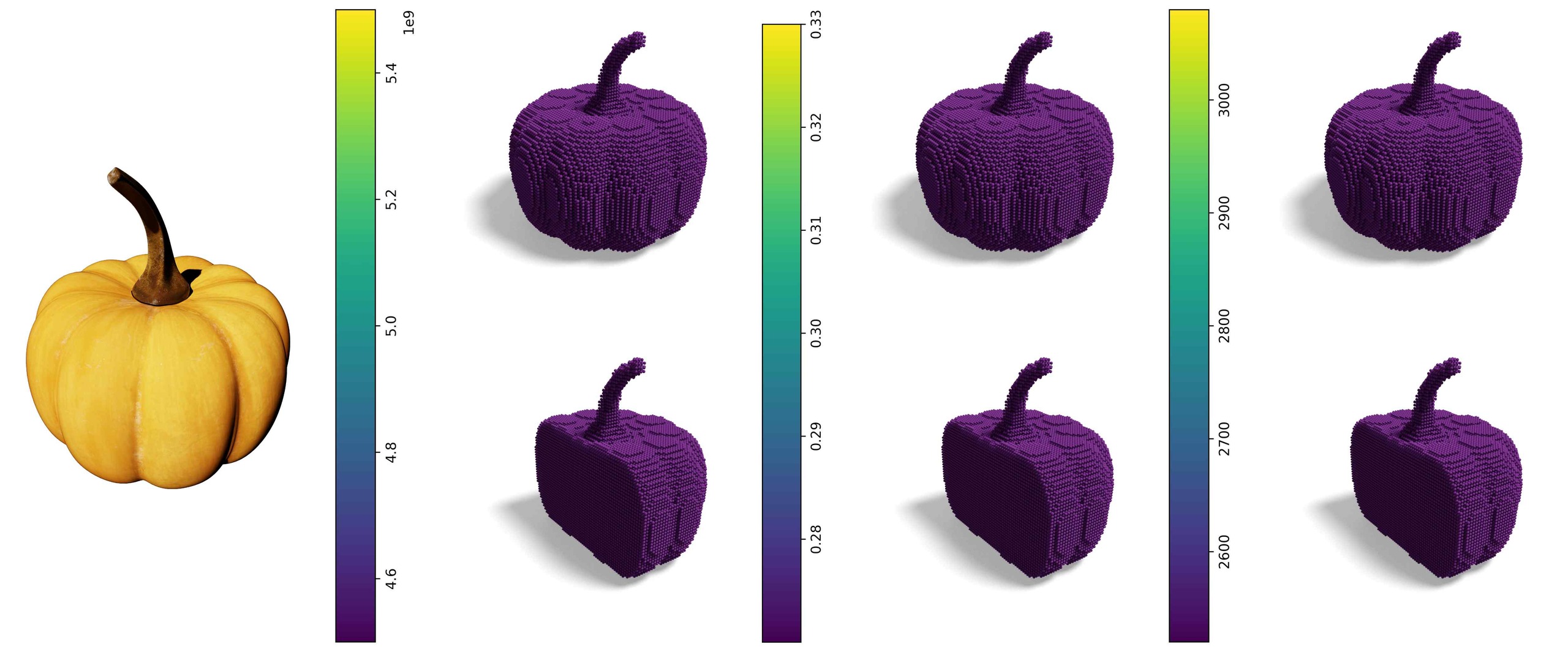} \\
    \end{tabular}
    \caption{\textbf{Inferred Mechanical Property Fields.} We show additional mechanical property fields and slice planes through mechanical property fields estimated by \ourmodel.}
    \label{fig:extrafields2}
\end{figure*}

\begin{figure*}[ht]
    \centering
    \setlength{\tabcolsep}{0pt}
    \renewcommand{\arraystretch}{0}
    \begin{tabular}{p{0.125\textwidth}p{0.125\textwidth}p{0.125\textwidth}p{0.125\textwidth}|p{0.125\textwidth}p{0.125\textwidth}p{0.125\textwidth}p{0.125\textwidth}}
    \centering Object & \centering Young's Modulus ($E$, Pa) & \centering Poisson's Ratio ($\nu$) & \centering Density ($\rho, \frac{kg}{m^3}$) & \centering Object & \centering Young's Modulus ($E$, Pa) & \centering Poisson's Ratio ($\nu$) & \centering Density ($\rho, \frac{kg}{m^3}$) \\
    \end{tabular}
    \begin{tabular}{@{}c@{\hspace{10pt}}c@{}}
    \textbf{\ourmodel} & \textbf{VoMP} \\[2pt]
    \includegraphics[width=0.5\textwidth]{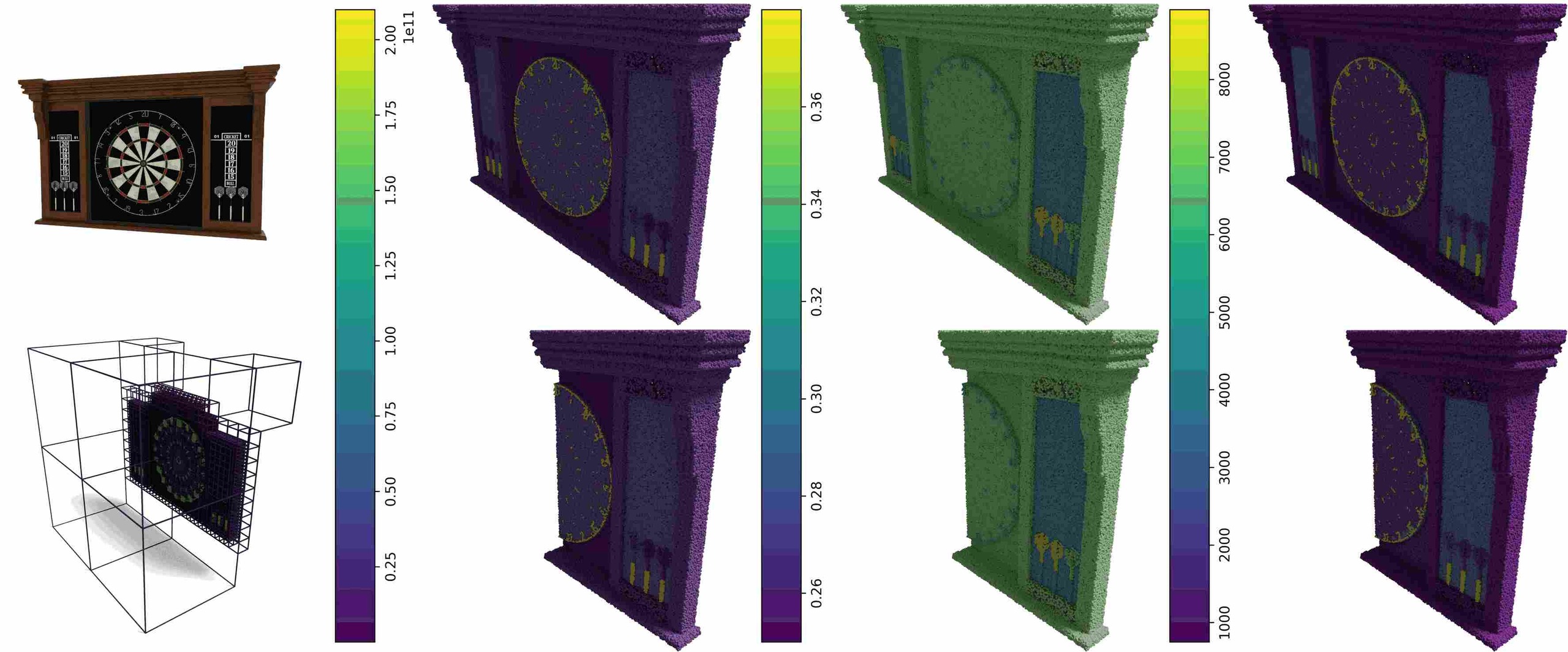} & 
    \includegraphics[width=0.5\textwidth]{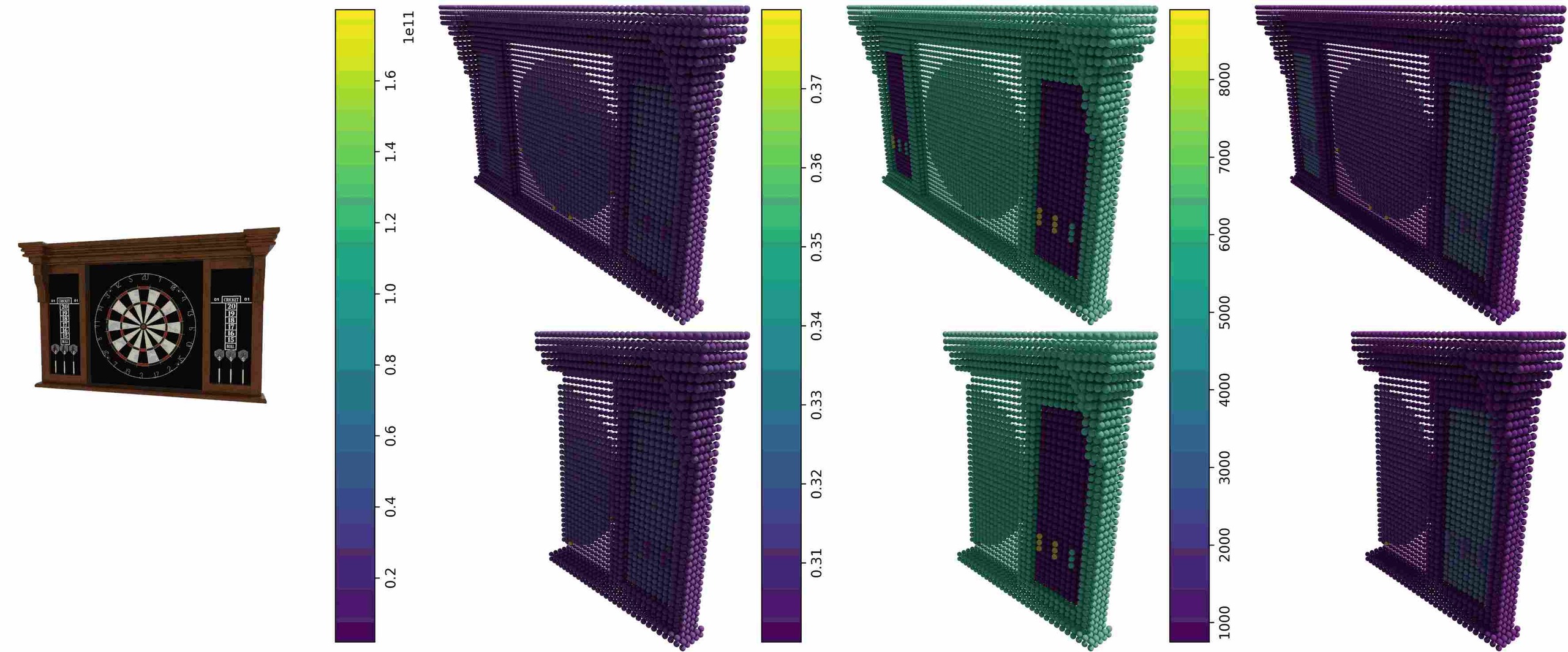} \\[2pt]
    \textbf{N2P} & \textbf{PUGS} \\[2pt]
    \includegraphics[width=0.5\textwidth]{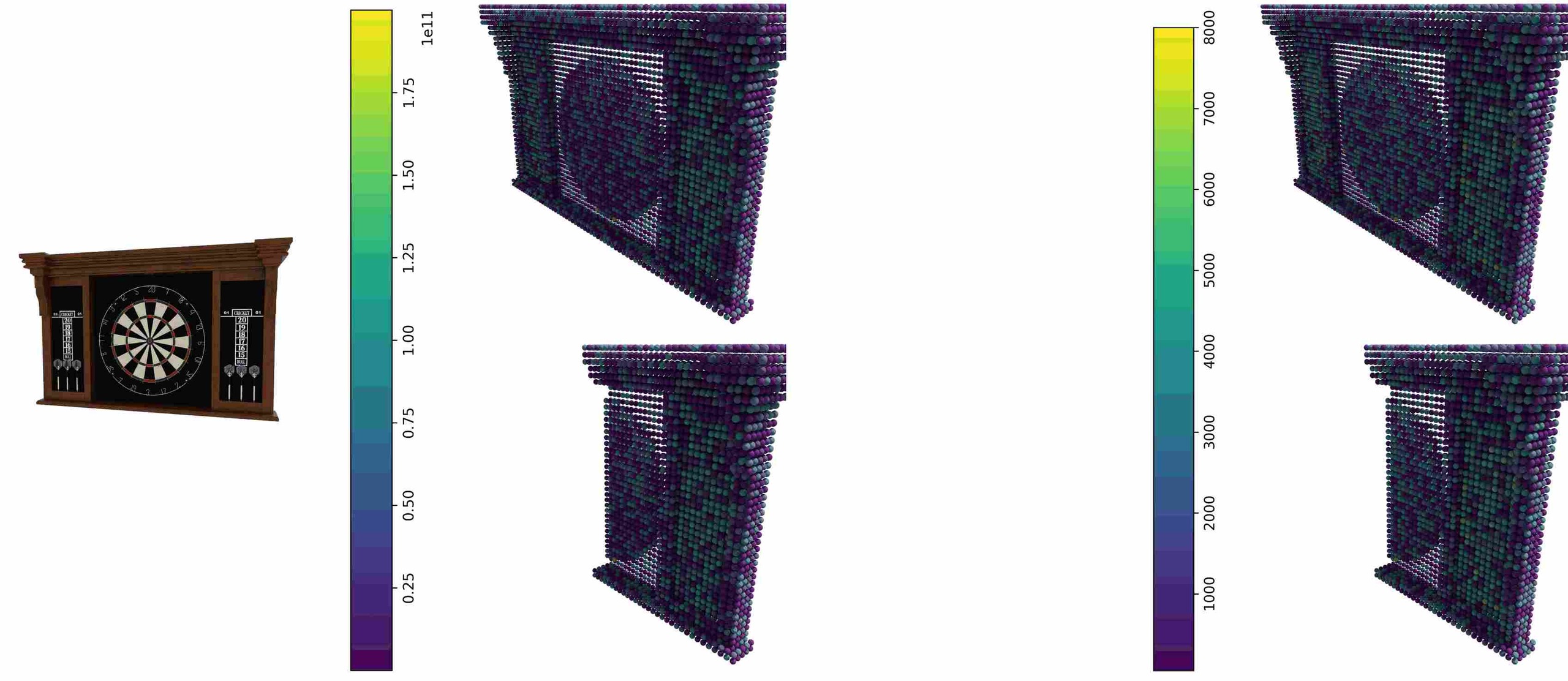} & 
    \includegraphics[width=0.5\textwidth]{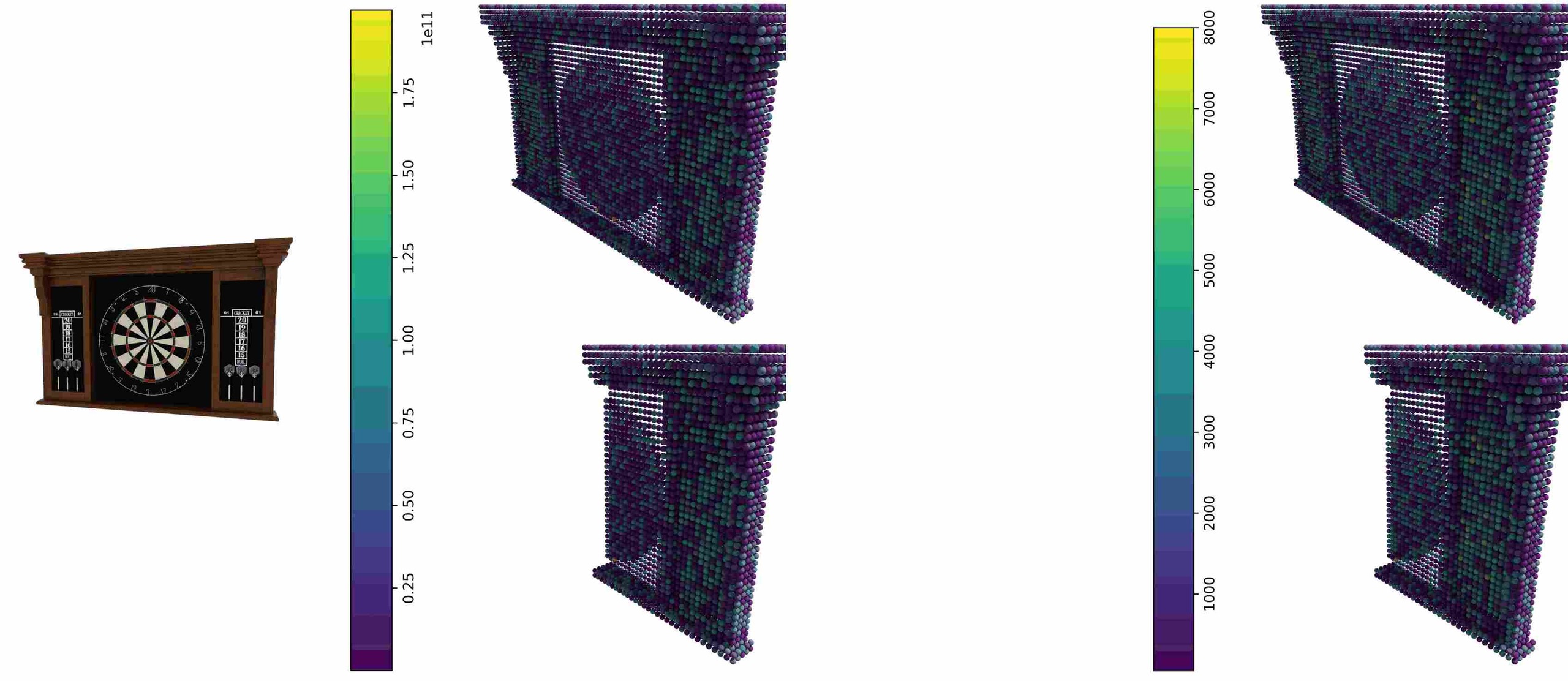} \\[2pt]
    \textbf{Phys4DGen} & \textbf{Pixie} \\[2pt]
    \includegraphics[width=0.5\textwidth]{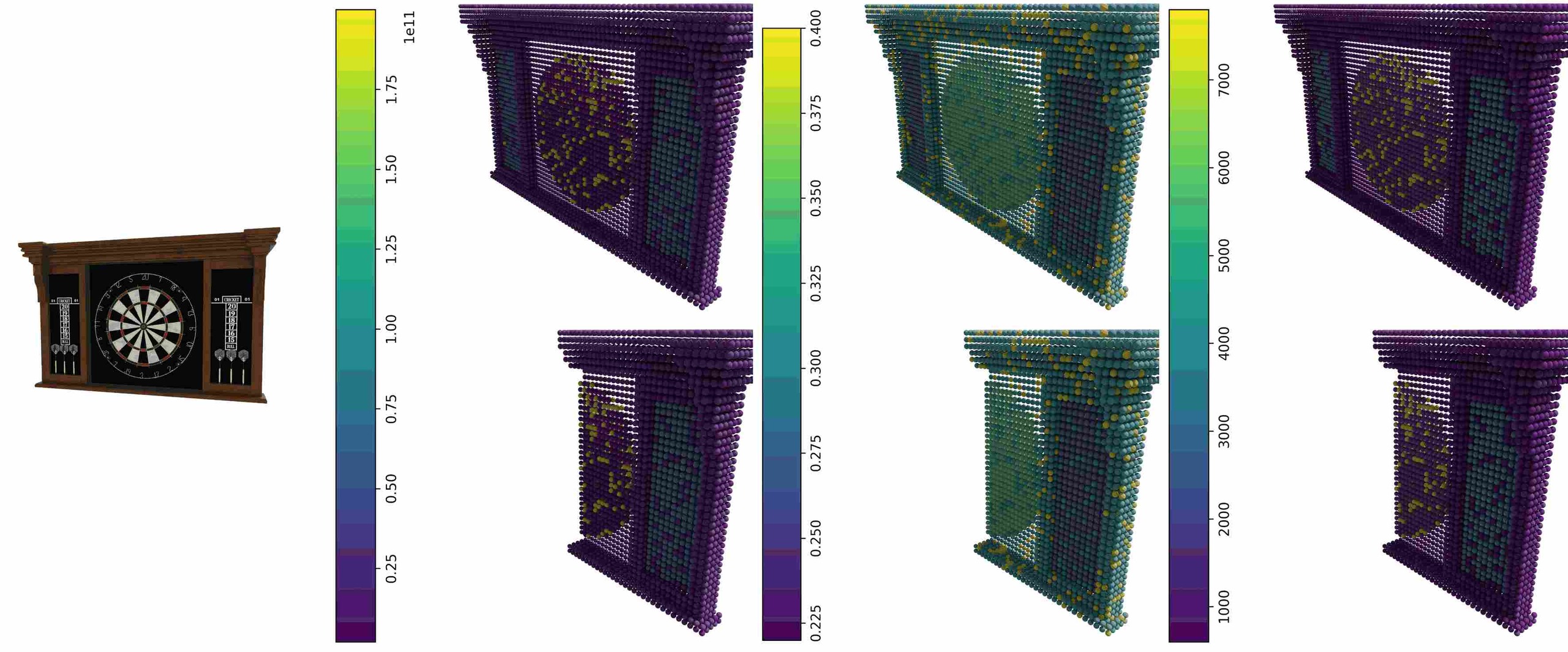} & 
    \includegraphics[width=0.5\textwidth]{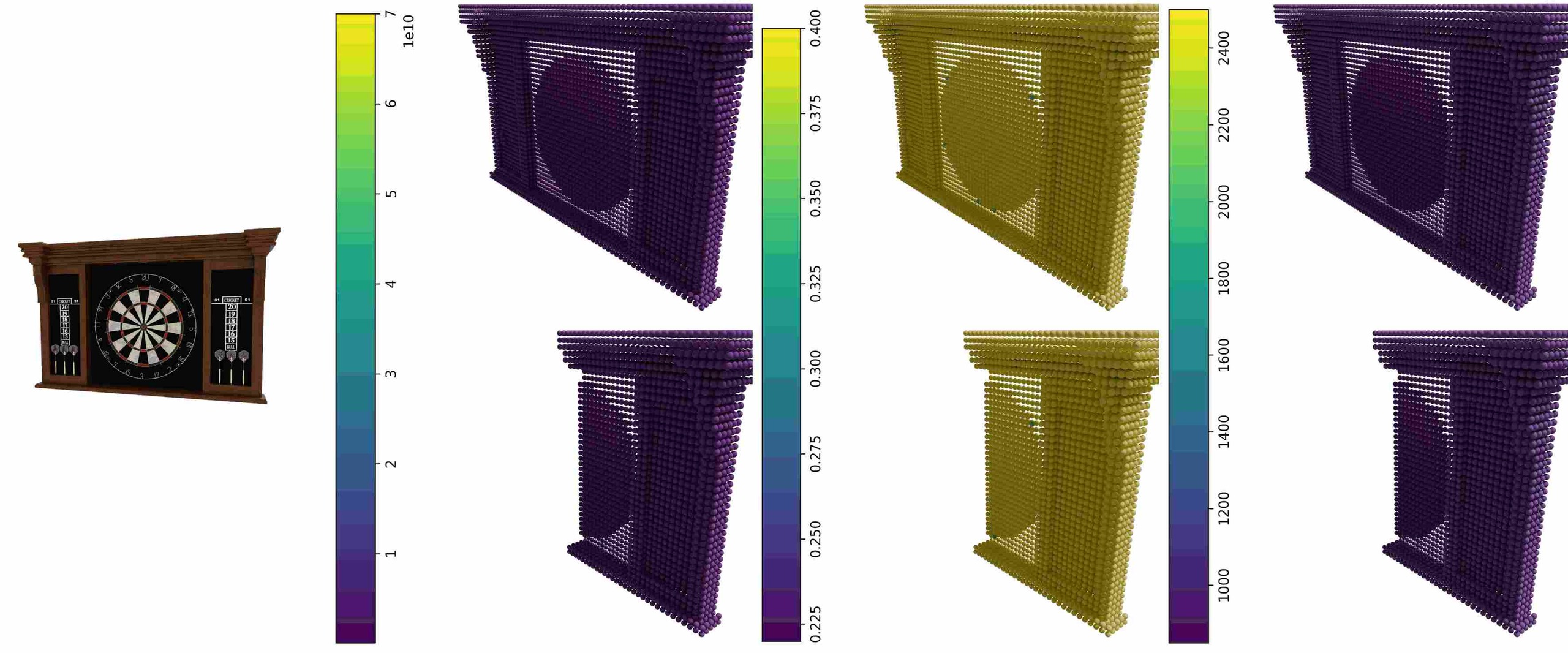} \\[2pt]
    \midrule
    \multicolumn{2}{c}{\textbf{Ground Truth}} \\[2pt]
    \multicolumn{2}{c}{\includegraphics[width=0.95\textwidth]{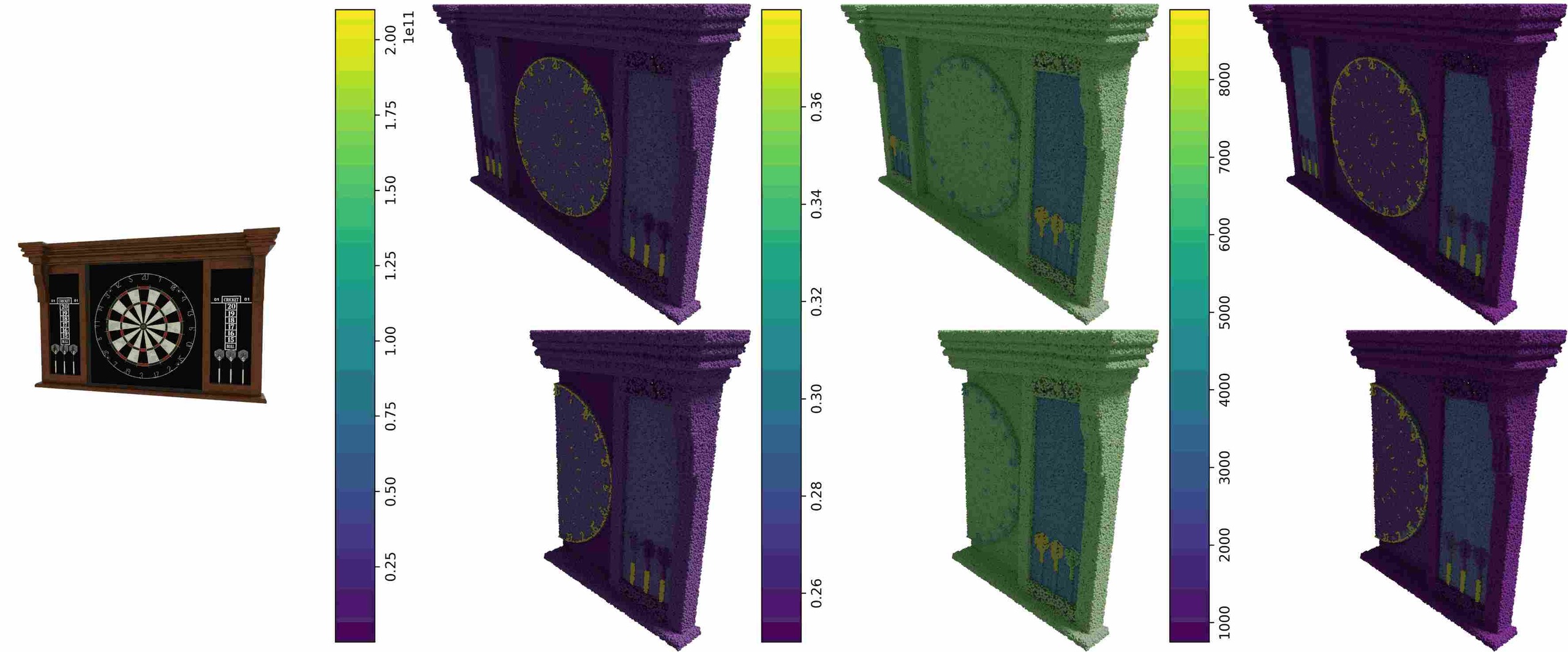}}
    \end{tabular}
    \caption{\textbf{Dartboard Comparison.} Mechanical property field comparisons across different methods.}
    \label{fig:dartboard_comparison}
\end{figure*}

\begin{figure*}[ht]
    \centering
    \setlength{\tabcolsep}{0pt}
    \renewcommand{\arraystretch}{0}
    \begin{tabular}{p{0.125\textwidth}p{0.125\textwidth}p{0.125\textwidth}p{0.125\textwidth}|p{0.125\textwidth}p{0.125\textwidth}p{0.125\textwidth}p{0.125\textwidth}}
    \centering Object & \centering Young's Modulus ($E$, Pa) & \centering Poisson's Ratio ($\nu$) & \centering Density ($\rho, \frac{kg}{m^3}$) & \centering Object & \centering Young's Modulus ($E$, Pa) & \centering Poisson's Ratio ($\nu$) & \centering Density ($\rho, \frac{kg}{m^3}$) \\
    \end{tabular}
    \begin{tabular}{@{}c@{\hspace{10pt}}c@{}}
    \textbf{\ourmodel} & \textbf{VoMP} \\[2pt]
    \includegraphics[width=0.5\textwidth]{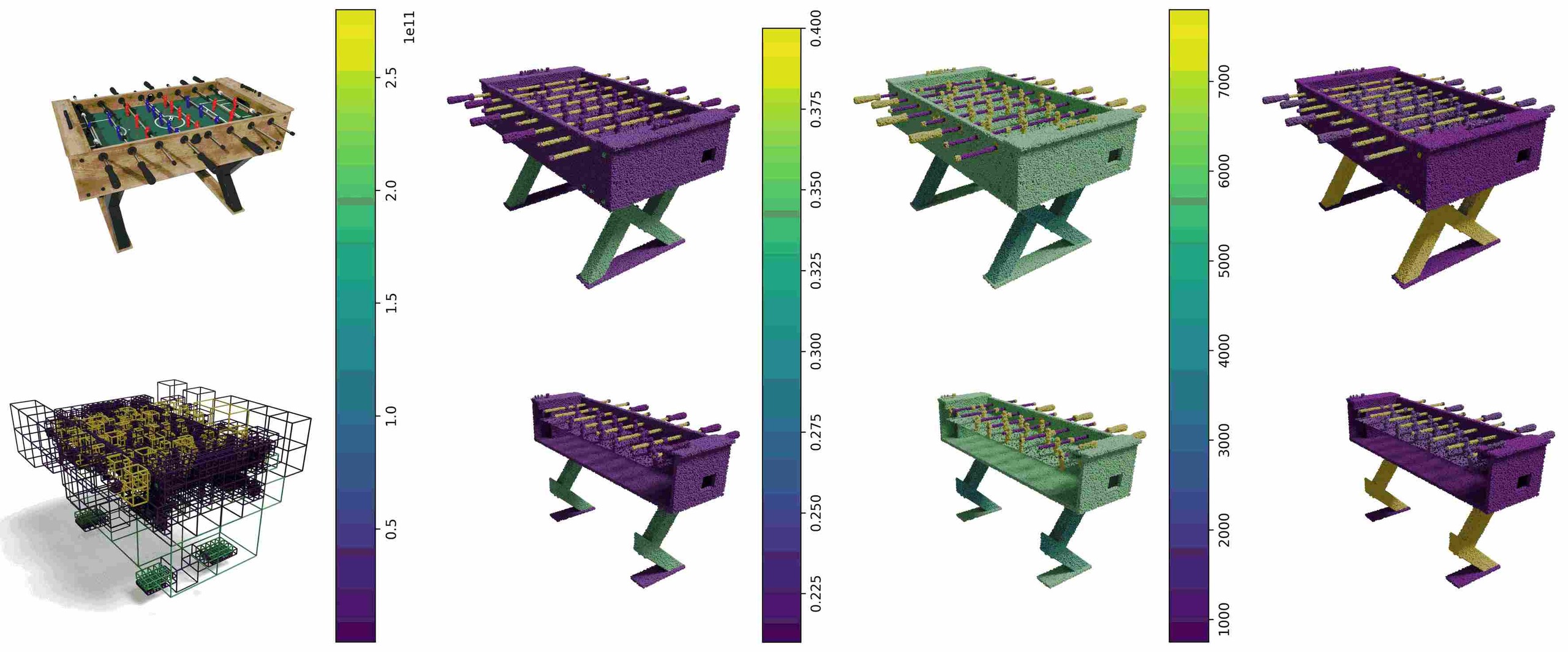} & 
    \includegraphics[width=0.5\textwidth]{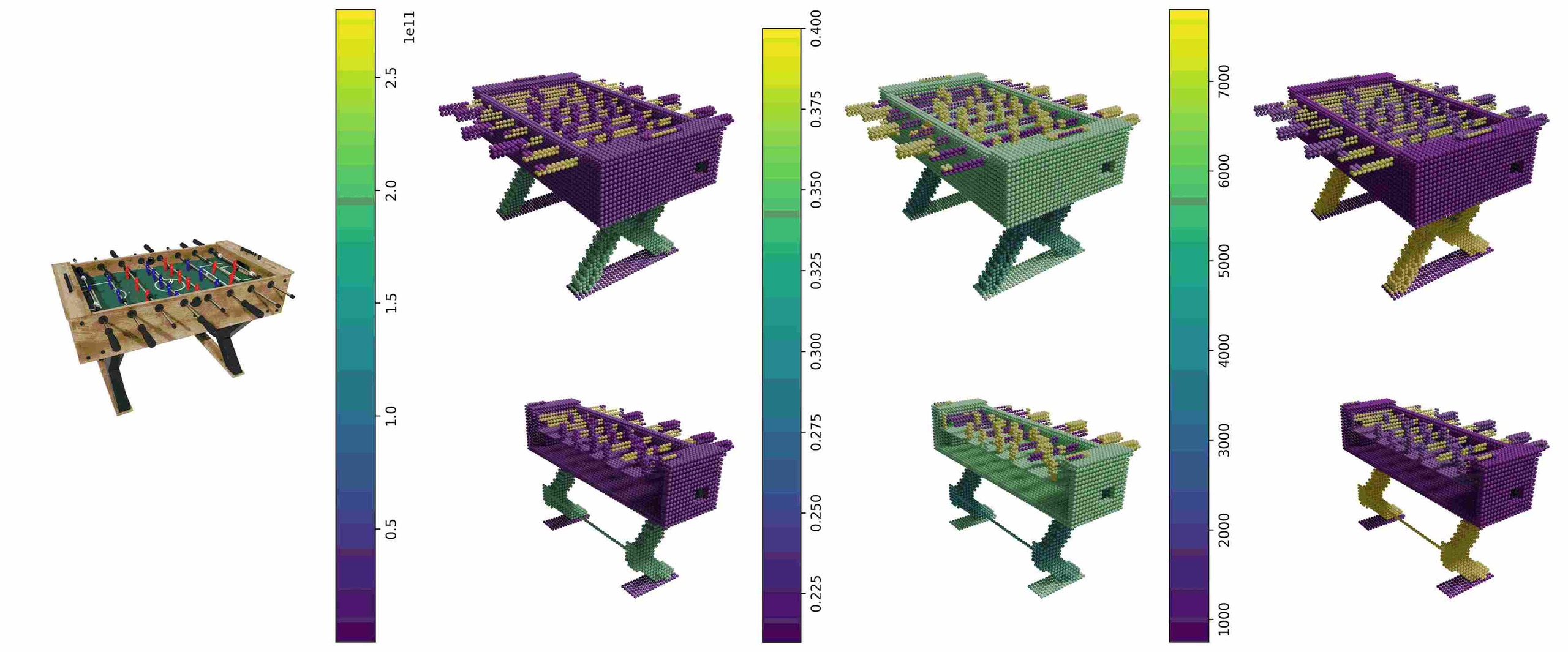} \\[2pt]
    \textbf{N2P} & \textbf{PUGS} \\[2pt]
    \includegraphics[width=0.5\textwidth]{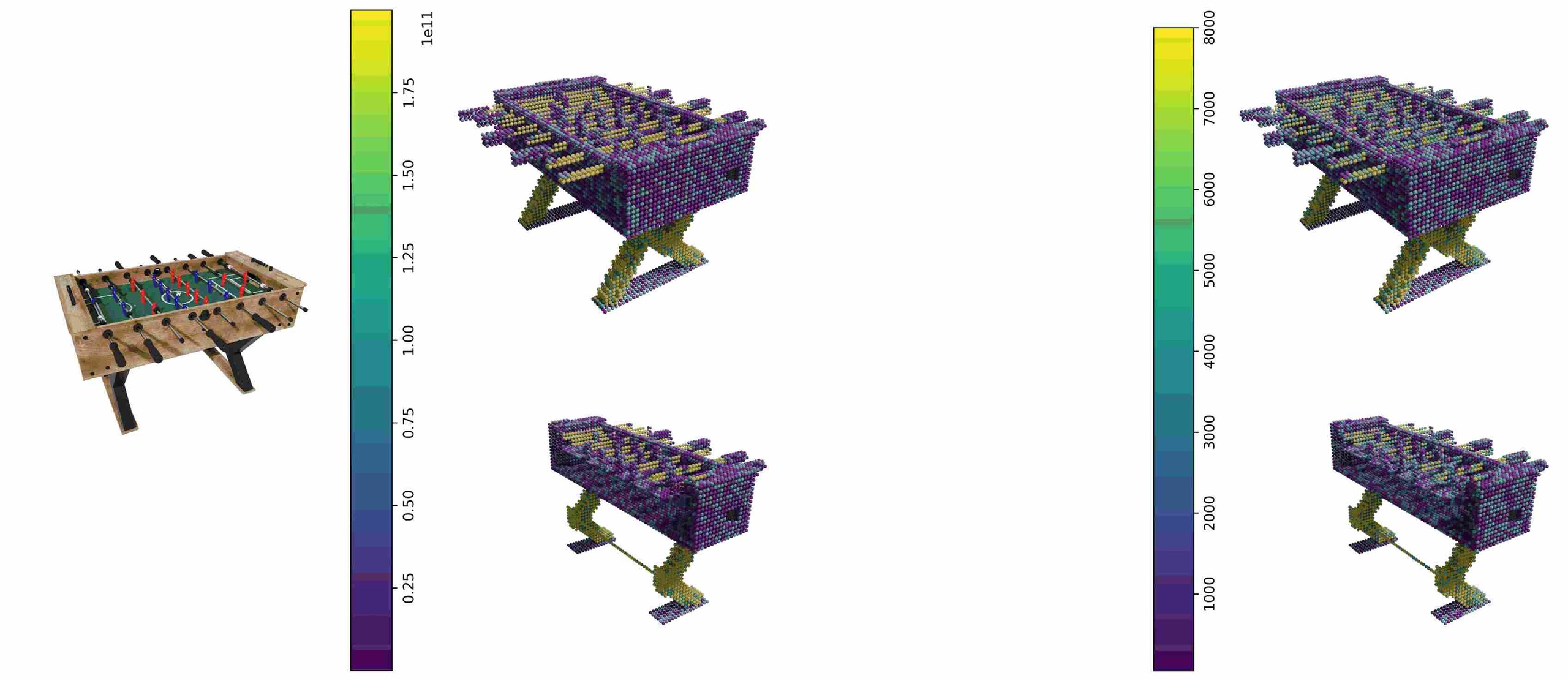} & 
    \includegraphics[width=0.5\textwidth]{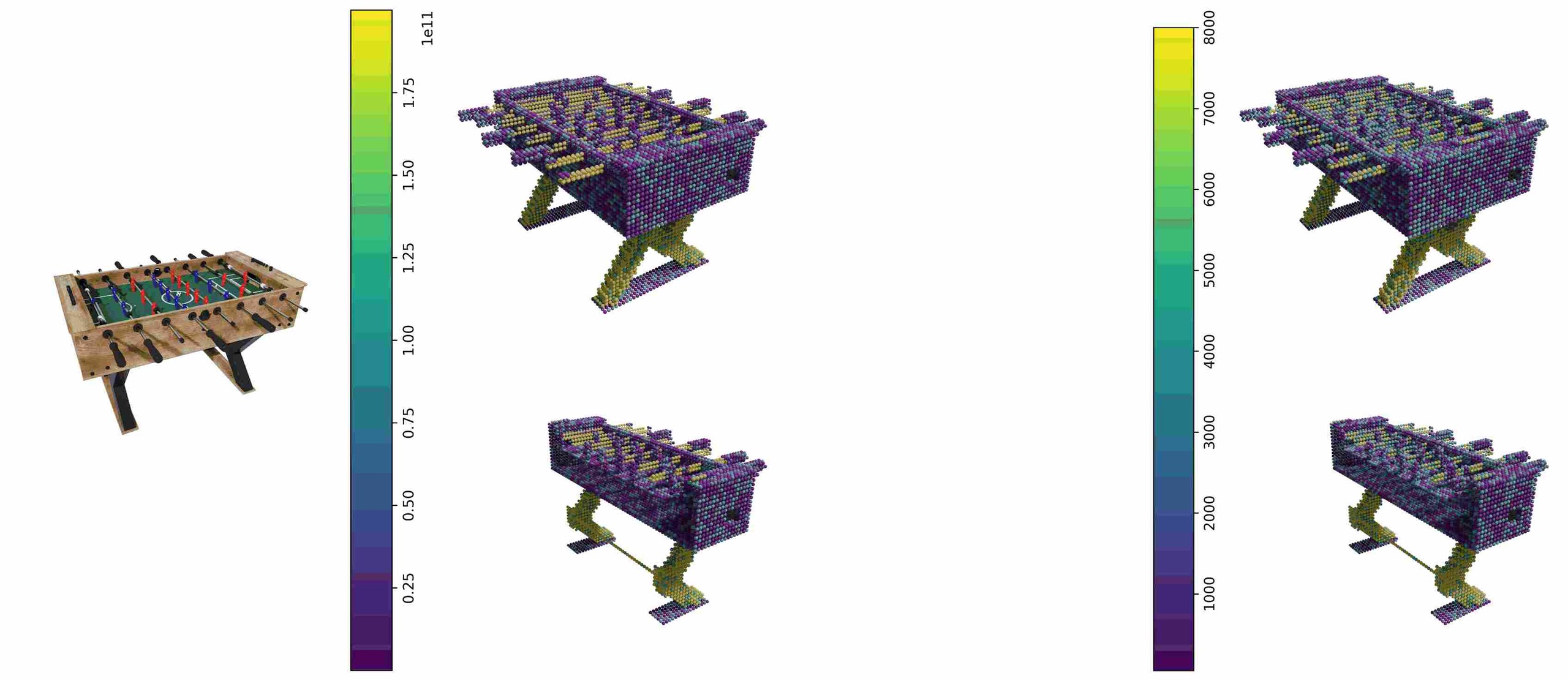} \\[2pt]
    \textbf{Phys4DGen} & \textbf{Pixie} \\[2pt]
    \includegraphics[width=0.5\textwidth]{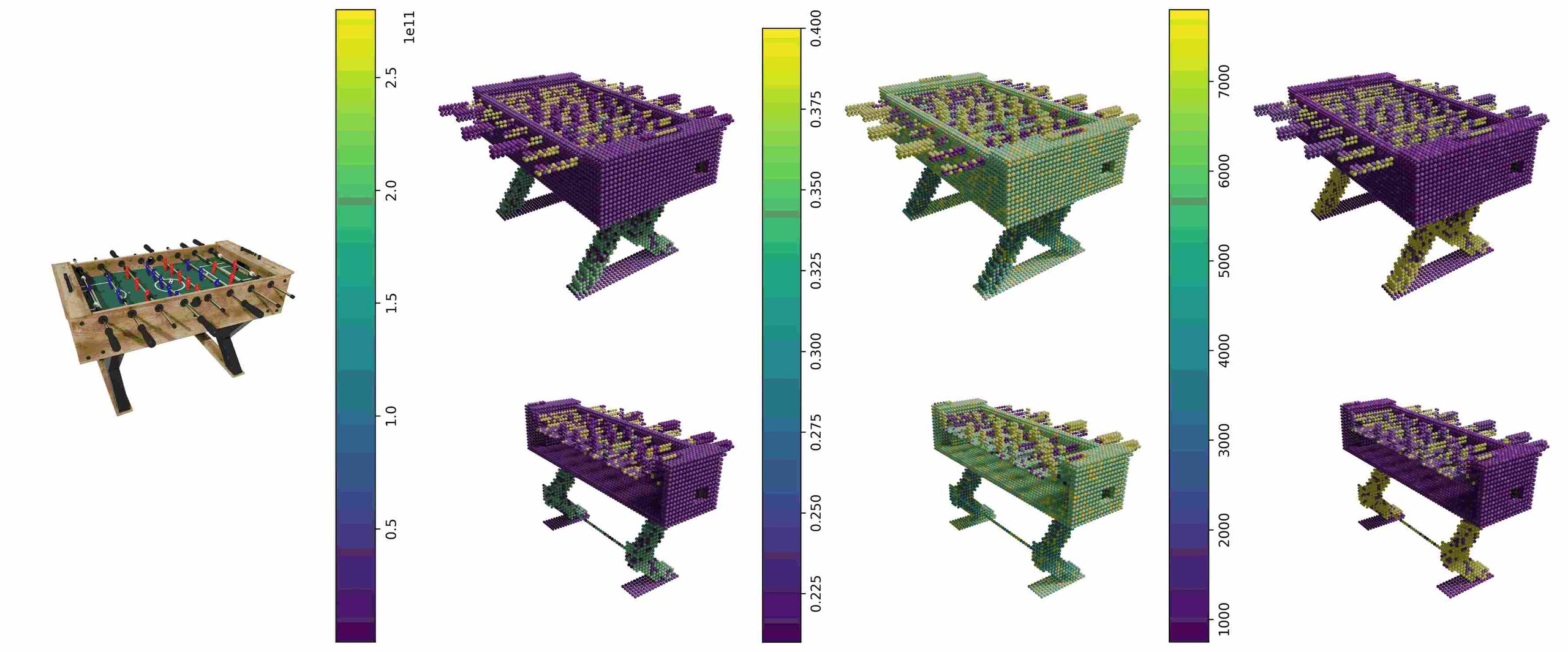} & 
    \includegraphics[width=0.5\textwidth]{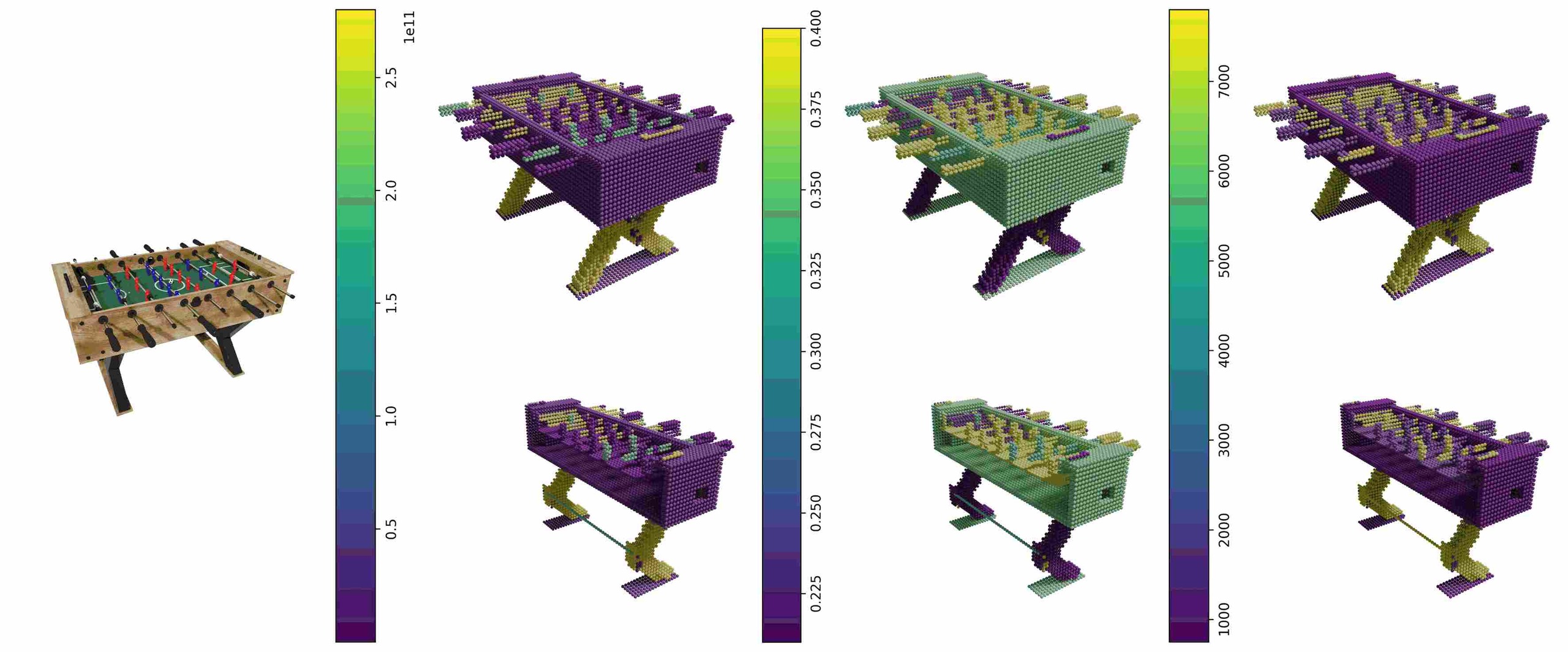} \\[2pt]
    \midrule
    \multicolumn{2}{c}{\textbf{Ground Truth}} \\[2pt]
    \multicolumn{2}{c}{\includegraphics[width=0.95\textwidth]{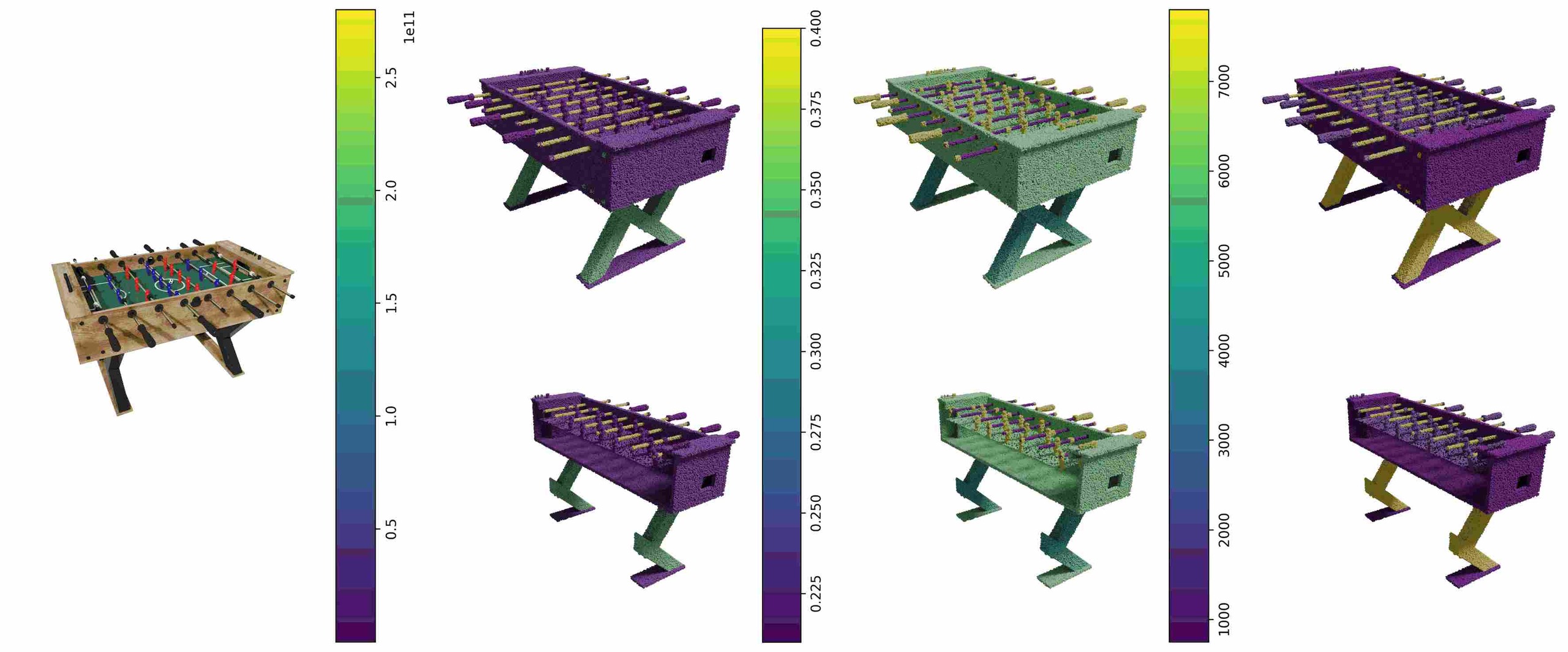}}
    \end{tabular}
    \caption{\textbf{Foosball Comparison.} Mechanical property field comparisons across different methods.}
    \label{fig:foosball_comparison}
\end{figure*}

\begin{figure*}[ht]
    \centering
    \setlength{\tabcolsep}{0pt}
    \renewcommand{\arraystretch}{0}
    \begin{tabular}{p{0.125\textwidth}p{0.125\textwidth}p{0.125\textwidth}p{0.125\textwidth}|p{0.125\textwidth}p{0.125\textwidth}p{0.125\textwidth}p{0.125\textwidth}}
    \centering Object & \centering Young's Modulus ($E$, Pa) & \centering Poisson's Ratio ($\nu$) & \centering Density ($\rho, \frac{kg}{m^3}$) & \centering Object & \centering Young's Modulus ($E$, Pa) & \centering Poisson's Ratio ($\nu$) & \centering Density ($\rho, \frac{kg}{m^3}$) \\
    \end{tabular}
    \begin{tabular}{@{}c@{\hspace{10pt}}c@{}}
    \textbf{\ourmodel} & \textbf{VoMP} \\[2pt]
    \includegraphics[width=0.5\textwidth]{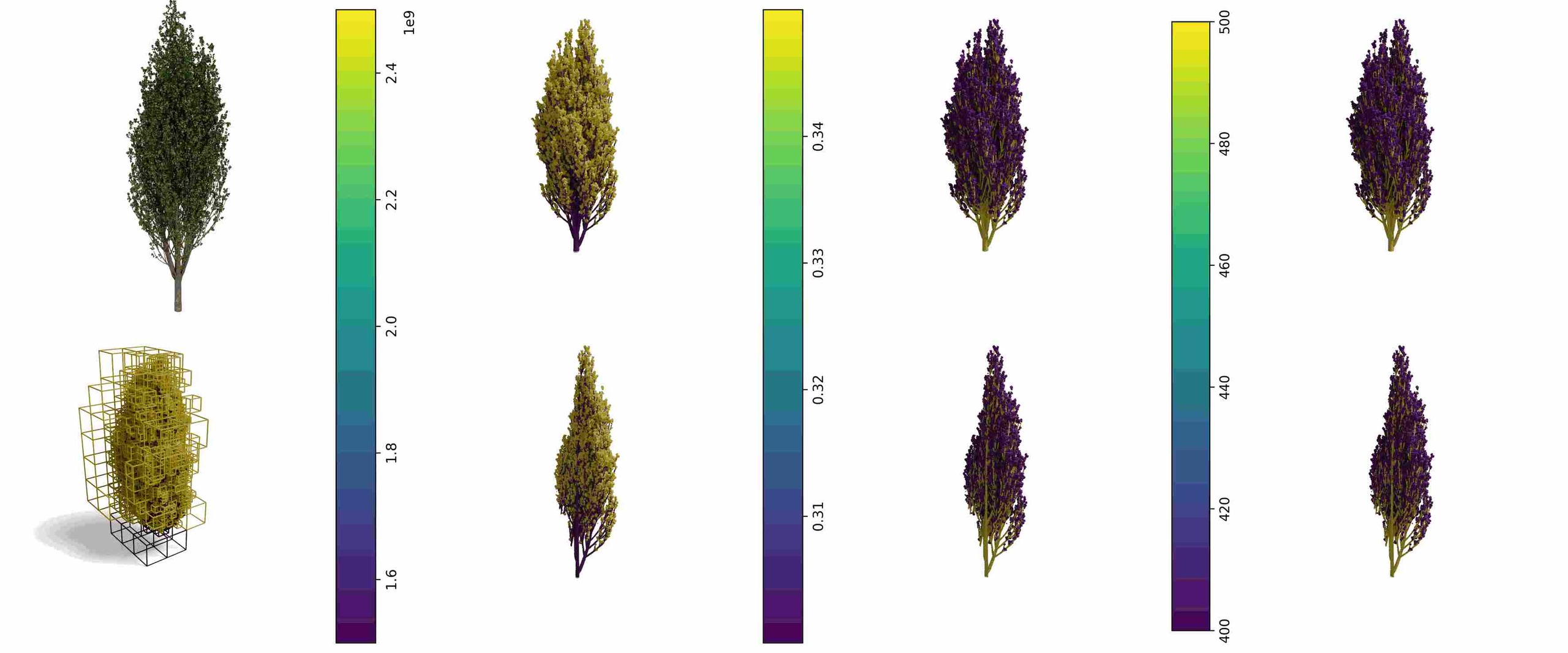} & 
    \includegraphics[width=0.5\textwidth]{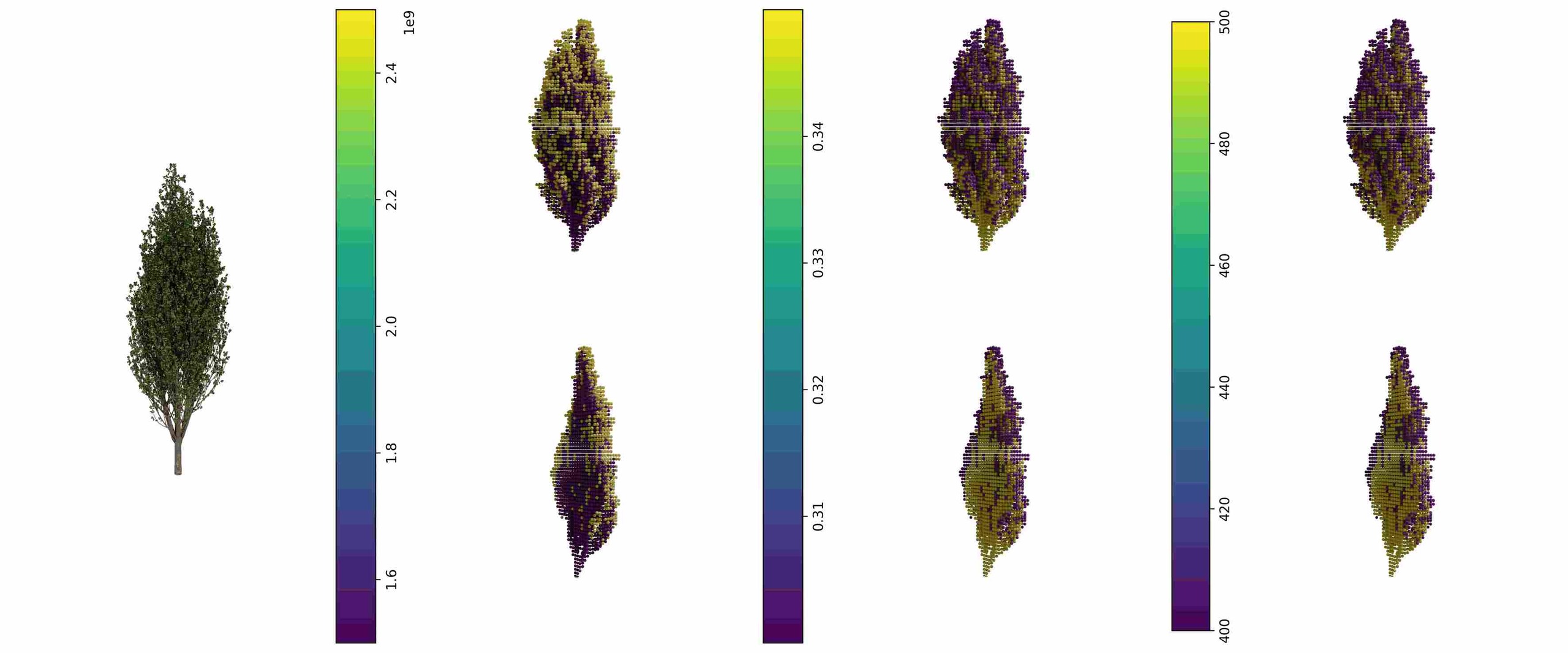} \\[2pt]
    \textbf{N2P} & \textbf{PUGS} \\[2pt]
    \includegraphics[width=0.5\textwidth]{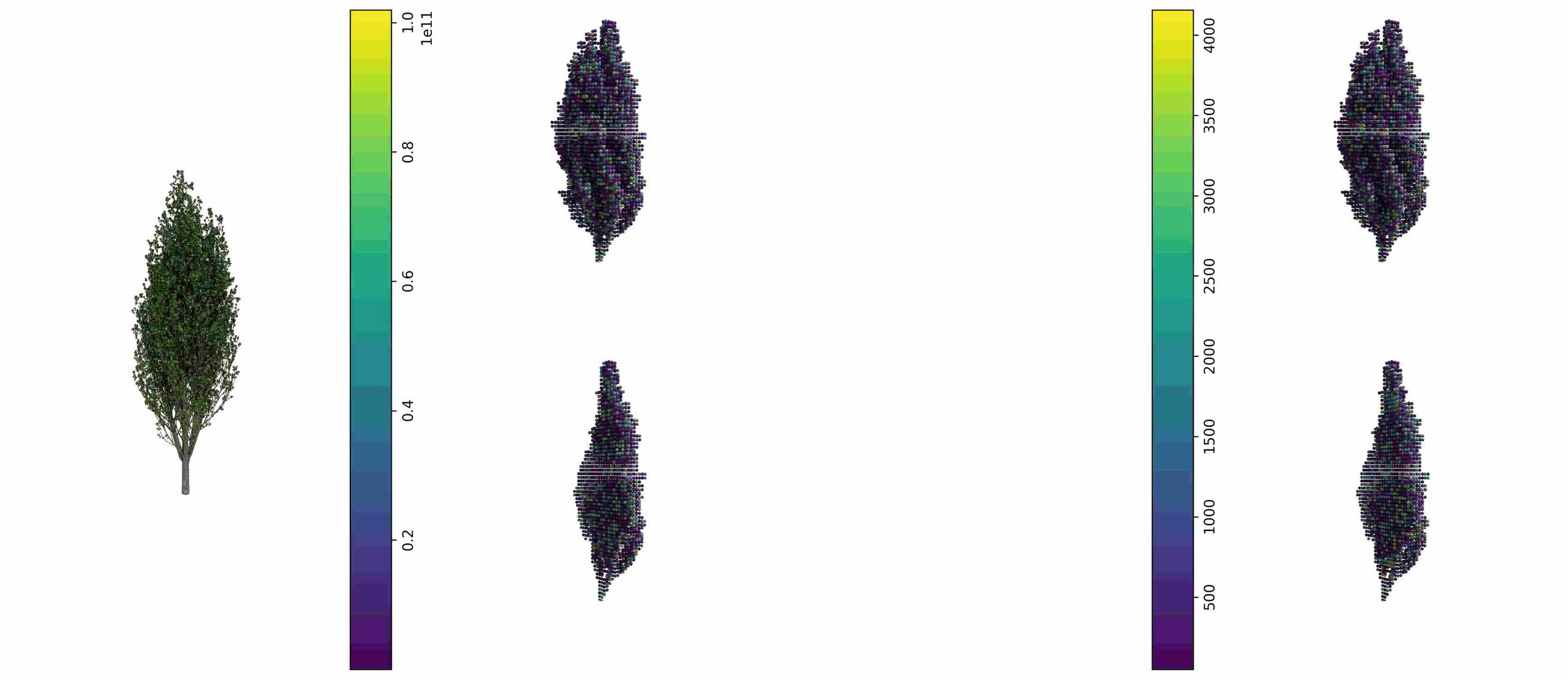} & 
    \includegraphics[width=0.5\textwidth]{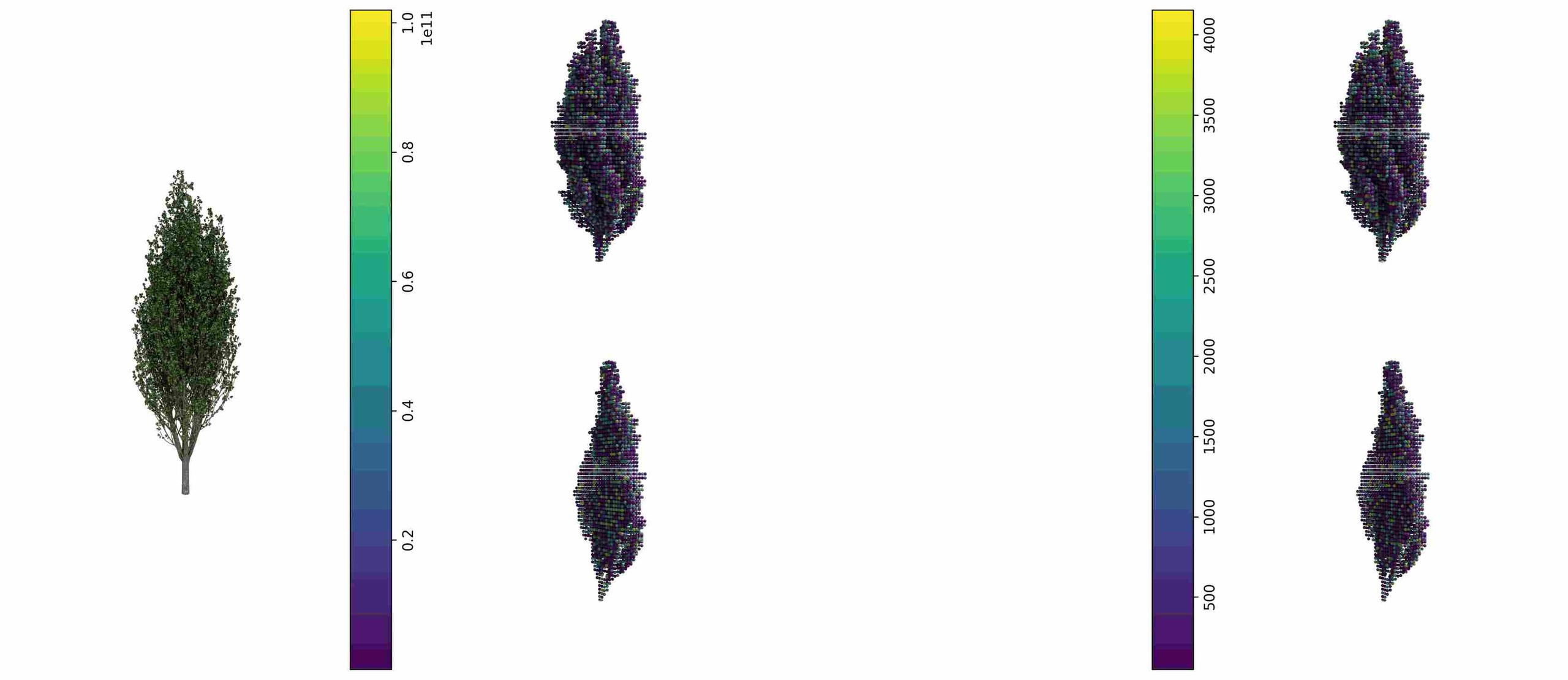} \\[2pt]
    \textbf{Phys4DGen} & \textbf{Pixie} \\[2pt]
    \includegraphics[width=0.5\textwidth]{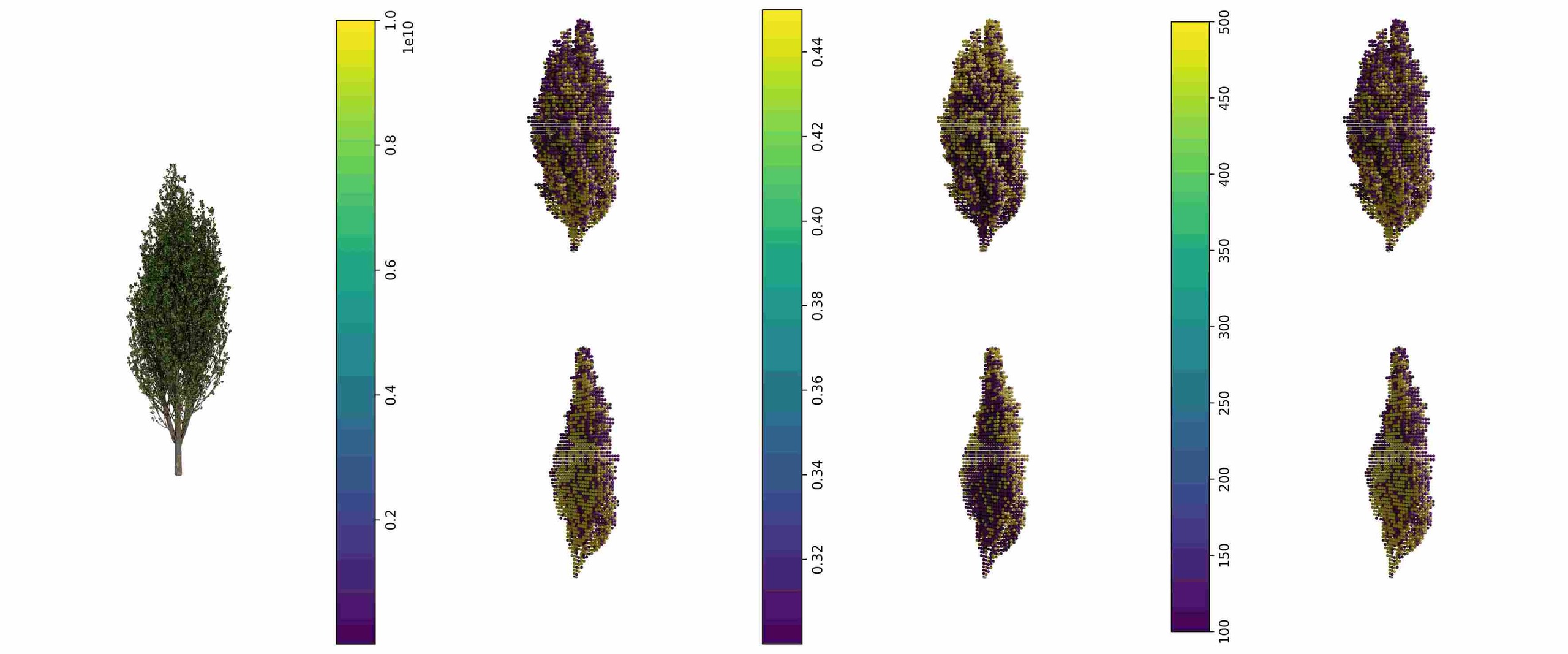} & 
    \includegraphics[width=0.5\textwidth]{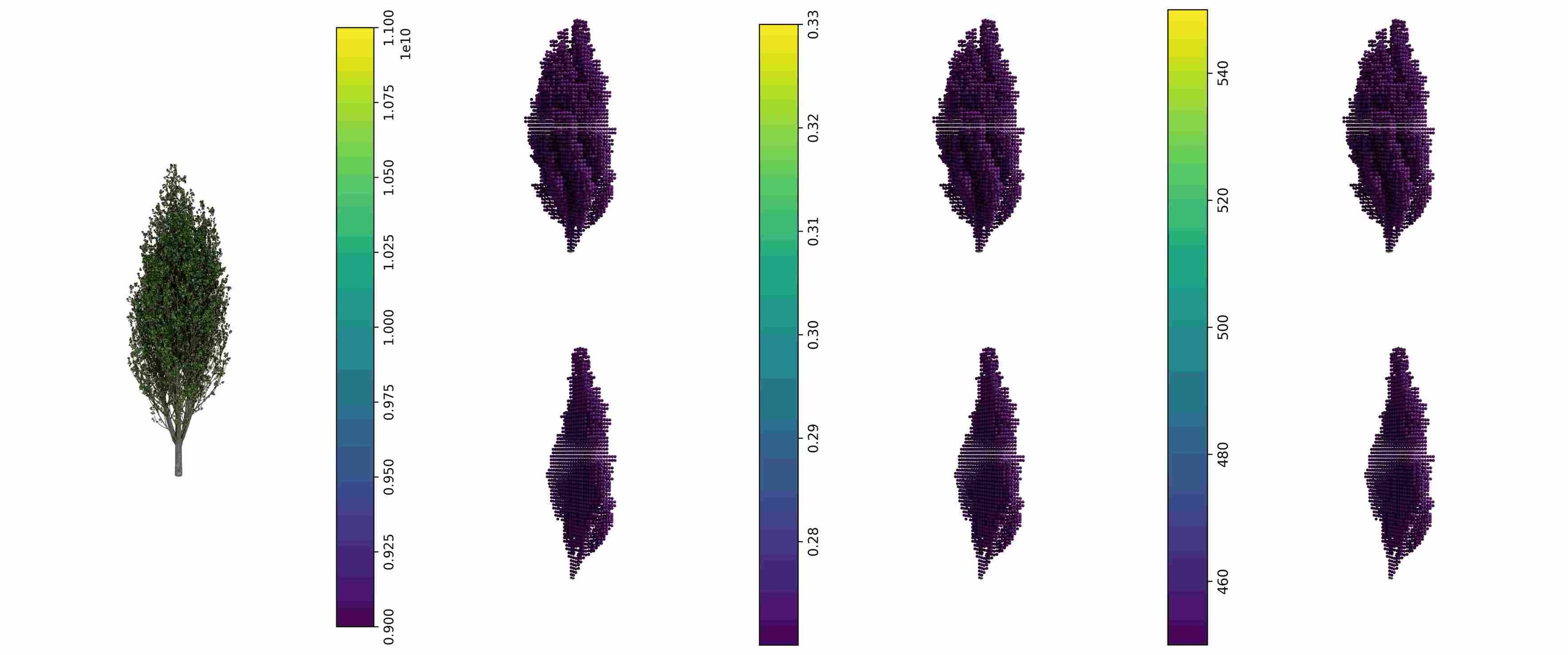} \\[2pt]
    \midrule
    \multicolumn{2}{c}{\textbf{Ground Truth}} \\[2pt]
    \multicolumn{2}{c}{\includegraphics[width=0.95\textwidth]{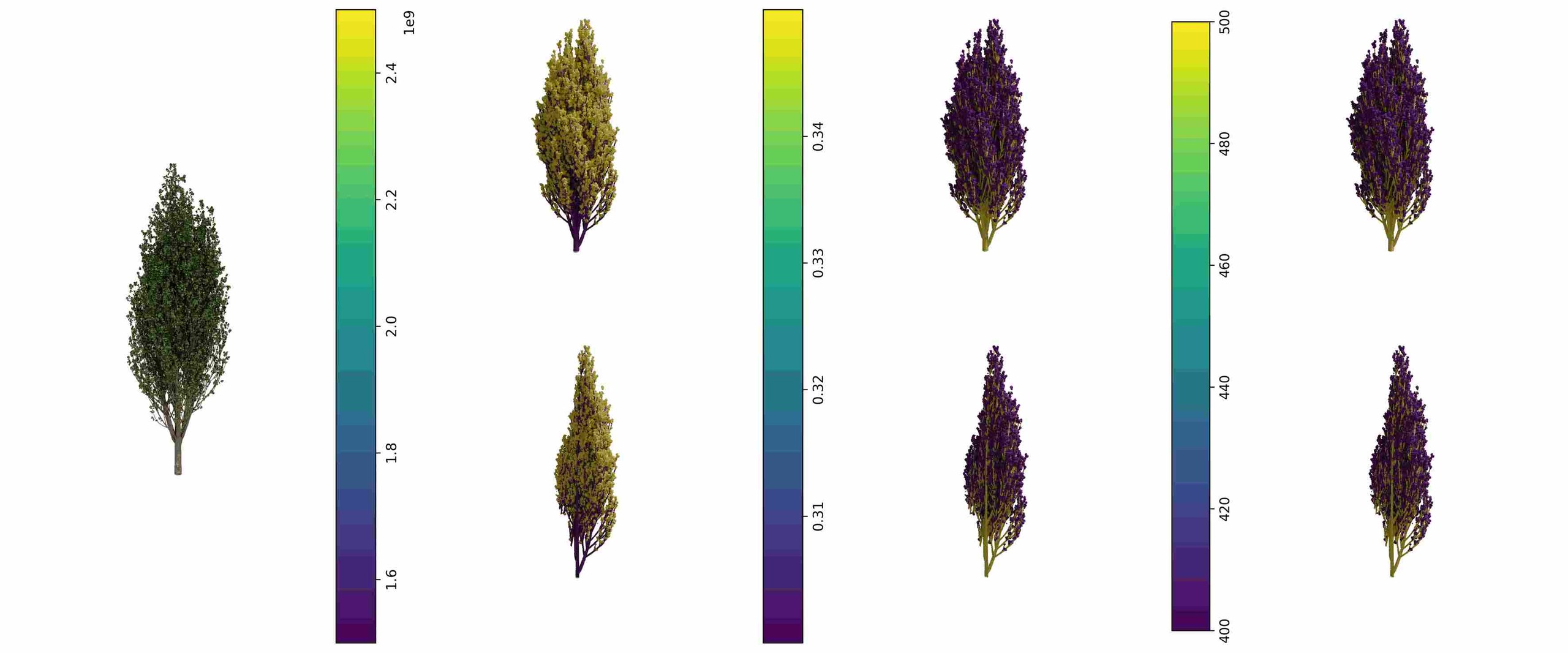}}
    \end{tabular}
    \caption{\textbf{Lombardy Poplar Comparison.} Mechanical property field comparisons across different methods.}
    \label{fig:lombardy_poplar_comparison}
\end{figure*}

\begin{figure*}[ht]
    \centering
    \setlength{\tabcolsep}{0pt}
    \renewcommand{\arraystretch}{0}
    \begin{tabular}{p{0.125\textwidth}p{0.125\textwidth}p{0.125\textwidth}p{0.125\textwidth}|p{0.125\textwidth}p{0.125\textwidth}p{0.125\textwidth}p{0.125\textwidth}}
    \centering Object & \centering Young's Modulus ($E$, Pa) & \centering Poisson's Ratio ($\nu$) & \centering Density ($\rho, \frac{kg}{m^3}$) & \centering Object & \centering Young's Modulus ($E$, Pa) & \centering Poisson's Ratio ($\nu$) & \centering Density ($\rho, \frac{kg}{m^3}$) \\
    \end{tabular}
    \begin{tabular}{@{}c@{\hspace{10pt}}c@{}}
    \textbf{\ourmodel} & \textbf{VoMP} \\[2pt]
    \includegraphics[width=0.5\textwidth]{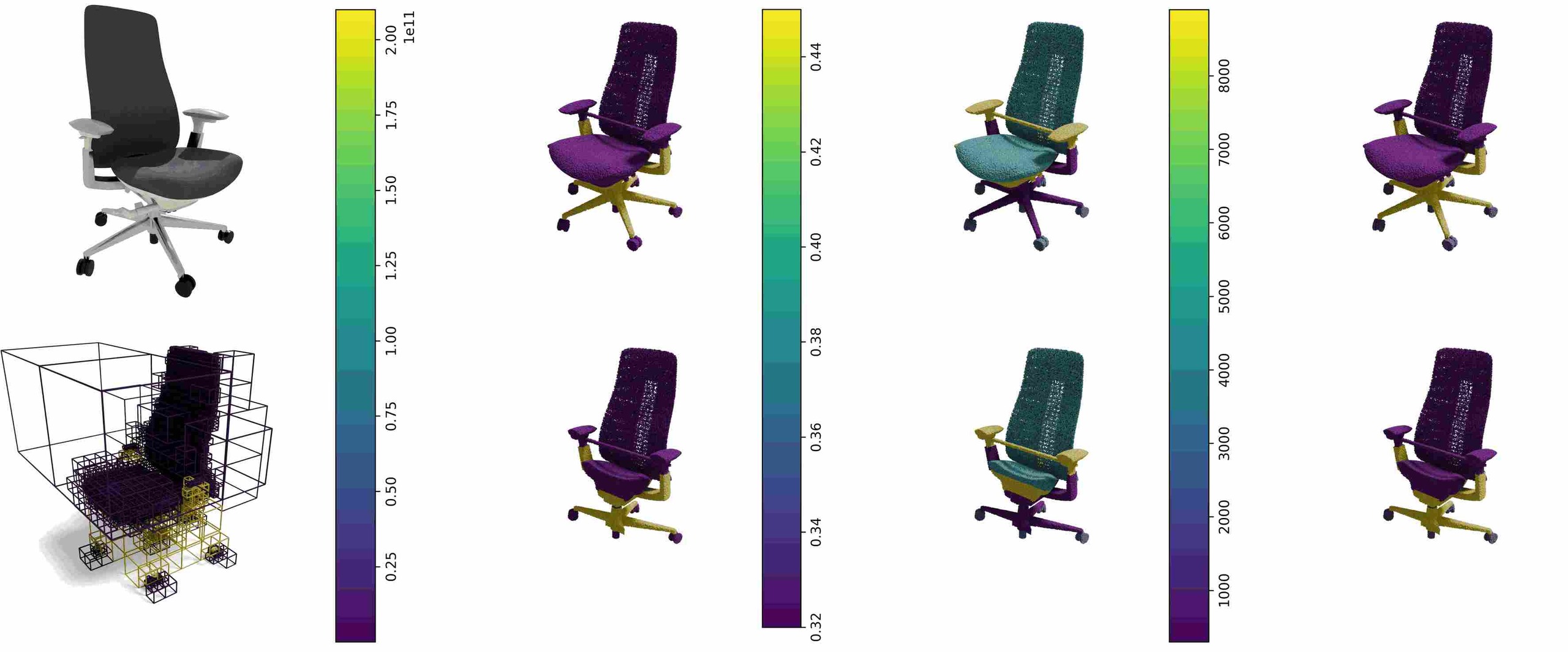} & 
    \includegraphics[width=0.5\textwidth]{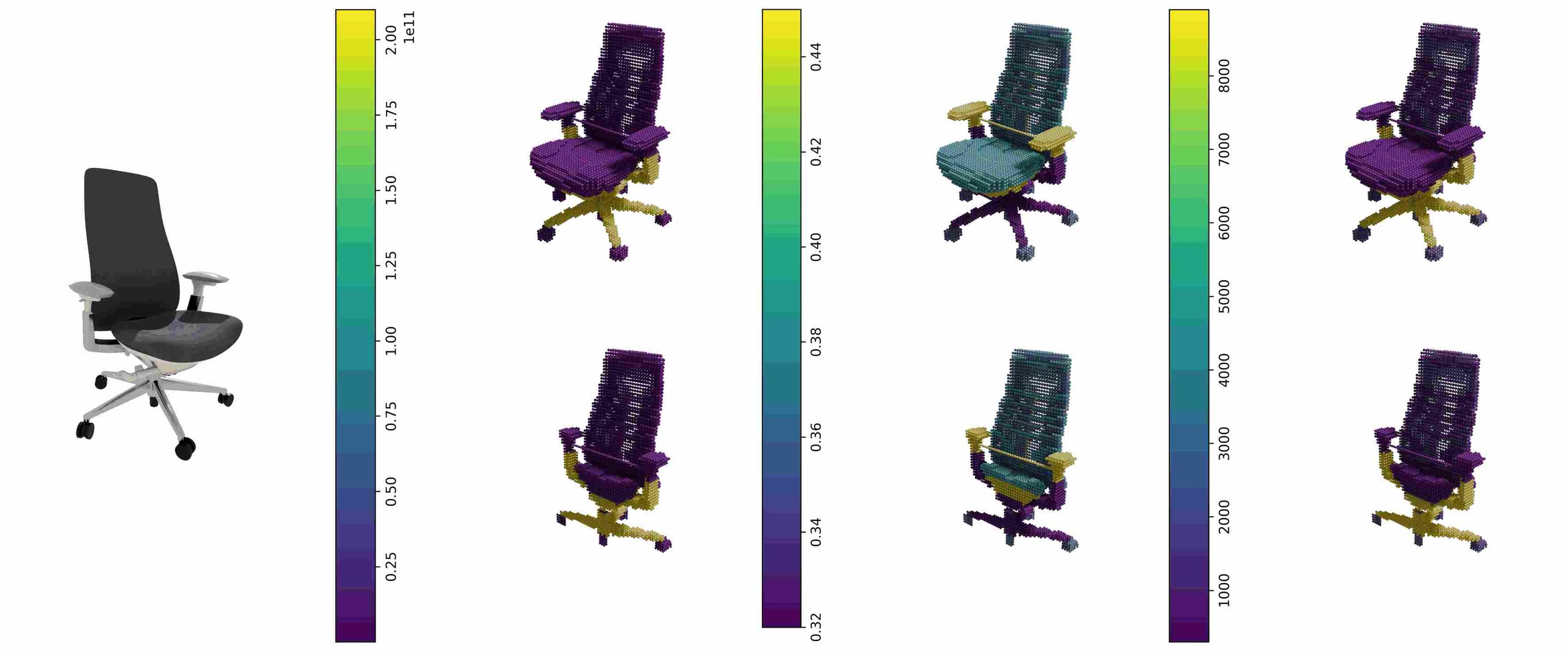} \\[2pt]
    \textbf{N2P} & \textbf{PUGS} \\[2pt]
    \includegraphics[width=0.5\textwidth]{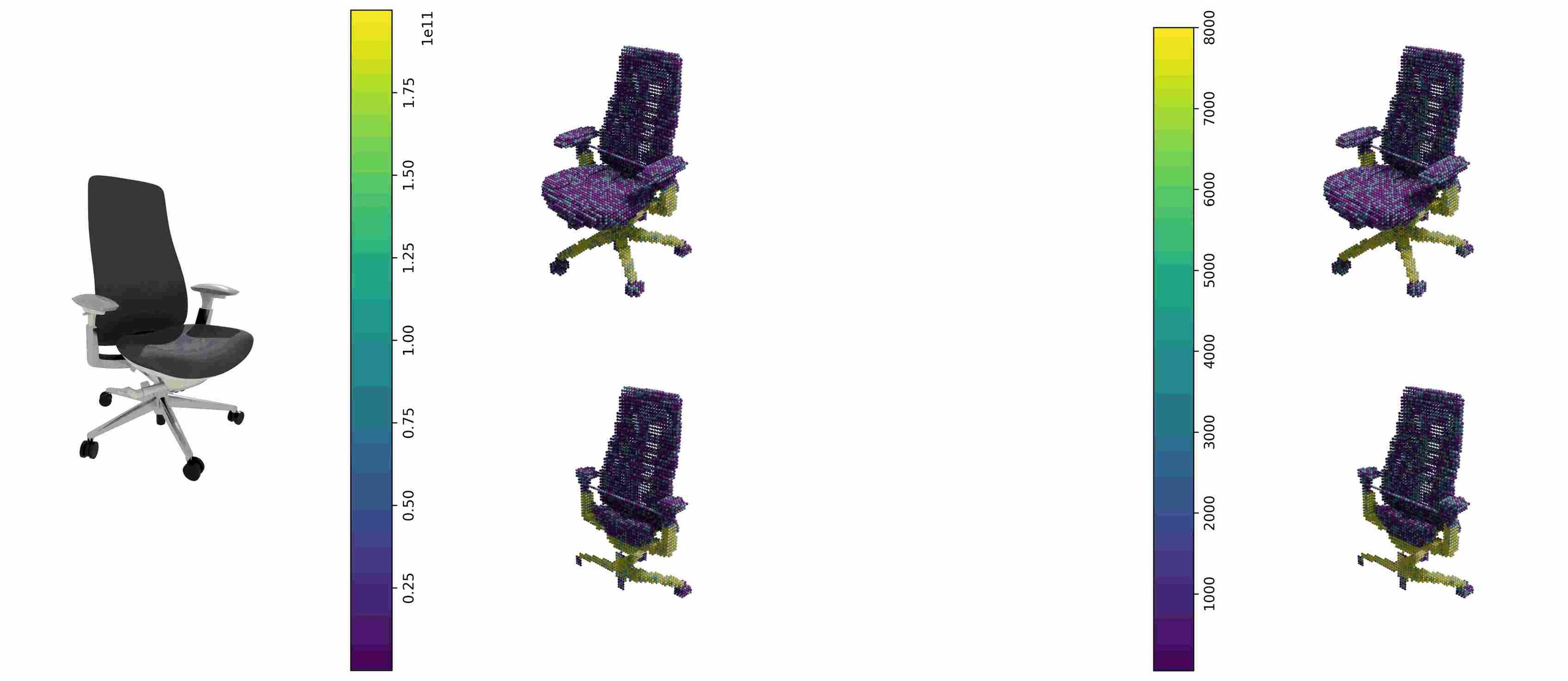} & 
    \includegraphics[width=0.5\textwidth]{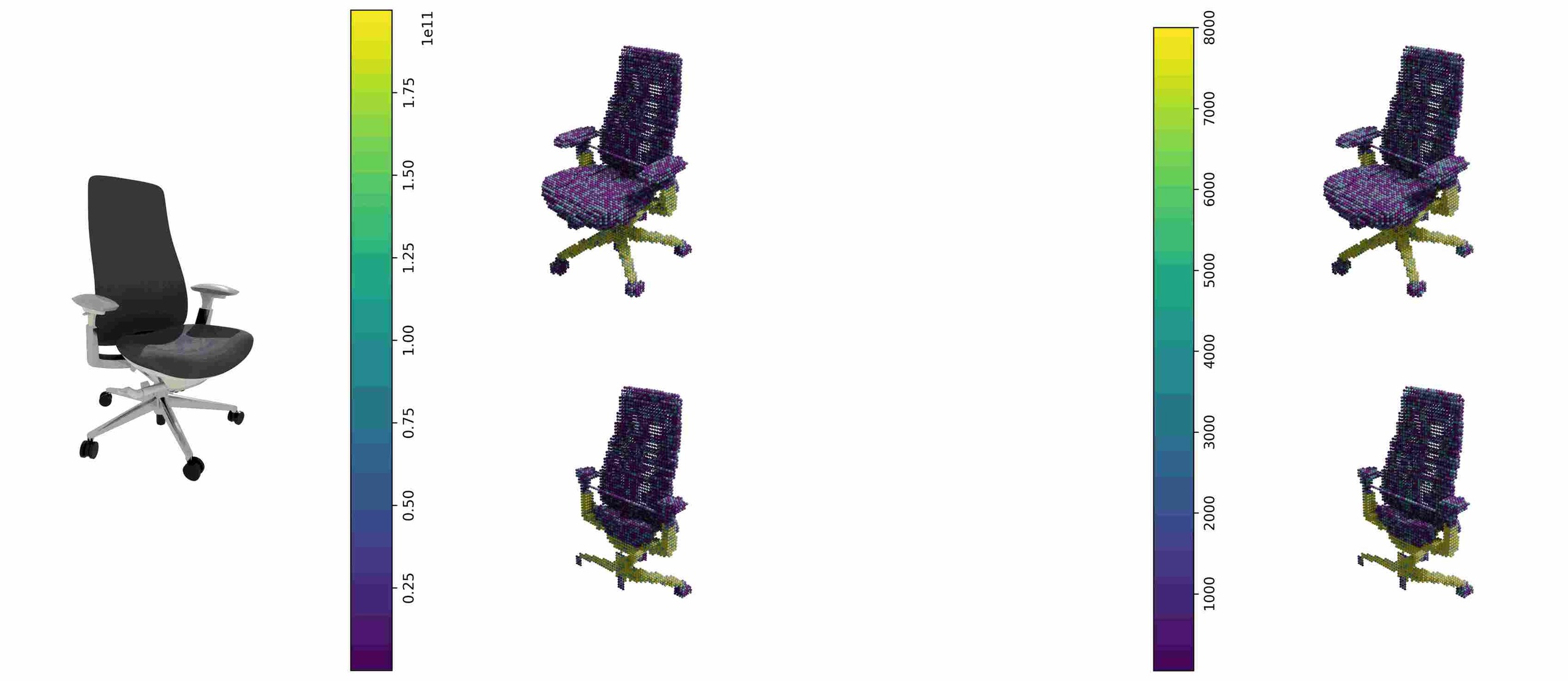} \\[2pt]
    \textbf{Phys4DGen} & \textbf{Pixie} \\[2pt]
    \includegraphics[width=0.5\textwidth]{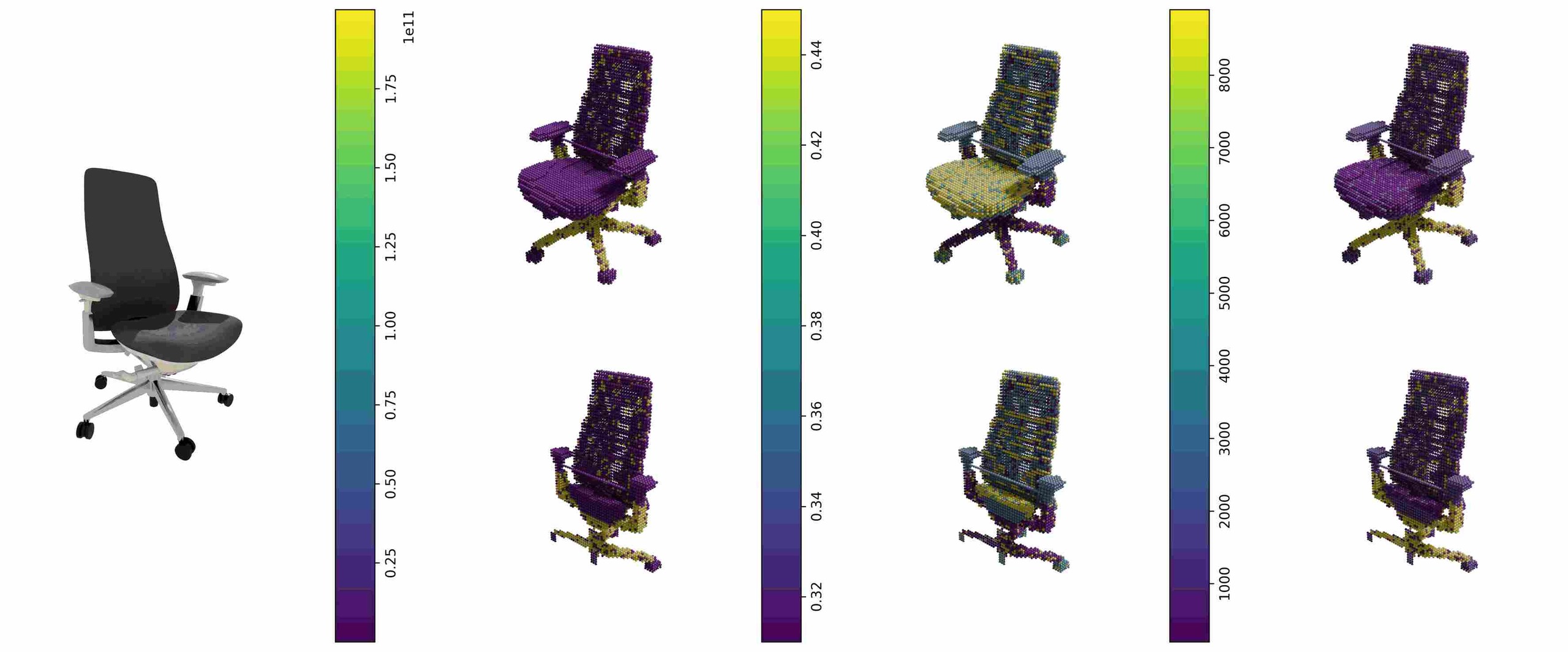} & 
    \includegraphics[width=0.5\textwidth]{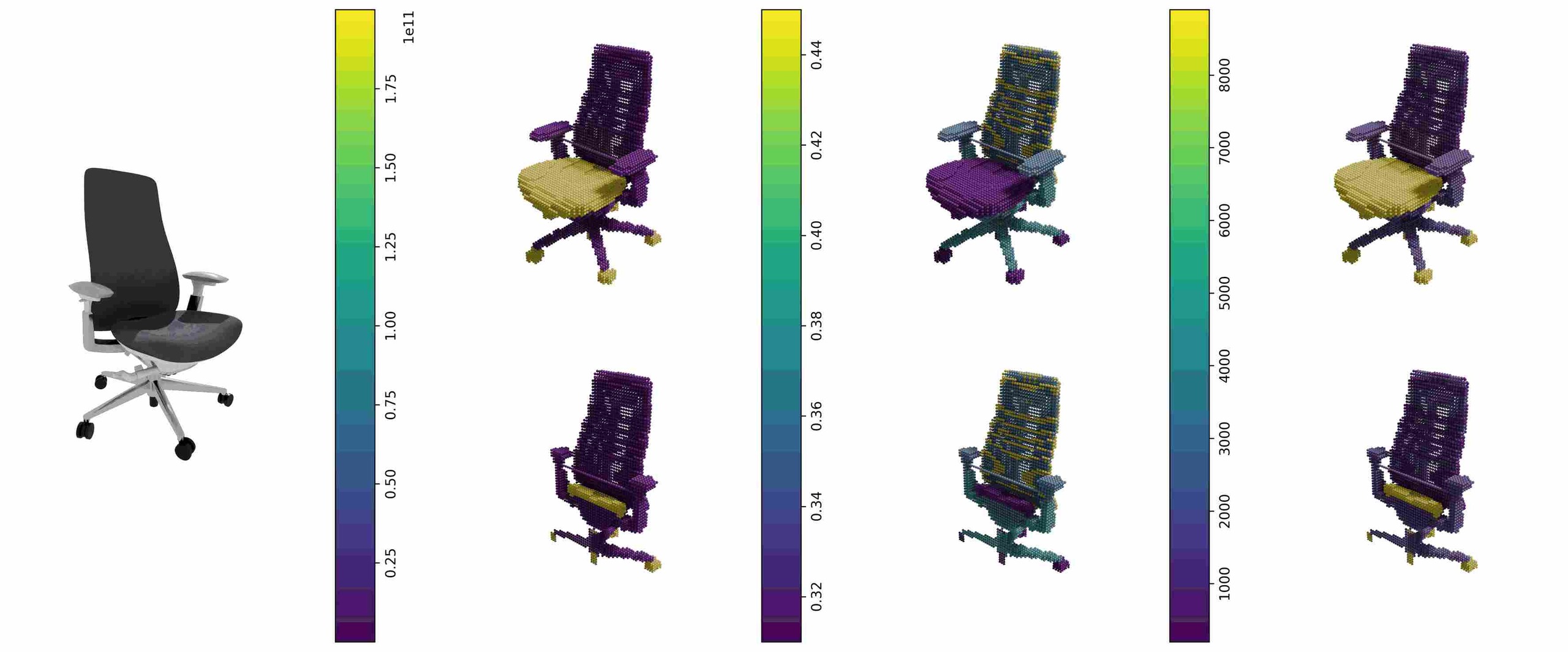} \\[2pt]
    \midrule
    \multicolumn{2}{c}{\textbf{Ground Truth}} \\[2pt]
    \multicolumn{2}{c}{\includegraphics[width=0.95\textwidth]{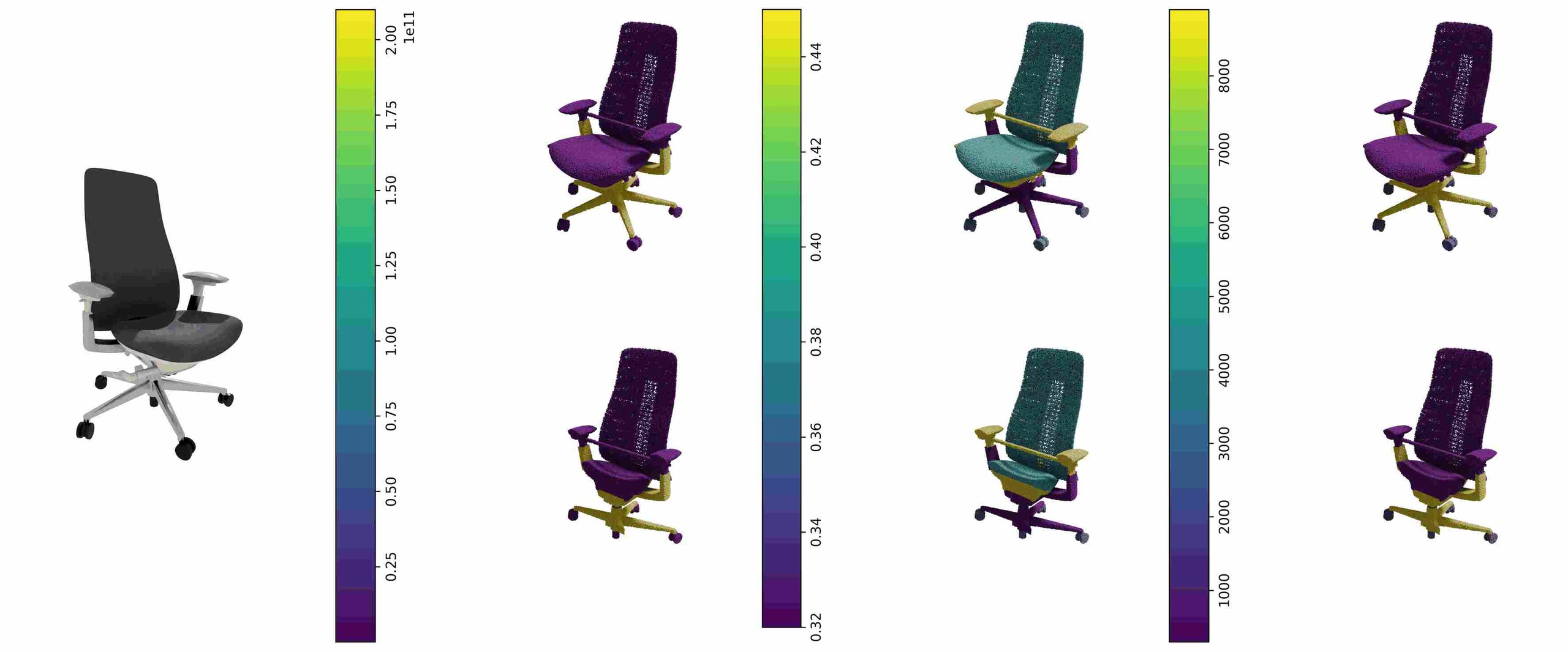}}
    \end{tabular}
    \caption{\textbf{Phineas Comparison.} Mechanical property field comparisons across different methods.}
    \label{fig:phineas_comparison}
\end{figure*}

\begin{figure*}[ht]
    \centering
    \setlength{\tabcolsep}{0pt}
    \renewcommand{\arraystretch}{0}
    \begin{tabular}{p{0.125\textwidth}p{0.125\textwidth}p{0.125\textwidth}p{0.125\textwidth}|p{0.125\textwidth}p{0.125\textwidth}p{0.125\textwidth}p{0.125\textwidth}}
    \centering Object & \centering Young's Modulus ($E$, Pa) & \centering Poisson's Ratio ($\nu$) & \centering Density ($\rho, \frac{kg}{m^3}$) & \centering Object & \centering Young's Modulus ($E$, Pa) & \centering Poisson's Ratio ($\nu$) & \centering Density ($\rho, \frac{kg}{m^3}$) \\
    \end{tabular}
    \begin{tabular}{@{}c@{\hspace{10pt}}c@{}}
    \textbf{\ourmodel} & \textbf{VoMP} \\[2pt]
    \includegraphics[width=0.5\textwidth]{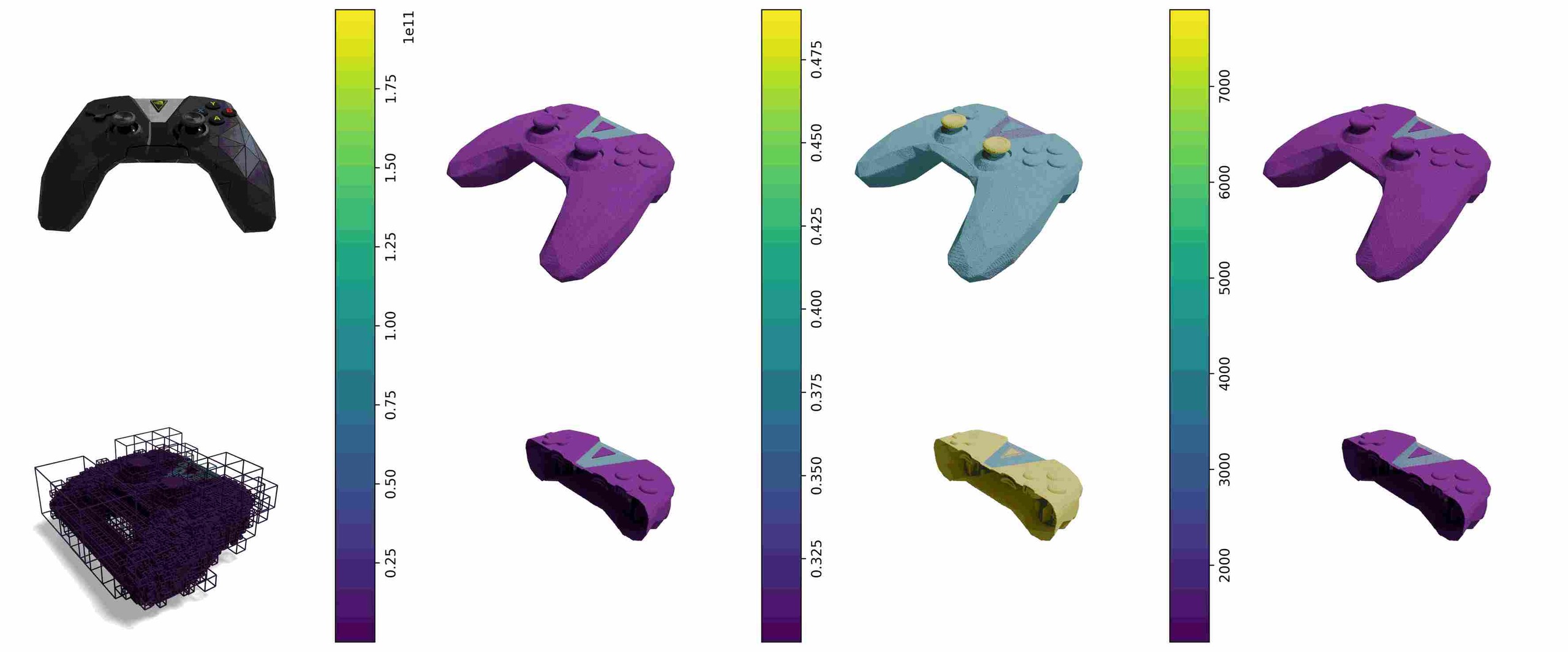} & 
    \includegraphics[width=0.5\textwidth]{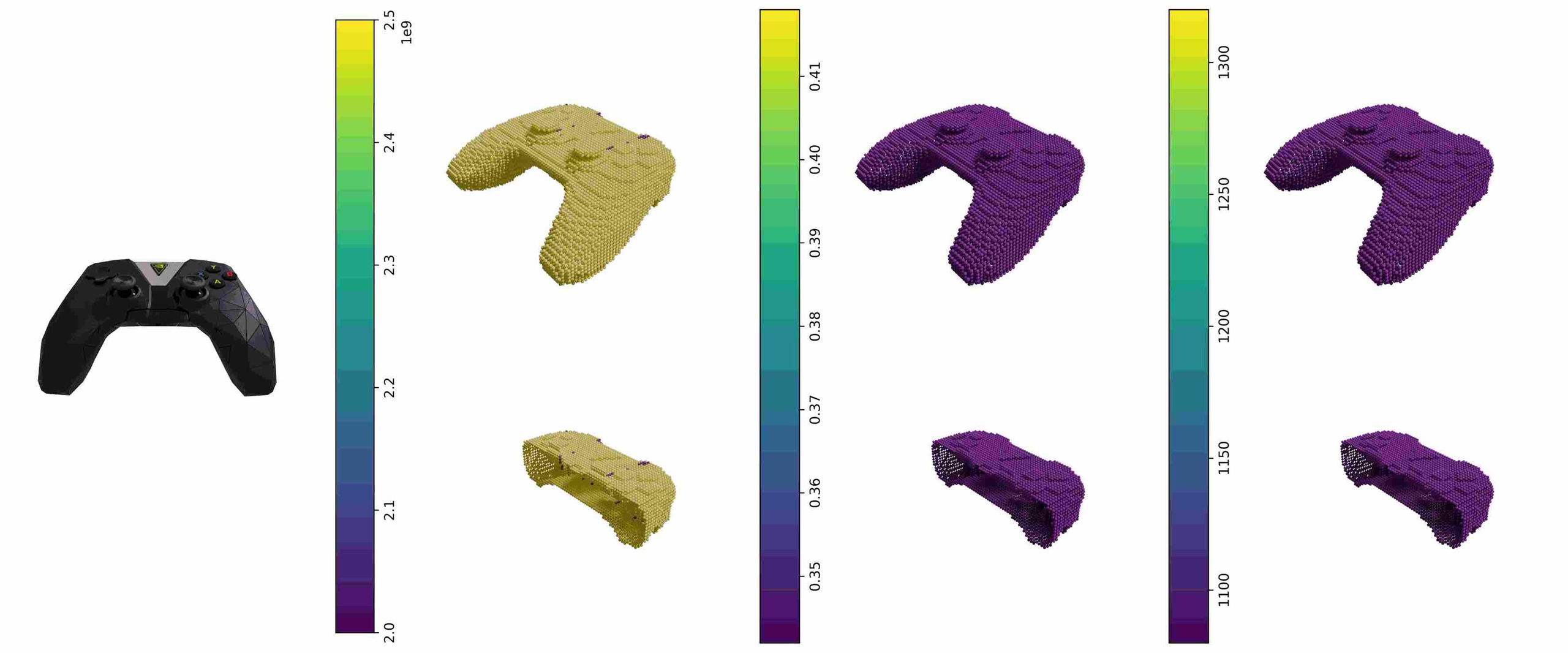} \\[2pt]
    \textbf{N2P} & \textbf{PUGS} \\[2pt]
    \includegraphics[width=0.5\textwidth]{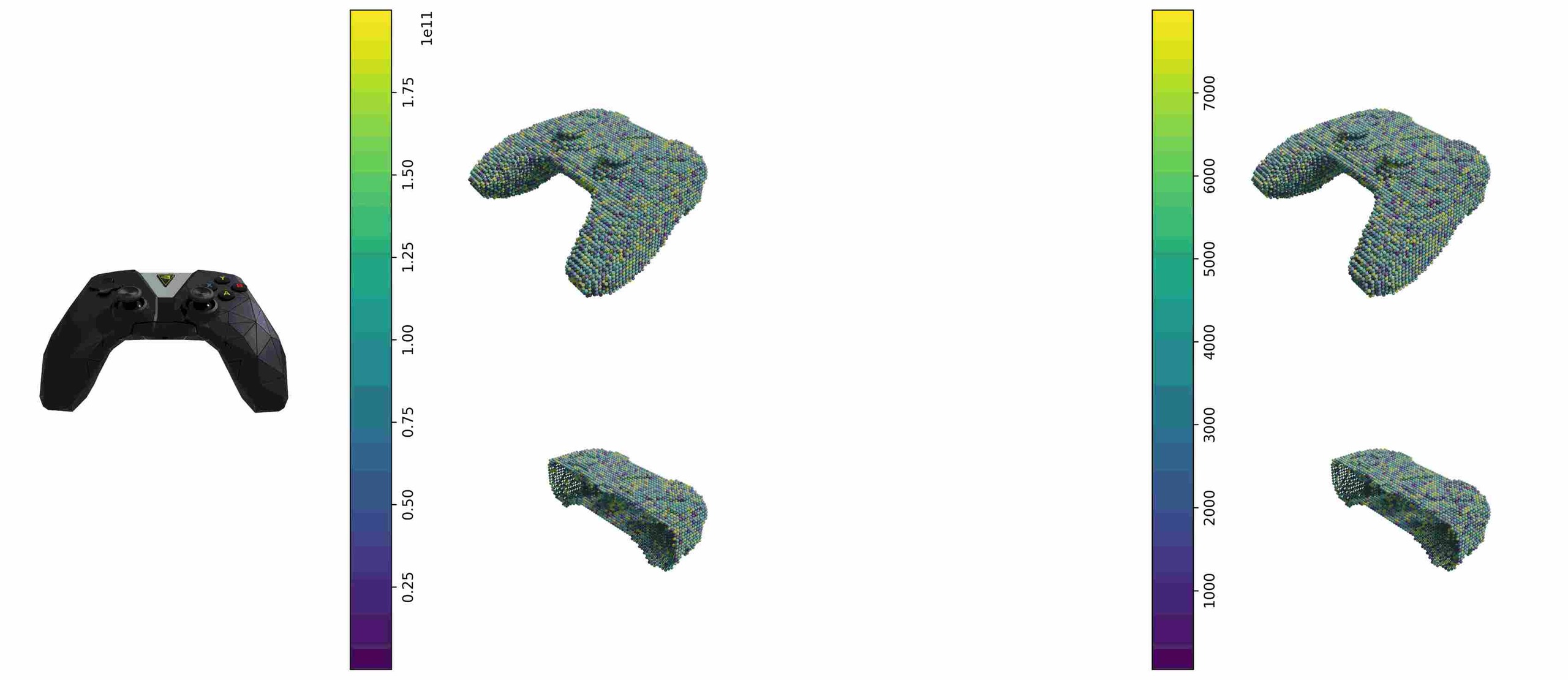} & 
    \includegraphics[width=0.5\textwidth]{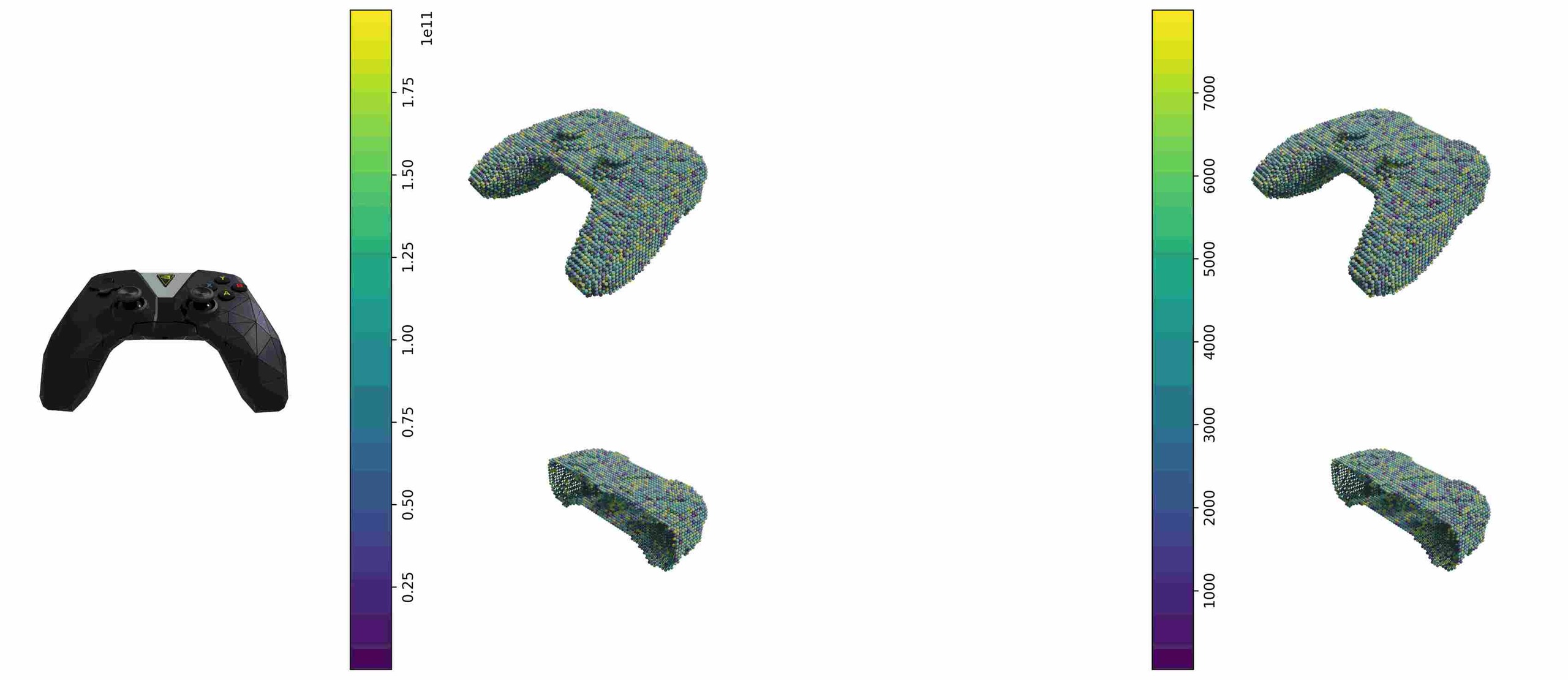} \\[2pt]
    \textbf{Phys4DGen} & \textbf{Pixie} \\[2pt]
    \includegraphics[width=0.5\textwidth]{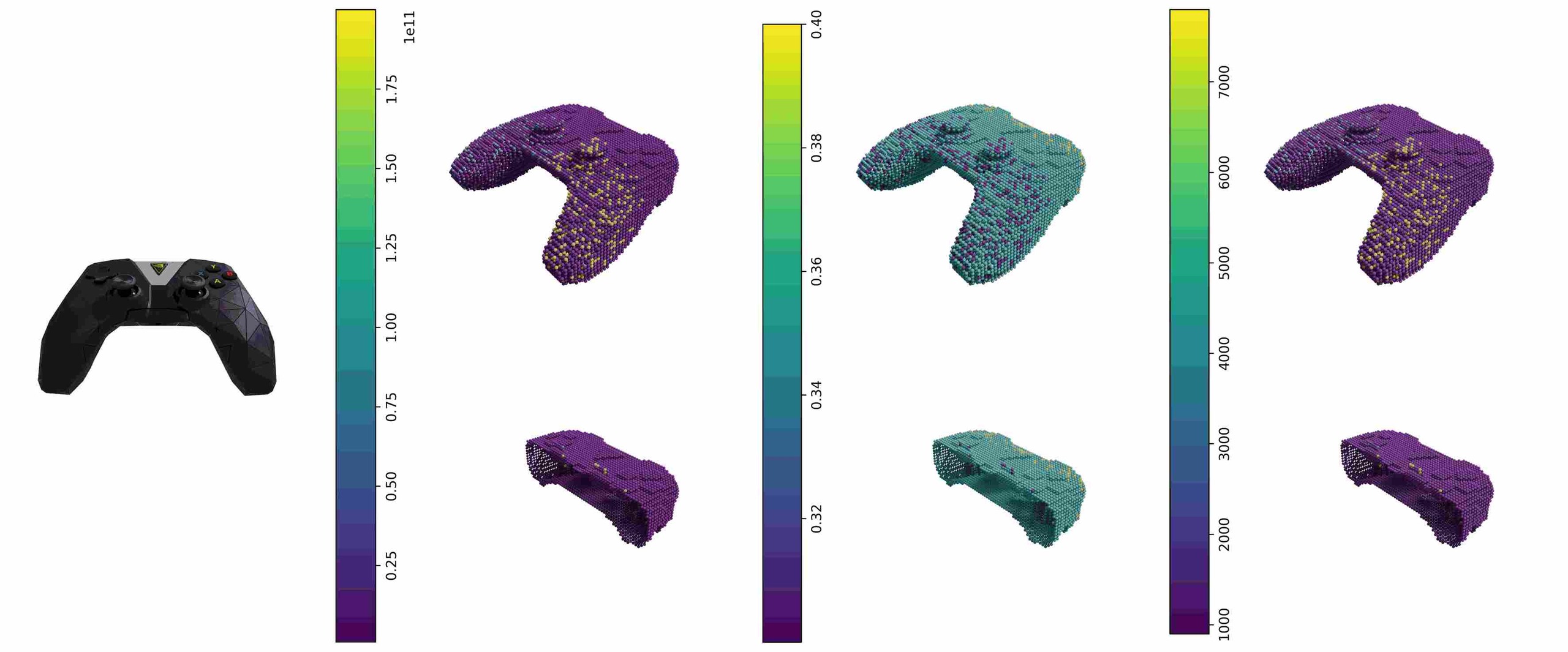} & 
    \includegraphics[width=0.5\textwidth]{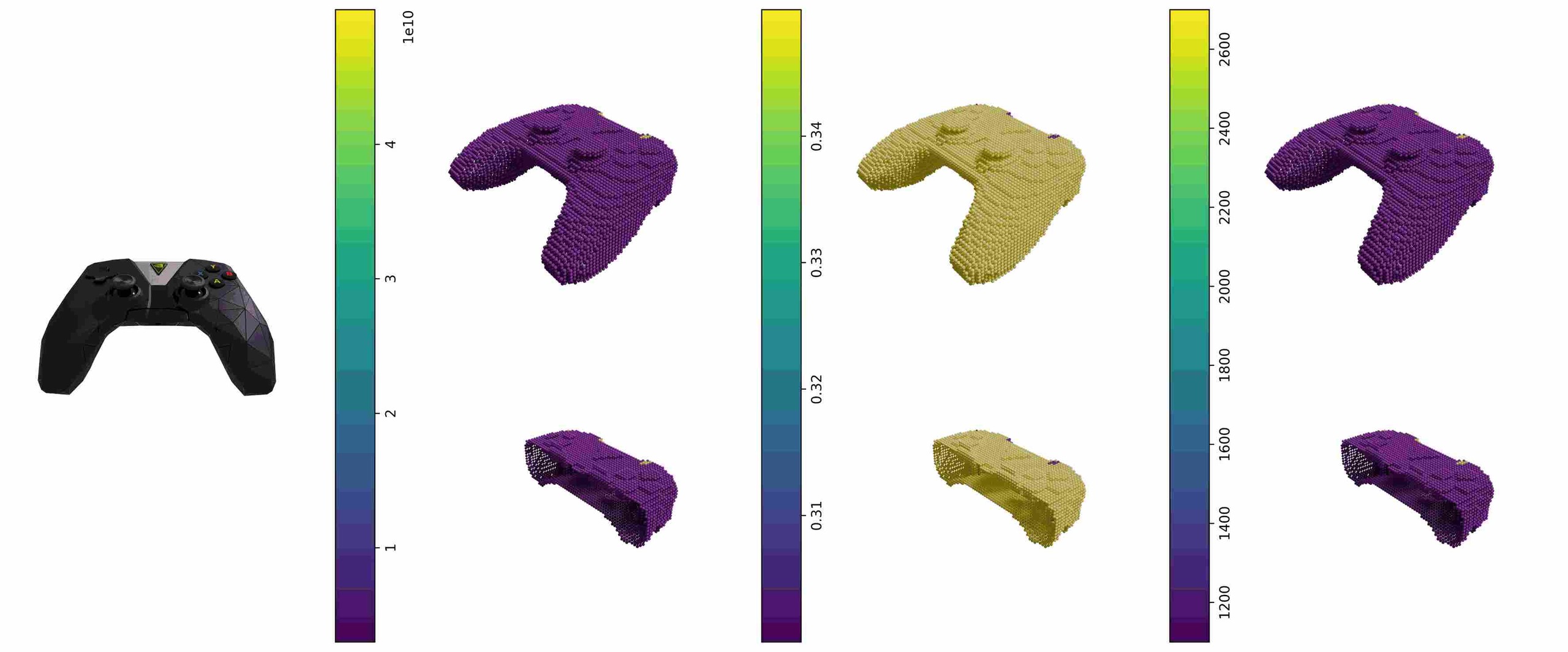} \\[2pt]
    \midrule
    \multicolumn{2}{c}{\textbf{Ground Truth}} \\[2pt]
    \multicolumn{2}{c}{\includegraphics[width=0.95\textwidth]{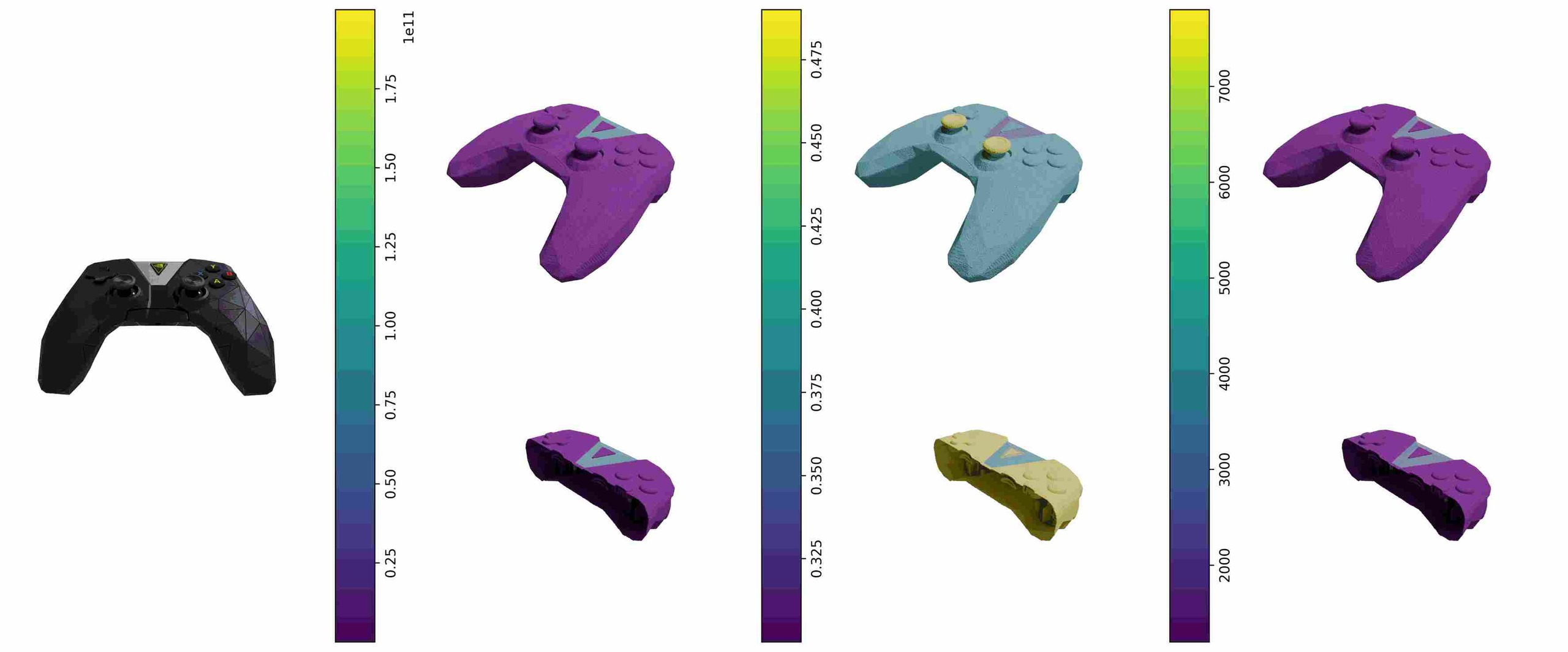}}
    \end{tabular}
    \caption{\textbf{Shield Controller Comparison.} Mechanical property field comparisons across different methods.}
    \label{fig:shield_controller_comparison}
\end{figure*}

We show additional mechanical property fields in~\Cref{fig:extrafields1,fig:extrafields2,fig:extrafields3}. We demonstrate additional comparisons with baseline methods in~\Cref{fig:dartboard_comparison,fig:phineas_comparison,fig:foosball_comparison,fig:lombardy_poplar_comparison,fig:shield_controller_comparison}.

%% file: figures/tb_scaling_object.tex
\begin{table*}
  \centering
  \caption{\textbf{Scaling with model size.} We report object-averaged errors at two query resolutions for all model sizes, using the same evaluation protocol as~\Cref{tab:voxel_object_metrics}. Larger models improve accuracy across $(E,\nu,\rho)$. We find that scaling test-time compute ($64^3 \to 1024^3$) is more effective at larger model sizes.}
  \label{tab:scale_object_avg}
  \begin{tabular}{lrrrrrr}
    \toprule
    \rowcolor{nvidiagreen!15}\textbf{Model} & \multicolumn{2}{c}{Young's Modulus Pa ($E$)} & \multicolumn{2}{c}{Poisson's Ratio ($\nu$)} & \multicolumn{2}{c}{Density $\frac{kg}{m^3}$ ($\rho$)} \\
    \cmidrule(r){2-3} \cmidrule(r){4-5} \cmidrule(r){6-7}
    \rowcolor{nvidiagreen!15} & ALDE ($\downarrow$) & ALRE ($\downarrow$) & ADE ($\downarrow$) & ARE ($\downarrow$) & ADE ($\downarrow$) & ARE ($\downarrow$) \\
    \midrule
    \rowcolor{gray!15} \multicolumn{7}{l}{Evaluation at $64^3$ resolution.} \\
    \textsc{S} & 0.5949 {\scriptsize{($\pm$0.31)}} & 0.0617 {\scriptsize{($\pm$0.04)}} & 0.0354 {\scriptsize{($\pm$0.01)}} & 0.1173 {\scriptsize{($\pm$0.04)}} & 215.7624 {\scriptsize{($\pm$151.54)}} & 0.1427 {\scriptsize{($\pm$0.11)}} \\
    \textsc{B} & 0.3828 {\scriptsize{($\pm$0.24)}} & 0.0397 {\scriptsize{($\pm$0.03)}} & 0.0232 {\scriptsize{($\pm$0.01)}} & 0.0768 {\scriptsize{($\pm$0.04)}} & 136.3824 {\scriptsize{($\pm$138.64)}} & 0.0902 {\scriptsize{($\pm$0.06)}} \\
    \textsc{B+} & 0.3625 {\scriptsize{($\pm$0.23)}} & 0.0376 {\scriptsize{($\pm$0.04)}} & 0.0223 {\scriptsize{($\pm$0.01)}} & 0.0741 {\scriptsize{($\pm$0.03)}} & 133.3584 {\scriptsize{($\pm$161.00)}} & 0.0882 {\scriptsize{($\pm$0.07)}} \\
    \textsc{L} & 0.3480 {\scriptsize{($\pm$0.29)}} & 0.0361 {\scriptsize{($\pm$0.04)}} & 0.0217 {\scriptsize{($\pm$0.01)}} & 0.0721 {\scriptsize{($\pm$0.02)}} & 131.2416 {\scriptsize{($\pm$154.22)}} & 0.0868 {\scriptsize{($\pm$0.06)}} \\
    \textsc{L+} & \underline{0.3355} {\scriptsize{($\pm$0.26)}} & \underline{0.0348} {\scriptsize{($\pm$0.03)}} & \underline{0.0211} {\scriptsize{($\pm$0.01)}} & \underline{0.0700} {\scriptsize{($\pm$0.04)}} & \underline{129.2781} {\scriptsize{($\pm$164.82)}} & \underline{0.0855} {\scriptsize{($\pm$0.07)}} \\
    \textsc{H} & \textbf{0.3278} {\scriptsize{($\pm$0.26)}} & \textbf{0.0340} {\scriptsize{($\pm$0.03)}} & \textbf{0.0205} {\scriptsize{($\pm$0.01)}} & \textbf{0.0680} {\scriptsize{($\pm$0.03)}} & \textbf{127.3125} {\scriptsize{($\pm$150.83)}} & \textbf{0.0842} {\scriptsize{($\pm$0.07)}} \\
    \midrule
    \rowcolor{gray!15} \multicolumn{7}{l}{Evaluation at $1024^3$ resolution.} \\
    \textsc{S} & 1.5898 {\scriptsize{($\pm$0.32)}} & 0.1649 {\scriptsize{($\pm$0.10)}} & 0.0439 {\scriptsize{($\pm$0.01)}} & 0.1457 {\scriptsize{($\pm$0.03)}} & 299.8296 {\scriptsize{($\pm$202.80)}} & 0.1983 {\scriptsize{($\pm$0.10)}} \\
    \textsc{B} & 1.1512 {\scriptsize{($\pm$0.29)}} & 0.1194 {\scriptsize{($\pm$0.10)}} & 0.0284 {\scriptsize{($\pm$0.01)}} & 0.0941 {\scriptsize{($\pm$0.04)}} & 179.1720 {\scriptsize{($\pm$228.62)}} & 0.1185 {\scriptsize{($\pm$0.10)}} \\
    \textsc{B+} & 1.0856 {\scriptsize{($\pm$0.33)}} & 0.1126 {\scriptsize{($\pm$0.10)}} & 0.0265 {\scriptsize{($\pm$0.01)}} & 0.0879 {\scriptsize{($\pm$0.04)}} & 173.7288 {\scriptsize{($\pm$176.72)}} & 0.1149 {\scriptsize{($\pm$0.10)}} \\
    \textsc{L} & 1.0008 {\scriptsize{($\pm$0.31)}} & 0.1038 {\scriptsize{($\pm$0.07)}} & 0.0241 {\scriptsize{($\pm$0.01)}} & 0.0798 {\scriptsize{($\pm$0.03)}} & 167.2272 {\scriptsize{($\pm$173.22)}} & 0.1106 {\scriptsize{($\pm$0.10)}} \\
    \textsc{L+} & \underline{0.9159} {\scriptsize{($\pm$0.23)}} & \underline{0.0950} {\scriptsize{($\pm$0.07)}} & \underline{0.0225} {\scriptsize{($\pm$0.01)}} & \underline{0.0745} {\scriptsize{($\pm$0.03)}} & \underline{161.7867} {\scriptsize{($\pm$174.03)}} & \underline{0.1070} {\scriptsize{($\pm$0.08)}} \\
    \textsc{H} & \textbf{0.8841} {\scriptsize{($\pm$0.27)}} & \textbf{0.0917} {\scriptsize{($\pm$0.07)}} & \textbf{0.0215} {\scriptsize{($\pm$0.01)}} & \textbf{0.0714} {\scriptsize{($\pm$0.03)}} & \textbf{158.4602} {\scriptsize{($\pm$176.28)}} & \textbf{0.1048} {\scriptsize{($\pm$0.08)}} \\
    \bottomrule
  \end{tabular}
\end{table*}

%% file: figures/appendix_scaling.tex
\input{figures/scaling}

\begin{figure}[tb]
\centering
\includegraphics[width=\linewidth]{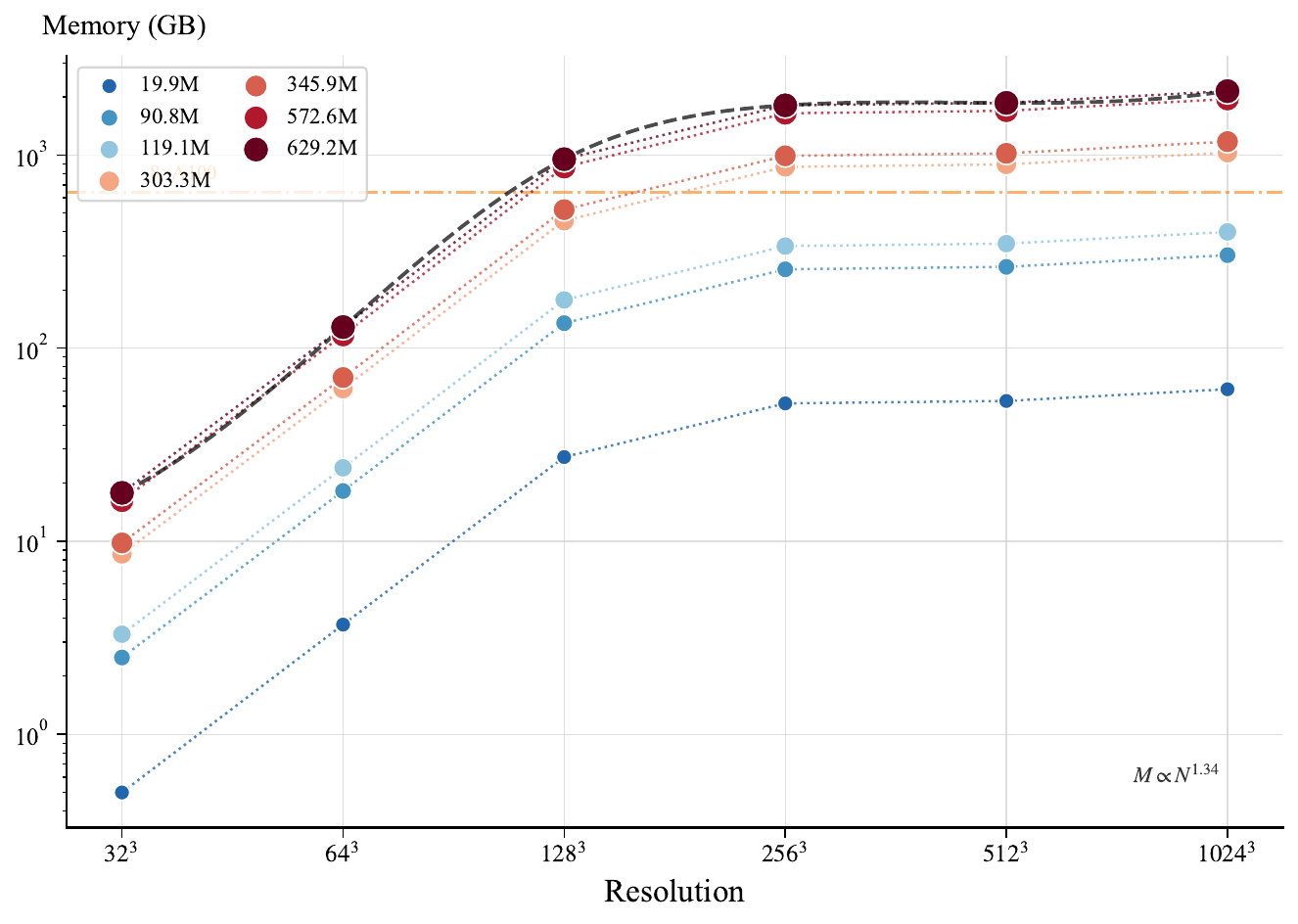}
\caption{%
\textbf{Memory Scaling.}
Peak GPU memory usage versus resolution.
We observe a sub-quadratic scaling relationship of $M \propto N^{1.35}$.
This efficient scaling allows for the generation of high-resolution $1024^3$ volumes within the memory constraints of standard hardware (e.g., 8$\times$A100, dashed orange line).
The curves for different model sizes remain parallel, suggesting that the memory overhead from model parameters is independent of the resolution-based scaling.
}
\label{fig:memory_scaling}
\end{figure}

\begin{figure}[tb]
    \centering
    \includegraphics[width=\linewidth]{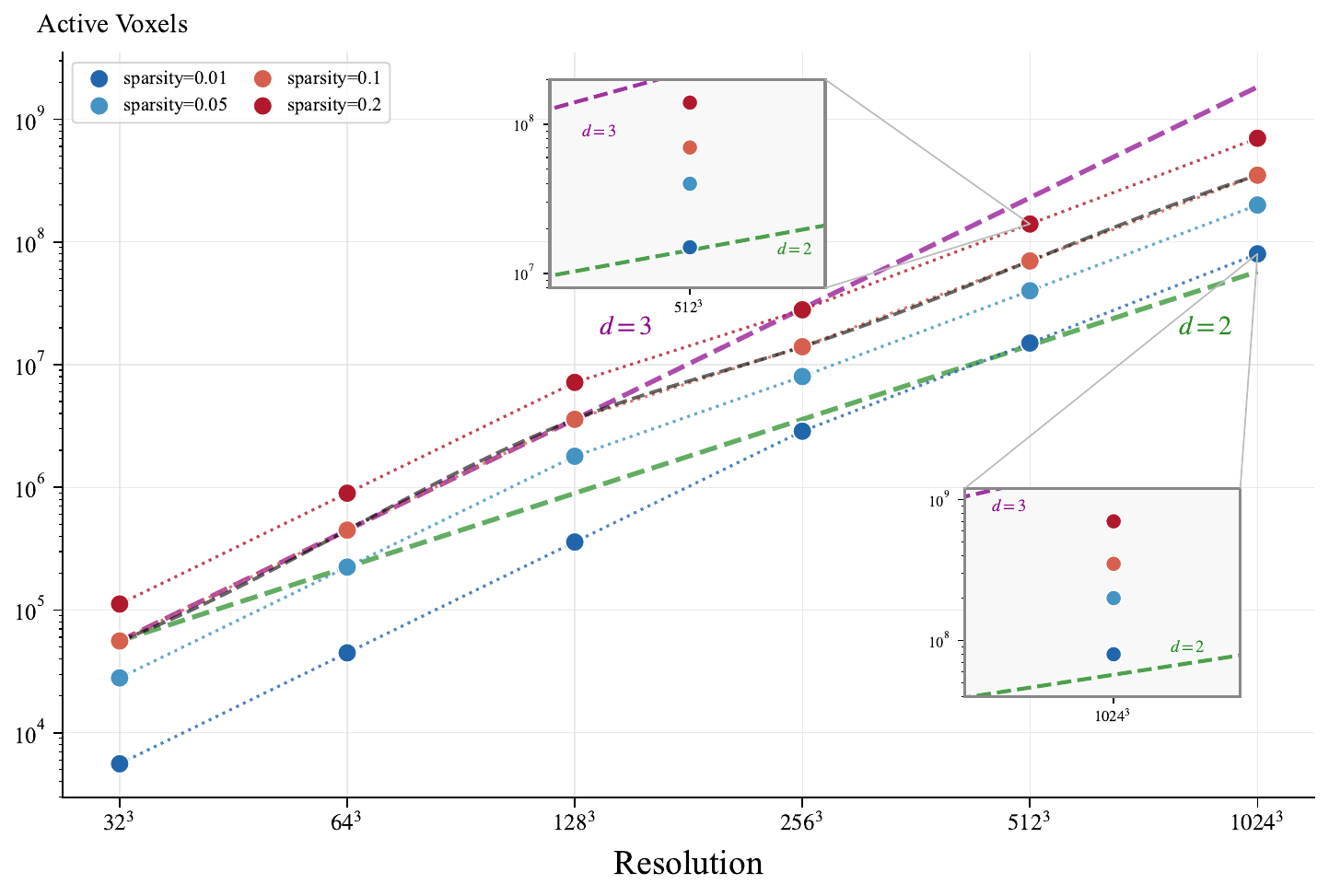}
    \caption{%
    \textbf{Effective Dimensionality of Adaptive Geometry.}
    Active voxel count as a function of resolution for varying sparsity thresholds.
    The slope of these curves represents the dimension $d$ of the generated geometry.
    We measure an effective dimensionality of $d \approx 2.48$ for our sparse adaptive volumetric geometry, which falls between surface scaling ($d=2$) and dense volumetric scaling ($d=3$). In some cases, \vacronym\ can represent a volume more efficiently than representing the same object's surface as a dense voxel grid.
    }
    \label{fig:fractal_dimension}
\end{figure}

\begin{figure*}[tb]
\centering
\includegraphics[width=\textwidth]{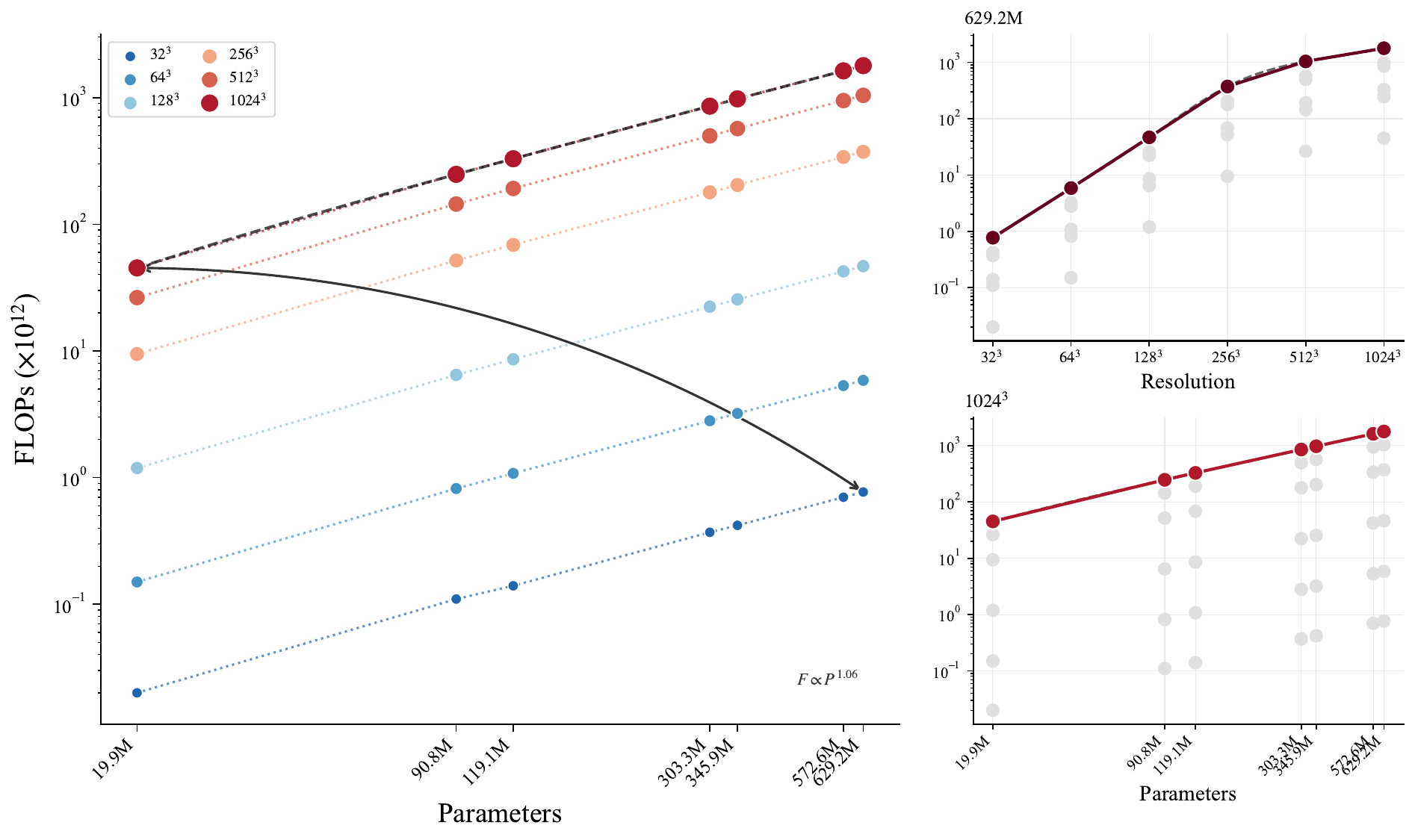}
\caption{%
\textbf{Parameter versus Resolution Sensitivity.}
\textbf{Left:} Total FLOPs as a function of model parameters for various resolutions.
\textbf{Top Right:} Resolution scaling for the Huge (665M) model.
\textbf{Bottom Right:} Parameter scaling at fixed $1024^3$ resolution.
We find that computational cost scales linearly with parameters ($F \propto P^{1.00}$), whereas it scales super-quadratically with resolution ($F \propto N^{2.32}$).
The vertical stratification in the left plot confirms that resolution is the dominant driver of compute; increasing model size from 69M to 665M increases cost by roughly $10\times$, whereas increasing resolution from $128^3$ to $1024^3$ increases cost by two orders of magnitude.
}
\label{fig:parameter_scaling}
\end{figure*}

\begin{figure*}[tb]
\centering
\includegraphics[width=\textwidth]{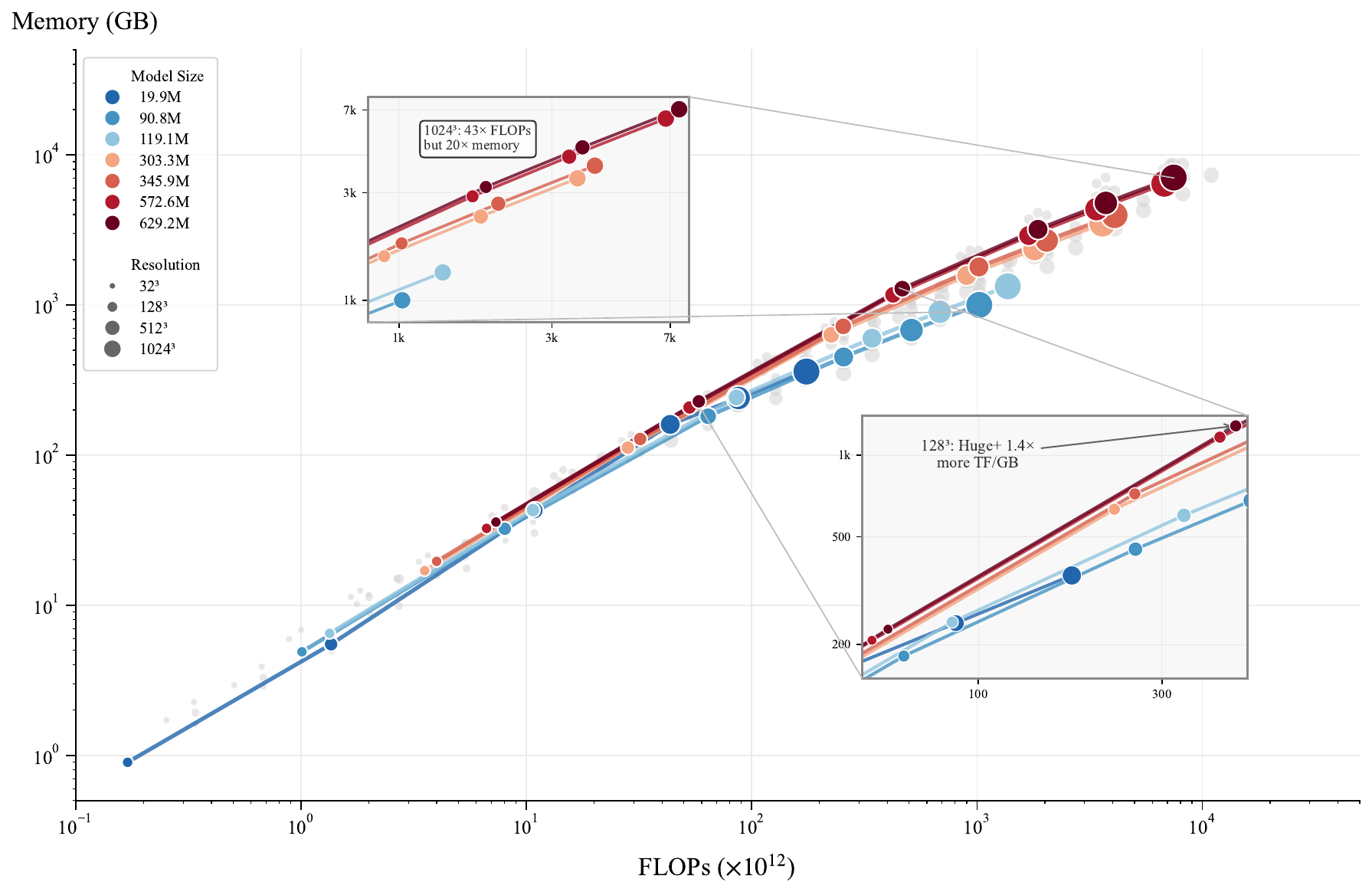}
\caption{%
\textbf{Compute-memory Pareto Frontier.}
Computational cost (FLOPs) versus peak memory usage.
The Pareto frontier shows the optimal trade-off between compute and memory.
\textbf{Insets:} At mid-compute budgets (lower-right inset, $128^3$ regime), a model larger than \textsc{H} achieves $1.4\times$ more TFLOPs per GB compared to smaller models.
At high-compute budgets (upper-left inset, $1024^3$ regime), scaling from the \textsc{S} to larger than \textsc{H} model yields $43\times$ more FLOPs for only $20\times$ more memory.
}
\label{fig:pareto_frontier}
\end{figure*}

%% file: figures/scaling.tex
\begin{figure*}[tb]
    \centering
    \includegraphics[width=\textwidth]{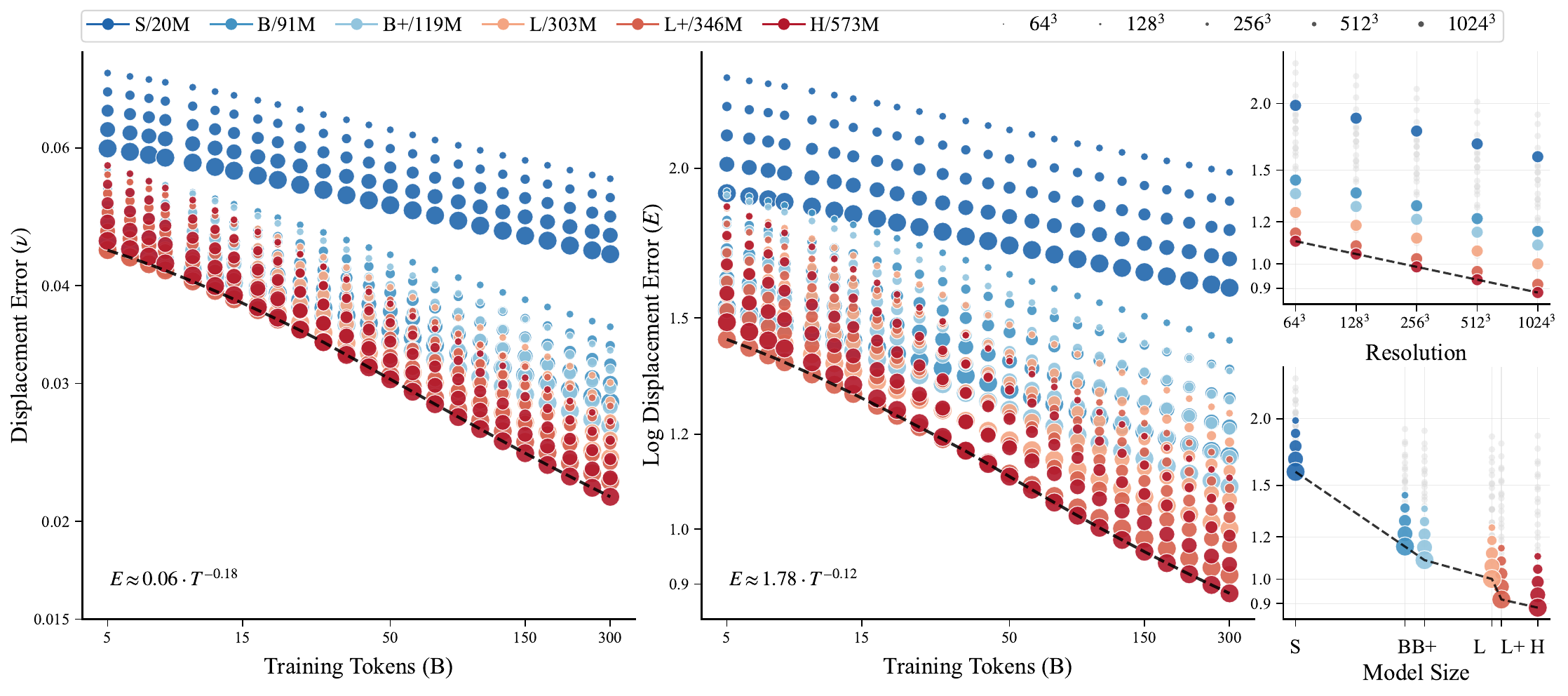}
    \caption{%
    \textbf{Scaling Model, Training, and Test-time Compute.}
    \emph{Left / Center:} We visualize the best runs of our sparse adaptive model across three independent axes: training tokens, test-time compute (output resolution), and model size.
    We show displacement errors for Poisson's ratio ($\nu$) and Young's modulus ($E$) as a function of training tokens, showing that larger models achieve lower error at a fixed training budget and that allocating additional test-time compute (higher resolution) consistently improves accuracy.
    \emph{Right:} We show the final training budget and show the error trend as a function of resolution (top) and model size (bottom).
    }
    \label{fig:full_scaling}
\end{figure*}

%% file: figures/fig_sim_results.tex
\begin{figure*}
\centering
  \includegraphics{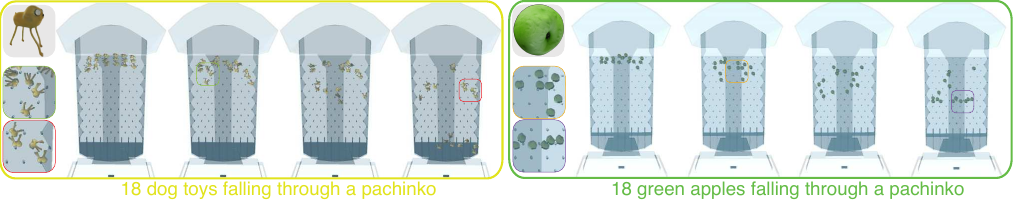}
  \includegraphics{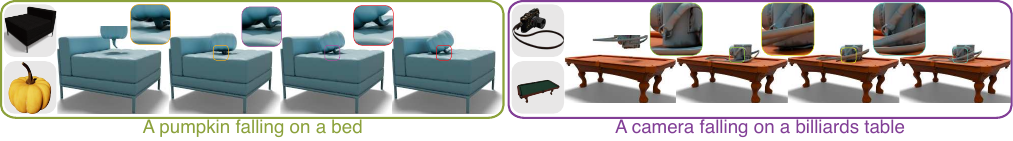}
  \includegraphics{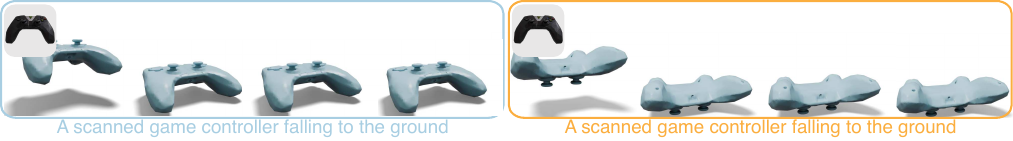}
  \includegraphics{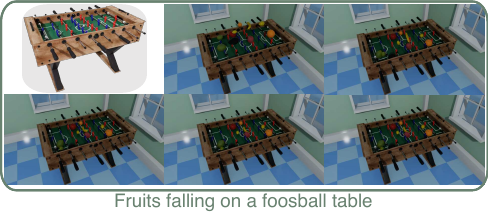}
  \label{fig:results}
  \caption{\textit{First:} We show realistic simulations for 18 Gaussian Splats falling through a pachinko machine mesh using generated properties (\video{04:44}). \textit{Second:} We show realistic simulations for meshes using predicted material values (\video{04:57}). \textit{Third:} In this example, we apply {\ourmodel} to this Gaussian Splat model that we captured using a commercial app. Our method converts this model into a simulation-ready asset, which is tetmeshed and simulated with FEM (\video{04:57}). \textit{Fourth:} We show realistic simulations for meshes using predicted material values (\video{04:44}).}
\end{figure*}

%% file: figures/tb_mech_properties_voxel.tex
\begin{table*}
  \centering
  \caption{\textbf{Mechanical Property Estimates (voxel-averaged).} of our method significantly outperform the baselines on all metrics and marginally outperforms the baseline even with low test-time compute ($64^3$). The metrics are averaged across all voxels.}
  \label{tab:voxel_avg_metrics}
  \resizebox{\textwidth}{!}{
  \begin{tabular}{lrrrrrr}
    \toprule
    \rowcolor{nvidiagreen!15}Method & \multicolumn{2}{c}{Young's Modulus Pa ($E$)} & \multicolumn{2}{c}{Poisson's Ratio ($\nu$)} & \multicolumn{2}{c}{Density $\frac{kg}{m^3}$ ($\rho$)} \\
    \cmidrule(r){2-3} \cmidrule(r){4-5} \cmidrule(r){6-7}
    \rowcolor{nvidiagreen!15}& ALDE ($\downarrow$) & ALRE ($\downarrow$) & ADE ($\downarrow$) & ARE ($\downarrow$) & ADE ($\downarrow$) & ARE ($\downarrow$) \\
    \midrule
    \rowcolor{gray!15} \multicolumn{7}{l}{Evaluation at $64^3$ resolution.} \\
    NeRF2Physics~\cite{zhai2024physicalpropertyunderstandinglanguageembedded} & 2.5719 {\scriptsize{($\pm$1.15)}} & 0.4122 {\scriptsize{($\pm$0.08)}} & - & - & 1354.9458 {\scriptsize{($\pm$1315.71)}} & 1.1496 {\scriptsize{($\pm$0.67)}} \\
    PUGS~\cite{shuai2025pugszeroshotphysicalunderstanding} & 3.8619 {\scriptsize{($\pm$2.01)}} & 0.4512 {\scriptsize{($\pm$0.11)}} & - & - & 3641.0715 {\scriptsize{($\pm$3320.78)}} & 4.0413 {\scriptsize{($\pm$4.16)}} \\
    Phys4DGen$^\star$~\cite{lin2025phys4dgenphysicscompliant4dgeneration} & 5.2977 {\scriptsize{($\pm$3.36)}} & 0.4825 {\scriptsize{($\pm$0.14)}} & 0.0394 {\scriptsize{($\pm$0.05)}} & 0.1425 {\scriptsize{($\pm$0.21)}} & 1285.9489 {\scriptsize{($\pm$1981.11)}} & 1.0445 {\scriptsize{($\pm$2.53)}} \\
    Pixie~\cite{le2025pixie} & 0.4073 {\scriptsize{($\pm$0.42)}} & 0.0462 {\scriptsize{($\pm$0.06)}} & 0.0272 {\scriptsize{($\pm$0.01)}} & 0.0904 {\scriptsize{($\pm$0.04)}} & \underline{110.7426} {\scriptsize{($\pm$294.88)}} & \underline{0.0899} {\scriptsize{($\pm$0.14)}} \\
    VoMP~\cite{dagli2025vomppredictingvolumetricmechanical} &  \underline{0.3765} {\scriptsize{($\pm$0.39)}} & 
    \underline{0.0421} {\scriptsize{($\pm$0.05)}} & 
    \underline{0.0250} {\scriptsize{($\pm$0.01)}} & 
    \underline{0.0837} {\scriptsize{($\pm$0.03)}} & 
    113.3807 {\scriptsize{($\pm$301.90)}} & 
    0.0908 {\scriptsize{($\pm$0.14)}}\\
    \midrule
    Ours-H (0.6B) & \textbf{0.3314} {\textbf{\scriptsize\textcolor{gray}{($\pm$0.34)}}} & \textbf{0.0342} {\textbf{\scriptsize\textcolor{gray}{($\pm$0.04)}}} & \textbf{0.0206} {\textbf{\scriptsize\textcolor{gray}{($\pm$0.01)}}} & \textbf{0.0687} {\textbf{\scriptsize\textcolor{gray}{($\pm$0.03)}}} & \textbf{96.4381} {\textbf{\scriptsize\textcolor{gray}{($\pm$248.62)}}} & \textbf{0.0806} {\textbf{\scriptsize\textcolor{gray}{($\pm$0.12)}}} \\
    \midrule
    \rowcolor{gray!15} \multicolumn{7}{l}{Evaluation at $1024^3$ resolution.} \\
    NeRF2Physics~\cite{zhai2024physicalpropertyunderstandinglanguageembedded} & 3.9814 {\scriptsize{($\pm$1.82)}} & 0.6127 {\scriptsize{($\pm$0.17)}} & - & - & 2548.9372 {\scriptsize{($\pm$1925.44)}} & 2.0861 {\scriptsize{($\pm$1.43)}} \\
    PUGS~\cite{shuai2025pugszeroshotphysicalunderstanding} & 6.3189 {\scriptsize{($\pm$2.97)}} & 0.7421 {\scriptsize{($\pm$0.22)}} & - & - & 6893.2247 {\scriptsize{($\pm$4628.35)}} & 6.1443 {\scriptsize{($\pm$6.21)}} \\
    Phys4DGen$^\star$~\cite{lin2025phys4dgenphysicscompliant4dgeneration} & 7.4136 {\scriptsize{($\pm$4.28)}} & 0.8317 {\scriptsize{($\pm$0.32)}} & 0.0789 {\scriptsize{($\pm$0.08)}} & 0.2912 {\scriptsize{($\pm$0.41)}} & 2962.5718 {\scriptsize{($\pm$3254.10)}} & 2.7415 {\scriptsize{($\pm$3.92)}} \\
    Pixie~\cite{le2025pixie} & 1.2289 {\scriptsize{($\pm$0.57)}} & 0.1394 {\scriptsize{($\pm$0.12)}} & 0.0418 {\scriptsize{($\pm$0.02)}} & 0.1412 {\scriptsize{($\pm$0.07)}} & 218.4621 {\scriptsize{($\pm$268.79)}} & 0.1627 {\scriptsize{($\pm$0.15)}} \\
    VoMP~\cite{dagli2025vomppredictingvolumetricmechanical} &  \underline{1.1284} {\scriptsize{($\pm$0.46)}} & \underline{0.1262} {\scriptsize{($\pm$0.10)}} & \underline{0.0334} {\scriptsize{($\pm$0.02)}} & \underline{0.1149} {\scriptsize{($\pm$0.06)}} & \underline{161.9243} {\scriptsize{($\pm$244.58)}} & \underline{0.1219} {\scriptsize{($\pm$0.13)}} \\
    \midrule
    Ours-H (0.6B) & \textbf{0.8614} {\textbf{\scriptsize\textcolor{gray}{($\pm$0.32)}}} & \textbf{0.0889} {\textbf{\scriptsize\textcolor{gray}{($\pm$0.09)}}} & \textbf{0.0207} {\textbf{\scriptsize\textcolor{gray}{($\pm$0.01)}}} & \textbf{0.0692} {\textbf{\scriptsize\textcolor{gray}{($\pm$0.03)}}} & \textbf{124.0773} {\textbf{\scriptsize\textcolor{gray}{($\pm$221.37)}}} & \textbf{0.1037} {\textbf{\scriptsize\textcolor{gray}{($\pm$0.10)}}} \\
    \bottomrule
  \end{tabular}
  }
\end{table*}

%% file: figures/tb_hard_voxels.tex
\begin{table*}
  \centering
  \caption{\textbf{\textsc{GVT-Hard} at $1024^3$ (voxel-averaged).} Global voxel-averaged errors on the challenging \textsc{GVT-Hard} subset. Most baselines degrade substantially under voxel averaging, while our gap between voxel and object aggregation remains small.}
  \label{tab:gvt_hard_voxel_avg}
  \resizebox{\textwidth}{!}{
  \begin{tabular}{lrrrrrr}
    \toprule
    \rowcolor{nvidiagreen!15}Method & \multicolumn{2}{c}{Young's Modulus Pa ($E$)} & \multicolumn{2}{c}{Poisson's Ratio ($\nu$)} & \multicolumn{2}{c}{Density $\frac{kg}{m^3}$ ($\rho$)} \\
    \cmidrule(r){2-3} \cmidrule(r){4-5} \cmidrule(r){6-7}
    \rowcolor{nvidiagreen!15}& ALDE ($\downarrow$) & ALRE ($\downarrow$) & ADE ($\downarrow$) & ARE ($\downarrow$) & ADE ($\downarrow$) & ARE ($\downarrow$) \\
    \midrule
    NeRF2Physics~\cite{zhai2024physicalpropertyunderstandinglanguageembedded} & 5.9300 {\scriptsize{($\pm$2.70)}} & 0.9500 {\scriptsize{($\pm$0.22)}} & - & - & 4683.9312 {\scriptsize{($\pm$2954.17)}} & 3.9045 {\scriptsize{($\pm$2.12)}} \\
    PUGS~\cite{shuai2025pugszeroshotphysicalunderstanding} & 10.2700 {\scriptsize{($\pm$4.40)}} & 1.2000 {\scriptsize{($\pm$0.35)}} & - & - & 10452.8837 {\scriptsize{($\pm$6890.34)}} & 9.2167 {\scriptsize{($\pm$6.55)}} \\
    Phys4DGen$^\star$~\cite{lin2025phys4dgenphysicscompliant4dgeneration} & 14.8200 {\scriptsize{($\pm$6.90)}} & 1.3500 {\scriptsize{($\pm$0.50)}} & 0.1383 {\scriptsize{($\pm$0.10)}} & 0.5000 {\scriptsize{($\pm$0.55)}} & 7421.3764 {\scriptsize{($\pm$5833.92)}} & 5.6281 {\scriptsize{($\pm$5.62)}} \\
    Pixie~\cite{le2025pixie} & 2.8200 {\scriptsize{($\pm$1.60)}} & 0.3200 {\scriptsize{($\pm$0.16)}} & 0.0662 {\scriptsize{($\pm$0.04)}} & 0.2200 {\scriptsize{($\pm$0.12)}} & 642.9148 {\scriptsize{($\pm$492.66)}} & 0.3392 {\scriptsize{($\pm$0.28)}} \\
    VoMP~\cite{dagli2025vomppredictingvolumetricmechanical} & \underline{2.5480} {\scriptsize{($\pm$1.45)}} & \underline{0.2850} {\scriptsize{($\pm$0.14)}} & \underline{0.0568} {\scriptsize{($\pm$0.03)}} & \underline{0.1900} {\scriptsize{($\pm$0.10)}} & \underline{571.3873} {\scriptsize{($\pm$463.21)}} & \underline{0.3184} {\scriptsize{($\pm$0.26)}} \\
    \midrule
    Ours-H (0.6B) & \textbf{1.2880} {\textbf{\scriptsize\textcolor{gray}{($\pm$0.42)}}} & \textbf{0.1330} {\textbf{\scriptsize\textcolor{gray}{($\pm$0.10)}}} & \textbf{0.0300} {\textbf{\scriptsize\textcolor{gray}{($\pm$0.02)}}} & \textbf{0.1000} {\textbf{\scriptsize\textcolor{gray}{($\pm$0.06)}}} & \textbf{254.6281} {\textbf{\scriptsize\textcolor{gray}{($\pm$237.45)}}} & \textbf{0.1651} {\textbf{\scriptsize\textcolor{gray}{($\pm$0.15)}}} \\
    \bottomrule
  \end{tabular}
  }
\end{table*}

%% file: text/appendix/sav.tex
\section{\vacronym: Our Sparse Adaptive Volumetric Voxels Backend}
\label{sec:savedetails}

We provide additional details about the \vacronym\ backend used for training and evaluation. Our goal is to ensure the representation is suitable as a generation target and as conditioning input, while being fast enough to evaluate material properties at high resolutions.

\subsection{Representation}
\label{sec:app_rep}

\paragraph{Coordinate System and Multi-Resolution Voxels.}
We operate on a normalized domain $\Omega \subset [-0.5,0.5)^3$ with finest grid resolution $G = 2^{L_{\max}}$. A level $\ell \in \{0,\dots,L_{\max}\}$ corresponds to voxel side length
\begin{equation}
s_\ell := \frac{2^\ell}{G},
\end{equation}
and a level grid size $G_\ell := G/2^\ell$. A level-$\ell$ voxel is indexed by $\mathbf{i}\in\{0,\dots,G_\ell-1\}^3$ and corresponds to the axis-aligned cell
\begin{equation}
V_{\ell,\mathbf{i}} := \prod_{\alpha\in\{x,y,z\}} \left[ -0.5 + i_\alpha s_\ell,\ -0.5 + (i_\alpha+1)s_\ell \right).
\end{equation}
Its geometric center is $\mathbf{c}_{\ell,\mathbf{i}} := -0.5 + (\mathbf{i}+0.5)\,s_\ell$.

\paragraph{Stored Nodes.}
An adaptive voxel tree stores a sparse subset of voxels at multiple levels. For each level $\ell$ we store a sparse index set $\mathcal{I}_\ell \subseteq \{0,\dots,G_\ell-1\}^3$ and associated features $\{\mathbf{f}_{\ell,\mathbf{i}}\in\mathbb{R}^d\}_{\mathbf{i}\in\mathcal{I}_\ell}$. We denote the resulting stored node set by
\begin{equation}
\mathcal{T} := \bigcup_{\ell=0}^{L_{\max}} \{(\ell,\mathbf{i},\mathbf{f}_{\ell,\mathbf{i}}) : \mathbf{i}\in\mathcal{I}_\ell\}.
\end{equation}

\paragraph{Finest-Available Query Operator.}
We interpret \vacronym\ as a representation of a function. Given a point $\mathbf{x}\in\Omega$, we define the queried feature as the \emph{finest available} voxel value that covers $\mathbf{x}$:
\begin{align}
(\ell^*(\mathbf{x}),\mathbf{i}^*(\mathbf{x})) &:= \arg\min_{\ell\in\{0,\dots,L_{\max}\}} \left\{ \ell\ :\ \left\lfloor \frac{\mathbf{x}+0.5}{s_\ell}\right\rfloor \in \mathcal{I}_\ell \right\}, \\
\mathcal{T}(\mathbf{x}) &:= \mathbf{f}_{\ell^*(\mathbf{x}),\,\mathbf{i}^*(\mathbf{x})}.
\label{eq:app_finest_available_query}
\end{align}
When $\mathcal{T}$ is a consistent hierarchy,~\Cref{eq:app_finest_available_query} is equivalent to the usual ``leaf voxel'' semantics. The operator also remains well-defined for \emph{partial} trees: if fine voxels are missing in a region, queries fall back to a coarser stored voxel and return its region-average feature. This behavior is used for level-wise supervision and test-time compute scaling, since truncating generation yields a valid coarser field.

\paragraph{Material Trees and Averaging under Truncation.}
For material prediction, we set $\mathcal{F}=\mathcal{M}$ with $d=3$ and $\mathbf{f}_{\ell,\mathbf{i}}=\mathbf{m}_{\ell,\mathbf{i}}=(E_{\ell,\mathbf{i}},\nu_{\ell,\mathbf{i}},\rho_{\ell,\mathbf{i}})$. Our value-range refinement rule stores the descendant mean at coarse voxels when refinement is not triggered. For a voxel $V_{\ell,\mathbf{i}}$ we denote its finest-level descendants by
\begin{equation}
\mathrm{desc}(\ell,\mathbf{i}) := \left\{ \mathbf{j}\in\mathcal{I}_0\ :\ \left\lfloor \mathbf{j}/2^\ell \right\rfloor = \mathbf{i} \right\},
\end{equation}
and define the descendant mean
\begin{equation}
\mathbf{m}_{\ell,\mathbf{i}} := \frac{1}{|\mathrm{desc}(\ell,\mathbf{i})|}\sum_{\mathbf{j}\in\mathrm{desc}(\ell,\mathbf{i})} \mathbf{m}_{0,\mathbf{j}}.
\end{equation}
Thus, if a region is represented only coarsely at inference time, the queried material is the physically meaningful average over that region.

\paragraph{Training-Only Internal Supervision Nodes.}
During training, we additionally store internal voxels that are known to be subdivided, solely to define structure supervision (keep vs.\ subdivide) and per-level losses. These internal nodes do not change the inferred field because queries always return the finest available voxel by~\Cref{eq:app_finest_available_query}. We therefore treat this as an auxiliary supervision scaffold, not a distinct inference-time representation.

\subsection{Baking DINO Features into \vacronym}\label{sec:app_baking_dino}
Here we provide details on how multi-view features of the input object are mapped to {\vacronym} to be ingested
by the Geometry Transformer (\S\ref{sec:method_encoder}).

We form the conditioning node features by reconstructing multi-view DINOv3~\cite{simeoni2025dinov3} patch-token features over a volumetric voxelization of the object. Let $\{\mathbf{p}_i\}_{i=1}^L$ denote the occupied finest-grid voxel centers and let $J$ denote the set of rendered views. For each view $j\in J$, let $\Pi_j:\mathbb{R}^3\to[-1,1]^2$ be the camera projection and let $d_{i,j}$ be the camera-space depth of $\mathbf{p}_i$ in view $j$. Let the DINOv3 patch-token map be $T_j\in\mathbb{R}^{d_{\mathrm{in}}\times n\times n}$ (feature dimension $d_{\mathrm{in}}$ on an $n\times n$ patch grid) and let $F_j:[-1,1]^2\to\mathbb{R}^{d_{\mathrm{in}}}$ denote bilinear sampling of $T_j$. At our target voxel resolution, many occupied voxels are weakly observed in some views; we therefore model the per-view reconstructed feature as $\tilde{\mathbf{f}}_{i,j}:=F_j(\Pi_j(\mathbf{p}_i))=\mathbf{f}^\star_i+\boldsymbol{\varepsilon}_{i,j}$ with $\mathbb{E}[\boldsymbol{\varepsilon}_{i,j}]=\mathbf{0}$ and depth-dependent variance $\mathbb{E}\|\boldsymbol{\varepsilon}_{i,j}\|_2^2 \propto 1+\alpha\,\bar d_{i,j}$. Using this model, we aggregate features using inverse-depth attenuation,
\begin{equation}
  \begin{aligned}
    \tilde{w}_{i,j} &= \frac{1}{1+\alpha\,\bar d_{i,j}}, \quad
    w_{i,j} = \frac{\tilde{w}_{i,j}}{\sum_{j'\in J}\tilde{w}_{i,j'}+\epsilon}, \\
    \mathbf{f}_i &= \sum_{j\in J} w_{i,j}\,F_j\!\big(\Pi_j(\mathbf{p}_i)\big),
  \end{aligned}
\label{eq:dino_attenuation}
\end{equation}
where $\bar d_{i,j}=d_{i,j}/(d_{\max}+\epsilon)$ normalizes depths by $d_{\max}:=\max_{i,j} d_{i,j}$ and $\alpha>0$ controls attenuation strength. The weights are normalized across views so that $\sum_{j\in J} w_{i,j}=1$ for each voxel $i$, yielding a depth-weighted average (the case $\alpha=0$ recovers uniform averaging from prior work~\cite{Wang_2023_CVPR, Dutt_2024_CVPR, Xiang_2025_CVPR, dagli2025vomppredictingvolumetricmechanical}).

We construct the conditioning tree $\mathcal{T}^{\mathrm{in}}$ by merging feature-homogeneous cells on the \vacronym\ grid. For a cell $V_{\ell,\bi}$ at level $\ell$ we sample a subset of finest-level occupied voxel centers $\mathcal{S}^{(K)}_{\ell,\bi}\subseteq \{\mathbf{p}_i:\mathbf{p}_i\in V_{\ell,\bi}\}$ and reconstruct their features $\{\mathbf{f}_i\}$. We then measure the within-cell feature similarity using the maximum pairwise distance in $\ell_2$-normalized feature space,
\begin{equation}
\delta_{\ell,\bi} :=
\max_{i,i'\in \mathcal{S}^{(K)}_{\ell,\bi}}
\left\|
\frac{\mathbf{f}_i}{\|\mathbf{f}_i\|_2}
-\frac{\mathbf{f}_{i'}}{\|\mathbf{f}_{i'}\|_2}
\right\|_2,
\label{eq:feat_uniformity}
\end{equation}
and consider the cell uniform if $\delta_{\ell,\bi}\le \tau_{\mathrm{feat}}$ for a fixed threshold $\tau_{\mathrm{feat}}>0$. Uniform cells are stored as leaves with pooled feature given by the mean of sampled (unnormalized) features,
\begin{equation}
\be_{\ell,\bi} := \frac{1}{|\mathcal{S}^{(K)}_{\ell,\bi}|}\sum_{i\in \mathcal{S}^{(K)}_{\ell,\bi}} \mathbf{f}_i,
\label{eq:feat_pool}
\end{equation}
while non-uniform cells are subdivided into occupied children and refined recursively (\Cref{alg:bottomup-tree}).

\subsection{Sparse Tensor Backend}
\label{sec:app_sparse_backend}

\paragraph{Per-Level Sparse Tensors.}
Each level $\ell$ is represented as a sparse tensor $\mathcal{S}_\ell := (\mathbf{C}_\ell,\mathbf{F}_\ell)$ with coordinates $\mathbf{C}_\ell\in\mathbb{Z}^{N_\ell\times 4}$ and features $\mathbf{F}_\ell\in\mathbb{R}^{N_\ell\times d}$. Each coordinate row has the form $(b,i_x,i_y,i_z)$ where $b$ is the batch index and $(i_x,i_y,i_z)\in\mathcal{I}_\ell$. For a single tree, $b=0$ for all rows. For GPU efficiency, we store coordinates as contiguous 32-bit integer tensors with four dimensions (batch id and integer grid indices). The features are stored contiguously in memory.

\paragraph{Hashing for Fast Lookup.}
For a level grid size $G_\ell$, we use a linear spatial hash,
\begin{equation}
h_\ell(i_x,i_y,i_z) := i_x G_\ell^2 + i_y G_\ell + i_z,
\end{equation}
which is unique on $\{0,\dots,G_\ell-1\}^3$. In our implementation, we realize membership tests by sorting hashes and applying binary search, yielding $O(\log N_\ell)$ lookup per query.

\paragraph{Batch Flattening across Trees and Levels.}
Given a batch of trees $\{\mathcal{T}_b\}_{b=0}^{B-1}$, we construct a single batched sparse tensor by concatenating all coordinates/features and writing the batch id into the first coordinate column. Because our encoder conditions on a mixed-level token set, we additionally maintain a per-token level vector $\boldsymbol{\ell}\in\mathbb{Z}^{N_{\mathrm{tot}}}$ aligned with the concatenated rows as we show in~\Cref{alg:flatten}.

\begin{algorithm}[tb]
\caption{Batch Flattening for Encoder Tokens}
\label{alg:flatten}
\begin{algorithmic}[1]
\REQUIRE Trees $\{\mathcal{T}_b\}_{b=0}^{B-1}$ with per-level sparse tensors $\{(\mathbf{C}^{(b)}_\ell,\mathbf{F}^{(b)}_\ell)\}$
\ENSURE Batched sparse tensor $(\mathbf{C}_{\mathrm{batch}},\mathbf{F}_{\mathrm{batch}})$ and per-token levels $\boldsymbol{\ell}$\vspace{1em}
\STATE $\texttt{coords}\leftarrow[]$
\STATE $\texttt{feats}\leftarrow[]$
\STATE $\texttt{levels}\leftarrow[]$
\FOR{$b=0,1,\dots,B-1$}
  \FOR{each occupied level $\ell$ in $\mathcal{T}_b$}
    \STATE $\mathbf{C}\leftarrow \mathbf{C}^{(b)}_\ell$
    \STATE set $\mathbf{C}[:,0]\leftarrow b$
    \STATE Append $\mathbf{C}$ to $\texttt{coords}$
    \STATE Append $\mathbf{F}^{(b)}_\ell$ to $\texttt{feats}$
    \STATE Append a vector of length $|\mathbf{C}|$ filled with $\ell$ to $\texttt{levels}$
  \ENDFOR
\ENDFOR
\STATE $\mathbf{C}_{\mathrm{batch}}\leftarrow \textsc{Cat}(\texttt{coords})$
\STATE $\mathbf{F}_{\mathrm{batch}}\leftarrow \textsc{Cat}(\texttt{feats})$
\STATE $\boldsymbol{\ell}\leftarrow \textsc{Cat}(\texttt{levels})$
\STATE \textbf{return} $(\mathbf{C}_{\mathrm{batch}},\mathbf{F}_{\mathrm{batch}}),\ \boldsymbol{\ell}$
\end{algorithmic}
\end{algorithm}

\paragraph{Implementation on Top of Sparse Tensor Libraries.}
We implement sparse voxel tensors using \texttt{spconv}~\cite{spconv2022} to store $(\mathbf{C}_\ell,\mathbf{F}_\ell)$ and to accelerate common sparse operators. For efficient batching, we maintain an ordering invariant: for each batch element $b$, all rows with $\mathbf{C}_\ell[:,0]=b$ form a contiguous slice. This induces a per-batch layout, $\{\mathrm{layout}_\ell[b]\}_{b=0}^{B-1}$ enabling fast extraction of a single item’s voxels and efficient broadcast along the batch dimension. When we update features without changing coordinates (a common pattern in neural blocks), we preserve all coordinate metadata and cached index mappings so coordinate-dependent preprocessing is reused instead of recomputed. Finally, we use a spatial cache keyed by scale/stride to reuse expensive coordinate-dependent computations (e.g., sorted hash permutations, window/serialization indices) across repeated operations without changing the stored representation.

\subsection{Core Operators}
\label{sec:app_ops}

We use four core operators throughout our pipeline: material tree construction (see~\Cref{alg:topdown-tree}), conditioning feature tree construction (see~\Cref{alg:bottomup-tree}), batched finest-available point queries (see~\Cref{alg:point-query}), and batch flattening into encoder tokens (see~\Cref{alg:flatten}). We serialize each tree by storing $G$, the occupied levels, and per-level coordinates and features; this is lossless with respect to~\Cref{eq:app_finest_available_query}.

\begin{algorithm}[tb]
\caption{Material Tree Construction via Value-Range Refinement}
\label{alg:topdown-tree}
\begin{algorithmic}[1]
\REQUIRE Finest-level occupied indices $\mathbf{I}_0\in\mathbb{Z}^{N_0\times 3}$, materials $\mathbf{M}_0\in\mathbb{R}^{N_0\times 3}$, resolution $G=2^{L_{\max}}$, tolerance $\boldsymbol{\tau}\in\mathbb{R}^3_+$
\ENSURE Stored material tree $\mathcal{T}=\bigcup_\ell \{(\ell,\mathbf{i},\mathbf{m}_{\ell,\mathbf{i}})\}$ (leaves plus internal supervision nodes)\vspace{1em}
\STATE \textcolor{nvidiagreen}{Phase 1: bottom-up statistics (aligned by parent hashing)}
\STATE $\mathbf{C}^{(0)}\leftarrow \mathbf{I}_0$
\STATE $\mathbf{m}^{(0)}\leftarrow \mathbf{M}_0$
\STATE $\mathbf{m}^{\min(0)}\leftarrow \mathbf{M}_0$
\STATE $\mathbf{m}^{\max(0)}\leftarrow \mathbf{M}_0$
\FOR{$\ell=1,2,\dots,L_{\max}$}
  \STATE $G_\ell \leftarrow G/2^\ell$
  \STATE \textcolor{nvidiagreen}{Parent coords}
  \STATE $\mathbf{P} \leftarrow \left\lfloor \mathbf{C}^{(\ell-1)}/2 \right\rfloor$
  \STATE $\mathbf{h} \leftarrow h_\ell(\mathbf{P})$
  \STATE \textcolor{nvidiagreen}{Group children by parent hash}
  \STATE $(\mathbf{u},\mathbf{inv},\mathbf{cnt}) \leftarrow \textsc{Unique}(\mathbf{h})$
  \STATE $\mathbf{C}^{(\ell)} \leftarrow \textsc{Unhash}(\mathbf{u},G_\ell)$
  \STATE $\mathbf{m}^{(\ell)} \leftarrow \textsc{ScatterSum}(\mathbf{m}^{(\ell-1)},\mathbf{inv}) / \mathbf{cnt}$
  \STATE $\mathbf{m}^{\min(\ell)} \leftarrow \textsc{ScatterMin}(\mathbf{m}^{\min(\ell-1)},\mathbf{inv})$
  \STATE $\mathbf{m}^{\max(\ell)} \leftarrow \textsc{ScatterMax}(\mathbf{m}^{\max(\ell-1)},\mathbf{inv})$
  \STATE \textcolor{nvidiagreen}{Descendant range}
  \STATE $\Delta^{(\ell)} \leftarrow \mathbf{m}^{\max(\ell)} - \mathbf{m}^{\min(\ell)}$
\ENDFOR
\STATE \textcolor{nvidiagreen}{Phase 2: coarse-to-fine selection of stored nodes}
\STATE $\ell_{\text{start}} \leftarrow \max\{\ell:\ |\mathbf{C}^{(\ell)}|>0\}$
\STATE \textcolor{nvidiagreen}{H[$\ell$]: hashes of subdivided voxels at level $\ell$}
\STATE $\mathcal{T}\leftarrow \emptyset$
\STATE $\mathcal{H}\leftarrow \emptyset$
\FOR{$\ell=\ell_{\text{start}}$ \textbf{downto} $0$}
  \IF{$\ell = \ell_{\text{start}}$}
    \STATE $\texttt{active}\leftarrow \texttt{True}^{|\mathbf{C}^{(\ell)}|}$
  \ELSE
    \STATE $\mathbf{p}\leftarrow \left\lfloor \mathbf{C}^{(\ell)}/2 \right\rfloor$
    \STATE $\mathbf{h}_p \leftarrow h_{\ell+1}(\mathbf{p})$
    \STATE $\texttt{active}\leftarrow \textsc{IsIn}(\mathbf{h}_p,\mathcal{H}[\ell+1])$
  \ENDIF
  \STATE $\mathbf{C}_a \leftarrow \mathbf{C}^{(\ell)}[\texttt{active}]$
  \STATE $\mathbf{m}_a \leftarrow \mathbf{m}^{(\ell)}[\texttt{active}]$
  \IF{$\ell=0$}
    \STATE Add all $(0,\mathbf{i},\mathbf{m}_a)$ to $\mathcal{T}$
  \ELSE
    \STATE $\texttt{subdiv}\leftarrow \textsc{Any}(\Delta^{(\ell)}[\texttt{active}] > \boldsymbol{\tau})$
    \STATE \textcolor{nvidiagreen}{Coarse leaves}
    \STATE Add all $(\ell,\mathbf{i},\mathbf{m}_a[\neg\texttt{subdiv}])$ to $\mathcal{T}$
    \STATE \textcolor{nvidiagreen}{Internal supervision}
    \STATE Add all $(\ell,\mathbf{i},\mathbf{m}_a[\texttt{subdiv}])$ to $\mathcal{T}$
    \STATE $\mathcal{H}[\ell] \leftarrow h_\ell(\mathbf{C}_a[\texttt{subdiv}])$
  \ENDIF
\ENDFOR
\STATE \textbf{return} $\mathcal{T}$
\end{algorithmic}
\end{algorithm}

\begin{algorithm}[tb]
\caption{Conditioning Feature Tree Construction via Lazy Refinement}
\label{alg:bottomup-tree}
\begin{algorithmic}[1]
\REQUIRE Finest-level occupied indices $\mathbf{I}_0\in\mathbb{Z}^{N_0\times 3}$, grid size $G=2^{L_{\max}}$, start level $\ell_{\mathrm{start}}$, samples per cell $K$, uniformity threshold $\tau_{\mathrm{feat}}>0$, max nodes $N_{\max}$, and voxel feature lifting as in~\Cref{eq:dino_attenuation}
\ENSURE Frontier-only feature tree $\mathcal{T}=\bigcup \{(\ell,\mathbf{i},\mathbf{e}_{\ell,\mathbf{i}})\}$\vspace{1em}
\STATE $\mathcal{T}\leftarrow \emptyset$
\STATE $\texttt{frontier}\leftarrow \{(\ell_{\mathrm{start}},\mathbf{c})\ :\ \mathbf{c}\in \textsc{Unique}(\lfloor \mathbf{I}_0/2^{\ell_{\mathrm{start}}}\rfloor)\}$
\WHILE{$|\texttt{frontier}|>0$ \AND $|\mathcal{T}|<N_{\max}$}
  \STATE Pop $(\ell,\mathbf{c})$ with maximal $\ell$ from \texttt{frontier}
  \STATE $\mathcal{S}\leftarrow \{n\ :\ \lfloor \mathbf{I}_0[n]/2^\ell\rfloor=\mathbf{c}\}$
  \IF{$|\mathcal{S}|=0$}
    \STATE \textbf{continue}
  \ENDIF
  \STATE $\mathcal{S}_{\mathrm{samp}} \leftarrow \textsc{Sample}(\mathcal{S},\min\{K,|\mathcal{S}|\})$
  \STATE $\mathbf{P}\leftarrow (\mathbf{I}_0[\mathcal{S}_{\mathrm{samp}}] + 0.5)/G - 0.5$
  \STATE $\mathbf{E}\leftarrow \textsc{LiftDINO}(\mathbf{P})$ \hfill (uses~\Cref{eq:dino_attenuation})
  \STATE \textcolor{nvidiagreen}{Row-normalize features: $\mathbf{E}_n \leftarrow \mathbf{E}/\|\mathbf{E}\|_2$}
  \STATE $\mathbf{D}\leftarrow \sqrt{2 - 2(\mathbf{E}_n\mathbf{E}_n^\top)}$
  \STATE $\texttt{uniform}\leftarrow \big(\max_{a\neq b}\mathbf{D}_{a,b}\le \tau_{\mathrm{feat}}\big)$
  \IF{$\texttt{uniform}$ \OR $\ell=0$}
    \STATE $\mathbf{e}_{\ell,\mathbf{c}} \leftarrow \textsc{Mean}(\mathbf{E})$
    \STATE Add $(\ell,\mathbf{c},\mathbf{e}_{\ell,\mathbf{c}})$ to $\mathcal{T}$
  \ELSE
    \FOR{each $\mathbf{o}\in\{0,1\}^3$}
      \STATE $\mathbf{c}'\leftarrow 2\mathbf{c}+\mathbf{o}$
      \IF{child cell $(\ell-1,\mathbf{c}')$ contains at least one occupied voxel}
        \STATE Add $(\ell-1,\mathbf{c}')$ to \texttt{frontier}
      \ENDIF
    \ENDFOR
  \ENDIF
\ENDWHILE
\STATE \textbf{return} $\mathcal{T}$
\end{algorithmic}
\end{algorithm}

\begin{algorithm}[tb]
\caption{Batched Point Query (Finest-Available)}
\label{alg:point-query}
\begin{algorithmic}[1]
\REQUIRE Query points $\mathbf{X}\in\Omega^{Q\times 3}$, per-level coordinate sets $\{\mathcal{I}_\ell\}$, per-level features $\{\mathbf{F}_\ell\}$, grid size $G=2^{L_{\max}}$
\ENSURE Queried features $\mathbf{R}\in\mathbb{R}^{Q\times d}$\vspace{1em}
\STATE $\mathbf{R}\leftarrow \mathbf{0}$
\STATE $\texttt{found}\leftarrow \texttt{False}^Q$
\FOR{$\ell=0,1,\dots,L_{\max}$}
  \STATE \textcolor{nvidiagreen}{Finest to coarsest}
  \IF{\textsc{All}(\texttt{found})}
    \STATE \textbf{break}
  \ENDIF
  \STATE $G_\ell \leftarrow G/2^\ell$
  \STATE \textcolor{nvidiagreen}{Level indices}
  \STATE $\mathbf{I}_q \leftarrow \left\lfloor (\mathbf{X}+0.5)/s_\ell \right\rfloor$
  \STATE $\mathbf{h}_q \leftarrow h_\ell(\mathbf{I}_q)$
  \STATE $\mathbf{h}_\ell \leftarrow h_\ell(\mathcal{I}_\ell)$
  \STATE $(\mathbf{h}_s,\mathbf{perm})\leftarrow \textsc{Sort}(\mathbf{h}_\ell)$
  \STATE $\mathbf{idx}\leftarrow \textsc{SearchSorted}(\mathbf{h}_s,\mathbf{h}_q)$
  \STATE clamp $\mathbf{idx}$ to valid range
  \STATE $\texttt{match}\leftarrow (\mathbf{h}_s[\mathbf{idx}] = \mathbf{h}_q)$
  \STATE $\texttt{upd}\leftarrow \texttt{match}\ \land\ \neg\texttt{found}$
  \STATE $\mathbf{R}[\texttt{upd}] \leftarrow \mathbf{F}_\ell[\mathbf{perm}[\mathbf{idx}[\texttt{upd}]]]$
  \STATE $\texttt{found}[\texttt{upd}] \leftarrow \texttt{True}$
\ENDFOR
\IF{$\neg\textsc{All}(\texttt{found})$}
  \STATE Assign remaining queries by nearest-neighbor over voxel centers across all stored nodes
\ENDIF
\STATE \textbf{return} $\mathbf{R}$
\end{algorithmic}
\end{algorithm}

%% file: text/appendix/ablations.tex
\section{Ablations}

\label{sec:app_ablations}

We provide an in-depth analysis motivating our Adaptive Geometry Transformer and Adaptive Material Generator training scheme by ablating each component.
Our ablations require changing the hyperparameters for fair comparisons; thus, for each ablation, we tune our hyperparameters within an
identical compute budget.
We run all the ablations at the \textsc{B} scale (\Cref{tab:model_scales}) which are reported in~\Cref{tab:ablation_1024}.
It is not possible to directly compare the results of the Material Gaussian Splats ablation with the baseline because the baseline is trained in a different way with a different architecture. Thus, we do our best to make it comparable to other ablations (\Cref{sec:app_gaussian_material_splats}).

\begin{table*}
  \centering
  \caption{\textbf{Architecture and training ablations at $1024^3$.} We evaluate all variants at \textsc{B} scale.}
  \label{tab:ablation_1024}
  \resizebox{\textwidth}{!}{
  \begin{tabular}{lrrrrrr}
    \toprule
    \rowcolor{nvidiagreen!15}Ablation & \multicolumn{2}{c}{Young's Modulus Pa ($E$)} & \multicolumn{2}{c}{Poisson's Ratio ($\nu$)} & \multicolumn{2}{c}{Density $\frac{kg}{m^3}$ ($\rho$)} \\
    \cmidrule(r){2-3} \cmidrule(r){4-5} \cmidrule(r){6-7}
    \rowcolor{nvidiagreen!15} & ALDE ($\downarrow$) & ALRE ($\downarrow$) & ADE ($\downarrow$) & ARE ($\downarrow$) & ADE ($\downarrow$) & ARE ($\downarrow$) \\
    \midrule
    \rowcolor{gray!15} \multicolumn{7}{l}{Initialization.} \\
    Scratch initialization & 1.3794 {\scriptsize{($\pm$0.35)}} & 0.1428 {\scriptsize{($\pm$0.11)}} & 0.0336 {\scriptsize{($\pm$0.02)}} & 0.1126 {\scriptsize{($\pm$0.06)}} & 219.4837 {\scriptsize{($\pm$265.00)}} & 0.1459 {\scriptsize{($\pm$0.12)}} \\
    VoMP~\cite{dagli2025vomppredictingvolumetricmechanical} initialization & 1.1916 {\scriptsize{($\pm$0.30)}} & 0.1236 {\scriptsize{($\pm$0.10)}} & 0.0292 {\scriptsize{($\pm$0.01)}} & 0.0968 {\scriptsize{($\pm$0.04)}} & 185.9342 {\scriptsize{($\pm$232.00)}} & 0.1214 {\scriptsize{($\pm$0.10)}} \\
    \midrule
    \rowcolor{gray!15} \multicolumn{7}{l}{Query embeddings.} \\
    w/o level embedding & 1.2619 {\scriptsize{($\pm$0.31)}} & 0.1307 {\scriptsize{($\pm$0.10)}} & 0.0311 {\scriptsize{($\pm$0.01)}} & 0.1032 {\scriptsize{($\pm$0.05)}} & 197.3824 {\scriptsize{($\pm$240.00)}} & 0.1289 {\scriptsize{($\pm$0.11)}} \\
    w/ RoPE~\cite{rope} level encoding & 1.1884 {\scriptsize{($\pm$0.30)}} & 0.1227 {\scriptsize{($\pm$0.10)}} & 0.0290 {\scriptsize{($\pm$0.01)}} & 0.0961 {\scriptsize{($\pm$0.04)}} & 184.5173 {\scriptsize{($\pm$233.00)}} & 0.1207 {\scriptsize{($\pm$0.10)}} \\
    w/o octant embedding & 1.4682 {\scriptsize{($\pm$0.40)}} & 0.1579 {\scriptsize{($\pm$0.12)}} & 0.0674 {\scriptsize{($\pm$0.04)}} & 0.2386 {\scriptsize{($\pm$0.18)}} & 419.6341 {\scriptsize{($\pm$540.00)}} & 0.6852 {\scriptsize{($\pm$0.55)}} \\
    \midrule
    \rowcolor{gray!15} \multicolumn{7}{l}{Structure and material supervision.} \\
    w/o empty-space supervision & 1.3106 {\scriptsize{($\pm$0.34)}} & 0.1358 {\scriptsize{($\pm$0.11)}} & 0.0329 {\scriptsize{($\pm$0.02)}} & 0.1096 {\scriptsize{($\pm$0.07)}} & 208.1756 {\scriptsize{($\pm$255.00)}} & 0.1374 {\scriptsize{($\pm$0.12)}} \\
    w/ leaf-only material supervision & 1.8893 {\scriptsize{($\pm$0.55)}} & 0.1967 {\scriptsize{($\pm$0.15)}} & 0.0386 {\scriptsize{($\pm$0.02)}} & 0.1298 {\scriptsize{($\pm$0.07)}} & 287.9042 {\scriptsize{($\pm$350.00)}} & 0.3218 {\scriptsize{($\pm$0.28)}} \\
    \midrule
    \rowcolor{gray!15} \multicolumn{7}{l}{Material parameterization.} \\
    w/o MatVAE~\cite{dagli2025vomppredictingvolumetricmechanical} & 1.6547 {\scriptsize{($\pm$0.48)}} & 0.1704 {\scriptsize{($\pm$0.13)}} & 0.0411 {\scriptsize{($\pm$0.03)}} & 0.1407 {\scriptsize{($\pm$0.10)}} & 268.7729 {\scriptsize{($\pm$330.00)}} & 0.2216 {\scriptsize{($\pm$0.19)}} \\
    \midrule
    \rowcolor{gray!15} \multicolumn{7}{l}{Material Gaussian Splats.} \\
    Material Gaussian Splats (\Cref{sec:app_gaussian_material_splats}) & \textbf{1.1015} {\textbf{\scriptsize\textcolor{gray}{($\pm$0.31)}}} & \textbf{0.1147} {\textbf{\scriptsize\textcolor{gray}{($\pm$0.10)}}} & \textbf{0.0273} {\textbf{\scriptsize\textcolor{gray}{($\pm$0.01)}}} & \textbf{0.0904} {\textbf{\scriptsize\textcolor{gray}{($\pm$0.04)}}} & \textbf{175.8321} {\textbf{\scriptsize\textcolor{gray}{($\pm$222.00)}}} & \textbf{0.1168} {\textbf{\scriptsize\textcolor{gray}{($\pm$0.10)}}} \\

    \midrule Ours-\textsc{B} & \underline{1.1512} {\scriptsize{($\pm$0.29)}} & \underline{0.1194} {\scriptsize{($\pm$0.10)}} & \underline{0.0284} {\scriptsize{($\pm$0.01)}} & \underline{0.0941} {\scriptsize{($\pm$0.04)}} & \underline{179.1720} {\scriptsize{($\pm$228.62)}} & \underline{0.1185} {\scriptsize{($\pm$0.10)}} \\
    \bottomrule
  \end{tabular}
  }
\end{table*}

\paragraph{Initialization.}
We initialize AGT from a pretrained TRELLIS~\cite{Xiang_2025_CVPR} encoder checkpoint (\Cref{sec:app_training}). We ablate this choice by training from scratch and by initializing from a VoMP~\cite{dagli2025vomppredictingvolumetricmechanical} geometry encoder checkpoint.

\paragraph{Query Embeddings.}
We ablate the discrete signals used to disambiguate candidate voxels during coarse-to-fine decoding. Removing the level embedding eliminates the explicit level index from the query (\Cref{eq:decoder_query}), while removing the octant embedding removes the child offset signal (\Cref{eq:octant_id,eq:decoder_query}). We also compare learned level embeddings to a deterministic RoPE-style~\cite{rope} level encoding.

\paragraph{Structure and Material Supervision.}
We ablate the two auxiliary supervision choices used to stabilize coarse-to-fine learning. First, we remove explicit supervision for \textsc{Empty} actions by computing $\cL_{\mathrm{struct}}$ only on non-empty ground-truth candidates (\Cref{eq:struct_loss}), which tests whether negative (empty-space) supervision is necessary for reliable structure prediction. Second, we restrict material supervision to the finest level by dropping coarse-level material losses (\Cref{eq:mat_loss}), testing whether coarse-level averages are needed to regularize long-range material assignments. We note that removing the coarse level material supervision also leads to not being able to scale test-time compute since lower resolutions no longer generate average material values.

\paragraph{Material Parameterization.}
We ablate the MatVAE constraint by directly regressing the normalized material triplet, removing the learned MatVAE decoder while keeping the same regression loss form (\Cref{eq:mat_loss}).

\subsection{Material Gaussian Splats}
\label{sec:app_gaussian_material_splats}

Our main method represents the material field as a piecewise-constant adaptive voxel tree, where each leaf voxel stores a constant material vector queried by the tree's spatial lookup. This discretization is effective for scaling to very high effective resolutions, but it ultimately ties the finest representable variation to the finest voxel size. To explore a complementary alternative for capturing sub-voxel material detail, we experimented with a fixed-grid 3D Gaussian Splatting~\cite{10.1145/3592433} inspired variant. We keep the training dataset and preprocessing identical to VoMP~\cite{dagli2025vomppredictingvolumetricmechanical}, including fixed-grid voxelization and per-voxel multi-view feature processing. We replace the single per-voxel material latent with a set of Gaussian primitives per voxel.

\paragraph{Representation.}
We voxelize the object on a fixed $64^3$ grid and process up to $L_N=32{,}768$ occupied voxels per object via stochastic subsampling when the occupied set is larger. Conditioned on the same per-voxel features as VoMP, for each occupied voxel $V_i$ with center $\mathbf{c}_i$ and side length $h$, the decoder predicts $N=32$ anisotropic 3D Gaussians with parameters $\{(\boldsymbol{\mu}_{i,k},\boldsymbol{\Sigma}_{i,k},\alpha_{i,k},\mathbf{z}_{i,k})\}_{k=1}^N$. Here $\boldsymbol{\mu}_{i,k}\in\mathbb{R}^3$ is a center constrained to lie inside $V_i$, $\boldsymbol{\Sigma}_{i,k}$ is a positive-definite covariance (parameterized by an anisotropic scale and a 3D rotation), $\alpha_{i,k}\in(0,1)$ is an opacity-like amplitude, and $\mathbf{z}_{i,k}\in\mathbb{R}^2$ is a 2D MatVAE latent code. Unlike appearance-focused splatting, no color is predicted; instead, each Gaussian carries a material latent which is decoded by the frozen MatVAE decoder. This is a direct extension of VoMP's per-voxel latent prediction: a single latent per voxel is replaced by $N$ latents tied to continuous Gaussian supports within the voxel.

\paragraph{Querying the Material Gaussian Splats.}
Querying $\hat{\mathcal{M}}(\bx)$ can be viewed as a local ``rendering'' operator analogous to 3D Gaussian splatting: we evaluate each Gaussian density at a 3D sample point and normalize contributions to obtain mixture weights. Given a query point $\bx\in V_i$, we form mixture weights from the Gaussian densities,
\begin{equation}
\begin{aligned}
\tilde{w}_{i,k}(\bx) &= \alpha_{i,k}\,\exp\!\left(-\tfrac{1}{2}(\bx-\boldsymbol{\mu}_{i,k})^\top \boldsymbol{\Sigma}_{i,k}^{-1}(\bx-\boldsymbol{\mu}_{i,k})\right),\\
w_{i,k}(\bx)&=\frac{\tilde{w}_{i,k}(\bx)}{\sum_{j=1}^N \tilde{w}_{i,j}(\bx)}.
\end{aligned}
\end{equation}
We then decode each latent through MatVAE and blend the resulting normalized material triplets,
\begin{equation}
\hat{\mathcal{M}}(\bx) = \sum_{k=1}^N w_{i,k}(\bx)\,g_{\textsc{MatVAE}}(\mathbf{z}_{i,k}).
\end{equation}
This defines a continuous, locally smooth material field within each voxel while preserving our use of MatVAE to constrain predicted properties.

\paragraph{Training.}
This ablation follows the supervised fixed-grid recipe of VoMP~\cite{dagli2025vomppredictingvolumetricmechanical}; the only change is to replace the single per-voxel latent with $N$ Gaussian-supported latents inside each occupied voxel. Let $\mathcal{V}$ denote the occupied voxel indices used in an iteration (after subsampling to $|\mathcal{V}|\le L_N$). For each $i\in\mathcal{V}$, we draw a fixed number $Q$ of sample points $\{\bx_{i,q}\}_{q=1}^Q\subset V_i$ and supervise the normalized material field at those points (normalization in~\Cref{sec:app_training}). Since $\hat{\mathcal{M}}(\bx_{i,q})$ is obtained by the differentiable rendering/query operator in the previous paragraph, training amounts to rendering materials at sampled points and minimizing the point-sampled regression loss
\begin{equation}
\cL_{\mathrm{GS}} =
\frac{1}{|\mathcal{V}|Q}
\sum_{i\in\mathcal{V}}
\sum_{q=1}^{Q}
\big\|\hat{\mathcal{M}}(\bx_{i,q}) - \mathbf{m}^{\star}(\bx_{i,q})\big\|^2_{\boldsymbol{\Lambda}},
\label{eq:gaussian_mat_loss}
\end{equation}
where $\mathbf{m}^{\star}(\bx)\in\mathbb{R}^3$ denotes the ground-truth normalized material triplet at $\bx$ and $\|\cdot\|^2_{\boldsymbol{\Lambda}}$ is the per-property weighted squared error defined in~\Cref{eq:mat_loss}. Empirically, this Gaussian refinement yields only modest gains over the voxel baseline while increasing per-voxel parameterization.

While ablating this model, we choose parameters under the same compute budget as other baselines to serve as a proxy for a fair comparison. This comparison is not perfect since Material Gaussian Splats model is trained with only Data Parallelism which was not possible for the other baselines due to their memory constraints during training. We find that the Material Gaussian Splats are able to outperform the \textsc{B} size, however it is unable to scale well and match performance of our higher model sizes.

%% file: text/appendix/metrics.tex
\section{Metrics}
\label{app:metrics}

We present an explanation of the metrics we use. We use the same metrics as in VoMP~\cite{dagli2025vomppredictingvolumetricmechanical}. 

\subsection{Metrics for Mass and Field Estimation}
\label{app:metricsfield}

To evaluate the accuracy of predicted scalar quantities such as object mass, as well as continuous scalar fields like density or stiffness, we use several commonly adopted metrics. Let $y$ denote a ground-truth scalar value or voxel-wise field (e.g., density), and $\hat{y}$ its predicted counterpart.

\paragraph{Absolute Difference Error (ADE).}  
The average absolute error between predicted and ground-truth values:
\begin{equation}
    \mathrm{ADE} = \frac{1}{N} \sum_{i=1}^N |y_i - \hat{y}_i|.
\end{equation}

This metric is scale-sensitive and reports the error in physical units (e.g., $\mathrm{kg/m^3}$ for density, $\mathrm{kg}$ for mass).

\paragraph{Absolute Log Difference Error (ALDE).}  
The average absolute error in logarithmic space:
\begin{equation}
    \mathrm{ALDE} = \frac{1}{N} \sum_{i=1}^N |\log y_i - \log \hat{y}_i|.
\end{equation}

This metric captures multiplicative error and is particularly useful for quantities that vary over several orders of magnitude.

\paragraph{Average Relative Error (ARE).}  
The mean relative deviation between predictions and ground truth:
\begin{equation}
    \mathrm{ARE} = \frac{1}{N} \sum_{i=1}^N \left| \frac{y_i - \hat{y}_i}{y_i} \right|.
\end{equation}

This dimensionless metric penalizes over- and under-estimates proportionally, making it appropriate for comparing across varying scales.

\paragraph{Minimum Ratio Error (MnRE).}  
A symmetric and bounded measure of relative accuracy:
\begin{equation}
\mathrm{MnRE} = \frac{1}{N} \sum_{i=1}^N \min\left( \frac{y_i}{\hat{y}_i}, \frac{\hat{y}_i}{y_i} \right).
\end{equation}

This metric ranges from $0$ to $1$ and is maximized when predictions are perfectly accurate. As suggested in prior work~\cite{mass_estimation_visual}, MnRE avoids bias toward systematic over- or under-estimation and reduces sensitivity to outliers, making it particularly effective for evaluating physical quantity predictions across heterogeneous samples.

\subsection{Metrics to Measure Differences in Mechanical Properties}
\label{app:metricsmech}

We evaluate mechanical-property estimates at query points inside the object volume. Let $\{(E_i,\nu_i,\rho_i)\}_{i=1}^N$ denote the ground-truth material triplets at the queried locations and let $\{(\hat{E}_i,\hat{\nu}_i,\hat{\rho}_i)\}_{i=1}^N$ denote the corresponding predictions.

\paragraph{Relative Error in $\log(E)$.}  
Relative error between predicted and true values of the logarithm of Young’s modulus $E$ reported in units of Pa. This captures relative error in material stiffness across several orders of magnitude.
\begin{equation}
    \mathrm{ARE}_{\log E} = \frac{1}{N}\sum_{i=1}^N \left|\log_{10}\hat{E}_i - \log_{10}E_i\right|.
\end{equation}

\paragraph{Relative Error in $\nu$.}  
Relative Error in linear space for Poisson’s ratio $\nu$, a dimensionless measure of lateral contraction under uniaxial loading.
\begin{equation}
    \mathrm{ARE}_{\nu} = \frac{1}{N}\sum_{i=1}^N \left|\hat{\nu}_i - \nu_i\right|.
\end{equation}

\paragraph{Relative Error  in $\rho$.}  
Relative Error between predicted and true values of material density $\rho$, reported in units of $\mathrm{kg/m^3}$.
\begin{equation}
    \mathrm{ARE}_{\rho} = \frac{1}{N}\sum_{i=1}^N \left|\hat{\rho}_i - \rho_i\right|.
\end{equation}

\paragraph{Displacement in $\log(E)$.}
We measure the mean absolute error in $\log_{10}$ space,
\begin{equation}
    \mathrm{ADE}_{\log E} = \frac{1}{N}\sum_{i=1}^N \left|\log_{10}\hat{E}_i - \log_{10}E_i\right|.
\end{equation}

\paragraph{Displacement in $\nu$.}
We measure the mean absolute error of Poisson's ratio in linear space,
\begin{equation}
    \mathrm{ADE}_{\nu} = \frac{1}{N}\sum_{i=1}^N \left|\hat{\nu}_i - \nu_i\right|.
\end{equation}

\paragraph{Displacement in $\rho$.}
We measure the mean absolute error of density in linear space,
\begin{equation}
    \mathrm{ADE}_{\rho} = \frac{1}{N}\sum_{i=1}^N \left|\hat{\rho}_i - \rho_i\right|.
\end{equation}

%% file: text/appendix/dataset.tex
\section{Dataset Details}
\label{sec:app_dataset}

This appendix summarizes dataset construction choices that affect the supervision targets and conditioning signals used in our experiments. We refer to our processed dataset as \gvt.

\subsection{Voxelizing for Training}
\label{sec:app_voxelize_train}

\paragraph{Assets.}
We construct \gvt\ from simulation-ready, textured USD assets pooled from multiple collections~\cite{nvidia_omniverse_simready, nvidia_omniverse_commercial_assets, nvidia_omniverse_residential_assets, nvidia_omniverse_vegetation_assets}. Each asset is assigned a stable instance identifier (SHA-256 hash of its canonical name within its source collection) and a semantic class label provided by source metadata.

\paragraph{Geometry Normalization and Occupancy.}
All assets are normalized to the coordinate convention of \Cref{sec:save}.
We voxelize the occupied volume at the finest grid resolution $G=1024$ (so $L_{\max}=10$) and represent geometry by the set of occupied finest-grid indices. We perform volumetric (solid) voxelization by first voxelizing the surface and then filling the interior, yielding occupancy throughout the object volume rather than only on the surface. For robustness, we cap the number of generated voxels per asset at $10^8$ during preprocessing; this cap is never active in typical cases.

\paragraph{Per-part Material Supervision.}
For supervision we require a material triplet \mattriplet\ at each occupied voxel (with $E$ in Pa, $\nu$ unitless, and $\rho$ in kg/m$^3$).
Each part of the object is assigned a constant material triplet, and each occupied voxel inherits the triplet of its part.

\subsection{Material Adaptive Tree for Training}
\label{sec:app_mat_tree}

From the voxelized material field at resolution $G=1024$, we build the supervision tree $\mathcal{T}^{\cM}$ using value-range refinement (\Cref{alg:topdown-tree}).
When multiple material samples map to the same voxel, we average them before refinement.
We use per-channel merge tolerances
\begin{equation}
  \boldsymbol{\tau} = (100.0,\ 0.01,\ 10.0),
\end{equation}
applied to $(E,\nu,\rho)$ in the units above, so that refinement is triggered only at boundaries where descendant ranges exceed these tolerances.
Coarser nodes store descendant means, enabling level-wise supervision as described in~\Cref{sec:savedetails}.

\subsection{Feature Adaptive Tree for Training and Inference}
\label{sec:app_feat_tree}

\paragraph{Multi-view Rendering.}
For each asset we render 150 RGB views at $512\times512$ resolution with transparent background and fixed lighting.
Cameras are distributed on a sphere using a low-discrepancy (Hammersley) sequence, with radius $2.1$ and horizontal field-of-view $40^\circ$.
For each view we record the camera-to-world transform and intrinsics so that features can be lifted consistently to the voxel grid.

\paragraph{DINOv3 Feature Lifting and Aggregation.}
We use DINOv3 ViT-H+/16~\cite{simeoni2025dinov3} patch-token features extracted at input resolution $512\times512$ with patch size $16$ (yielding a $32\times32$ patch grid) and feature dimension $d_{\mathrm{in}}=1280$. We apply ImageNet mean/std normalization to each view prior to the DINO forward pass.
We lift patch-token features to voxels and aggregate across views as described in \Cref{sec:method_encoder}, \Cref{sec:app_baking_dino} and \Cref{eq:dino_attenuation}, using inverse-depth attenuation with $\alpha=2.0$.

\paragraph{Adaptive Feature Tree Hyperparameters.}
To bound preprocessing cost at $G=1024$, we cap the number of occupied finest-grid voxels used for feature lifting to $100{,}000$ via stochastic subsampling when the occupied set is larger.
We build $\mathcal{T}^{\mathrm{in}}$ using the lazy refinement procedure in~\Cref{alg:bottomup-tree} with start level $\ell_{\mathrm{start}}=8$, samples per cell $K=16$, and a maximum of $N_{\max}=100{,}000$ stored nodes. We use the \emph{max-distance} uniformity test in normalized feature space (\Cref{eq:feat_uniformity}) with threshold $\tau_{\mathrm{feat}}=0.01$.

\subsection{Dataset Statistics}
\label{sec:app_dataset_stats}

\begin{figure*}[t]
\centering
\begin{subfigure}[t]{0.49\textwidth}
\centering
\includegraphics[width=\textwidth]{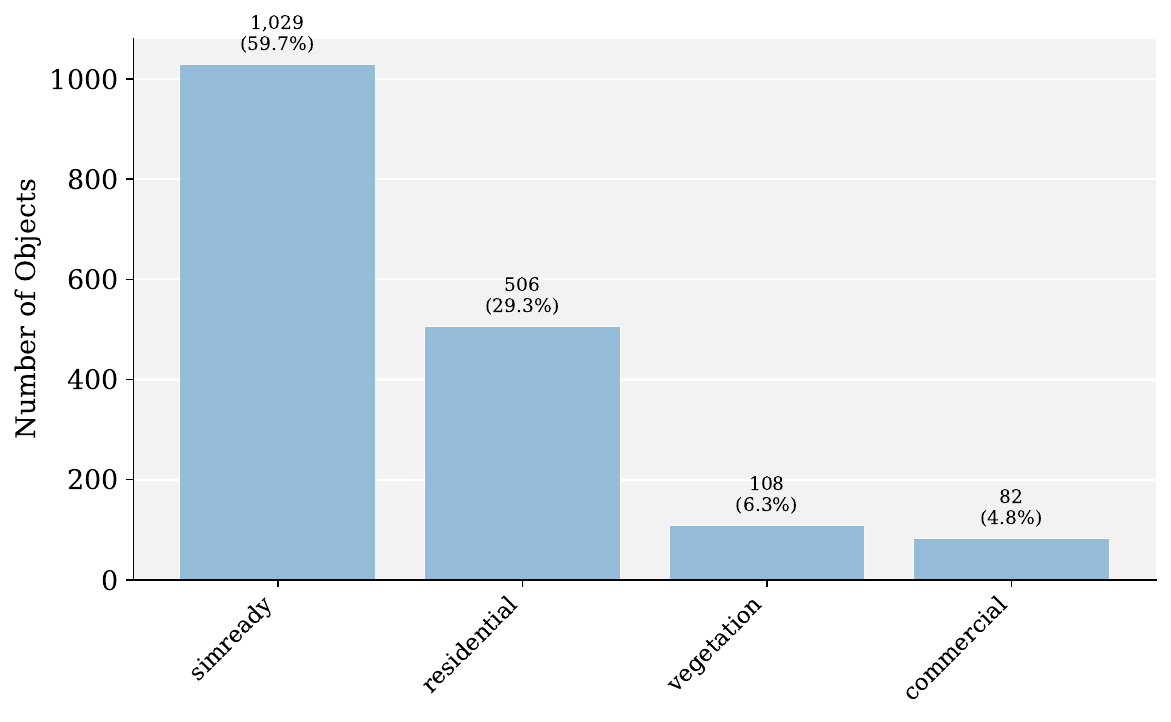}
\caption{Source-collection distribution.}
\end{subfigure}
\begin{subfigure}[t]{0.49\textwidth}
\centering
\includegraphics[width=\textwidth]{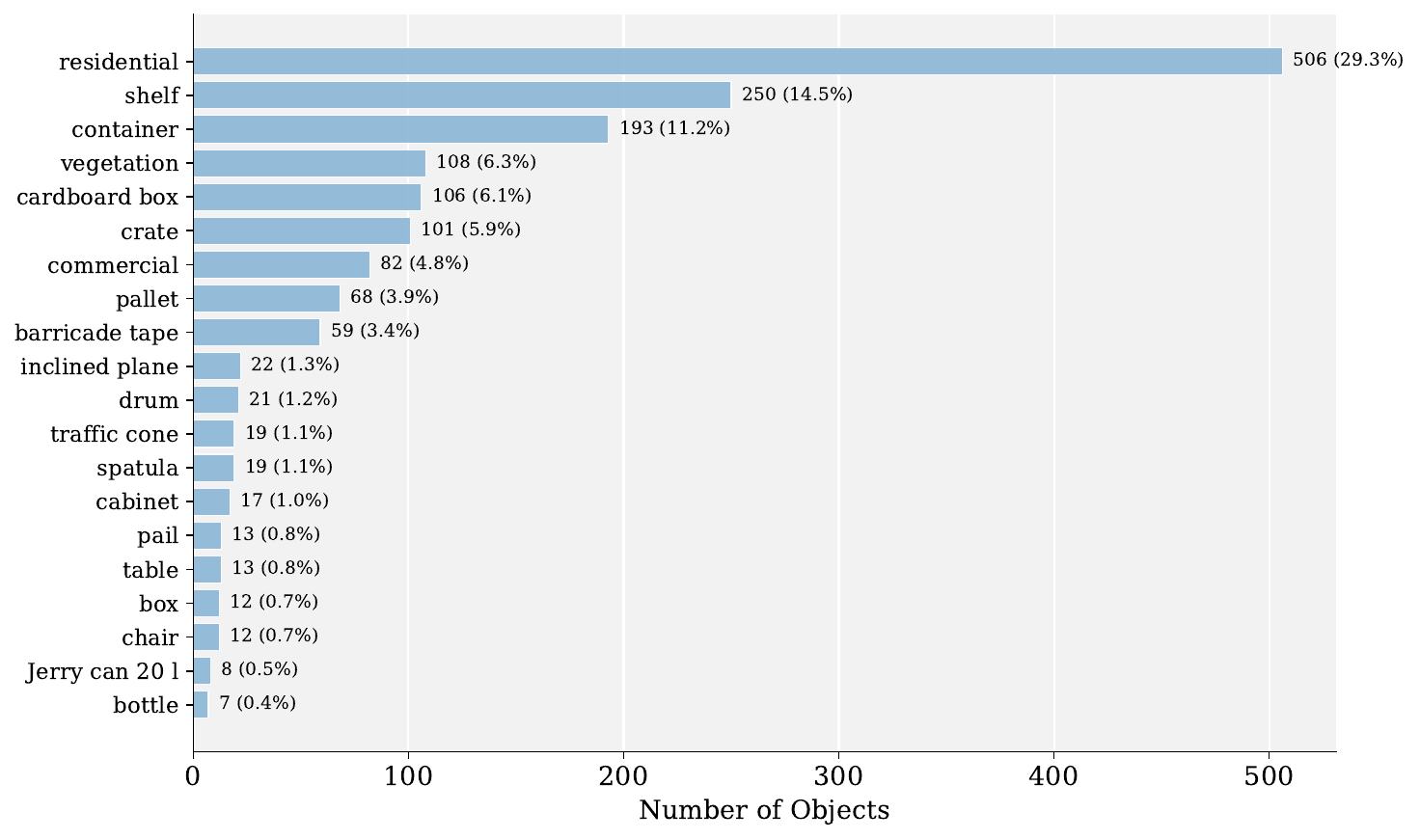}
\caption{Semantic class distribution (top classes).}
\end{subfigure}
\vspace{4pt}
\begin{subfigure}[t]{0.49\textwidth}
\centering
\includegraphics[width=\textwidth]{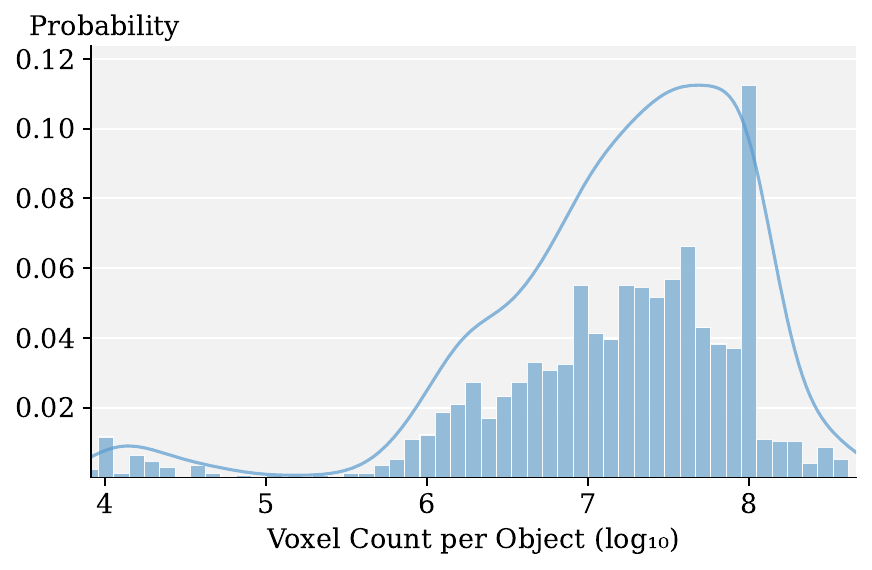}
\caption{Occupied voxel count per object.}
\end{subfigure}
\begin{subfigure}[t]{0.49\textwidth}
\centering
\includegraphics[width=\textwidth]{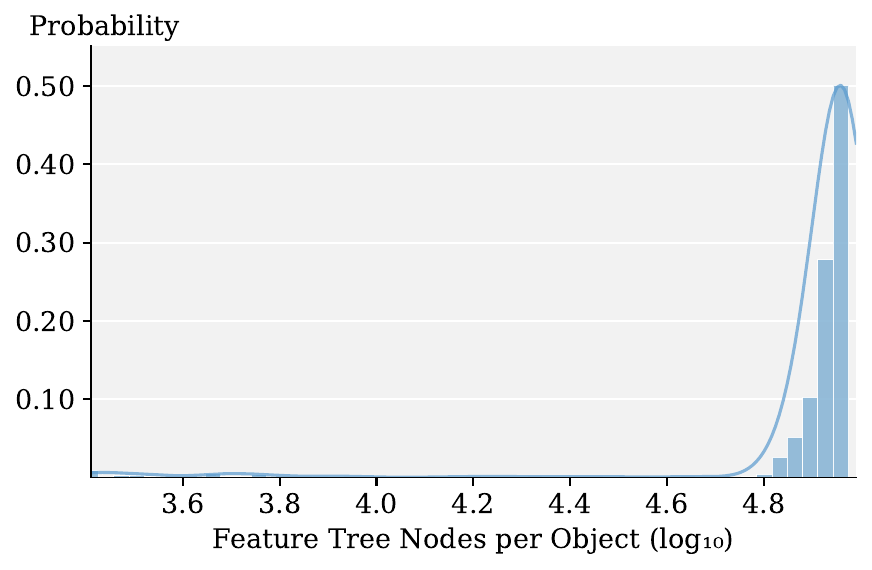}
\caption{Feature-tree nodes per object.}
\end{subfigure}
\caption{Dataset statistics for \gvt\ computed after preprocessing at $G=1024$.}
\label{fig:gvt_stats}
\end{figure*}

We report statistics of \gvt\ after the preprocessing steps in \Cref{sec:app_voxelize_train,sec:app_mat_tree,sec:app_feat_tree} at finest resolution $G=1024$. \gvt\ uses the same assets as GVM~\cite{dagli2025vomppredictingvolumetricmechanical} but slightly increases the number of assets to 1{,}725 high-quality objects. These objects like GVM~\cite{dagli2025vomppredictingvolumetricmechanical} span four source collections~\cite{nvidia_omniverse_simready, nvidia_omniverse_commercial_assets, nvidia_omniverse_residential_assets, nvidia_omniverse_vegetation_assets}, dominated by SimReady (59.7\%) and Residential (29.3\%) assets (\Cref{fig:gvt_stats}a). The dataset contains 55 semantic classes with a long-tailed distribution; the most frequent classes are \emph{residential} (29.3\%), \emph{shelf} (14.5\%), and \emph{container} (11.2\%) (\Cref{fig:gvt_stats}b).

\begin{figure*}[t]
\centering
\begin{subfigure}[t]{\textwidth}
\centering
\includegraphics[width=\textwidth]{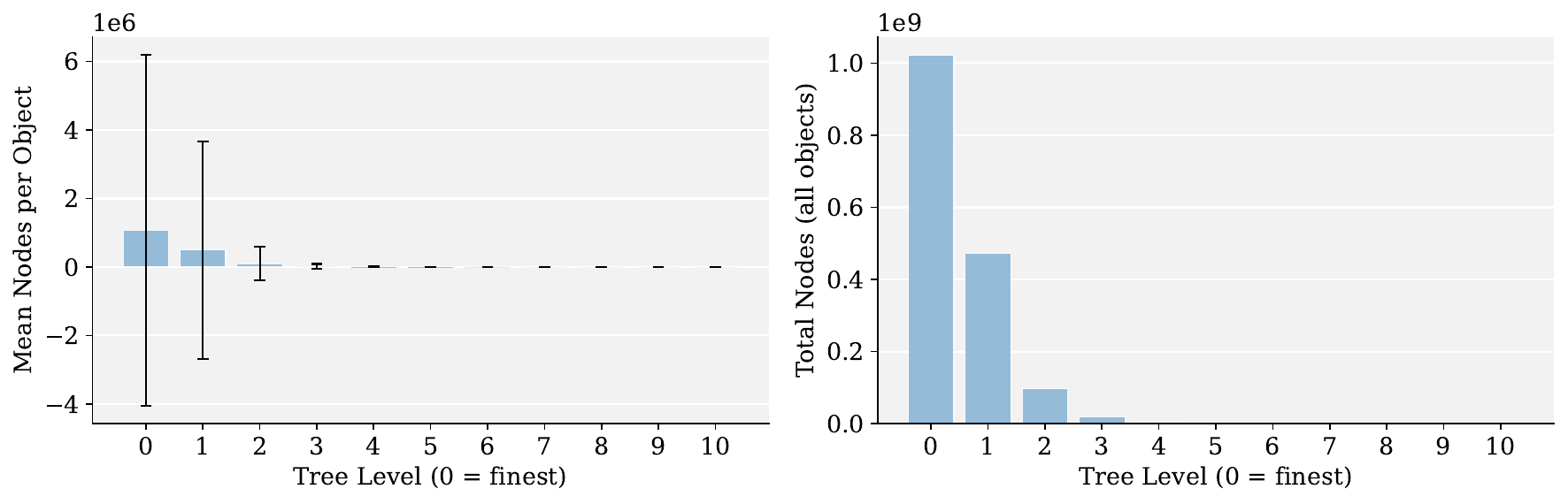}
\caption{Material tree.}
\end{subfigure}
\vspace{4pt}
\begin{subfigure}[t]{\textwidth}
\centering
\includegraphics[width=\textwidth]{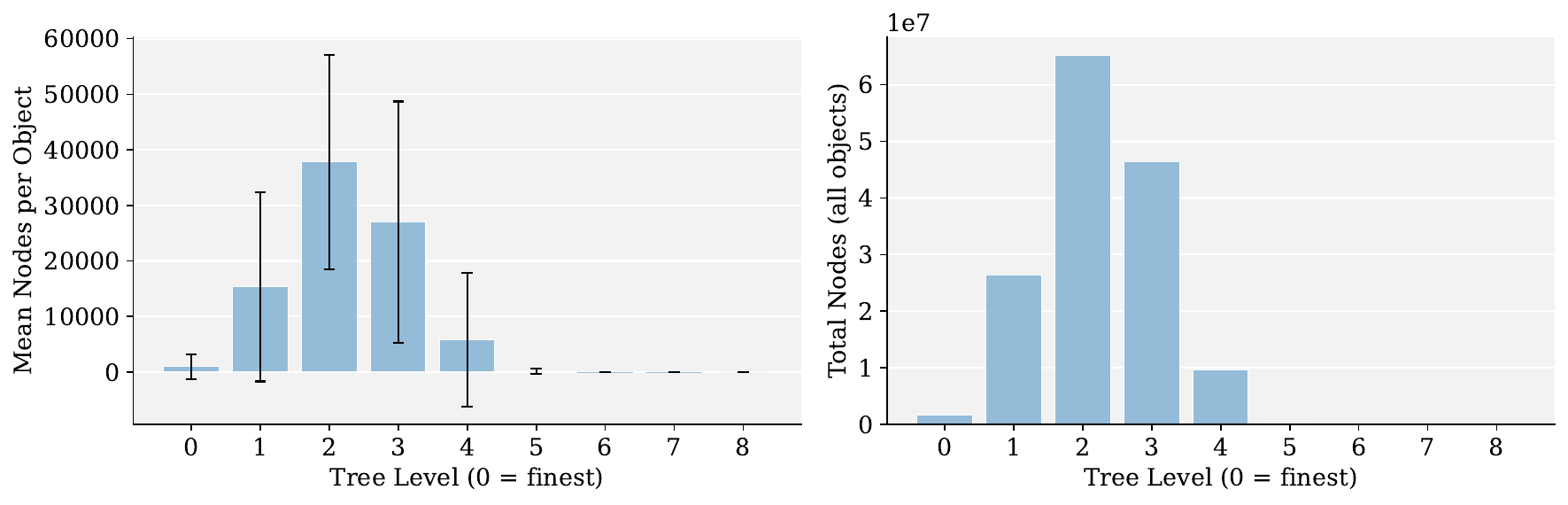}
\caption{Feature tree.}
\end{subfigure}
\caption{Distribution of \vacronym\ tree nodes across levels in \gvt. Each plot reports (left) the mean nodes per object at each level and (right) the total nodes aggregated over all objects.}
\label{fig:gvt_tree_levels}
\end{figure*}

The volumetric setting induces very large and heavy-tailed occupancy: objects contain 22.5M occupied voxels on median, and the 5th--95th percentile range spans 0.88M--123.0M occupied voxels (\Cref{tab:gvt_summary},~\Cref{fig:gvt_stats}c). At $G=1024$, this corresponds to $\approx 86.7$k input tokens per object (nodes in $\mathcal{T}^{\mathrm{in}}$) and up to $398{,}112$ output tokens per object (decoder candidates $\sum_{\ell=0}^{L_{\max}}|\cC_\ell|$ in \Cref{sec:method_decoder} under our per-level cap), which are the token counts we train on.
This motivates the adaptive discretization and candidate-only computation described in the main text. For sparsity statistics,~\Cref{fig:gvt_stats}d shows the distribution of adaptive feature-tree sizes (mean 86.7k nodes/object).~\Cref{fig:gvt_tree_levels} summarizes how material and feature tree nodes distribute across levels. Reported material-property summary statistics are computed by first averaging each property over the nodes of an object and then taking moments across objects.

\begin{table}[t]
\centering
\caption{Summary statistics of \gvt\ after preprocessing at finest resolution $G=1024$.}
\label{tab:gvt_summary}
\resizebox{\linewidth}{!}{
\begin{tabular}{lr}
\toprule
\rowcolor{nvidiagreen!15}
Statistic & Value \\
\midrule
\rowcolor{gray!15}\multicolumn{2}{l}{Dataset} \\
Objects & 1{,}725 \\
Source collections & 4 \\
Semantic classes & 55 \\
\midrule
\rowcolor{gray!15}\multicolumn{2}{l}{Voxels} \\
Total occupied voxels (all objects) & 75{,}266{,}550{,}963 \\
Occupied voxels/object (mean) & 43{,}632{,}783 \stdfmt{58{,}967{,}607} \\
Occupied voxels/object (median) & 22{,}545{,}066 \\
Occupied voxels/object (p5--p95) & 883{,}455--123{,}006{,}818 \\
Occupied voxels/object (min--max) & 8{,}200--460{,}452{,}213 \\
\midrule
\rowcolor{gray!15}\multicolumn{2}{l}{Material tree} \\
Nodes/object (mean) & 936{,}665 \stdfmt{6{,}434{,}929} \\
Nodes/object (median) & 6{,}637 \\
Nodes/object (p95) & 3{,}018{,}837 \\
Nodes/object (max) & 115{,}021{,}795 \\
Levels/object (mean) & 6.6 \\
Total nodes (all objects) & 1{,}615{,}747{,}465 \\
\midrule
\rowcolor{gray!15}\multicolumn{2}{l}{Materials} \\
Young's modulus $E$ (mean) & $3.48\times 10^{10}$ Pa \\
Poisson's ratio $\nu$ (mean) & 0.337 \\
Density $\rho$ (mean) & 1949.0 kg/m$^3$ \\
\midrule
\rowcolor{gray!15}\multicolumn{2}{l}{Feature tree} \\
Nodes/object (mean) & 86{,}719 \stdfmt{16{,}488} \\
Nodes/object (median) & 91{,}101 \\
Nodes/object (p5--p95) & 71{,}255--96{,}615 \\
Nodes/object (max) & 97{,}940 \\
Levels/object (mean) & 6.7 \\
Total nodes (all objects) & 149{,}503{,}385 \\
Feature dimension ($d_{\mathrm{in}}$) & 1280 \\
\bottomrule
\end{tabular}
}
\end{table}

%% file: text/appendix/implementation.tex
\section{Additional Details on Training}
\label{sec:app_training}

\subsection{Network Design}

\begin{figure}
    \centering
    \includegraphics{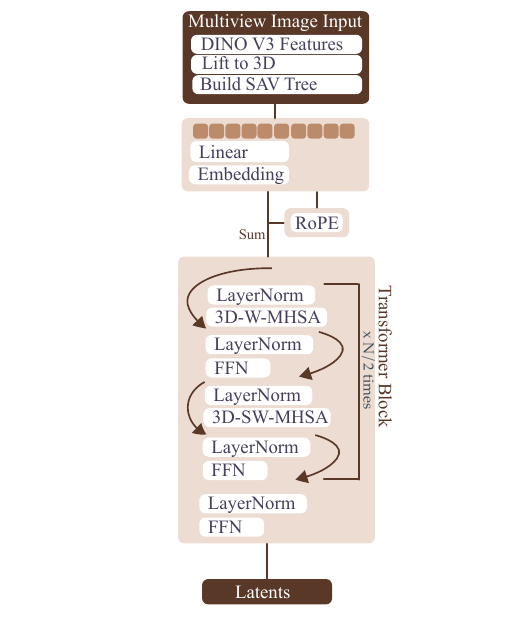}
    \caption{Encoder Network}
    \label{fig:arch_encoder}
\end{figure}

\begin{figure}
    \centering
    \includegraphics{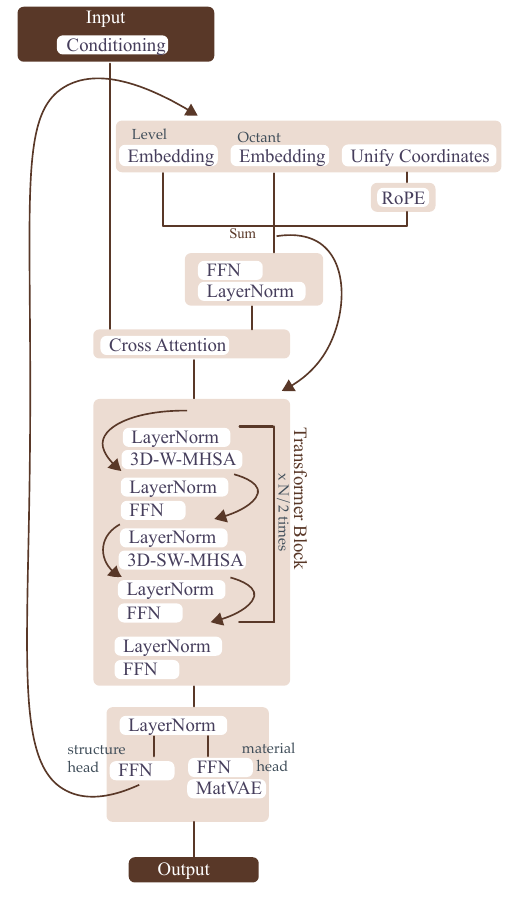}
    \caption{Decoder Network}
    \label{fig:arch_decoder}
\end{figure}

See \Cref{sec:method} for the definitions of the Adaptive Geometry Transformer (AGT) and Adaptive Material Generator (AMG), including the conditioning tree construction, unified coordinates, and the coarse-to-fine generation procedure. Here we provide the architectural parameterization used in our experiments, together with output-preserving implementation choices that are needed to train at high effective resolutions.

AGT is a sparse 3D Transformer operating on unified coordinates (\Cref{eq:unified_coords}). We use a pre-norm block with residual connections~\cite{he2015deepresiduallearningimage}:
\begin{equation}
\begin{aligned}
\bh &\leftarrow \bh + \mathrm{MSA}\big(\mathrm{LN}(\bh)\big),\\
\bh &\leftarrow \bh + \mathrm{MLP}\big(\mathrm{LN}(\bh)\big),
\end{aligned}
\end{equation}
where $\mathrm{MSA}$ is multi-head self-attention and $\mathrm{MLP}$ is a two-layer feedforward network (MLP ratio $4$ with GELU activations). Self-attention uses sparse 3D shifted-window attention~\cite{swin, swinv2, Xiang_2025_CVPR} with RoPE~\cite{rope} on unified coordinates. We use width $d_{\mathrm{model}}=768$ with $12$ heads (head dimension $64$), and apply per-head RMS normalization~\cite{10.5555/3454287.3455397} to queries and keys before the dot-product attention. AGT is initialized from a pretrained TRELLIS encoder \texttt{slat\_enc\_swin8\_B\_64l8} checkpoint~\cite{Xiang_2025_CVPR} and fine-tuned end-to-end.

AMG details the block structure of the coarse-to-fine generator described in \Cref{sec:method_decoder}. Each AMG block applies full cross-attention from level-$\ell$ candidates to AGT latents, followed by sparse shifted-window self-attention among candidates at that level, and uses the same $(d_{\mathrm{model}}, \text{heads}, \text{head dim}, \text{MLP ratio})$ as AGT. We share AMG block weights across refinement levels so that parameter count is independent of $L_{\max}$ while compute scales linearly with the number of levels. The query construction and output heads are summarized in \Cref{tab:amg_arch_details}.

AMG predicts a 2D latent that is decoded by the frozen MatVAE decoder from VoMP~\cite{dagli2025vomppredictingvolumetricmechanical}.

For numerical stability under mixed precision and large token counts, LayerNorm~\cite{ba2016layernormalization} is computed in float32 and cast back to the activation dtype. To bound peak memory without changing outputs, we evaluate token-wise operators (LayerNorm~\cite{ba2016layernormalization} and the position-wise MLP) in exact chunks when processing very large token sets. Shifted-window attention~\cite{swin, swinv2} and RoPE~\cite{rope} depend only on the discrete sparse coordinates; since the same coordinate sets are reused across many Transformer blocks and, in AMG, repeatedly across refinement levels, we cache coordinate-dependent quantities such as RoPE $\{\cos,\sin\}$ factors and window-partition index maps and reuse them across blocks. \Cref{tab:agt_arch_details,tab:amg_arch_details} summarize the block-level architectures.

\begin{table}[t]
    \centering
    \caption{Architectural details of the Adaptive Geometry Transformer (AGT). Shifted-window self-attention uses window size $32$ (unified-grid units) and RoPE on unified coordinates; per-head Q/K RMS normalization is applied before dot-product attention. The number of blocks $B_{\mathrm{enc}}$ varies by model scale (see~\Cref{tab:model_scales}).}
    
	\begin{tabular}{c|c}
        \toprule
        \rowcolor{nvidiagreen!15}
        Stage & Block \\
        \midrule
        \textbf{In\_proj} & Token embedding $\bh^0_{\ell,\bi}$ (\Cref{eq:agt_token_embed}) \\
        \midrule
        \textbf{Stem} & $\left[\begin{matrix}\operatorname{LayerNorm}\\\operatorname{QK\text{-}RMSNorm}\\\operatorname{ShiftWinSelfAttn}(12\times64)\\\operatorname{LayerNorm}\\\operatorname{Linear}(768,3072)\\\operatorname{GELU}\\\operatorname{Linear}(3072,768)\end{matrix}\right]\times B_{\mathrm{enc}}$ \\
        \bottomrule
    \end{tabular}
    \label{tab:agt_arch_details}
\end{table}

\begin{table}[t]
    \centering
    \caption{Architectural details of the Adaptive Material Generator (AMG). Each block applies cross-attention to AGT latents followed by shifted-window self-attention among candidates. AMG blocks are shared across all refinement levels, so the number of blocks $B_{\mathrm{dec}}$ is independent of $L_{\max}$ (see~\Cref{tab:model_scales}).}
	\begin{tabular}{c|c}
        \toprule
        \rowcolor{nvidiagreen!15}
        Stage & Block \\
        \midrule
        \textbf{Query} & Query embedding $\mathbf{q}_{\ell,\bi}$ (\Cref{eq:decoder_query}) \\
        \midrule
        \textbf{Stem} & $\left[\begin{matrix}\operatorname{LayerNorm}\\\operatorname{QK\text{-}RMSNorm}\\\operatorname{CrossAttn}(12\times64)\\\operatorname{LayerNorm}\\\operatorname{QK\text{-}RMSNorm}\\\operatorname{ShiftWinSelfAttn}(12\times64)\\\operatorname{LayerNorm}\\\operatorname{Linear}(768,3072)\\\operatorname{GELU}\\\operatorname{Linear}(3072,768)\end{matrix}\right]\times B_{\mathrm{dec}}$ \\
        \midrule
        \multirow{2}{*}{\textbf{Out\_proj}} & $\operatorname{Linear}(768, 3)$ (structure logits) \\
        & $\operatorname{Linear}(768, 2)$ (material latent) \\
        \bottomrule
    \end{tabular}
    \label{tab:amg_arch_details}
\end{table}

\begin{table*}[t]
\centering
\caption{Model scale configurations.}
\label{tab:model_scales}
\begin{tabular}{lcccccccc}
\toprule
\rowcolor{nvidiagreen!15}
Scale & Encoder blocks $B_{\mathrm{enc}}$ & Decoder blocks $B_{\mathrm{dec}}$ & Params (M) & $d_{\mathrm{in}}$ & $d_{\mathrm{model}}$ & Heads & Window & $d_z$ \\
\midrule
\textsc{S}  & 1  & 1  & 19.9  & 1280 & 768 & 12 & 32 & 2 \\
\textsc{B}  & 6  & 6  & 90.8  & 1280 & 768 & 12 & 32 & 2 \\
\textsc{B+} & 8  & 8  & 119.1 & 1280 & 768 & 12 & 32 & 2 \\
\textsc{L}  & 21 & 21 & 303.3 & 1280 & 768 & 12 & 32 & 2 \\
\textsc{L+} & 24 & 24 & 345.9 & 1280 & 768 & 12 & 32 & 2 \\
\textsc{H}  & 40 & 40 & 572.6 & 1280 & 768 & 12 & 32 & 2 \\
\bottomrule
\end{tabular}
\end{table*}

\subsection{Training Recipe}

We provide additional details needed to reproduce optimization and normalization, together with the full hyperparameter tables.

We train on normalized material targets to balance gradients across properties. We apply a log-minmax transform to Young's modulus and density and a minmax transform to Poisson's ratio:
\begin{align}
E' &= \frac{\log_{10}(\max(E,\epsilon)) - \log_{10}(E_{\min})}{\log_{10}(E_{\max})-\log_{10}(E_{\min})}, \\
\nu' &= \frac{\nu - \nu_{\min}}{\nu_{\max}-\nu_{\min}}, \\
\rho' &= \frac{\log_{10}(\max(\rho,\epsilon)) - \log_{10}(\rho_{\min})}{\log_{10}(\rho_{\max})-\log_{10}(\rho_{\min})},
\end{align}
with a small $\epsilon$ to avoid $\log(0)$. We reuse the normalization bounds computed for MatVAE training to ensure the latent decoder and the field generator operate in a consistent normalized space.

We optimize the multi-level objective in~\Cref{eq:train_loss} with level weights $\omega_\ell=\gamma^\ell$. We use $\lambda_{\mathrm{struct}}=1$ and report $\lambda_{\mathrm{mat}}$ and the per-property weight matrix $\boldsymbol{\Lambda}=\mathrm{diag}(\lambda_E,\lambda_\nu,\lambda_\rho)$ in~\Cref{tab:train_hparams,tab:train_hparams_large}.

We optimize with AdamW~\cite{kingma2017adammethodstochasticoptimization, loshchilov2018decoupled} for $T$ optimization steps using a linear warmup followed by linear decay. With peak learning rate $\eta_{\mathrm{peak}}$ and warmup length $T_{\mathrm{w}}$, we linearly ramp from $\eta_{\mathrm{peak}}/T_{\mathrm{w}}$ (at step $0$) to $\eta_{\mathrm{peak}}$, and then decay linearly to $\eta_{\mathrm{end}}=0.01\,\eta_{\mathrm{peak}}$ by step $T$. We clip the global gradient norm and maintain an exponential moving average of the Adaptive Geometry Transformer and Adaptive Material Generator weights. Training uses bfloat16 mixed precision for the forward pass, while optimizer state is maintained in FP32. We summarize the hyperparameters in~\Cref{tab:train_hparams,tab:train_hparams_large}.
We present details on inference-time decoding, candidate capping, and teacher forcing is provided in~\Cref{alg:amg_inference,alg:candidate_capping,alg:teacher_forcing}.

\begin{algorithm}[tb]
\caption{Inference-Time Coarse-to-Fine Decoding}
\label{alg:amg_inference}
\begin{algorithmic}[1]
\REQUIRE Conditioning latents $\mathbf{H}^{\mathrm{in}}$ from AGT; max level $L_{\max}$; initial candidate set $\cC_{L_{\max}}=\{(L_{\max},(0,0,0))\}$.
\ENSURE Generated material tree $\mathcal{T}^{\cM'}$.\vspace{0.5em}
\STATE Initialize $\mathcal{T}^{\cM'}\leftarrow\emptyset$
\STATE Initialize $\cC_{L_{\max}}\leftarrow\{(L_{\max},(0,0,0))\}$
\STATE Initialize parent-to-child state for the root as $\mathbf{0}$ (as in \Cref{eq:decoder_query})
\FOR{$\ell=L_{\max},L_{\max}-1,\dots,0$}
  \STATE Run AMG at level $\ell$ on candidates $\cC_\ell$ (queries by \Cref{eq:decoder_query}) to obtain logits $\{\mathbf{a}_{\ell,\bi}\}$ and latent means $\{\hat{\bz}_{\ell,\bi}\}$.
  \STATE $\hat{s}_{\ell,\bi}\leftarrow \arg\max \mathrm{softmax}(\mathbf{a}_{\ell,\bi})$ for each candidate.
  \STATE Enforce boundary constraints: at $\ell=L_{\max}$, disallow \textsc{Empty}; at $\ell=0$, disallow \textsc{Subdivide}.
  \STATE Add all non-empty candidates to $\mathcal{T}^{\cM'}$ with their predicted latents $\hat{\bz}_{\ell,\bi}$ (MatVAE decoding in \Cref{sec:method_decoder}).
  \IF{$\ell>0$}
    \STATE $\cC_{\ell-1} \leftarrow \bigcup_{(\ell,\bi)\in\cC_\ell:\ \hat{s}_{\ell,\bi}=\textsc{Subdivide}} \mathrm{Children}(\ell,\bi)$ %
    \IF{$\cC_{\ell-1}=\emptyset$}
      \STATE \textbf{break}
    \ENDIF
  \ENDIF
\ENDFOR
\STATE \textbf{return} $\mathcal{T}^{\cM'}$
\end{algorithmic}
\end{algorithm}

\begin{algorithm}[tb]
\caption{Window-Based Candidate Capping}
\label{alg:candidate_capping}
\begin{algorithmic}[1]
\REQUIRE Candidate grid indices $\mathbf{I}\in\mathbb{Z}^{N\times 3}$ and aligned parent-to-child states $\mathbf{P}\in\mathbb{R}^{N\times d}$ (set to $\mathbf{0}$ at the root); cap $N_{\max}$; minimum window size $w$.
\ENSURE Subsampled candidates $(\mathbf{I}',\mathbf{P}')$ and selected indices $\pi$.\vspace{0.5em}
\IF{$N\le N_{\max}$}
  \STATE $\pi\leftarrow\{1,\dots,N\}$
  \STATE \textbf{return} $(\mathbf{I},\mathbf{P}),\ \pi$
\ENDIF
\STATE \textcolor{nvidiagreen}{Compute spatial bounding box}:
\STATE $\mathbf{i}_{\min}\leftarrow \min(\mathbf{I})$
\STATE $\mathbf{i}_{\max}\leftarrow \max(\mathbf{I})$
\STATE $\boldsymbol{\delta}\leftarrow \mathbf{i}_{\max}-\mathbf{i}_{\min}+\mathbf{1}$
\STATE \textcolor{nvidiagreen}{Choose a centered window sized to contain $\approx N_{\max}$ points.}
\STATE $r\leftarrow N_{\max}/N$
\STATE $\alpha\leftarrow r^{1/3}$
\STATE $\mathbf{w}\leftarrow \max(w,\ \lfloor \alpha\,\boldsymbol{\delta}\rfloor)$
\STATE $\mathbf{c}\leftarrow \lfloor(\mathbf{i}_{\min}+\mathbf{i}_{\max})/2\rfloor$
\STATE $\mathbf{o}\leftarrow \mathrm{clip}(\mathbf{c}-\lfloor \mathbf{w}/2\rfloor,\ \mathbf{i}_{\min},\ \mathbf{i}_{\max}-\mathbf{w}+\mathbf{1})$
\STATE $\mathbf{o}^+\leftarrow \mathbf{o}+\mathbf{w}$
\STATE \textcolor{nvidiagreen}{Keep candidates in the window.}
\STATE $\pi\leftarrow\{n:\ \mathbf{o}\le \mathbf{I}_n < \mathbf{o}^+\}$ \hfill (componentwise)
\IF{$|\pi|>N_{\max}$}
  \STATE Truncate $\pi$ to its first $N_{\max}$ indices.
\ENDIF
\IF{$|\pi|<N_{\max}/4$}
  \STATE Fallback $\pi\leftarrow \{1,\dots,\min(N,N_{\max})\}$.
\ENDIF
\STATE $\mathbf{I}'\leftarrow \mathbf{I}[\pi]$
\STATE $\mathbf{P}'\leftarrow \mathbf{P}[\pi]$
\STATE \textbf{return} $(\mathbf{I}',\mathbf{P}'),\ \pi$
\end{algorithmic}
\end{algorithm}

\begin{algorithm}[tb]
\caption{Teacher-Forced Candidate Expansion and Loss Evaluation During Training}
\label{alg:teacher_forcing}
\begin{algorithmic}[1]
\REQUIRE Conditioning latents $\mathbf{H}^{\mathrm{in}}$ from AGT; ground-truth material tree $\mathcal{T}^{\cM}$ with per-level voxel sets $\{\cV^{\star}_\ell\}$; max level $L_{\max}$; candidate cap $|\cC_\ell|_{\max}$; level weights $\omega_\ell$; loss weights $(\lambda_{\mathrm{struct}},\lambda_{\mathrm{mat}})$ and $\boldsymbol{\Lambda}$.
\ENSURE Loss $\cL$ as in \Cref{eq:train_loss}.\vspace{0.5em}
\STATE \textcolor{nvidiagreen}{Initialize frontier}
\STATE $\cC_{L_{\max}} \leftarrow \{(L_{\max},(0,0,0))\}$
\STATE $\texttt{sum\_CE}\leftarrow 0$
\STATE $\texttt{sum\_MSE}\leftarrow 0$
\STATE $\texttt{cnt\_C}\leftarrow 0$
\STATE $\texttt{cnt\_P}\leftarrow 0$
\FOR{$\ell=L_{\max},L_{\max}-1,\dots,0$}
  \STATE \textcolor{nvidiagreen}{Candidate capping} If $|\cC_\ell|>|\cC_\ell|_{\max}$, subsample candidates with a contiguous spatial window (\Cref{alg:candidate_capping}).
  \STATE Run AMG at level $\ell$ on candidates $\cC_\ell$ conditioned on $\mathbf{H}^{\mathrm{in}}$ to obtain logits $\{\mathbf{a}_{\ell,\bi}\}$ and latents $\{\bz_{\ell,\bi}\}$.
  \FOR{each $(\ell,\bi)\in\cC_\ell$}
    \STATE Compute ground-truth structure label $s^\star_{\ell,\bi}$ by \Cref{eq:gt_labels}.
    \STATE $\texttt{sum\_CE} \mathrel{+}= \omega_\ell\big(-\log \mathrm{softmax}(\mathbf{a}_{\ell,\bi})(s^\star_{\ell,\bi})\big)$
    \STATE $\texttt{cnt\_C}\mathrel{+}=1$
    \IF{$s^\star_{\ell,\bi}\neq\textsc{Empty}$}
      \STATE Decode $\bz_{\ell,\bi}$ with frozen MatVAE to obtain $\hat{\mathbf{m}}_{\ell,\bi}$ and compare to target $\mathbf{m}^\star_{\ell,\bi}$.
      \STATE $\texttt{sum\_MSE} \mathrel{+}= \omega_\ell\,\big\|\hat{\mathbf{m}}_{\ell,\bi}-\mathbf{m}^\star_{\ell,\bi}\big\|^2_{\boldsymbol{\Lambda}}$
      \STATE $\texttt{cnt\_P}\mathrel{+}=1$
    \ENDIF
  \ENDFOR
  \IF{$\ell>0$}
    \STATE \textcolor{nvidiagreen}{Teacher-forced expansion}
    \STATE $\cC_{\ell-1} \leftarrow \bigcup_{(\ell,\bi)\in\cC_\ell:\ s^\star_{\ell,\bi}=\textsc{Subdivide}}\mathrm{Children}(\ell,\bi)$ \hfill (see \Cref{eq:teacher_forcing_candidates})
  \ENDIF
\ENDFOR
\STATE $\cL_{\mathrm{struct}} \leftarrow \texttt{sum\_CE}/\texttt{cnt\_C}$
\STATE $\cL_{\mathrm{mat}}\leftarrow \texttt{sum\_MSE}/\texttt{cnt\_P}$
\STATE \textbf{return} $\cL \leftarrow \lambda_{\mathrm{struct}}\cL_{\mathrm{struct}} + \lambda_{\mathrm{mat}}\cL_{\mathrm{mat}}$
\end{algorithmic}
\end{algorithm}

\begin{table*}[t]
\centering
\caption{Training hyperparameters for \textsc{S}, \textsc{B}, and \textsc{B+}.}
\label{tab:train_hparams}
\resizebox{\textwidth}{!}{%
\begin{tabular}{lccc}
\toprule
\rowcolor{nvidiagreen!15}
Hyperparameter & \textsc{S} & \textsc{B} & \textsc{B+} \\
\midrule
\rowcolor{gray!15}\multicolumn{4}{l}{Parallelism} \\
GPUs ($W$) & 16 & 16 & 16 \\
ZeRO-3 shard group size ($S_{\mathrm{shard}}$) & 16 & 16 & 16 \\
DDP replica count ($R=W/S_{\mathrm{shard}}$) & 1 & 1 & 1 \\
\midrule
\rowcolor{gray!15}\multicolumn{4}{l}{Optimization} \\
Optimizer & AdamW & AdamW & AdamW \\
AdamW $(\beta_1,\beta_2)$ & $(0.9,0.999)$ & $(0.9,0.999)$ & $(0.9,0.999)$ \\
AdamW $\epsilon$ & $10^{-8}$ & $10^{-8}$ & $10^{-8}$ \\
Weight decay & $0.01$ & $0.01$ & $0.01$ \\
Training steps ($T$) & $60{,}000$ & $60{,}000$ & $60{,}000$ \\
Batch size / GPU & 1 & 1 & 1 \\
Global batch size ($W\times$ batch/GPU) & 16 & 16 & 16 \\
Gradient accumulation steps & 1 & 1 & 1 \\
Gradient clipping (global norm) & $1.0$ & $1.0$ & $1.0$ \\
EMA decay & $0.9999$ & $0.9999$ & $0.9999$ \\
Precision (forward) & bf16 & bf16 & bf16 \\
\midrule
\rowcolor{gray!15}\multicolumn{4}{l}{Learning-rate schedule} \\
Schedule & linear warmup + linear decay & linear warmup + linear decay & linear warmup + linear decay \\
Warmup init LR ($\eta_{\mathrm{peak}}/T_{\mathrm{w}}$) & $10^{-7}$ & $10^{-7}$ & $10^{-7}$ \\
LR peak ($\eta_{\mathrm{peak}}$) & $2\times10^{-4}$ & $2\times10^{-4}$ & $2\times10^{-4}$ \\
Warmup steps ($T_{\mathrm{w}}$) & $2{,}000$ & $2{,}000$ & $2{,}000$ \\
LR end ($\eta_{\mathrm{end}}$) & $2\times10^{-6}$ & $2\times10^{-6}$ & $2\times10^{-6}$ \\
\midrule
\rowcolor{gray!15}\multicolumn{4}{l}{Loss} \\
Structure loss $\cL_{\mathrm{struct}}$ & 3-way cross-entropy & 3-way cross-entropy & 3-way cross-entropy \\
Structure loss weight $\lambda_{\mathrm{struct}}$ & $1$ & $1$ & $1$ \\
Material loss $\cL_{\mathrm{mat}}$ & $\ell_2$ on normalized triplets & $\ell_2$ on normalized triplets & $\ell_2$ on normalized triplets \\
Level-weight base $\gamma$ (for $\omega_\ell=\gamma^\ell$) & $1.4$ & $1.4$ & $1.4$ \\
Material loss weight $\lambda_{\mathrm{mat}}$ & $75$ & $75$ & $75$ \\
Per-property weights $\boldsymbol{\Lambda}$ & $(1,1,3)$ & $(1,1,3)$ & $(1,1,3)$ \\
\midrule
\rowcolor{gray!15}\multicolumn{4}{l}{Sparsity and exact chunking (output-preserving)} \\
Max candidates / level ($|\cC_\ell|_{\max}$) & $36{,}192$ & $36{,}192$ & $36{,}192$ \\
Window-attention chunk cap (tokens, $N_{\mathrm{chunk}}^{\mathrm{attn}}$) & $262{,}144$ & $262{,}144$ & $262{,}144$ \\
MatVAE decode chunk ($N_{\mathrm{chunk}}^{\mathrm{MatVAE}}$) & $128$ & $128$ & $128$ \\
\bottomrule
\end{tabular}
}
\end{table*}

\begin{table*}[t]
\centering
\caption{Training hyperparameters for \textsc{L}, \textsc{L+}, and \textsc{H}.}
\label{tab:train_hparams_large}
\resizebox{\textwidth}{!}{%
\begin{tabular}{lccc}
\toprule
\rowcolor{nvidiagreen!15}
Hyperparameter & \textsc{L} & \textsc{L+} & \textsc{H} \\
\midrule
\rowcolor{gray!15}\multicolumn{4}{l}{Parallelism} \\
GPUs ($W$) & 32 & 32 & 32 \\
ZeRO-3 shard group size ($S_{\mathrm{shard}}$) & 32 & 32 & 32 \\
DDP replica count ($R=W/S_{\mathrm{shard}}$) & 1 & 1 & 1 \\
\midrule
\rowcolor{gray!15}\multicolumn{4}{l}{Optimization} \\
Optimizer & AdamW & AdamW & AdamW \\
AdamW $(\beta_1,\beta_2)$ & $(0.9,0.999)$ & $(0.9,0.999)$ & $(0.9,0.999)$ \\
AdamW $\epsilon$ & $10^{-8}$ & $10^{-8}$ & $10^{-8}$ \\
Weight decay & $0.01$ & $0.01$ & $0.01$ \\
Training steps ($T$) & $60{,}000$ & $60{,}000$ & $60{,}000$ \\
Batch size / GPU & 1 & 1 & 1 \\
Global batch size ($W\times$ batch/GPU) & 32 & 32 & 32 \\
Gradient accumulation steps & 1 & 1 & 1 \\
Gradient clipping (global norm) & $1.0$ & $1.0$ & $1.0$ \\
EMA decay & $0.9999$ & $0.9999$ & $0.9999$ \\
Precision (forward) & bf16 & bf16 & bf16 \\
\midrule
\rowcolor{gray!15}\multicolumn{4}{l}{Learning-rate schedule} \\
Schedule & linear warmup + linear decay & linear warmup + linear decay & linear warmup + linear decay \\
Warmup init LR ($\eta_{\mathrm{peak}}/T_{\mathrm{w}}$) & $10^{-7}$ & $10^{-7}$ & $10^{-7}$ \\
LR peak ($\eta_{\mathrm{peak}}$) & $2\times10^{-4}$ & $2\times10^{-4}$ & $2\times10^{-4}$ \\
Warmup steps ($T_{\mathrm{w}}$) & $2{,}000$ & $2{,}000$ & $2{,}000$ \\
LR end ($\eta_{\mathrm{end}}$) & $2\times10^{-6}$ & $2\times10^{-6}$ & $2\times10^{-6}$ \\
\midrule
\rowcolor{gray!15}\multicolumn{4}{l}{Loss} \\
Structure loss $\cL_{\mathrm{struct}}$ & 3-way cross-entropy & 3-way cross-entropy & 3-way cross-entropy \\
Structure loss weight $\lambda_{\mathrm{struct}}$ & $1$ & $1$ & $1$ \\
Material loss $\cL_{\mathrm{mat}}$ & $\ell_2$ on normalized triplets & $\ell_2$ on normalized triplets & $\ell_2$ on normalized triplets \\
Level-weight base $\gamma$ (for $\omega_\ell=\gamma^\ell$) & $1.4$ & $1.4$ & $1.4$ \\
Material loss weight $\lambda_{\mathrm{mat}}$ & $75$ & $75$ & $75$ \\
Per-property weights $\boldsymbol{\Lambda}$ & $(1,1,3)$ & $(1,1,3)$ & $(1,1,3)$ \\
\midrule
\rowcolor{gray!15}\multicolumn{4}{l}{Sparsity and exact chunking (output-preserving)} \\
Max candidates / level ($|\cC_\ell|_{\max}$) & $36{,}192$ & $36{,}192$ & $36{,}192$ \\
Window-attention chunk cap (tokens, $N_{\mathrm{chunk}}^{\mathrm{attn}}$) & $262{,}144$ & $262{,}144$ & $262{,}144$ \\
MatVAE decode chunk ($N_{\mathrm{chunk}}^{\mathrm{MatVAE}}$) & $128$ & $128$ & $128$ \\
\bottomrule
\end{tabular}
}
\end{table*}

\subsection{Distributed Training}

We train with Hybrid Sharded Data Parallelism (HSDP) \emph{i.e.} ZeRO-3~\cite{rajbhandari2020zeromemoryoptimizationstraining}/FSDP-2~\cite{zhao2023pytorchfsdpexperiencesscaling} + Distributed Data Parallelism (DDP). We adapt Megatron-LM's Megatron-FSDP~\cite{shoeybi2020megatronlmtrainingmultibillionparameter} implementation for our training. We denote an inner group of size $S_{\mathrm{shard}}$ with data-parallel replication across $R$ outer replicas. For a world size $W$, we set $R=W/S_{\mathrm{shard}}$.  We summarize the parallelism-related hyperparameters in~\Cref{tab:train_hparams,tab:train_hparams_large}.

Adaptive trees vary substantially in depth across objects, and many samples terminate early due to no further \textsc{Subdivide} decisions at coarse levels. Under HSDP, every rank must execute an identical sequence of collectives (parameter gathers and gradient reduce-scatters) in the same order. We therefore enforce a fixed-level schedule during teacher forcing: if a sample has no real candidates at a level, decoding continues with a single dummy candidate for the remaining finer levels, and those dummy levels are masked out of the loss. This keeps collective order aligned across ranks while making it much more computationally efficient to train with diverse data.

We apply global gradient-norm clipping in the sharded setting by first aggregating the squared norm across ranks in the data-parallel group and then scaling local shards accordingly. The total number of input tokens equals the number of nodes in the adaptive feature tree, and the total number of decoder candidates equals the sum of candidate voxels across all levels. Candidate capping bounds the number of candidates processed at a level by selecting a contiguous spatial window, which controls attention memory while preserving locality; pseudo-code is provided in~\Cref{alg:candidate_capping}. For additional memory safety at high resolutions, we compute windowed self-attention in chunks, and we decode MatVAE~\cite{dagli2025vomppredictingvolumetricmechanical} outputs in chunks when the number of non-empty candidates is large.

\section{Additional Implementation Details}
\label{sec:app_implementation}

\subsection{Compute}
\label{sec:app_compute}

We report training compute in A100-80GB GPU-days, defined as the number of GPUs multiplied by wall-clock days for an end-to-end training run of AGT+AMG.~\Cref{tab:compute} lists the compute required for each model scale in~\Cref{tab:model_scales}.

\begin{table}[tb]
    \centering
    \caption{Training compute by model scale, reported in A100-80GB GPU-days.}
    \label{tab:compute}
    \begin{tabular}{lc}
        \toprule
        \rowcolor{nvidiagreen!15}
        Scale & GPU-days \\
        \midrule
        \textsc{S}  & 83 \\
        \textsc{B}  & 92 \\
        \textsc{B+} & 92 \\
        \textsc{L}  & 166 \\
        \textsc{L+} & 166 \\
        \textsc{H}  & 172 \\
        \bottomrule
    \end{tabular}
\end{table}

\subsection{Simulation and Rendering}

\subsection{Baselines}
\label{app:impl:baselines}

\paragraph{Converting Hardness to Young's Modulus.} NeRF2Physics~\cite{zhai2024physicalpropertyunderstandinglanguageembedded} does not estimate a numerical value of Young's Modulus, but instead predicts Shore A-Shore D hardness. Thus, to compare our method with NeRF2Physics~\cite{zhai2024physicalpropertyunderstandinglanguageembedded} we convert these Shore hardness values to average Young's Modulus values.

\begin{itemize}
\item {\bf Shore A.} For Shore A hardness, we follow~\cite{astm2240} and use:
\begin{equation}
E_{\text{MPa}} = e^{(S_A \times 0.0235) - 0.6403}
\end{equation}
where $S_A$ is the Shore A hardness value and $E_{\text{MPa}}$ is Young's modulus in megapascals.
\item {\bf Shore D.} For Shore D hardness, we follow~\cite{astm2240} and use:
\begin{equation}
E_{\text{MPa}} = e^{((S_D + 50) \times 0.0235) - 0.6403}
\end{equation}
where $S_D$ is the Shore D hardness value and $E_{\text{MPa}}$ is Young's modulus in megapascals.
\end{itemize}

\paragraph{Point or Voxel Sampling.} The baselines NeRF2Physics~\cite{zhai2024physicalpropertyunderstandinglanguageembedded} and PUGS~\cite{shuai2025pugszeroshotphysicalunderstanding} in their methods sample points from the NeRF or Gaussian splat, respectively, and predict mechanical properties at those points. To ensure fair comparisons, we explicitly make these methods work on the same set of points in the object on which our method is evaluated. Pixie~\cite{le2025pixie} and VoMP~\cite{dagli2025vomppredictingvolumetricmechanical} work on a fixed lower resolution voxel grid of $64^3$. When evaluating Pixie~\cite{le2025pixie} and VoMP~\cite{dagli2025vomppredictingvolumetricmechanical} at higher resolutions, we sample voxel centers and interpolate the mechanical properties at those points from the lower resolution voxel grid.

\paragraph{Implementation details of Baselines.} The baseline NeRF2Physics~\cite{zhai2024physicalpropertyunderstandinglanguageembedded} uses \texttt{gpt-3.5-turbo} for certain parts of their pipeline. We replace \texttt{gpt-3.5-turbo} in their pipeline with a better performing model, GPT-4o~\cite{openai2024gpt4technicalreport}. The baseline Phys4DGen~\cite{lin2025phys4dgenphysicscompliant4dgeneration} does not have code available. Thus, we follow the evaluation pipeline from VoMP~\cite{dagli2025vomppredictingvolumetricmechanical} and faithfully reproduce the parts, "Material Grouping and Internal Discovery" and "MLLMs-Guided Material Identification". We reproduce these parts of their pipeline using GPT-4o~\cite{openai2024gpt4technicalreport} for the MLLMs-Guided Material Identification. Furthermore, we obtained the prompts from the authors of Phys4DGen~\cite{lin2025phys4dgenphysicscompliant4dgeneration} and use the same prompts.

\section{Additional Details on the Simulations}

We experiment with Simplicits~\cite{10.1145/3658184}, a reduced order simulator~\cite{10.1145/3658184, Xiang_2026_CVPR}, an accurate finite-element method (FEM)~\cite{10.1145/3735126, gipc2024} simulator, and Isaac~\cite{NVIDIA_Isaac_Sim} for our simulations with generated mechanical properties. We share details on these simulations below.

\subsection{Material Interpolation Scheme For Simulation}
Material values in {\ourmodel} are returned on a sparse voxel grid. Simulators, however, need material values are arbitrary query locations (such as mesh vertices or monte-carlo sampled cubature points). We used nearest-neighbor interpolation on the material voxel field to source material parameters for aribtrary query points in the field. 

\subsection{Evaluating on FEM and Simplicits Simulations}
Our FEM and Simplicits simulation pipelines for evaluating predicted materials follows the same setup as introduced in the original VoMP paper \cite{dagli2025vomppredictingvolumetricmechanical}. Specifically, the predicted volumetric material parameters, \mattriplet, are converted into simulation-ready Lam\'e parameters and passed to the object's constitutive model. The FEM and Simplicits solver implementation details are identical to the original VoMP paper, as well as solver hyperparameters and pre-processing steps (such as mesh construction and simplicits basis construction). Our FEM hyperparameters are listed in~\Cref{fem:hyperpram}. All out simulations are run on an RTX A6000 - 48 GB.

\subsection{Evaluating on IsaacSim}
Additional simulations are conducted in NVIDIA Isaac Sim~\cite{NVIDIA_Isaac_Sim}, a robotics simulation platform built on NVIDIA Omniverse that leverages the PhysX physics engine~\cite{physx} for rigid body dynamics. Tb.~\ref{tab:physx_hyperparam} details the key hyperparameters used in our simulation pipeline. PhysX employs a Temporal Gauss-Seidel (TGS) iterative solver~\cite{physx} for constraint resolution. The simulation runs at 120 Hz, maintaining real-time performance. Contact handling is enhanced through Continuous Collision Detection (CCD) on the object meshes, which prevents fast-moving objects from tunneling. Material properties, including friction coefficients and restitution values, are tuned to approximate realistic object interactions observations.

\begin{table*}[tb]
\centering
\caption{Hyperparameters for FEM simulation.}
\label{fem:hyperpram}
\begin{tabular}{llll}
\toprule
\rowcolor{nvidiagreen!15}Hyperparameter & Value & Hyperparameter & Value \\
\midrule Time Integrator & Backward Euler & Linear Solver & pre-conditioned CG\\
Nonlinear Solver & Newton's w/ line search & \quad Linear tolerance & $10^{-3}$\\
\quad Newton max iters. & 1024 & Line search & \\
\quad Velocity tol.& 0.05 $ms^{-1}$ & \quad max iters & 8 \\
\quad CCD tol. & 1.0 & Collision &\\
\quad Transform rate tol. & 0.1/s & \quad Friction & 0.5\\
$dt$ & 0.02 & \quad Contact Resistance & 1.0\\
Gravity & $[0.0, -9.8, 0.0]$ & \quad $\hat{d}$ & 0.01\\
\bottomrule
\end{tabular}
\end{table*}

\begin{table}[tb]
\centering
\caption{Hyperparameters for PhysX simulation.}
\label{tab:physx_hyperparam}
\begin{tabular}{ll}
\toprule
\rowcolor{nvidiagreen!15}Hyperparameter & Value \\
\midrule 
Solver Type & PhysX TGS \\
Time Step ($dt$) & 1/120 s (0.00833 s) \\
Render Interval & 8 steps \\
Position Iterations & 4 \\
Velocity Iterations & 1 \\
Relaxation & 0.75 \\
Warm Start & 0.4 \\
Gravity & $[0.0, 0.0, -9.81]$ m/s$^2$ \\
\midrule
\multicolumn{2}{l}{\textit{Contact \& Collision}} \\
Enable CCD & True \\
Contact Offset & 0.02 m \\
Rest Offset & 0.01 m \\
Max Extraction Velocity & 100.0 m/s \\
Shape Collision Distance & 0.0 m \\
Shape Collision Margin & 0.0 m \\
Shape Appx & Cvx Decomp \\
\midrule
\multicolumn{2}{l}{\textit{Object Material Properties}} \\
Static Friction & 0.5 \\
Dynamic Friction & 0.4 \\
Restitution & 0.1 \\
\bottomrule
\end{tabular}
\end{table}

\section{Additional Related Works}

For completeness, we include other tangentially related works here. To address the trade-off between computational efficiency and representational fidelity, recent methods across various modalities have increasingly adopted adaptive, some form of multi-scale strategies different than us. In the video and language modeling, approaches use dynamic tokenization, where models adaptively compress visual tokens~\cite{DynamicVLM}, utilize multi-scale language units~\cite{10891515, barbere2024dynamic, nawrot-etal-2023-efficient}, or learn space-time tokens to reduce temporal redundancy~\cite{Liu_2017_ICCV, yan2025elastictokadaptivetokenizationimage, NEURIPS2021_6a30e32e}. Similarly, for static images, many Vision Transformers handle adaptive patch sizes, either by training a single model for variable resolutions~\cite{p2022wavemix, Hu_Cheng_Lu_Cao_Wei_Liu_Li_2024, 10.1007/978-3-031-43415-0_5, NEURIPS2021_64517d84, WANG2024106235, Beyer_2023_CVPR, Chen_2021_ICCV} or by dynamically mixing patch sizes within a single inference pass to allocate compute to complex regions~\cite{NEURIPS2021_747d3443, fan2024vitarvisiontransformerresolution, Havtorn_2023_ICCV, Chen_Lin_Li_Shen_Wu_Chao_Ji_2023, An_Zhao_Gong_Wang_Wang_Yang_2024, Ronen_2023_CVPR}. In 3D generation, some works generate geometry in a coarse-to-fine scheme from low-resolution priors or learn features over a 3D space in a coarse-to-fine scheme~\cite{10.1145/3721238.3730601, ren2024xcubelargescale3dgenerative, ren2024scubeinstantlargescalescene, Ghadai_2019_CVPR_Workshops, 10580279}.

Beyond static property prediction, extensive research explores recovering physical attributes through dynamic interaction or observation. This includes estimating parameters from video sequences~\cite{davis2015visual, Mottaghi_2016_CVPR, 10.1007/3-540-47969-4_37, chen2025vid2sim, liu2024physics3d, xue20233dintphysgeneralized3dgroundedvisual, li2025freegave, 5459407, yildirim2interpreting, li2023pacnerfphysicsaugmentedcontinuum, pmlr-v119-li20j, BMVC2016_39, NIPS2015_d09bf415, NIPS2017_4c56ff4c, Xia_2024_CVPR, xu2019densephysnetlearningdensephysical, feng2024pienerfphysicsbasedinteractiveelastodynamics, lin2024phy124fastphysicsdriven4d} or through direct physical manipulation of real-world objects~\cite{yu2024octopiobjectpropertyreasoning, 10.1145/383259.383268, lang2003scanning, lloyd2001robotic, pai2008acme, pai2000robotics, 10160731, pinto2016curiousrobotlearningvisual}. There are works that focus on generation, several works enforce physical plausibility such as gravitational stability during the synthesis of new shapes~\cite{lin2025omniphysgs, NEURIPS2024_d7af02c8, chen2024atlas3dphysicallyconstrainedselfsupporting, ni2024phyreconphysicallyplausibleneural, Yang_2024_CVPR, Mezghanni_2022_CVPR, Chen_2025_CVPR, cao2025physx-3d, cao2025sophylearninggeneratesimulationready}. However, unlike our approach, these methods focus on creating new geometry rather than augmenting existing assets with mechanical fields. Finally, alternative techniques bypass explicit property estimation entirely by directly predicting surface displacements~\cite{zhang2024adaptigraph, li2023robocook} or articulation parameters~\cite{Xia_2025_CVPR, goyal2025geopard, song2025magicarticulate, aygun2024saor, werby2025articulated, li2020category}.